\documentclass{article}
\pdfoutput=1
\usepackage{graphicx}
\usepackage[utf8]{inputenc}
\usepackage{amsmath}
\usepackage{amsfonts}
\usepackage{hyperref}
\hypersetup{colorlinks,citecolor=blue,urlcolor=blue}
\usepackage{upgreek}
\usepackage[T1]{fontenc}
\usepackage{bm}
\usepackage{mathtools}
\usepackage{amssymb,amsmath,amsthm}
\usepackage[english]{babel}
\usepackage{booktabs}
\usepackage{caption}
\usepackage{enumitem}
\usepackage[margin=1in]{geometry}
\usepackage{xcolor}
\usepackage{enumitem}
\usepackage{chngcntr}
\usepackage{subcaption}
\DeclareCaptionLabelFormat{adja-page}{\hrulefill\\#1 #2 \emph{(previous page)}}

\usepackage[utf8]{inputenc}
\usepackage{makecell}
\usepackage{mathtools}
\usepackage{tikz}
\usepackage{url}
\usepackage{verbatim}
\usepackage[capitalise]{cleveref}
\usepackage{stevemath}
\usepackage{FCmath}
\usepackage{tmpMath}
\usepackage[round,authoryear]{natbib}
\usepackage{prerex}
\usetikzlibrary{fit}
\usepackage{algorithm,algcompatible}

\usepackage{authblk}
\allowdisplaybreaks

\allowdisplaybreaks

\begin{document}

\title{Exact Asymptotics for Linear Quadratic Adaptive Control}

\author{Feicheng Wang}
\author{Lucas Janson}
\affil{Department of Statistics, Harvard University}

    
\date{}

\maketitle
\begin{abstract}


Recent progress in reinforcement learning has led to remarkable performance in a range of applications, but its deployment in high-stakes settings remains quite rare. One reason is 
a limited understanding of the behavior of reinforcement algorithms, both in terms of their regret and their ability to learn the underlying system dynamics---existing work is focused almost exclusively on characterizing \emph{rates}, with little attention paid to the constants multiplying those rates that can be critically important in practice.
To start to address this challenge, we study perhaps the simplest non-bandit reinforcement learning problem: 
linear quadratic adaptive control (LQAC)
. By carefully combining recent finite-sample performance bounds for the LQAC problem with a particular (less-recent) martingale central limit theorem, we are able to derive asymptotically-\emph{exact} expressions for the regret, estimation error, and prediction error of a 
rate-optimal stepwise-updating LQAC algorithm. 
In simulations on both stable and unstable systems, we find that our asymptotic theory also describes the algorithm's finite-sample behavior remarkably well.



\end{abstract}


\section{Introduction}
\subsection{Problem statement}

Many dynamic systems such as robots, 
power grids, 
or living cells 
can be described at any given time $t$ by a system state $x_t$ that depends on both its previous state $x_{t-1}$ and some internal or external control $u_{t-1}$ that is applied to direct the system to achieve its desired function. Both adaptive control and reinforcement learning address the problem of choosing the controls $u_t$ when the system dynamics, i.e., the relationship between $x_{t+1}$ and $(x_{t},u_{t})$, are \emph{unknown}. But the behavior of the algorithms developed in these fields has been characterized only coarsely, even in the simplest systems, preventing their deployment in high-stakes applications that require precise guarantees on safety and performance.

In this paper we will consider a canonical model for such systems, the discrete-time linear dynamical system: 
\begin{equation}
\label{eq:LinearModel}
    x_{t+1} = A x_t + B u_t + \varepsilon_t,
\end{equation}
where $x_t \in \mathbb{R}^{n}$ represents the state of the system at time $t$ and starts at some initial state $x_0$, $u_t \in \mathbb{R}^d$ represents the action or control applied at time $t$, $\varepsilon_t \iid \calN(0, \sigma ^2 I_n)$
is the system noise, and $A\in\mathbb{R}^{n\times n}$ and $B\in\mathbb{R}^{n\times d}$ are matrices determining the system's linear dynamics; the fact that they do not depend on $t$ makes this a \emph{time-homogeneous} dynamical model. The states $x_t$ and controls $u_t$ are assumed to have been transformed so that $x_t$ closer to zero represents the system better-performing its function, and $u_t$ closer to zero represents lower control cost/effort. The goal is to find an algorithm $U$ that, at each time $t$, outputs a control $u_t = U(H_t)$ that is computed using the entire thus-far-observed history of the system $H_t = \{x_t, u_{t-1}, x_{t-1},\dots,u_1,x_1,u_0\}$ to maximize the system's function while minimizing control effort.

We formalize this tradeoff by augmenting the linear dynamics \eqref{eq:LinearModel} with the popular \emph{quadratic} cost function, so that
at every time $t$, the system incurs the cost $x_t^\top Qx_t + u_t^\top Ru_t$, for some known positive-definite matrices $Q\in\mathbb{R}^{n\times n}$ and $R\in\mathbb{R}^{d\times d}$. 
In order to abstract away finite-sample issues arising from different time horizons $T$, we will focus on the infinite-horizon problem, which seeks to minimize the expected average limiting cost:
\begin{equation}
\label{eq:cost definition}
    \mathcal{J}(U) = \lim_{T\rightarrow\infty}
    \mathbb{E}\mathcal{J}(U,T),
    \hspace{1cm}
    \mathcal{J}(U,T) = \frac{1}{T}
    \sum_{t=1}^{T} \left(x_t^\top Qx_t + u_t^\top Ru_t\right).
\end{equation}
When the system dynamics $A$ and $B$ are known, the cost-minimizing algorithm is known and called the linear-quadratic regulator (LQR): $U^*(H_t) = Kx_t$, where $K\in\mathbb{R}^{d\times n}$ is the efficiently-computable solution to a system of equations that only depend on $A$, $B$, $Q$, and $R$; we will review the exact expressions for $K$ in Section~\ref{section:review on the LQR problem}. Like the Gaussian linear model in regression and supervised learning, the aforementioned linear-quadratic problem is foundational to control theory because it is conceptually simple yet it provides a remarkably good description for some real-world systems (e.g., 
biological systems \citep{priess2014solutions}, aircraft flight control \citep{choi1999lqr}, or power supply \citep{shabaani2003application}), and insights from its study often translate to innovations and improved understanding in far-more-complex models. 

In this paper we consider the case when the system dynamics $A$ and $B$ are unknown, which we call \emph{linear-quadratic adaptive control} (LQAC),
to distinguish it from the LQR setting when $A$ and $B$ are assumed known. Intuitively, one might hope that after enough time observing a system controlled by almost any algorithm, one should be able to estimate $A$ and $B$ (and hence $K$) fairly well and thus be able to apply an algorithm quite close to $U^*$. Indeed the key challenge in LQAC, as in any reinforcement learning problem, is to trade off \emph{exploration} (actions that help estimate $A$ and $B$) with \emph{exploitation} (actions that minimize cost). We will quantify the cost of an LQAC algorithm by its \emph{average regret}:\footnote{Not to be confused with the more-common \emph{cumulative regret}, given by $T\mathcal{R}(U,T)$. Since one is simply $T$ times the other, it makes no mathematical difference which one is considered, but we prefer a regret formulation that does not diverge to infinity.}
\begin{equation*}
    \mathcal{R}(U,T) = \mathcal{J}(U,T)-\mathcal{J}(U^*,T).
\end{equation*}
A flurry of recent work has proposed new algorithms for LQAC and studied their regret and estimation error; we review this literature in Section~\ref{subsection: Related Works}. These studies have produced
finite-sample bounds (in terms of the problem parameters) on various performance metrics which capture the rates at which those metrics depend on various values, especially time $T$.
These recent breakthroughs have 
advanced the field significantly, but two significant hurdles remain to using their insights to enable reliable, safe, high-performance reinforcement learning.
\begin{itemize}
    \item Many of the benefits of a theoretical characterization of the performance of an algorithm (e.g., its regret or estimation error) involve quantifying \emph{differences}, such as the difference in performance between
    two algorithms applied to the same system
    or the performance difference between applying the same algorithm to two different systems.
    But a difference between two rigorous, but loose, bounds that have the same rate can be misleading,
    since the difference in the looseness of the bounds can overwhelm the difference in the true performance. 
    
    \item 
    When an expression characterizing an algorithm's performance depends explicitly on the system dynamics (in our case, $A$ and $B$), it
    cannot actually be evaluated in practice because the system dynamics are by assumption unknown.
    Thus in order to enable certain critical aspects of reinforcement learning such as safety, non-stationarity detection, and generalization to new systems, 
    there is a pressing need to characterize algorithmic behavior in terms
    only of \emph{observable} quantities. 
\end{itemize}

\subsection{Our contribution}

This paper presents asymptotically-exact expressions for a number of quantities of interest for a simple LQAC algorithm that achieves the optimal rate of regret. That is, we prove that the performance of the algorithm converges \emph{exactly} to the expressions we present. We have two types of results: asymptotically-exact expressions in terms of non-random system \emph{parameters}, and asymptotically-exact expressions in terms of only \emph{observable} random variables. 

\paragraph{Theory for a rate-optimal algorithm with stepwise-update estimates.}
The LQAC algorithm we consider in all of the theory in this paper is very simple and intuitive, using a least-squares estimate of the system dynamics at each time point to estimate the optimal controller $K$ and adding a vanishing exploration noise to that certainty-equivalent control which can be tuned to achieve the optimal rate of regret. All our theory is for a single system trajectory (no independent restarts), and in contrast to existing literature on LQAC we allow our algorithm to update its estimate of the dynamics at \emph{every} time step, although we show our theoretical results can easily be extended to the more common setting of logarithmic updating as well.

\paragraph{Asymptotically-exact expressions characterizing LQAC performance metrics.}
For a number of different performance metrics of interest for the LQAC problem, we provide asymptotically-exact expressions (a) purely in terms of the non-random, unknown system parameters, and (b) purely in terms of the random, observable system history. In particular, we provide both types (a) and (b) of asymptotically-exact expressions for 
\begin{itemize}
    \item[(i)] the \emph{regret} at any current or future time point, 
    \item[(ii)] the distribution of the \emph{estimation error} of the least-squares estimate of system dynamics $A$ and $B$, and
    \item[(iii)] the distribution of the \emph{prediction error} of the least-squares estimate of a future state.
\end{itemize}
We further use (ii) to derive the estimation error of the least-squares estimate of the optimal controller $K$, and to identify a function of the dynamics, $A+BK$, that can be estimated at a much faster rate than just $A$ or $B$ (although to reiterate, our expressions characterize not just the rates but the exact constants multiplying those rates as well). Our observable expressions for (ii) and (iii) immediately give us asymptotically-exact online confidence regions for the system dynamics (and optimal controller $K$) and prediction regions for a future state, respectively. 

\paragraph{Numerical validation of our theory}
We apply our algorithm to both a stable and an unstable simulated system to compare our asymptotic expressions to the performance metrics they characterize, and we find quite good agreement, even at very early time steps.

\subsection{Related work}
\label{subsection: Related Works}
Our study of the asymptotics of the LQAC problem has connections with many works across control theory, machine learning, and statistics, and we defer a more thorough exposition of related work to \cref{subsection: Background}, while here only focusing on the most relevant literature.


The LQAC algorithm we consider in this paper falls into the class of algorithms which has been referred to as \emph{certainty equivalent} controllers in the literature. The key idea is to estimate the system dynamics and then apply a control that would be optimal if the estimate were correct. Following this strategy blindly is known to be inconsistent \citep{becker1985adaptive,lai1982iterated}, but a simple fix is to add a vanishing noise term, which was shown by \citet{dean2018regret} to achieve $\logO(T^{-1/3})$ average regret and later by \citet{faradonbeh2018input,faradonbeh2018optimality,mania2019certainty} to achieve $\logO(T^{-1/2})$ average regret. The recent work of \citet{simchowitz2020naive} refined the existing regret bounds and showed $\logO(T^{-1/2})$ to be the \emph{optimal} rate of average regret. To our knowledge, all LQAC algorithms that have been proved to achieve the optimal rate of regret update their estimate of the system dynamics \emph{logarithmically} often,\footnote{The only exception is \citet{abeille2018improved}, whose Thompson sampling algorithm updates its estimates at every step, but their proof only holds for scalar systems ($n=1$).} and their bounds on regret and estimation error hold in finite samples but have conservative constants multiplying the rate.

There is work on \emph{system identification} and in particular on \emph{optimal experimental design} that relates to our characterization of the estimation error of the learned system dynamics. These works focus mainly on minimizing estimation error with little or no consideration for the regret, and hence only consider algorithms with average regret bounded away from zero as this allows the optimal rate of estimation error of $\calO(T^{-1/2})$. For such algorithms (which essentially correspond to our \cref{alg:myAlg} with $\beta=1$), these works do provide asymptotically-exact expressions for the estimation error \citep{LjungSystem,bombois2006least, gerencser2009identification,hjalmarsson2009system,wahlberg2010optimal,6425920,stojanovic2014adaptive,stojanovic2016optimal,7574358}. More recent work provides finite-sample bounds on the estimation error of such algorithms, but with conservative constants multiplying the rate \citep{abbasi2011online,simchowitz2018learning,sarkar2019finite,dean2019sample,oymak2019non,sarkar2019finite,khosravi2020nonlinear,sattar2020non,foster2020learning,zheng2020non,sun2020finite}. 
The main distinction between our paper and all these related works is that we consider a stepwise-updating, regret-rate-optimal LQAC algorithm and provide characterizations of the regret, estimation error, and prediction error that are asymptotically-\emph{exact}. To achieve these results, our proofs combine recent finite-sample bounds \citep{dean2018regret,mania2019certainty} with martingale central limit theorems developed in the statistics literature \citep{lai1982least,anderson1992asymptotic}.

\subsection{Preliminaries}
\label{section:review on the LQR problem}
We make the following mild assumption on $A$ and $B$, without which \emph{no} algorithm could even achieve finite average regret.
\begin{assumption}[Stability]
\label{asm:InitialStableCondition}
    Assume the system is \emph{stabilizable}, i.e., there exists $K_0$ such that the spectral radius (maximum absolute eigenvalue) of $A+BK_0$ is strictly less than 1.
\end{assumption}
\noindent Under \cref{asm:InitialStableCondition}, there is a unique optimal controller that can be computed from $A$ and $B$, given by the linear feedback controller $u_t = K x_t$, where 
\begin{equation}
\label{eq:ControllerK}
K = - (R + B^\top P B)^{-1}B^\top P A  .
\end{equation}
Here $P$ is the unique positive definite solution to the discrete algebraic Riccati equation (DARE): 
\begin{align}
\label{eq:riccati}
  P = A^\top P A - A^\top P B (R + B^\top P B)^{-1} B^\top P A + Q
\end{align}

\section{Algorithm}
\label{section: Algorithm}
The algorithm whose performance we characterize in \cref{section: Main Theorem} is given in \cref{alg:myAlg}. At the end of each step in line~\ref{line:controller}, we apply a plug-in version of the LQR controller, $\Kh_tx_t$, plus added exploration noise 
that vanishes asymptotically
with variance 
$\tau^2t^{-(1-\beta)}\log^\alpha(t)$.
Larger $\beta$ corresponds to more exploration noise, and we will see that $\beta=1/2$ gives the optimal rate of regret and is the only $\beta$ value for which a nonzero $\alpha$ is needed in our theory.\footnote{$\beta = 1$ and $\alpha=0$ would make the added exploration noise non-vanishing and give the optimal rate of system identification estimation error; see \cref{sec:app_ext_beta1} for the extension of our results to the case of $\beta=1$.} 
$\Kh_t$ is taken as the solution to the DARE (\cref{eq:ControllerK,eq:riccati}) with inputs $\Ah_{t-1}, \Bh_{t-1}$ computed in line~\ref{line:ols}.
Line~\ref{line:check} then checks whether the state or controller is too large, and if so, $\Kh_t$ is set to $K_0$, which by assumption stabilizes the system. The cutoffs for `too large' are determined by inputs $C_x$ and $C_K$, with the latter assumed to be greater than $\norm{K}$. We will prove (\cref{prop:one_epoch_estimate_withMyalg}) the cutoffs are only breached, and hence $K_0$ applied, finitely often with probability 1, and none of $K_0$, $C_x$, or $C_K$ appear in any of our expressions characterizing the asymptotic performance of \cref{alg:myAlg}. We note that $\Kh_t$ is computed from $\Ah_{t-1}$ and $\Bh_{t-1}$ as opposed to $\Ah_t$ and $\Bh_t$---we expect this to have little impact on the performance but it is needed for the proof of the key \cref{lem:bmsb}.

Since the algorithm asymptotically always just applies a noisy plug-in version of the LQR controller, it is simple, intuitive, and computationally efficient.\footnote{The least squares estimator can be computed efficiently in a recursive manner \citep{engel2004kernel}.} All our theory and experimental results are exactly based on \cref{alg:myAlg} without any modification, and in particular, we always analyze a single trajectory (no independent restarts) and our estimates of $A$ and $B$ are updated \emph{stepwise}, i.e., at every time step. This last point is a significant departure from existing literature which focuses on \emph{logarithmic} updating. We show in \cref{fig:estimation rate one half,fig:estimation rate one half unstable} that updating stepwise reduces regret compared to updating logarithmically often, but in fact our theory also applies to a logarithmically-updated version of \cref{alg:myAlg}, as made precise in the following remark.

\begin{remark}[Logarithmically-updated estimates]
\label{remark: Logarithmically-update estimates}
All our theoretical results in \cref{section: Main Theorem} also hold when $\Ah_t$ and $\Bh_t$ are only updated $\Theta(\log(t))$ times per $t$ steps. More precisely, assume $\{t_i\}_{i=1}^\infty$ are the times at which $\Kh_{t}$ is updated. As long as there exists a constant $C$ such that $\limsup\limits_{i\rightarrow \infty}\frac{t_{i+1}}{t_i} \le C$, all results in \cref{section: Main Theorem} still hold.
\end{remark}

\begin{center}
    \begin{algorithm}[ht]
    \caption{Stepwise Noisy Certainty Equivalent Control}
    \begin{algorithmic}[1]
      \REQUIRE{Initial state $x_0$, stabilizing control matrix $K_0$, 
        scalars $C_{x} > 0$, $C_{K} > \norm{K}$, $\tau^2 > 0$, $\beta \in [1/2,1)$, and $\alpha>3/2$ when $\beta=1/2$.}
        \STATE Let $u_0 = K_0x_0 + \tau w_0$ and $u_1 = K_0x_1 + \tau w_1$, with $w_0,w_1\stackrel{iid}{\sim}\calN(0,I_d)$.
        \FOR{$t = 2,3,\dots$}
            \STATE Compute
            \begin{equation}
                \label{eq: AhBh estimator}
                (\Ah_{t-1}, \Bh_{t-1}) \in \argmin_{(A', B')} \sum_{k=0}^{t-2} \ltwonorm{x_{k+1} - A' x_k - B' u_k}^2
            \end{equation}
            and if stabilizable, plug them into the DARE (\cref{eq:ControllerK,eq:riccati}) to compute $\Kh_t$, otherwise set $\Kh_t=K_0$.
            \label{line:ols}
            \STATE If 
            $\norm{x_{t}} > C_x\log(t)$ or $\norm{\Kh_t} > C_K$, reset $\Kh_t = K_0$.\label{line:check}
            \STATE Let
            \begin{equation}\label{eq:Myinput}
            u_t = \Kh_tx_t + \eta_t,\hspace{1cm} \eta_t =  \tau\sqrt{t^{-(1-\beta)}\log^\alpha(t)}\,w_t,\hspace{1cm}w_t \stackrel{iid}{\sim} \calN(0, I_\inputdim)
            \end{equation}\label{line:controller}
        \ENDFOR
    \end{algorithmic}
    \label{alg:myAlg}
    \end{algorithm}
\end{center}

\section{Theoretical results}
\label{section: Main Theorem}

Almost all of our asymptotic results are based on the following new result which shows that the \emph{Gram matrix} 
$
\sum_{i=0}^{t-1}
\begin{bmatrix}
        x_i\\
        u_i\\
\end{bmatrix}
\begin{bmatrix}
        x_i\\
        u_i\\
\end{bmatrix}^\top 
\in \mathbb{R}^{(n+d) \times (n+d)}$ is asymptotically equal in a certain sense to the deterministic matrix $D_tD_t^\top$, where
\begin{equation}
\label{eq:D_t Definition}
        D_t := 
        t^{\beta/2}\log^{\alpha/2}(t)
\left[
    \begin{array}{cc}
    I_n & 0\\
    K & I_d\\
    \end{array}
\right]
\left[
    \begin{array}{cc}
    C_t^{1/2} & 0\\
    0 & \sqrt{\frac{\tau^2}{\beta}} I_d\\
    \end{array}
\right]
,\end{equation}
and
\begin{equation*}
C_t = t^{1-\beta}\log^{-\alpha}(t) \sum_{p=0}^{\infty}(A+BK) ^{p}((A+BK) ^{p})^\top \sigma^2  + \frac{\tau^2}{\beta}\sum_{q=0 }^{\infty}(A+BK) ^{q}BB^\top ((A+BK) ^{q})^\top
.\end{equation*}

\begin{thm}\label{thm:main tool}
\cref{alg:myAlg} applied to a system described by \cref{eq:LinearModel} under \cref{asm:InitialStableCondition} satisfies
\begin{equation}
\label{eq: Dt -1 Gram Dt -1}
D_t^{-1}
\sum_{i=0}^{t-1}
\begin{bmatrix}
        x_i\\
        u_i\\
\end{bmatrix}
\begin{bmatrix}
        x_i\\
        u_i\\
\end{bmatrix}^\top  
(D_t^\top)^{-1} \convP I_{n+d}
.\end{equation}
\end{thm}
The proof of \cref{thm:main tool} can be found at \cref{The proof of thm:main tool}. The main idea was to first prove \cref{eq: Dt -1 Gram Dt -1} under the simplifying approximation that $\Kh_t = K$, and then to derive novel uniform rate bounds on the estimation error $\Kh_t - K$ by extending existing bounds \citep{mania2019certainty,dean2018regret} to the setting of stepwise update. \cref{thm:main tool} is the key ingredient that will allow us to asymptotically exactly characterize many of the important properties of \cref{alg:myAlg}.

\subsection{Parametric expressions}
\label{subsection: Parametric expressions}
We have three different types of asymptotically-exact expressions characterizing the system performance in terms of only the non-random problem parameters (i.e., the algorithm, system, and cost function parameters): the regret (\cref{sec:regretexpr}), the distribution of the estimation error $[\Ah_t-A,\Bh_t-B]$ (\cref{sec:esterrexpr}), and the distribution of the prediction error $(\Ah_tx_t+\Bh_tu_t) - (Ax_t+Bu_t)$ (\cref{sec:prederrexpr}).

\subsubsection{Asymptotically exact expression for the regret (parametric)}\label{sec:regretexpr}
Our first result in fact does not follow from \cref{thm:main tool} but requires instead a careful decomposition of the regret paired with novel rate bounds.

\begin{thm}
\label{thm:regret}
The average regret of the controller $U$ defined by \cref{alg:myAlg} applied through time horizon $T$ to a system described by \cref{eq:LinearModel} under \cref{asm:InitialStableCondition} satisfies, as $T \to \infty$,
\begin{equation}
\label{eq:regret my alg}
\frac{\mathcal{R}(U,T)}{\tau^2\beta^{-1} \Tr(B^\top PB +R)T^{\beta-1}\log^\alpha(T)} \convP 1,
\end{equation}
with $\beta = 1/2$ therefore achieving the optimal rate \citep{simchowitz2020naive} of
$\mathcal{R}(U,T) = \logO_p(T^{-1/2})$.
\end{thm}
The proof can be found at \cref{The proof of thm:regret}. 
To our knowledge, this is the first time an LQAC algorithm's regret has been characterized asymptotically \emph{exactly}, i.e., \cref{eq:regret my alg} not only captures the rate but also the constant multiplying that rate. With an exact expression for the asymptotic regret, a user can understand exactly how the regret of \cref{alg:myAlg} depends on the system parameters, and would be able to compare this expression directly with exact expressions for other algorithms (if they existed).

\subsubsection{Asymptotic distribution of the estimation error (parametric)}\label{sec:esterrexpr}
\cref{thm:main tool} provides the key ingredient in a martingale central limit theorem (CLT) for the estimators $\Ah_t, \Bh_t$ \citep{anderson1992asymptotic}, 
which gives the exact asymptotic distribution of the estimation error in terms of only the system parameters. 

\begin{thm}\label{thm:main CLT}
\cref{alg:myAlg} applied to a system described by \cref{eq:LinearModel} under \cref{asm:InitialStableCondition} satisfies, as $t \to \infty$,
\begin{equation}
\label{eq:final Conclusion}
    \vvector \left(    
    \begin{bmatrix}
        \Ah_t - A,\Bh_t- B
    \end{bmatrix} D_t\right) \convD 
    \calN(0, \sigma^2 I_{\statedim(n+d)} )
.\end{equation}
\end{thm}
The proof of \cref{thm:main CLT} can be found at \cref{The proof of thm:main CLT}. 
Again, to our knowledge, this is the first time an LQAC algorithm's estimation error has been characterized asymptotically \emph{exactly}
and, similarly, such a result can help a user understand exactly how the distribution of the estimation error of \cref{alg:myAlg} depends on the system parameters.

\begin{remark}[A convergence rate disparity]
\label{remark: fast convergence rate}
Plugging the definition of $D_t$ \cref{eq:D_t Definition} into \cref{eq:final Conclusion} gives different convergence rates for two  different parts of $[\Ah_t - A, \Bh_t - B]$. In particular, as $t \to \infty$,
    \begin{equation}
    \label{eq: fast slow rate CLT}
       \vvector \left(    
    \begin{bmatrix}
        t^{\beta/2} \log^{\alpha/2}(t)C_t^{1/2}(\Ah_t - A + (\Bh_t- B)K), & \sqrt{\frac{\tau^2}{\beta}}t^{\beta/2} \log^{\alpha/2}(t)(\Bh_t- B)
    \end{bmatrix} 
    \right) \convD 
    \calN(0, \sigma^2 I_{\statedim(n+d)} ).
    \end{equation}
    Thus $\Ah_t - A + (\Bh_t- B)K$ converges at the rate of $\left(t^{\beta/2} \log^{\alpha/2}(t)C_t^{1/2}\right)^{-1} = \calO_p(t^{-1/2})$ for any $\beta$, while $\Bh_t- B$ converges at the slower $\beta$-dependent rate of $\calO_p(t^{-\beta/2}\log^{-\alpha/2}(t))$. The faster convergence rate of $\Ah_t - A + (\Bh_t- B)K$ implies strong dependency between $\Ah_t - A$ and $\Bh_t- B$: $\Ah_t - A \approx -(\Bh_t- B)K$.
\end{remark}


\begin{remark}[Regret-estimation trade-off]
Because of the asymptotic linear relationship $\Ah_t - A \approx -(\Bh_t- B)K$, the estimation error $[\Ah_t - A, \Bh_t - B]$ can be characterized by the asymptotic variance of $\vvector[\Bh_t- B]$:  $\frac{\beta\sigma^2}{\tau^2}t^{-\beta}\log^{-\alpha}(t) I_{nd}$. Combining this with \cref{thm:regret} gives the following asymptotic identity that precisely characterizes a fundamental regret-estimation trade-off for \cref{alg:myAlg} with any $\beta$: as $t \to \infty$,
\begin{equation*}
     t\mathcal{R}(U,t) \cdot \Cov(\vvector(\Bh_t- B)) \convP \Tr(B^\top PB +R) \sigma^2 I_{nd}
.\end{equation*}
\end{remark}

Because $K$ is a function of $[A, B]$ (and asymptotically, $\Kh_t$ is the same function of $[\Ah_{t-1}, \Bh_{t-1}]$), by the Delta method, we can use its matrix of derivatives $\frac{dK}{d[A, B]} := \frac{d \,\vvector(K)}{d \,\vvector([A, B])} \in \mathbb{R}^{nd \times n(n+d)}$ to translate the asymptotic distribution of $[\Ah_t - A, \Bh_t - B]$ from \cref{thm:main CLT} to the asymptotic distribution of $\Kh_t - K$.

\begin{corr}
\label{corr: K CLT parametric}
Assume $A+BK$ is full rank. Then \cref{alg:myAlg} applied to a system described by \cref{eq:LinearModel} under \cref{asm:InitialStableCondition} satisfies, as $t \to \infty$,
\begin{equation}
    \label{eq: K CLT parametric}
    \sqrt{\frac{\tau^2}{\sigma^2\beta}}t^{\beta/2} \log^{\alpha/2}(t)
    \left(\left(\frac{dK}{d[A,B]}\right) 
    \left(\begin{bmatrix}
    -K^\top \\
    I_d
    \end{bmatrix}
    \otimes 
    I_n
    \right)
    \right)^{-1}
    \vvector\left(\Kh_t - K\right)
    \convD 
    \calN(0, I_{nd})
.\end{equation}
\end{corr}
The proof of \cref{corr: K CLT parametric} can be found at \cref{subsection: The proof of K CLT parametric}. \cref{eq: K CLT parametric} quantifies the distance from the current control matrix $\Kh_t$ to the optimal control matrix $K$, 
and shows implicitly but asymptotically exactly how the distribution of that distance depends on the system dynamics. 

\subsubsection{Asymptotic distribution of the prediction error (parametric)}\label{sec:prederrexpr}
If we consider the entire history $\{x_i, u_i\}_{i=0}^{t}$ to be the input of the prediction rule whose goal is to predict the next state $x_{t+1}$, then the optimal (in terms of mean squared error) prediction is given by $\E [x_{t+1} \,|\, \{x_i, u_i\}_{i=0}^{t} ] = Ax_t + Bu_t$, and a natural choice at time $t$ would be to use the least-squares prediction rule given by $\Ah_t x_t + \Bh_t u_t$.
By combining \cref{thm:main CLT}'s asymptotic distribution for $[\Ah_t - A, \Bh_t- B]$  with a careful handling of the asymptotic dependence between $(x_t, u_t)$ and $[\Ah_t - A, \Bh_t- B]$, we can derive the asymptotic distribution of the error $\Ah_tx_t + \Bh_t u_t - (Ax_t + Bu_t)$ of the least-squares prediction rule. 
\begin{thm}\label{thm:prediction CLT parametric}
\cref{alg:myAlg} applied to a system described by \cref{eq:LinearModel} under \cref{asm:InitialStableCondition} satisfies, as $t \to \infty$,
\begin{align}
\label{eq:prediction CLT parametric}
    \left(
    x_t^\top\left( \sum_{p=0}^{\infty}(A+BK) ^{p}
    \left((A+BK) ^{p}\right)^\top\right)^{-1}x_t  
    +
     \beta \sigma^2
     \lnorm{w_t}^2
    \right)^{-1/2} 
     t^{1/2}  \left((\Ah_t - A)x_t + (\Bh_t- B)u_t\right) 
     \convD \calN(0,I_n).
\end{align}
\end{thm}
The proof of \cref{thm:prediction CLT parametric} can be found at \cref{The proof of thm:prediction CLT parametric}. This expression is parametric in the sense that the first parenthetical only depends on the system parameters and the random variables $x_t$ and $w_t$ that are used by the algorithm in the time step immediately before the prediction is made.
Note that the convergence rate of $\logO_p(t^{-1/2})$ does not depend on $\beta$, as foreshadowed by \cref{remark: fast convergence rate}, but the constant in the convergence does depend on $\beta$. 
Thus, \cref{eq:prediction CLT parametric} shows that the optimal asymptotic prediction error is attained at $\beta = 1/2$ ($x_t$'s asymptotic distribution does not depend on $\beta$, so asymptotically the only $\beta$ dependence is in the term $\beta\sigma\|w_t\|^2$), a conclusion we could not have reached had we only considered the rate. 
\cref{thm:prediction CLT parametric} can easily be extended to characterize the full prediction error of $x_{t+1} - (\Ah_tx_t + \Bh_t u_t)$ by simply adding $\sigma^2$ to the first parenthetical.




\subsection{Observable expressions}
\label{subsection: Observable expressions}
The previous subsection provides three asymptotically-exact expressions (regret, estimation error, and prediction error) in terms of only the system parameters; in this subsection, we provide three analogous asymptotically exact expressions in terms of only observable random variables.

\subsubsection{Asymptotically exact expression for the regret (observable)}
Define $\Ph_t$ as the plug-in estimator using \cref{eq:riccati}:
\begin{align*}
  \Ph_t = \Ah_t^\top \Ph_t \Ah_t - \Ah_t^\top \Ph_t \Bh_t (R + \Bh_t^\top \Ph_t \Bh_t)^{-1} \Bh_t^\top \Ph_t \Ah_t + Q.
\end{align*}
Then by consistency of $\Ah_{t}$ and $\Bh_{t}$ (see \cref{thm:main CLT}), and therefore also $\Ph_t$, the plug-in version of \cref{eq:regret my alg} is an immediate corollary of \cref{thm:regret}.

\begin{corr}
\label{corr:regret}
The average regret of the controller $U$ defined by \cref{alg:myAlg} applied through time horizon $T$ to a system described by \cref{eq:LinearModel} under \cref{asm:InitialStableCondition} satisfies, as $t \to \infty$ and $T \to \infty$,
\begin{equation}
\label{eq:regret my alg observable}
\frac{\mathcal{R}(U,T)}{\tau^2\beta^{-1} \Tr(\Bh_{t}^\top \Ph_t \Bh_{t} +R)T^{\beta-1}\log^\alpha(T)} \convP 1
.\end{equation}
\end{corr}

The proof of \cref{corr:regret} can be found at \cref{The proof of corr:regret}.
Notice when $t \le T$, \cref{corr:regret} tells us that we can consistently estimate the regret at a future time point. Furthermore, the Delta method applied to \cref{thm:main CLT} gives the asymptotic distribution of the denominator in \cref{eq:regret my alg observable}. 


\subsubsection{Asymptotic distribution of the estimation error (observable)}


Combining the asymptotic equivalence of Gram matrix and $D_tD_t^\top$ from \cref{thm:main tool}, the asymptotic distribution of the estimation error from \cref{thm:main CLT}, and Slutsky's theorem immediately produces the following very useful corollary. 

\begin{corr}\label{thm:main}
\cref{alg:myAlg} applied to a system described by \cref{eq:LinearModel} under \cref{asm:InitialStableCondition} satisfies
\begin{equation*}
    \Tr\left( 
    \begin{bmatrix}
        \Ah_t - A,\Bh_t- B
    \end{bmatrix} 
    \sum_{i=0}^{t-1}
    \begin{bmatrix}
            x_i\\
            u_i\\
    \end{bmatrix}
    \begin{bmatrix}
            x_i\\
            u_i\\
    \end{bmatrix}^\top
    \begin{bmatrix}
        \Ah_t - A,\Bh_t- B
    \end{bmatrix} ^\top 
    \right)
    \convD \sigma^2\chi^2_{n(n+d)}
.\end{equation*}
\end{corr}
The proof of \cref{thm:main} can be found at \cref{The proof of thm:main}. The reason it is useful is it allows us to construct an asymptotically exact ellipsoidal confidence region for the system dynamics $A$ and $B$. In particular, the following confidence region has asymptotic coverage exactly $1-\alpha$ and is entirely and efficiently computable from data observable through time $t$:

\begin{equation}
\label{eq: ellipsoid 2}
\left\{
A, B \,:\,
    \sigma^{-2}
            \Tr\left( 
    \begin{bmatrix}
        \Ah_t - A,\Bh_t- B
    \end{bmatrix} 
    \sum_{i=0}^{t-1}
    \begin{bmatrix}
            x_i\\
            u_i\\
    \end{bmatrix}
    \begin{bmatrix}
            x_i\\
            u_i\\
    \end{bmatrix}^\top
    \begin{bmatrix}
        \Ah_t - A,\Bh_t- B
    \end{bmatrix} ^\top 
    \right)
    \le \chi^2_{n(n+d), 1-\alpha}
\right\},
\end{equation}
where $\chi^2_{n(n+d), 1-\alpha}$ is the $1-\alpha$ quantile of a $\chi^2_{n(n+d)}$ random variable. To our knowledge, this is the first asymptotically exact confidence region for the system dynamics in the LQAC problem. 
Note the confidence region in \cref{eq: ellipsoid 2} is identical to the confidence region one would compute if the data points $\{x_i, u_i\}_{i=0}^{t-1}$ were i.i.d., but the theory that led us to this result is far more challenging than in the i.i.d. setting.

Analogously to \cref{corr: K CLT parametric}, we can also use the Delta method to derive a confidence region for $K$.

\begin{corr}
\label{corr: K confidence region}
Assume $A+BK$ is full rank. Then \cref{alg:myAlg} applied to a system described by \cref{eq:LinearModel} under \cref{asm:InitialStableCondition} satisfies
\begin{equation*}
    \label{eq: K confidence region}
    \vvector(
        \Kh_t - K
    ) ^\top 
        \left(
        \left(\frac{dK}{d[A, B]}\right)_t
    \left(
\sum_{i=0}^{t-1}
    \begin{bmatrix}
            x_i\\
            u_i\\
    \end{bmatrix}
    \begin{bmatrix}
            x_i\\
            u_i\\
    \end{bmatrix}^\top
\otimes I_n\right)^{-1}
    \left(\frac{dK}{d[A, B]}\right)_t^\top
    \right)^{-1}
    \vvector(
        \Kh_t - K
    ) 
    \convD \sigma^2\chi^2_{nd}
,\end{equation*}
where $\left(\frac{dK}{d[A, B]}\right)_t \in \mathbb{R}^{nd \times n(n+d)}$ is defined as $\frac{dK}{d[A, B]}$ evaluated at $\Ah_{t-1}, \Bh_{t-1}$. 
\end{corr}
The proof of \cref{corr: K confidence region} can be found at \cref{subsection: The proof of K confidence region}. 
\cref{corr: K confidence region} gives the following asymptotically exact ellipsoidal $1-\alpha$ confidence region for $K$:
\begin{equation*}
\left\{
K \,:\,
\sigma^{-2}
    \vvector(
        \Kh_t - K
    ) ^\top 
        \left(
        \left(\frac{dK}{d[A, B]}\right)_t
    \left(
\sum_{i=0}^{t-1}
    \begin{bmatrix}
            x_i\\
            u_i\\
    \end{bmatrix}
    \begin{bmatrix}
            x_i\\
            u_i\\
    \end{bmatrix}^\top
\otimes I_n\right)^{-1}
    \left(\frac{dK}{d[A, B]}\right)_t^\top
    \right)^{-1}
    \vvector(
        \Kh_t - K
    )
    \le 
    \chi^2_{nd, 1-\alpha}
\right\}.
\end{equation*}

\subsubsection{Asymptotic distribution of the prediction error (observable)}
We can obtain an observable expression for the asymptotic distribution of the prediction error as a direct corollary of \cref{thm:main tool,thm:prediction CLT parametric}.



\begin{corr}
\label{thm: prediction CLT}
\cref{alg:myAlg} applied to a system described by \cref{eq:LinearModel} under \cref{asm:InitialStableCondition} satisfies: 
\begin{align*}
\begin{split}
    \left(
    \sigma^2
\begin{bmatrix}
        x_t \\
        u_{t}
    \end{bmatrix}^\top
    \left(
    \sum_{i=0}^{t-1}
    \begin{bmatrix}
            x_i\\
            u_i\\
    \end{bmatrix}
    \begin{bmatrix}
            x_i\\
            u_i\\
    \end{bmatrix}
    ^\top
    \right)^{-1}
    \begin{bmatrix}
        x_t \\
        u_{t}
    \end{bmatrix}
    \right)^{-1/2} \left((\Ah_{t} - A)x_t + 
    (\Bh_{t}- B)u_{t}\right) 
    \convD \calN(0, I_n)
.\end{split}
\end{align*}
\end{corr}
The proof can be found in \cref{The proof of thm: prediction CLT}, and is a special case of a more general result that allows the users to choose their own desired input by replacing $u_t = \Kh_t x_t + \eta_t$ with $u_t = \Kh_t x_t + \xi_t$ for any $\xi_t$ constant or independent of the data. Again, \cref{thm: prediction CLT} can easily be extended to characterize the full prediction error of $x_{t+1} - (\Ah_tx_t + \Bh_t u_t)$ by simply adding $\sigma^2$ to the first parenthetical, leading to the following prediction region:
\begin{equation}
\label{eq: prediction region observable}
    \left\{
    x_{t+1}  \,: \,
    \sigma^{-2}        
    \left(
    1+
    \begin{bmatrix}
            x_t \\
            u_{t}
        \end{bmatrix}^\top
\left(
    \sum_{i=0}^{t-1}
    \begin{bmatrix}
            x_i\\
            u_i\\
    \end{bmatrix}
    \begin{bmatrix}
            x_i\\
            u_i\\
    \end{bmatrix}
    ^\top
    \right)^{-1}
        \begin{bmatrix}
            x_t \\
            u_{t}
        \end{bmatrix}
        \right)^{-1}
          \norm{(\Ah_t x_t + \Bh_t u_t) - x_{t+1}}^2
    \le \chi^2_{n, 1-\alpha}
    \right\}.
\end{equation}
Having at each time $t$ a computable region with a high probability of containing the next state $x_{t+1}$ is a crucial ingredient in ensuring the \emph{safety} of a learning system, as it both provides a warning about where the system will be next and gives the system the opportunity to change or cancel the control $u_t$ if the prediction region intersects an unsafe part of the state space.

As an additional application of the prediction region \cref{eq: prediction region observable}, since $x_{t+1}$ is observed at the next time step, we can use the agreement between our prediction region and the true $x_{t+1}$ to test certain assumptions about our system. For instance, the hypothesis test which rejects if $x_{t+1}$ does not fall within the prediction region constructed at time $t$ constitutes a asymptotically valid level-$\alpha$ test of our stationary linear dynamics encoded in \cref{eq:LinearModel}. For instance, if we are confident about the linearity of our system but worried that it may be non-stationary, we could use this test to detect whether the dynamics have changed within the first $t+1$ time steps, and more generally, such tests could be strung together to constitute a change detection algorithm \citep{grunwald2019safe,wang2020online}. 

Note that the naive prediction region
\begin{equation}
\label{eq: prediction region observable simple}
    \left\{
    x_{t+1}  \,: \,
    \sigma^{-2}        
          \norm{(\Ah_t x_t + \Bh_t u_t) - x_{t+1}}^2
    \le \chi^2_{n, 1-\alpha}
    \right\}.
\end{equation}
also has asymptotically exact coverage even though it ignores the estimation error in $[\Ah_t,\Bh_t]$. However, our experiments show that our prediction region from \cref{eq: prediction region observable} achieves much better finite-sample coverage by accounting for the estimation error of $[\Ah_t,\Bh_t]$; see \cref{fig:Prediction Coverage True one half}.

\section{Experiments}
\label{section: Experiments}
We verify our algorithm's performance in one stable and one unstable dynamical system. 
We focus on comparing the finite sample performance of our algorithm to our theoretical predictions, and defer comparison between our algorithm and other existing algorithms for future work (see \citet{dean2018regret} for a comparison between an algorithm similar to our algorithm except it updates $\Kh_t$ logarithmically often and other algorithms which we will review in \cref{subsection: Background}). In the main text, we will only display the figures with $\beta = 1/2$ and $\alpha = 2$ in the stable system; the remaining figures and details of the experimental setup can be found in \cref{section: Additional Experiments}. \footnote{Source code for reproducing our results can be found at \url{https://github.com/Feicheng-Wang/LQAC_code}.}


\subsection{A representative simulation}
\label{subsection: Partial experiment result}
\cref{fig:Summary stable system} summarizes the results of our experiment with $\beta = 1/2$ and $\alpha = 2$ in a stable system (for the analogous figure in an unstable system see \cref{fig:Summary unstable system}). The main takeaways are:



\begin{itemize}
    \item \cref{fig:Compare Regret one half} shows that \cref{alg:myAlg}'s stepwise update leads to lower regret than update logarithmically often, although the difference is small compared with the variability of the regret. The difference is qualitatively similar but quantitatively larger in the unstable system, and the difference can be quite large for poor choices of $K_0$, but pretty robust for choices of $C_K$; see \cref{fig:Compare Regret one half unstable,fig:log regret three choices unstable system,fig:log regret CK}.
    

    \item \cref{fig:regret ratio one half} verifies that the ratio of the true observed regret with either of our regret expressions in \cref{thm:regret} and \cref{corr:regret} is converging to 1. Note that the large confidence band is due to the huge variance in the regret itself. The analogous plots for $\beta \neq 1/2$ and the unstable system can be found at \cref{fig:regret ratio one half unstable,fig:regret_stable_unstable}; larger $\beta$ speeds up the convergence speed.
    
    
    
    \item \cref{fig:estimation rate one half} verifies the convergence rate disparity in \cref{remark: fast convergence rate} that $\Ah_t-A$, $\Bh_t-B$, and $\Kh_t -K$ have a slow convergence rate $\logO(t^{-\beta/2})$, while $\Ah_t-A + (\Bh_t-B) K$ has a fast convergence rate $\logO(t^{-1/2})$
    ;  see \cref{fig:estimation rate one half unstable}.
    
    
    \item 
    \cref{fig:Coverage one half} shows that, the finite sample coverage of our confidence regions and prediction region closely matches our asymptotic theory in \cref{thm:main,corr: K confidence region,thm: prediction CLT}.
    Also \cref{fig:Prediction Coverage True one half} shows that our prediction region \cref{eq: prediction region observable} have better finite sample coverage than the naive region \cref{eq: prediction region observable simple}. In this simulation, the observable expressions have slightly better coverage. 
    Similar results hold for other choices of $\beta$ and the unstable systems (\cref{fig:Coverage one half unstable,fig:log regret no Cx}).
    
\end{itemize}

\begin{figure}[H]
\centering

\begin{subfigure}{.45\textwidth}
  \captionsetup{justification=centering}
  \caption{Benefit of stepwise updates}
  \includegraphics[width = \linewidth]{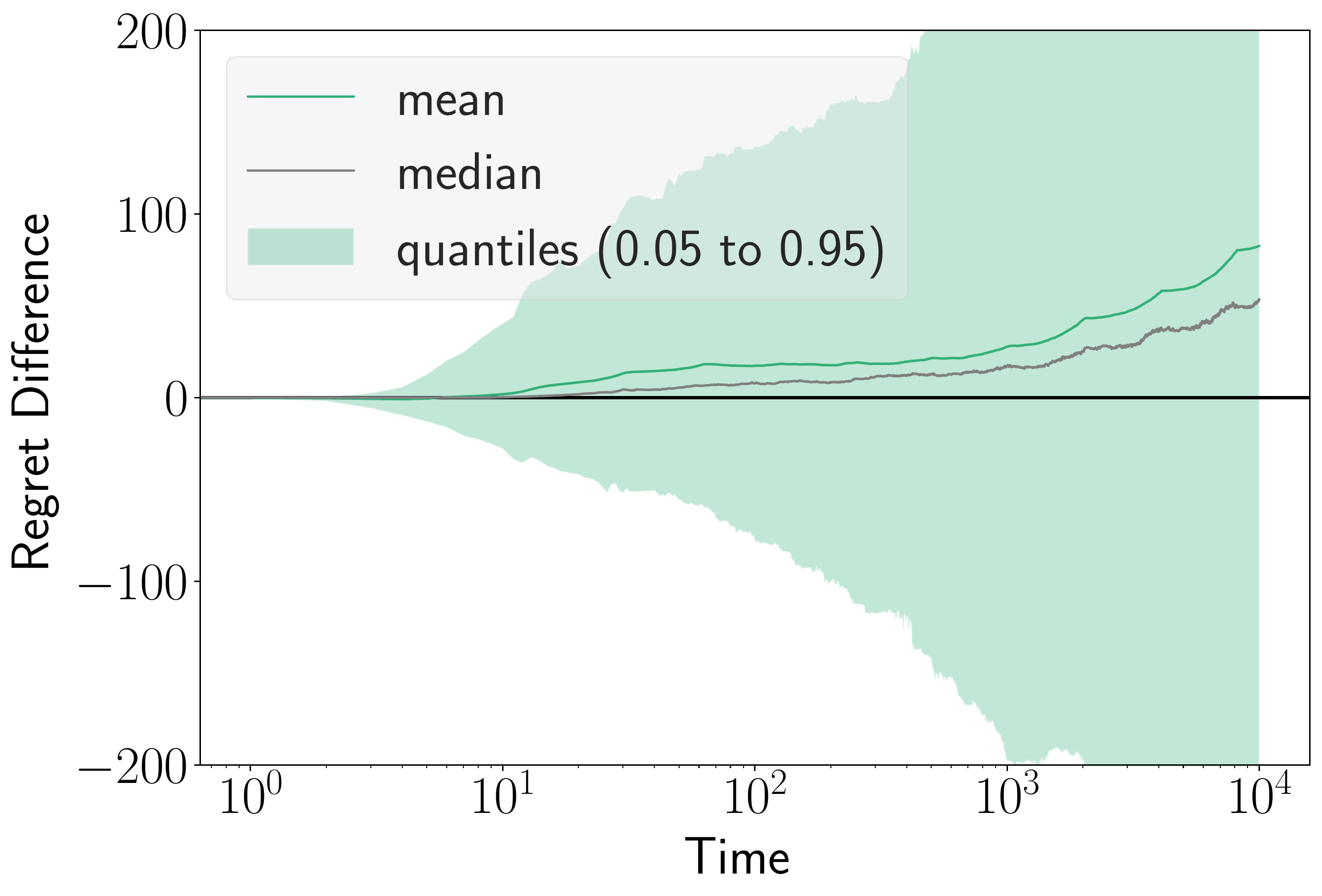}
  \label{fig:Compare Regret one half}
\end{subfigure}
\begin{subfigure}{.45\textwidth}
  \caption{Regret Ratio}
  \includegraphics[width = \linewidth]{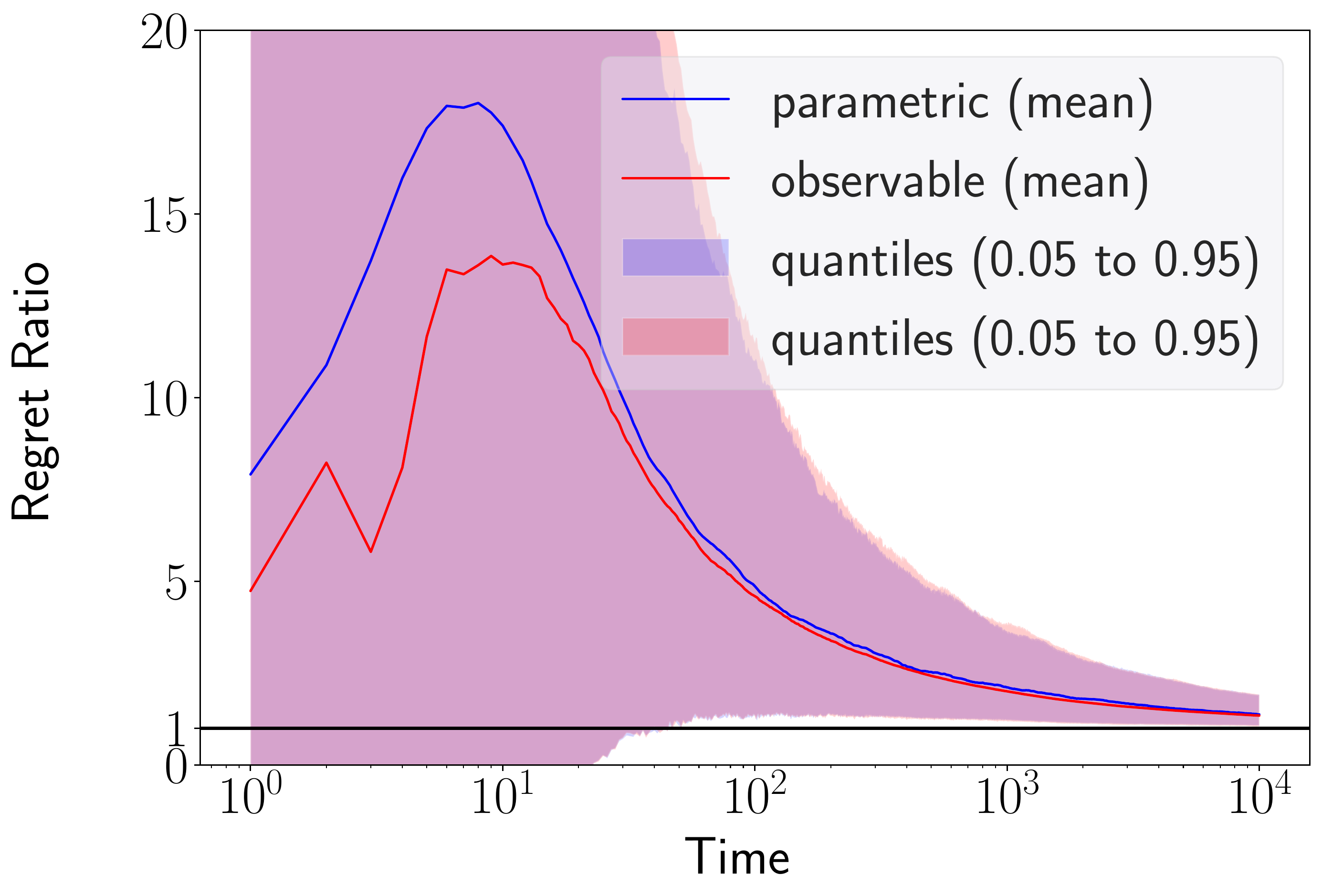}
  \label{fig:regret ratio one half}
\end{subfigure}

\begin{subfigure}{.45\textwidth}
\captionsetup{justification=centering}
  \caption{Differing Convergence Rates}
  \includegraphics[width = \linewidth]{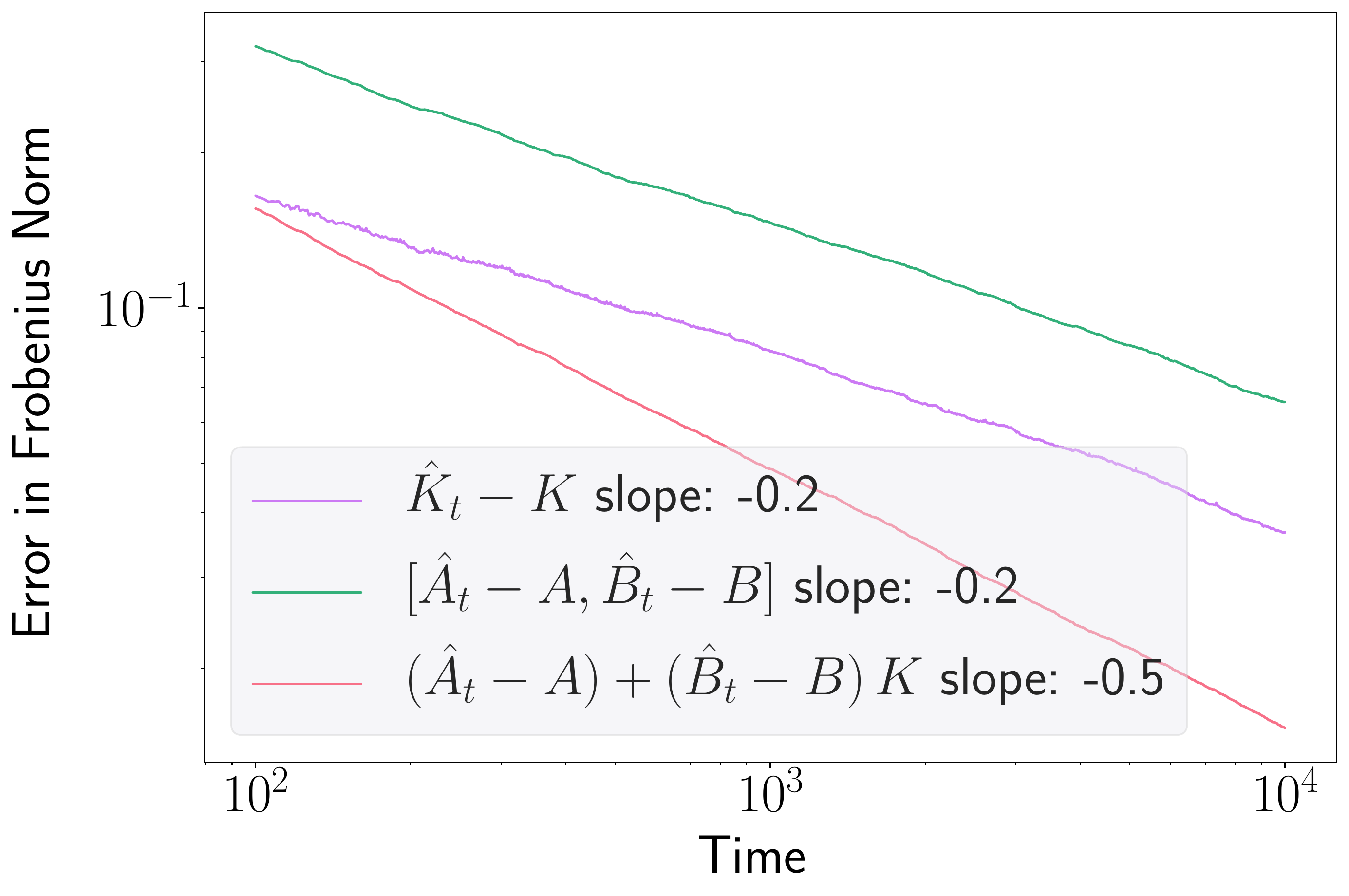}
  \label{fig:estimation rate one half}
\end{subfigure}
\begin{subfigure}{.45\textwidth}
  \caption{Confidence Region Coverage}
  \includegraphics[width = \linewidth]{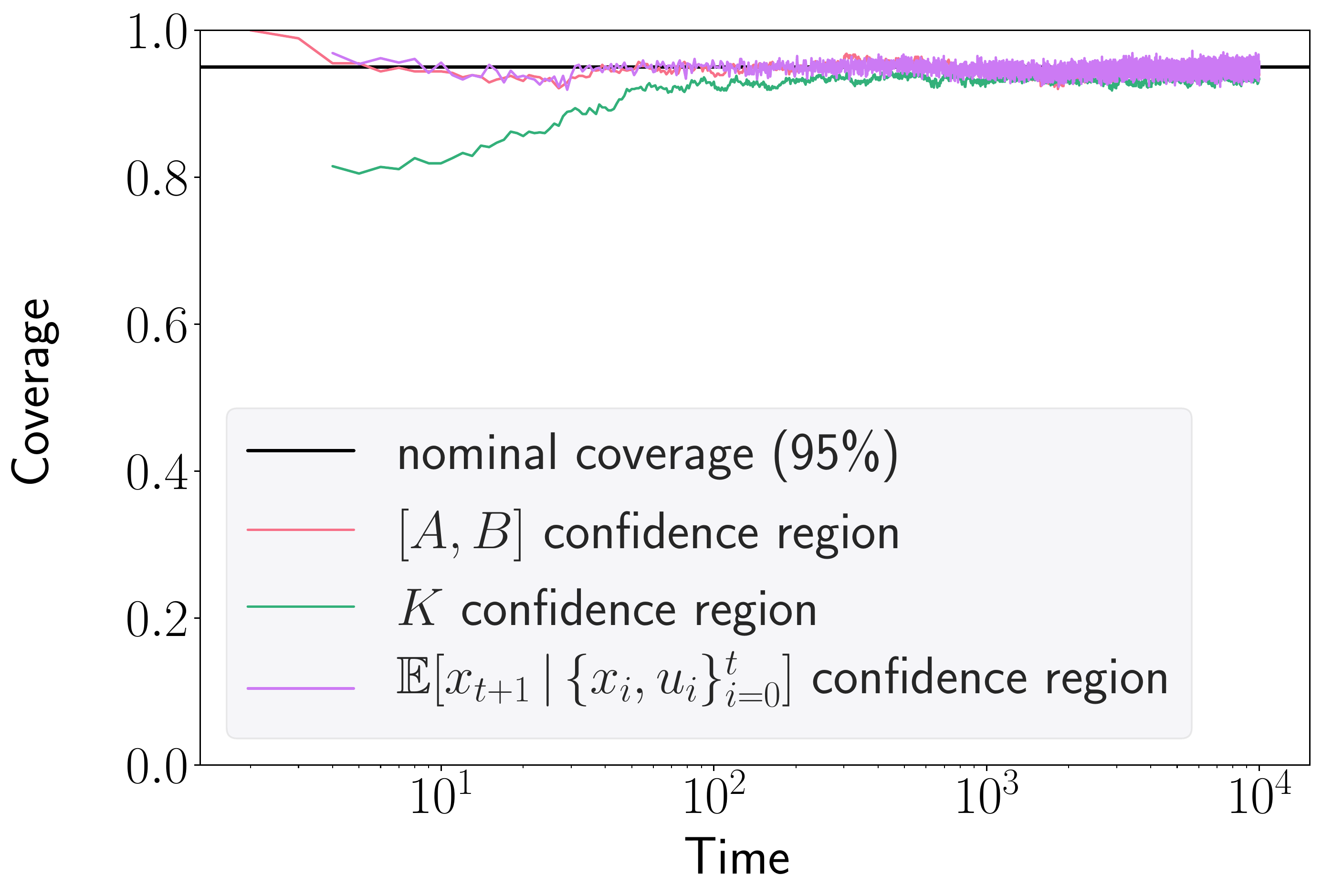}
  \label{fig:Coverage one half}
\end{subfigure}


\begin{subfigure}{.45\textwidth}
  \caption{Prediction Region Coverage}
  \includegraphics[width = \linewidth]{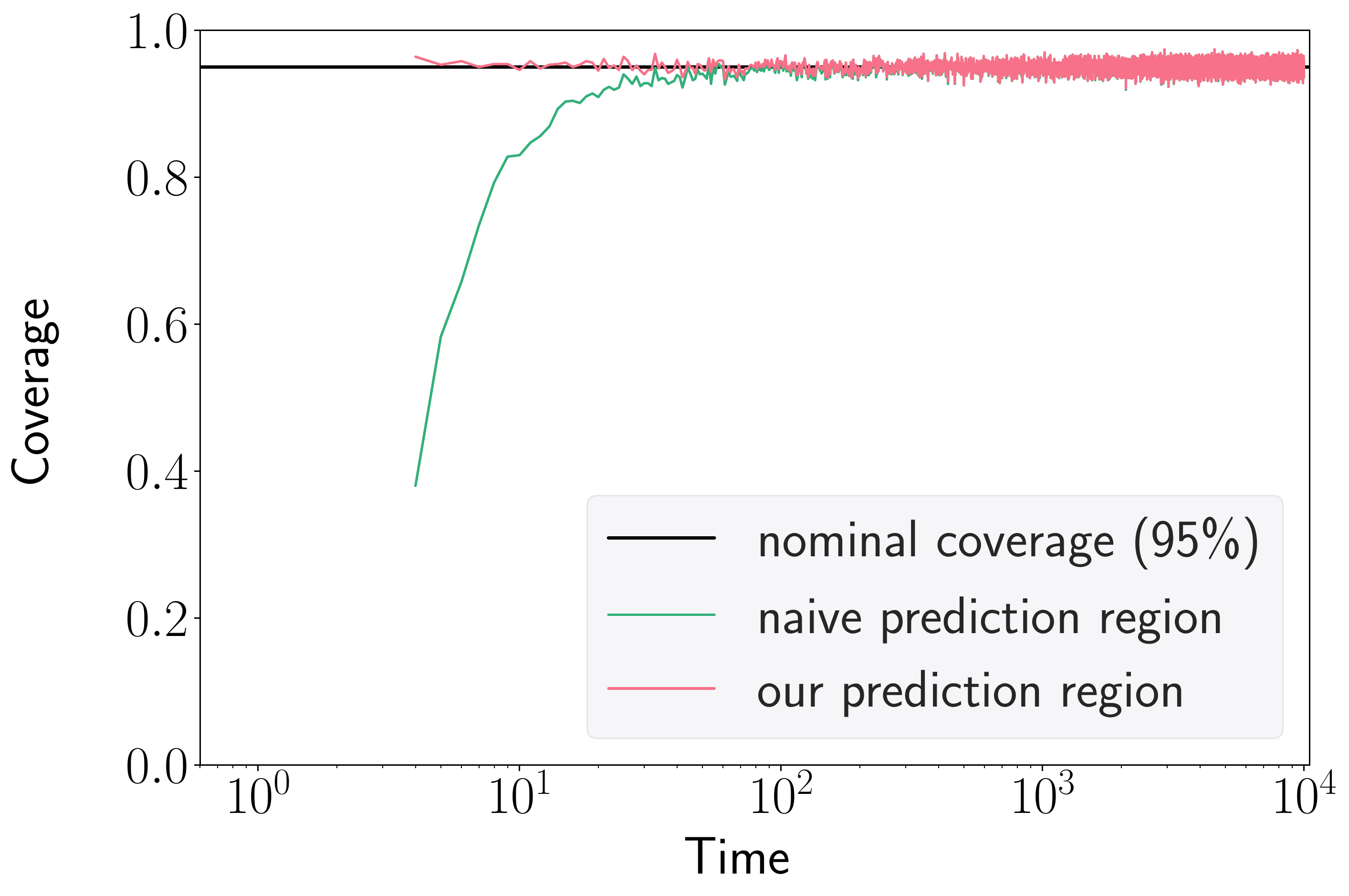}
  \label{fig:Prediction Coverage True one half}
\end{subfigure}
\caption[]{\textit{(See next page for caption)}}
\end{figure}

\begin{figure}[H]
\ContinuedFloat
\caption{Summary of 1000 independent experiments applying \cref{alg:myAlg} with $\beta = 1/2$, $\alpha = 2$, $C_x = 1$, and $C_K = 5$ on the stable system described in \cref{subsection: easy setting}. (a) Difference between the regret of \cref{alg:myAlg} using stepwise and logarithmic updates. 
(b) The ratio of the empirical regret and our parametric or observable expressions for the regret. 
(c) The average Frobenius norm of various estimation errors considered in this paper, with slopes fitted on a log-log scale so that the estimation error is $\logO(t^{\text{slope}})$. 
The effect of $\alpha$ was removed from the slopes of $\Kh_t-K$ and $[\Ah_t-A,\Bh_t-B]$ by dividing the error by $\log^{\alpha/2}(t)$.
(d) Coverage of our 95\% confidence regions for $[A,B]$, $K$, and $\E [x_{t+1} \,|\, \{x_i, u_i\}_{i=0}^{t} ] = Ax_t + Bu_t$. 
(e) Coverage of our 95\% prediction region for $ x_{t+1} \,|\, \{x_i, u_i\}_{i=0}^{t}$, along with coverage of the naive prediction region given in \cref{eq: prediction region observable simple}.
}
    \label{fig:Summary stable system}
\end{figure}

\section{Detailed review of related work}
\label{subsection: Background}
The LQAC problem lies at the intersection of adaptive control and reinforcement learning and has drawn considerable attention in the past decade. This line of work differs from much of the work in reinforcement learning that is based on games or other virtual simulators that can be rerun infinitely many times (\citet{vinyals2017starcraft}, \citet{silver2017mastering}) because it is run in one-shot. However, many real-world applications cannot be easily restarted over and over again, and repeating experiments can be prohibitively expensive. Aside from the CE approach taken in this paper and reviewed in \cref{subsection: Related Works}, we classify LQAC algorithms into two broad categories:

\begin{itemize}
    \item \textbf{Optimism in the Face of Uncertainty}: This method uses non-convex optimization to repeatedly select a near optimal control (in the regret sense) from a confidence set, achieves the optimal rate of regret \citep{abbasi2011regret,ibrahimi2012efficient,faradonbeh2017finite}. 
    Later \citet{cohen2019learning} extended this work by replacing non-convex optimization with semi-definite programming and still achieves the optimal regret. 
    
    

    
    \item \textbf{Thompson Sampling}: 
    Starting with a prior distribution for the system parameters, one can use Bayes' rule to update a posterior distribution online and can use samples from that posterior to choose controls that balance exploration and exploitation.
    The pioneering work \citep{abeille2017thompson} applying this idea to LQAC demonstrated a suboptimal $\logO(T^{-1/3})$ average regret, which is later improved to the optimal rate $\logO(T^{-1/2})$ by \citet{ouyang2017learning,faradonbeh2018optimality}. \citet{abeille2018improved} is the only work which we know of that achieves the optimal rate with stepwise updates, although their proofs only apply in scalar systems (i.e., $n=1$).

\end{itemize}

\paragraph{Logarithmic Regret}
    We pause here to clarify that any result achieving logarithmic regret is in a different setting from ours (in our setting, a lower bound of $\logO(T^{-1/2})$ was proven in \citet{simchowitz2020naive}). For example, when the system parameters $A$ and $B$ are known or partially known, a logarithmic rate of regret is achievable due to the extra information in $A$ and $B$ which allows faster estimation of $K$ \citep{foster2020logarithmic,cassel2020logarithmic}. Or,  when the states are only partially observed, although the controller receives less information, the optimal controller also has less information, which turns out to allow a logarithmic rate of regret \citep{lale2020logarithmic,tsiamis2020online}. As a final example, when the cost is not an explicit function of the controls $u_t$, a logarithmic rate of regret is achievable using a controller called a self-tuning regulator, which is similar to our certainty equivalent controller
    except that it targets a different optimal controller $U^*$ (because the cost function is different) and applies constant size probing steps logarithmically often \citep{lai1986extended,lai1986asymptotically,guo1991astrom,guo1995convergence}.

\paragraph{Sequential Analysis and Time Series}
Establishing asymptotic normality is common in sequential analysis \citep{lai2001sequential} and time series or state space model analysis \citep{kohn1986prediction,pedroni2004panel}, but the focus in these fields is on stationary and Markovian time series (although we assume our system is stabilizable, the data generated by applying our adaptive controller to that system is non-Markovian and non-stationary as the controller depends on the whole history) and on simpler forms of dependence than we consider.

\section{Discussion}
This paper's main contributions are asymptotically exact expressions for the regret and the distributions of the estimation and prediction errors of a stepwise updating noisy certainty equivalent control algorithm in terms of either the system parameters or observable random variables. These results improve the field's understanding of the LQAC problem and open up a number of new research directions: 

\begin{enumerate}
    \item \textbf{Theoretical improvements}. Our simulations support our suspicion that all of our results except for \cref{thm:regret} and \cref{corr:regret} hold under more general version of \cref{alg:myAlg} that allows $\beta = 1/2$ and $\alpha = 0$, the summation in Line \ref{line:ols} to go up to $t-1$, and the removal of Line \ref{line:check}. We expect such extensions to require significantly stronger theoretical machinery, and we hope that future work will prove these extensions and analogues to \cref{thm:regret} and \cref{corr:regret} which account for an expected additional term of order $\calO(T^{-1/2})$.
    

    \item \textbf{Safe reinforcement learning}. 
    Existing work in safe reinforcement learning relies heavily on prediction regions derived from Bayesian inference \citep{berkenkamp2017safe,koller2018learning}. Our \cref{thm: prediction CLT} provides a tight frequentist asymptotic prediction region that, unlike Bayesian inference, does not assume a prior on the system parameters, providing a potential starting point for new safe reinforcement learning algorithms.


\item \textbf{Non-stationarity reinforcement learning}. As mentioned in the last paragraph of \cref{section: Main Theorem}, our prediction region can be used for change point detection in non-stationary systems. Many existing work designed for reinforcement learning algorithms in the non-stationary environment relies on some form of change point detection, although they focus on discrete state and action spaces \citep{da2006dealing,auer2009near,padakandla2019reinforcement}. Thus, our work may be useful for designing new reinforcement learning algorithms in non-stationary settings with continuous state and action spaces.

\end{enumerate}


\section*{Acknowledgements}
We are grateful to Na Li, Haoyi Yang, and Yue Li for helpful discussions regarding this project.



\clearpage
\bibliography{ref}
\clearpage
\appendix
\counterwithin{figure}{section}
\counterwithin{theorem}{section}





\section{Preliminaries}
\subsection{Notation}

Let us first review the definition of $\calO(\cdot)$, and generalize the notation to contain relative constants $\theta$, as well as introducing a new notation representing constant functions that we know exactly the order as well as the coefficient in front of the largest order term.
\begin{defn}
\label{defn: Big O notation}
Let $f$ and $g$ both be real valued function, and suppose $g(x)$ is strictly positive for any $x$ large enough. Then
\begin{enumerate}
    \item $f(x) = \calO(g(x))$ if and only if $\exists x_0$, $\abs{f(x)} \le Mg(x)$ for any $x \ge x_0$.
     \item $f(x) = \logO(g(x))$ if and only if $\exists x_0$ and $\exists k \in \mathbb{Z}$, $\abs{f(x)} \le Mg(x)\log^k(g(x))$ for any $x \ge x_0$.

    \item $f(x) = \calO(g(x))$ is a fixed function with regard to $x$ such that $\exists C > 0$, and $\lim_{x\to\infty}\abs{f(x)/g(x)} = C$ 
    \item $f(x) = \calO(\theta;g(x))$ is a fixed function with regard to $x$ such that $\exists C(\theta) > 0$, and $\lim_{x\to\infty}\abs{f(x)/g(x)} = C(\theta)$ 
    
    \item
    For a set of random variables $X_n$ and a corresponding set of constants $a_n$, the notation
    \[X_n = o_p(a_n).\]
    means that the set of values $X_n/a_n$ converges to zero in probability as $n$ approaches an appropriate limit. Equivalently, $X_n = o_p(a_n)$ can be written as $X_n/a_n = o_p(1)$, where $X_n = o_p(1)$ is defined as 
    \[X_n \convP 0.\]
    
    \item
    For a set of random variables $X_n$ and $Y_n$, where $Y_n$ is almost surely non-zero, the notation
    \[X_n = o(Y_n) \as \]
    means that 
    \begin{equation*}
        X_n/Y_n \asConv 0
    .\end{equation*}
    
    \item
    The notation
    \[X_n = \calO_p(a_n).\]
    means that the set of values $X_n/a_n$ is stochastically bounded. That is, for any $\epsilon > 0$, there exists a finite $M > 0$ and a finite $N > 0$ such that,
    \[\P(\abs{X_n/a_n} > M) < \epsilon, \forall n > N.\]
    
    \item
    \label{itm: big O as defn}
    \[X_n = \calO(a_n) \as\]
    if for almost every $\omega \in \Omega$, there exists a number $C(\omega)$ such that $\abs{X_n(\omega)} \le C(\omega)a_n$. In other words, $X_n = \calO(a_n) \as$ if there exists a random variable $C$ such that $\abs{X_n} \le Ca_n \as$ Equivalently, 
    \begin{equation*}
        X_n = \calO(a_n) \as \equivalent \limsup_{n \to \infty} \frac{\abs{X_n}}{a_n} < \infty \as
    \end{equation*}
    
    \item
    The notation
    \[X_n = \logO_p(a_n).\]
    means that the set of values $X_n/a_n$ is stochastically bounded up to a constant order of $log(a_n)$. That is, for any $\epsilon > 0$, there exists a finite $M > 0$ , a finite $k \in \mathbb{Z}$, and a finite $N > 0$ such that,
    \[\P(\abs{X_n/\log^k(a_n)a_n} > M) < \epsilon, \forall n > N.\]
\end{enumerate}

\textbf{All these definitions can be generalized to vectors or matrices with entry-wise definition.} Without extra specification, all norms $\norm{\cdot}$ (for both vectors and matrices) are meant to be $L_2$ norm $\norm{\cdot}_2$, i.e., operator-2 norm for the matrix.

\end{defn}
Some relationships between these notations are worth keeping in mind: (see Eq.(7) and Eq.(8) in \citet{janson2011probability}) 
\begin{equation}
\label{eq:small o as to p}
X_n = o(a_n) \as \Longrightarrow X_n = o_p(a_n)
.\end{equation}

\begin{equation}
\label{eq:big o as to p}
X_n = O(a_n) \as \Longrightarrow X_n = O_p(a_n)
.\end{equation}

To carefully track down the constant chosen manually, when we state order bounds like $\calO(\theta;g(x))$, $\theta$ should not contain variables such as $\delta$ which are set fixed when we prove high probability bounds but could be varying later, but could contain global constants such as $A$, $B$, $K$, $P$, $Q$, $R$, dimension $\inputdim$, $\statedim$ and $C_x$, $C_u$, $\tau$, $\beta$ that are fixed throughout the whole algorithm.

In order to differentiate $\calO(\cdot)$ from fixed constants, we denote $\calO(\theta)$ as constant terms which could be potentially varying and only related with $\theta$. That means for the same $\calO(\theta)$ symbol in two different places, they can be different constants. One special symbol is $\calO(1)$ which represents constant that does not rely on any parameters.

\subsection{Extending results to $\beta=1$}\label{sec:app_ext_beta1}
Although the main text only considered vanishing exploration noise (i.e., $\beta<1$), for completeness (and because it is straightforward to do so) we will also consider the case of $\beta=1$ and $\alpha\le 0$ for all of our results.

\subsection{Proof dependency tree}
\begin{figure}[p]
    \makebox[\linewidth]{
        \includegraphics[width=1\linewidth]{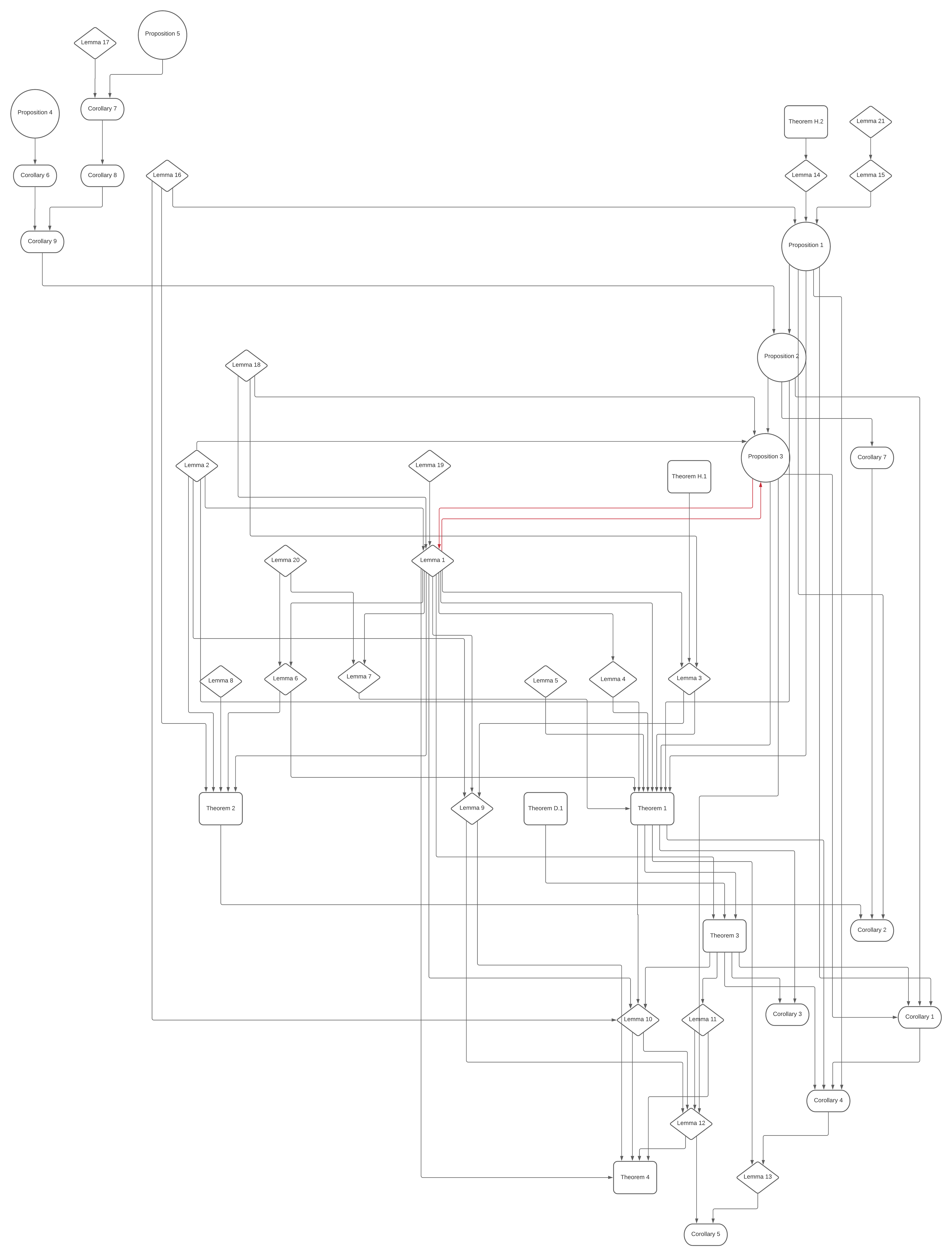}
    }
    \caption{Proof dependency tree}
    \label{fig: Proof dependency tree}
\end{figure}

In order to make the proof more readable and easier to understand, we put the proof outlines first and summarize most useful middle steps by lemmas. These lemmas' proofs often involve more technical details and is deferred to later parts in the appendix. While this may help readers have better understanding in the high level ideas behind the long proof, we realize that it may also cause loops in the proof structure. Thus, we provide a tree (\cref{fig: Proof dependency tree}) which describes the exact proof dependency structure to make sure that there is no circular argument. In \cref{fig: Proof dependency tree}, all conclusions lies in a perfect tree graph except for the loop marked in red between \cref{lemma: Hi prob bounds in theorem 2} and \cref{prop:one_epoch_estimate_withMyalg}. This is not a contradiction because the proof of \cref{prop:one_epoch_estimate_withMyalg} only relies on a subset of conclusions in \cref{lemma: Hi prob bounds in theorem 2}: \cref{eq:bound on eta_p,eq: stochastic bound on B eta p plus  varepsilon p}, which do not require \cref{prop:one_epoch_estimate_withMyalg} to hold. Some of the proofs relies on \cref{eq: sum eta_t}, which is not included in the graph but still 
self-consistent (does not rely on other results in the paper).

\clearpage

\clearpage

\section{The proof of \cref{thm:main tool}}
\label{The proof of thm:main tool}
\begin{thm*}
\cref{alg:myAlg} applied to a system described by \cref{eq:LinearModel} under \cref{asm:InitialStableCondition} satisfies
\begin{equation}
\label{eq: more general CLT condition 1}
D_t^{-1}
\sum_{i=0}^{t-1}
\begin{bmatrix}
        x_i\\
        u_i\\
\end{bmatrix}
\begin{bmatrix}
        x_i\\
        u_i\\
\end{bmatrix}^\top  
(D_t^\top)^{-1} \convP I_{n+d}
.\end{equation}
\end{thm*}
\subsection{Proof Outline}
\begin{proof}
Let us first examine the Gram matrix $\sum_{i=0}^{t-1} \begin{bmatrix}         x_i\\         u_i\\ \end{bmatrix} \begin{bmatrix}         x_i\\         u_i\\ \end{bmatrix}^\top $. Denote
\begin{equation}
    \label{eq: defn Mt}
\begin{split}
    M_t :=& \sum_{i=1}^{t-1}x_ix_i^\top/t^\beta\log^{\alpha}(t)  \\
\end{split}
,\end{equation}
and 
\begin{equation}
    \label{eq: defn Delta t}
\begin{split}
    \Delta_t :=& \sum_{i=1}^{t-1}u_ix_i^\top/t^\beta\log^{\alpha}(t)  - K M_t \\
    =& \sum_{i=1}^{t-1}((\Kh_t -K)x_i + \eta_i)x_i^\top/t^\beta\log^{\alpha}(t) 
.\end{split}
\end{equation}
We will show that 
\begin{equation*}
    \sum_{i=0}^{t-1}u_iu_i^\top/t^{\beta}\log^{\alpha}(t) = 
    KM_tK ^\top + \Delta_t K^\top  + K\Delta_t^\top  + \frac{\tau^2}\beta  I_d + o_p(1)
,\end{equation*}
and thus we can write our Gram matrix as
\begin{equation*}
\begin{split}
\sum_{i=1}^{t-1} \begin{bmatrix}         x_i\\         u_i\\ \end{bmatrix} \begin{bmatrix}         x_i\\         u_i\\ \end{bmatrix}^\top  /t^{\beta}\log^{\alpha}(t)  =& 
    \left[
                \begin{array}{cc}
                \sum_{i=0}^{t-1}x_ix_i^\top  & \sum_{i=1}^{t-1}x_iu_i^\top \\
                \sum_{i=0}^{t-1}u_ix_i^\top  & \sum_{i=1}^{t-1}u_iu_i^\top \\
                \end{array}
                \right]/t^{\beta}\log^{\alpha}(t)  \\
= &
    \left[
    \begin{array}{cc}
    M_t & M_tK ^\top + \Delta_t^\top  \\
    KM_t+ \Delta_t & KM_tK ^\top + \Delta_t K^\top  + K\Delta_t^\top  + \frac{\tau^2}\beta  I_d \\
    \end{array}
    \right]    + o_p(1) \\
=& 
\left[
    \begin{array}{cc}
    I_n & 0\\
    K & I_d\\
    \end{array}
\right]\left[
    \begin{array}{cc}
    M_t & \Delta_t^\top\\
    \Delta_t & \frac{\tau^2}\beta I_d\\
    \end{array}
\right]\left[
    \begin{array}{cc}
    I_n & K^\top \\
    0 & I_d\\
    \end{array}
\right] + o_p(1) .
\end{split}
\end{equation*}
Therefore, in order to satisfy 
\[D_t^{-1}\sum_{i=0}^{t-1} \begin{bmatrix}         x_i\\         u_i\\ \end{bmatrix} \begin{bmatrix}         x_i\\         u_i\\ \end{bmatrix}^\top  (D_t^\top)^{-1}  \convP I_{n+d}, \]
we can pick $D_t^{-1} := 
\left[
    \begin{array}{cc}
    C_t^{-1/2} & 0\\
    0 & \sqrt{\frac{\beta}{\tau^2}} I_d\\
    \end{array}
\right]
\left[
    \begin{array}{cc}
    I_n & 0\\
    -K & I_d\\
    \end{array}
\right]/t^{\beta/2}\log^{\alpha/2}(t) $. $C_t$ is a deterministic matrix which satisfies $C_t^{-1/2}M_t^{1/2} \convP I_\statedim$ and $C_t^{-1/2}\Delta_t = o_p(1)$ (we will give $C_t$'s exact expression in \cref{eq: Ct definition}). With this choice of $D_t^{-1}$, we have

\begin{equation*}
\begin{split}
    & D_t^{-1}\sum_{i=0}^{t-1} \begin{bmatrix}         x_i\\         u_i\\ \end{bmatrix} \begin{bmatrix}         x_i\\         u_i\\ \end{bmatrix}^\top  (D_t^\top)^{-1} \\  
    =& \left[
    \begin{array}{cc}
    C_t^{-1/2} & 0\\
    0 & \sqrt{\frac{\beta}{\tau^2}} I_d\\
    \end{array}
\right]
\left[
    \begin{array}{cc}
    I_n & 0\\
    -K & I_d\\
    \end{array}
\right]
\left(\sum_{i=0}^{t-1} \begin{bmatrix}         x_i\\         u_i\\ \end{bmatrix} \begin{bmatrix}         x_i\\         u_i\\ \end{bmatrix}^\top /t^{\beta}\log^\alpha(t)
\right)
 \left[
    \begin{array}{cc}
    I_n & -K^\top\\
    0 & I_d\\
    \end{array}
\right]
 \left[
    \begin{array}{cc}
    C_t^{-1/2} & 0\\
    0 & \sqrt{\frac{\beta}{\tau^2}} I_d\\
    \end{array}
\right] \\
=& 
\left[
    \begin{array}{cc}
    C_t^{-1/2} & 0\\
    0 & \sqrt{\frac{\beta}{\tau^2}} I_d\\
    \end{array}
\right]
\left[
    \begin{array}{cc}
    I_n & 0\\
    -K & I_d\\
    \end{array}
\right]
\left(
\left[
    \begin{array}{cc}
    I_n & 0\\
    K & I_d\\
    \end{array}
\right]\left[
    \begin{array}{cc}
    M_t & \Delta_t^\top\\
    \Delta_t & \frac{\tau^2}\beta I_d\\
    \end{array}
\right]\left[
    \begin{array}{cc}
    I_n & K^\top \\
    0 & I_d\\
    \end{array}
\right]
+ o_p(1)
\right)\\
& \quad \cdot
 \left[
    \begin{array}{cc}
    I_n & -K^\top\\
    0 & I_d\\
    \end{array}
\right]
 \left[
    \begin{array}{cc}
    C_t^{-1/2} & 0\\
    0 & \sqrt{\frac{\beta}{\tau^2}} I_d\\
    \end{array}
\right]  \\
=& 
\left[
    \begin{array}{cc}
    C_t^{-1/2} & 0\\
    0 & \sqrt{\frac{\beta}{\tau^2}} I_d\\
    \end{array}
\right]
\left[
    \begin{array}{cc}
    I_n & 0\\
    -K & I_d\\
    \end{array}
\right]
\left[
    \begin{array}{cc}
    I_n & 0\\
    K & I_d\\
    \end{array}
\right]\left[
    \begin{array}{cc}
    M_t & \Delta_t^\top\\
    \Delta_t & \frac{\tau^2}\beta I_d\\
    \end{array}
\right]\left[
    \begin{array}{cc}
    I_n & K^\top \\
    0 & I_d\\
    \end{array}
\right] \\
& \quad \cdot
 \left[
    \begin{array}{cc}
    I_n & -K^\top\\
    0 & I_d\\
    \end{array}
\right]
 \left[
    \begin{array}{cc}
    C_t^{-1/2} & 0\\
    0 & \sqrt{\frac{\beta}{\tau^2}} I_d\\
    \end{array}
\right] + o_p(1) \quad  \text{(we can move $o_p(1)$ outside because $C_t^{-1/2} \to 0$)}\\
=& 
\left[
    \begin{array}{cc}
    C_t^{-1/2} & 0\\
    0 & \sqrt{\frac{\beta}{\tau^2}} I_d\\
    \end{array}
\right]
\left[
    \begin{array}{cc}
    M_t & \Delta_t^\top\\
    \Delta_t & \frac{\tau^2}\beta I_d\\
    \end{array}
\right]
 \left[
    \begin{array}{cc}
    C_t^{-1/2} & 0\\
    0 & \sqrt{\frac{\beta}{\tau^2}} I_d\\
    \end{array}
\right] + o_p(1) \\
=& 
\left[
    \begin{array}{cc}
    C_t^{-1/2}M_tC_t^{-1/2} & \sqrt{\frac{\beta}{\tau^2}} C_t^{-1/2}\Delta_t^\top \\
    \sqrt{\frac{\beta}{\tau^2}} \Delta_t C_t^{-1/2} &  I_d\\
    \end{array}
\right]
  + o_p(1) \\
 =& I_{n+d} + o_p(1) 
.\end{split}
\end{equation*}

\paragraph{Components needing further explanation}
In the final step of the above derivation there are still several points that remains unclear, namely
\begin{itemize}
    \item $ \sum_{i=0}^{t-1}u_iu_i^\top/t^{\beta}\log^{\alpha}(t) = 
    KM_tK ^\top + \Delta_t K^\top  + K\Delta_t^\top  + \frac{\tau^2}\beta  I_d + o_p(1)$,
    
    \item $C_t^{-1/2}M_t^{1/2} \convP I_\statedim$,  and
    \item $C_t^{-1/2}\Delta_t = o_p(1)$.
\end{itemize}
As we will see, the order of $\Delta_t$ is decided by the convergence rate of $\Kh_t -K$. Because of that, the first step in our proof is to identify the convergence rate of $\Kh_t -K$. Then we will prove the three remaining points in The proof of \cref{eq: more general CLT condition 1}. 
To summarize,
 our proof can be mainly separated into two big steps:
 \begin{enumerate}
     \item Identify the convergence rate of $\Kh_t -K$. (see \cref{first part of proving main thm})
     \item Prove \cref{eq: more general CLT condition 1} holds:
    \[D_t^{-1}\sum_{i=0}^{t-1} \begin{bmatrix}         x_i\\         u_i\\ \end{bmatrix} \begin{bmatrix}         x_i\\         u_i\\ \end{bmatrix}^\top  (D_t^\top)^{-1}  \convP I_{n+d}.\]
     
     \begin{itemize}
         \item      Summarize uniform high probability bound for some random variables, which will serve as basic tools for later proof. (see \cref{tools main thm})
         \item Prove $C_t^{-1/2}M_t^{1/2} \convP I_\statedim$. (see \cref{2nd part of proving main thm})
         \item Prove $C_t^{-1/2}\Delta_t = o_p(1)$. (see \cref{3rd part of proving main thm})

         \item Prove $\sum_{i=0}^{t-1}u_iu_i^\top/t^\beta = 
            KM_tK ^\top + \Delta_t K^\top  + K\Delta_t^\top  + \frac{\tau^2}\beta  I_d + o_p(1)$. (see \cref{4th part of proving main thm})
     \end{itemize}

 \end{enumerate}
 Now we will examine these steps in order.
\end{proof}

\subsection{Convergence rate of $\Kh_t - K$}
\label{first part of proving main thm}
As said in the previous part, the main purpose of this section is to derive the convergence rate of $\Kh_t - K$, which is one crucial step in our proof. Denote the stabilizing controller computed by Line \ref{line:ols} \cref{alg:myAlg} as $\Kt_{t+1}$, i.e., 
\begin{align*}
    \Kt_{t+1} = 
    \begin{cases}
    \text{Solve DARE \cref{eq:ControllerK,eq:riccati} with } A = \Ah_{t}, B = \Bh_t,  &\text{for }(\Ah_{t}, \Bh_{t}) \text{ stabilizable}\\
    K_0, &\text{for }(\Ah_{t}, \Bh_{t}) \text{ not stabilizable}\\
    \end{cases}
.
\end{align*}
By Line \ref{line:check} \cref{alg:myAlg}, $\Kh_{t+1}$ can be written as:
\begin{align*}
    \Kh_{t+1} = 
    \begin{cases}
    K_0,  &\text{when }\norm{x_{t}} > C_x\log(t) \text{ or } \norm{\Kh_t} > C_K\\
    \Kt_{t+1}, &\text{otherwise}\\
    \end{cases}
.
\end{align*}


In particular, the proof can be separated into three parts: 
\begin{enumerate}
    \item Derive the convergence rate of $\Ah_t$ and $\Bh_t$.
    \item Show that $\Kt_{t+1}$ enjoy the same convergence rate as $\Ah_t$ and $\Bh_t$.
    \item Show that $\Kh_{t+1}$ is only different from $\Kt_{t+1}$ finitely often, and as a result, $\Kh_{t+1}$ also enjoy the same convergence rate as $\Ah_t$ and $\Bh_t$.
\end{enumerate}
Correspondingly we have the following three propositions:
\begin{prop}[Similar to Proposition C.1 in \citet{dean2018regret}]
\label{prop:one_epoch_estimate}
Let $x_0 \in \R^{\statedim}$ be any initial state. Assume \cref{asm:InitialStableCondition} is satisfied. When applying $\cref{alg:myAlg}$,
  \begin{equation*}
      \max\left\{ \norm{\Ah_t - A}, \norm{\Bh_t - B}\right\} = 
    \calO(t^{-\frac{\beta}{2}} \log^{\frac{-\alpha + 1}{2}}(t)) \as
 \end{equation*}

\end{prop}
The proof of \cref{prop:one_epoch_estimate} can be found in \cref{section: The proof of one_epoch_estimate}. 

\begin{prop}
\label{prop:one_epoch_estimate_withK}
Let $x_0 \in \R^{\statedim}$ be any initial state. Assume \cref{asm:InitialStableCondition} is satisfied. When applying $\cref{alg:myAlg}$, 
\begin{equation*}
\max\left\{ \norm{\Ah_t - A}, \norm{\Bh_t - B}, \norm{\Kt_{t+1} - K}\right\}
     = \calO(t^{-\frac{\beta}{2}} \log^{\frac{-\alpha + 1}{2}}(t)) \as
\end{equation*}
\end{prop}
The proof of \cref{prop:one_epoch_estimate_withK} can be found in \cref{section: The proof of one_epoch_estimate_withK}.

\begin{prop}
\label{prop:one_epoch_estimate_withMyalg}
Let $x_0 \in \R^{\statedim}$ be any initial state. Assume \cref{asm:InitialStableCondition} is satisfied. When applying \cref{alg:myAlg}, 
\begin{equation}
\label{eq:uniform high probability bound for Kh}
\max\left\{ \norm{\Ah_t - A}, \norm{\Bh_t - B}, \norm{\Kh_{t+1} - K}\right\}
     = \calO(t^{-\frac{\beta}{2}} \log^{\frac{-\alpha + 1}{2}}(t)) \as
\end{equation}
\end{prop}

The proof of \cref{prop:one_epoch_estimate_withMyalg} can be found in \cref{section: The proof of one_epoch_estimate_withMyalg}.

\cref{prop:one_epoch_estimate,prop:one_epoch_estimate_withK,prop:one_epoch_estimate_withMyalg} all hold additionally for a version of \cref{alg:myAlg} that only updates logarithmically often; see \cref{sec: The proof of Propositions}. The takeaway from this section is the uniform bound for $\norm{\Kh_{t+1} - K}$ \cref{eq:uniform high probability bound for Kh}, which is the only property of $\Kh_t$ we need for the rest of the proof.

\subsection{Proving \cref{eq: more general CLT condition 1}}

\subsubsection{Uniform Bounds}
\label{tools main thm}
In this section we will show several basic uniform bounds that will be used frequently in the later The proof of \cref{thm:main tool}. 
\begin{lemma}
\label{lemma: Hi prob bounds in theorem 2}
~
\begin{itemize}
    \item
    \begin{equation}
    \label{eq:bound on eta_p}
    \norm{\varepsilon_t}, \norm{\eta_t} = \calO(\log^{1/2}(t))  \as
    \end{equation}
    
    \item 
    \begin{equation}
    \label{eq: stochastic bound on B eta p plus  varepsilon p}
    \norm{B\eta_t+\varepsilon_t}  = \calO(\log^{1/2}(t))  \as
    \end{equation}
    \end{itemize}
    Assume \cref{eq:uniform high probability bound for Kh}, then:
    \begin{itemize}
    \item 
    \begin{equation}
    \label{eq: stochastic bound delta_t}
        \norm{\delta_t} = \norm{\Kh_{t} - K} =
        \calO(t^{-\frac{\beta}{2}} \log^{\frac{-\alpha + 1}{2}}(t)
        ) \as
    \end{equation}
    
    \item 
    For $t > q$,
    \begin{equation}
    \label{eq: stochastic bound Lhat product}
        \norm{(L + B\delta_{t-1}) \cdots (L + B\delta_{q})} =
        \calO(\rhoL^{t-q}) \as
    \end{equation}
    
    \item
    \begin{equation}
    \label{eq: bound norm xi by log t} 
        \norm{x_t}, \norm{u_t} =
        \calO(\log^{1/2}(t)) \as
    \end{equation}
    
\end{itemize}
where $\delta_t := \Kh_{t} - K$, $L := A+BK$, and $\rhoL := \frac{2 + \rho(L)}{3}$. \textbf{Additionally, when $t=0, 1$ all these terms are bounded by $\calO(1) \as$}
\end{lemma}

The proof can be found in \cref{The proof of lemma: Hi prob bounds in theorem 2}. Following \cref{defn: Big O notation} \cref{itm: big O as defn}, \cref{lemma: Hi prob bounds in theorem 2} presents uniform upper bounds for $t \ge 0$. 
We will see that all states $x_t$ and actions $u_t$ can be expressed in recursive summations, which can be bounded easily if we have uniform upper bound for each of their components.


Let us briefly explain why these orders makes sense. 
\begin{itemize}
    \item The first two inequalities come from the tail bound for standard Gaussian random variables, whose maximum scales as $\log^{1/2}(t)$.
    
    \item The third inequality \cref{eq: stochastic bound delta_t} directly follows from \cref{eq:uniform high probability bound for Kh}. 
    
    \item The fourth inequality \cref{eq: stochastic bound Lhat product} holds with exponential decay because the $L$ has spectual radius $<1$ and by \cref{eq: stochastic bound delta_t} $\delta_t$ is shrinking to $0$.
    
    \item The fifth inequality \cref{eq: bound norm xi by log t} holds because the system is stabilizable and the effect of previous states and actions are exponentially decaying, leaving the main factor in the norm to come from the recent system noises. By the first two inequalities $\norm{x_t}$ is uniformly bounded by $\log^{1/2}(t)$ scale.
\end{itemize}

\subsubsection{Showing $C_t^{-1/2}M_t^{1/2} \convP I_n$}
\label{2nd part of proving main thm}

We wish to show that $M_t = \sum_{i=0}^{t-1}x_ix_i^\top/t^\beta\log^\alpha(t) = 
C_t (1 + o_p(1))$, where 
\begin{equation}
\label{eq: Ct definition}
    C_t = \log^{-\alpha}(t)
t^{1-\beta} \sum_{p=0}^{\infty}L ^{p}(L ^{p})^\top \sigma^2  + \frac{\tau^2}{\beta}\sum_{q=0 }^{\infty}L ^{q}BB^\top (L ^{q})^\top
\end{equation}
Recall the system definition \cref{eq:LinearModel}:
\[x_{t+1} = A x_t + B u_t + \varepsilon_t.\]
and the input \cref{eq:Myinput}
\[u_t = \Kh_tx_t + \eta_t.\]
Recursively applying these two equations produces the following formula for $x_t$ in terms of $x_0$, $\{\varepsilon_p\}_{p=0}^{t-1}$, and $\{\eta_p\}_{p=0}^{t-1}$.
\begin{lemma}
\label{lemma: StateExpansion}
For any $t \ge 1$,
\begin{equation}
\label{eq: StateExpansion}
    x_{t} = \sum_{p=0}^{t-1}(A+B \Kh_{t-1})\cdots(A+B \Kh_{p+1})(B \eta_p+\varepsilon_p) +(A+B \Kh_{t-1})\cdots(A+B K_0)x_0
,\end{equation}
and 
\begin{equation*}
u_{t} = \sum_{p=0}^{t-1}\Kh_t(A+B \Kh_{t-1})\cdots(A+B \Kh_{p+1})(B \eta_p+\varepsilon_p) +\Kh_t(A+B \Kh_{t-1})\cdots(A+B K_0)x_0 + \eta_t   
.\end{equation*}
Here when $p = t-1$, we define the product $(A+B \Kh_{t-1})\cdots(A+B \Kh_{p+1}) := I_n$.
\end{lemma}
The proof can be found in \cref{The proof of lemma: StateExpansion}. As a result, we can rewrite $\sum_{i=0}^{t-1}x_ix_i^\top$ into a summation in terms of $\{\varepsilon_i, \eta_i\}_{i=0}^{t-1}$. First consider the terms without $x_0$.
\begin{align*}
\sum_{i=1}^{t-1}\sum_{p=0}^{i-1}\sum_{q=0}^{i-1}\left[(A+B \Kh_{i-1})\cdots(A+B \Kh_{p+1})(B \eta_p+\varepsilon_p)\right]
\left[(A+B \Kh_{i-1})\cdots(A+B \Kh_{q+1})(B \eta_q+\varepsilon_q)\right]^\top
.\end{align*}
This whole expression can be separated into four components with the following bounds:
\begin{lemma}
\label{lemma: four components xtxt}
Assume \cref{eq:uniform high probability bound for Kh}, then:

\begin{enumerate}

\item 
\begin{align*}
    &\sum_{i=1}^{t-1}\sum_{p=0}^{i-1}\sum_{q=0}^{i-1}\left[(A+B K)^{i-p-1}\right](B \eta_p+\varepsilon_p)(B \eta_q+\varepsilon_q)^\top 
\left[(A+B K)^{i-q-1}\right]^\top  \\
&\hspace{5cm}= t^\beta \log^\alpha(t) (C_t +o_p(1))
.\end{align*}

\item
\begin{equation*}
\begin{aligned}
 \sum_{i=1}^{t-1}\sum_{p=0}^{i-1}\sum_{q=0}^{i-1}\left[(A+B \Kh_{i-1})\cdots(A+B \Kh_{p+1}) - (A+B K)^{i-p-1}\right]&\\
 \cdot (B \eta_p+\varepsilon_p)(B \eta_q+\varepsilon_q)^\top 
\left[(A+B K)^{i-q-1}\right]^\top & =  \calO(t^{1-\beta/2}\log^{\frac{-\alpha + 3}{2}}(t)) \as
\end{aligned}
\end{equation*}

\item
\begin{align*}
 &\sum_{i=1}^{t-1}\sum_{p=0}^{i-1}\sum_{q=0}^{i-1}\left[ (A+B K)^{i-p-1}\right](B \eta_p+\varepsilon_p)(B \eta_q+\varepsilon_q)^\top 
\left[(A+B \Kh_{i-1})\cdots(A+B \Kh_{q+1}) - (A+B K)^{i-q-1}\right]^\top  \\
 & 
\hspace{8cm} = \calO(t^{1-\beta/2}\log^{\frac{-\alpha + 3}{2}}(t)) \as
\end{align*}

\item 
\begin{align*}
    &\sum_{i=1}^{t-1}\sum_{p=0}^{i-1}\sum_{q=0}^{i-1}\left[(A+B \Kh_{i-1})\cdots(A+B \Kh_{p+1}) - (A+B K)^{i-p-1}\right] 
(B \eta_p+\varepsilon_p)(B \eta_q+\varepsilon_q)^\top \\
 & \hspace{3cm} \cdot
\left[(A+B \Kh_{i-1})\cdots(A+B \Kh_{q+1}) - (A+B K)^{i-q-1}\right]^\top  = \calO(t^{1-\beta/2}\log^{\frac{-\alpha + 3}{2}}(t)) \as
\end{align*}
\end{enumerate}
\end{lemma}
The proof can be found in \cref{The proof of lemma: four components xtxt}. 

It remains to consider the remaining terms with $x_0$, which is relatively straight-forward, since the effect of the initial state is exponentially decaying when $t \to \infty$.

\begin{lemma}
\label{lemma: Term with starting point xtxt}
Assume \cref{eq:uniform high probability bound for Kh}, then
\begin{enumerate}
    \item $\sum_{i=0}^{t-1}\left[ (A+B \Kh_{i-1})\cdots(A+B K_{0})x_0\right]
    \left[\sum_{q=0}^{i-1}(A+B \Kh_{i-1})\cdots(A+B \Kh_{q+1})(B\eta_q+\varepsilon_q)\right]^T = \logO(1) \as$
    
    \item $\sum_{i=0}^{t-1}\left[ (A+B \Kh_{i-1})\cdots(A+B K_{0})x_0\right]
    \left[ (A+B \Kh_{i-1})\cdots(A+B K_{0})x_0\right]^T = \calO(1) \as$
\end{enumerate}
\end{lemma}
The proof can be found in \cref{The proof of lemma: Term with starting point xtxt}. As mentioned in \cref{eq:big o as to p}, $\calO \as$ notation is stronger than $\calO_p$ notation.
Summing up all the results in \cref{lemma: four components xtxt} and  \cref{lemma: Term with starting point xtxt} we can finally conclude that
\[ 
\sum_{i=0}^{t-1} x_ix_i^\top = 
t^\beta \log^\alpha(t) (C_t +o_p(1)) + \calO_p(t^{1-\beta/2}\log^{\frac{-\alpha + 3}{2}}(t))
.\]
Thus
\begin{equation}
\label{eq:Cov xx}
M_t = \sum_{i=0}^{t-1}x_ix_i^\top/t^\beta\log^\alpha(t) = C_t +o_p(1) + \calO_p(t^{1-3\beta/2}\log^{\frac{-3\alpha + 3}{2}}(t))
,\end{equation}
where $C_t$ is defined in \cref{eq: Ct definition}
This is already very close to our objective $C_t^{-1/2}M_t^{1/2} \convP I_n$, but we still need to show that $C_t$ is an invertible matrix. $C_t$ is already a positive semi-definite (PSD) matrix  because it is a weighted summation of PSD matrices $L^p(L ^{p})^\top$ and $L ^{q}BB^\top (L ^{q})^\top$. The only thing we need to ensure is that $C_t$ is a full rank matrix. And that is indeed true because the $p=0$ term is the identity matrix, and adding more PSD matrices $L^p(L ^{p})^\top$ and $L ^{q}BB^\top (L ^{q})^\top$ will not change its positive definite nature. Following \cref{eq: Ct definition}, 
we have (because $\beta < 1$ or $\beta = 1 $ and $\alpha \le 0$)
\begin{equation}
\label{eq: Ct order}
    C_t = 
        \log^{-\alpha}(t)t^{1-\beta} 
        \sum_{p=0}^{\infty}L ^{p}\left(\sigma^2I_n + 1_{\{\beta=1,\alpha=0\}}\tau^2 BB^\top\right)(L ^{p})^\top (I_n + o(1))
.\end{equation}
Thus 
\begin{equation}
\label{eq: Ct inverse order}
    C_t^{-1} 
    = t^{\beta-1}\log^\alpha(t) 
    \left(        \sum_{p=0}^{\infty}L ^{p}\left(\sigma^2I_n + 1_{\{\beta=1,\alpha=0\}}\tau^2 BB^\top\right)(L ^{p})^\top
    \right)^{-1}
    (I_n + o(1))
    = \calO(t^{\beta-1}\log^\alpha(t))
.\end{equation}
Noticing that 
\[\calO(t^{\beta-1}\log^\alpha(t)) \calO_p(t^{1-3\beta/2}\log^{\frac{-3\alpha + 3}{2}}(t)) = \calO_p(t^{-\beta/2}\log^{\frac{-\alpha + 3}{2}}(t)) = o_p(1),\]
we have from \cref{eq:Cov xx}
\begin{equation}
\label{eq: Ct -1 Mt convP I_n}
    C_t^{-1}M_t \convP I_n
.\end{equation}

With the help of the following lemma we conclude that $ C_t^{-1/2}M_t^{1/2} \convP I_n$.

\begin{lemma}
\label{lemma:AtBtConvergeIp}
Assume we have two matrix sequences $\{A_t\}_{t=1}^\infty$ and $\{B_t\}_{t=1}^\infty$, where $A_t$ and $B_t$ are $p \times p$ positive definite matrices, then 
\[A_t^{2}B_t^2 \convP I_p.\]
iff
\[A_tB_t \convP I_p.\]
\end{lemma}
The proof can be found in \cref{The proof of lemma:AtBtConvergeIp} (Thanks for the help from Haoyi Yang and Yue Li in proving this lemma). 

\subsubsection{Proving $C_t^{-1/2}\Delta_t = o_p(1)$}
\label{3rd part of proving main thm}
Recall the definition of $\Delta_t$ from \cref{eq: defn Delta t}:
\begin{equation*}
\Delta_t := \left(\sum_{i=0}^{t-1}(\Kh_i-K )x_ix_i^\top + \sum_{i=0}^{t-1}\eta_ix_i^\top\right)
/t^\beta \log^\alpha(t)
.\end{equation*}
The order of $\Delta_t$ depends on the order of its two components:

\begin{lemma}
\label{lemma: three parts u_tx_t}
Assume \cref{eq:uniform high probability bound for Kh}, then
\begin{enumerate}
    \item $
    \sum_{i=0}^{t-1}(\Kh_i-K )x_ix_i^\top =  
    \calO(t^{1-\beta/2}\log^{\frac{-\alpha + 3}{2}}(t)) \as
    $
    \item $ \sum_{i=0}^{t-1}\eta_ix_i^\top = 
    o\left(t^{\beta/2}\log^{\frac{\alpha+3}{2}}(t) \right) \as $
\end{enumerate}
\end{lemma}
The proof can be found in \cref{The proof of lemma: three parts u_tx_t}.
The first term has larger order than the second term when 
$1/2 \le \beta < 1$ or $\beta = 1$ and $\alpha \le 0$.
As a result, we have 
\begin{align}
\label{eq: Delta t order}
\begin{split}
    \Delta_t &= \calO(t^{1-3\beta/2}\log^{\frac{-3\alpha + 3}{2}}(t)) \as \quad \text{(when } \beta \in [1/2, 1)\text{)}
\end{split}
\end{align}
Observe from \cref{eq: Ct inverse order}:
\begin{equation*}
    C_t^{-1} = \calO(t^{\beta-1}\log^\alpha(t))
.\end{equation*}
Then when $\beta > 1/2$ or $\beta = 1/2, \alpha > 3/2$
\begin{align*}
    C_t^{-1/2}\Delta_t =& \calO(t^{-1/2+\beta/2}\log^{\alpha/2}(t)t^{1-3\beta/2}\log^{\frac{-3\alpha + 3}{2}}(t)) \\
    =& \calO(t^{1/2-\beta}\log^{\frac{-2\alpha + 3}{2}}(t)) \\
    =& o(1) \as
\end{align*}

\subsubsection{Proving $\sum_{i=0}^{t-1}u_iu_i^\top/
t^\beta \log^\alpha(t)
= 
            KM_tK ^\top + \Delta_t K^\top  + K\Delta_t^\top  + \frac{\tau^2}\beta  I_d + o_p(1)$}
\label{4th part of proving main thm}

Finally we need to check
\[\sum_{i=0}^{t-1}u_iu_i^\top  = \sum_{i=0}^{t-1}((K +\delta_i)x_i+\eta_i)((K +\delta_i)x_i+\eta_i)^\top ,\]
where $\delta_i = \Kh_i - K$.
There are six different kinds of terms in the above equation, namely $\sum_{i=0}^{t-1} Kx_ix_i^TK^\top $,
$\sum_{i=0}^{t-1} Kx_ix_i^\top \delta_i^\top $ and $\sum_{i=0}^{t-1} \delta_ix_ix_i^TK^\top $,
$\sum_{i=0}^{t-1} Kx_i\eta_i^\top $ and $\sum_{i=0}^{t-1} \eta_ix_i^TK^\top $,
$\sum_{i=0}^{t-1} \delta_ix_ix_i^\top \delta_i^\top $,
$\sum_{i=0}^{t-1} \delta_ix_i\eta_i^\top $ and $\sum_{i=0}^{t-1} \eta_ix_i^\top \delta_i^\top$,
and $\sum_{i=0}^{t-1} \eta_i\eta_i^\top $. The first three terms can be written as
\[\sum_{i=0}^{t-1} Kx_ix_i^TK^\top/t^\beta \log^\alpha(t)  = K M_t K^\top,\]
and
\[\left(\sum_{i=0}^{t-1} Kx_ix_i^\top \delta_i^\top + \sum_{i=0}^{t-1} \delta_ix_ix_i^TK^\top + 
\sum_{i=0}^{t-1} Kx_i\eta_i^\top + \sum_{i=0}^{t-1} \eta_ix_i^TK^\top \right)/t^\beta \log^\alpha(t)  = K \Delta_t^T + \Delta_t K^T.\]
The remaining terms can be summarized by
\begin{lemma}
\label{lemma: six parts u_tu_t}
Assume \cref{eq:uniform high probability bound for Kh}, then
\begin{enumerate}
    \item $\sum_{i=0}^{t-1} \delta_ix_ix_i^\top \delta_i^\top = \calO(t^{1-\beta}\log^{-\alpha+2}(t)) \as$
    \item $\sum_{i=0}^{t-1}\delta_ix_i\eta_i^\top = (\sum_{i=0}^{t-1} \eta_ix_i^\top \delta_i^\top)^\top =  
    o\left(\log^{2}(t)\right) \as$
    \item $\sum_{i=0}^{t-1}\eta_i\eta_i^\top  = t^\beta\frac{\tau^2}{\beta}\log^\alpha(t)(I_d + o_p(1)) $
\end{enumerate}
\end{lemma}
The proof of \cref{lemma: six parts u_tu_t} can be found in \cref{The proof of lemma: six parts u_tu_t}.
Combining all parts in \cref{lemma: six parts u_tu_t} we have when $\beta > 1/2$ or $\beta = 1/2, \alpha > 1$, the third item dominates the other two. To sum up, we have
\begin{equation}
\label{eq: uu decomp}
    \sum_{i=0}^{t-1}u_iu_i^\top/t^\beta\log^\alpha(t) = 
            KM_tK ^\top + \Delta_t K^\top  + K\Delta_t^\top  + \frac{\tau^2}\beta  I_d + o_p(1)
.\end{equation}

\paragraph{Summary}
Now we have completed all missing proof pieces in the proof of \cref{eq: more general CLT condition 1}, which finishes The proof of \cref{thm:main tool}.

\section{The proof of \cref{thm:regret}}
\label{The proof of thm:regret}

\begin{thm*}
The average regret of the controller $U$ defined by \cref{alg:myAlg} applied through time horizon $T$ to a system described by \cref{eq:LinearModel} under \cref{asm:InitialStableCondition} satisfies, as $T \to \infty$,
\begin{equation*}
\frac{\mathcal{R}(U,T)}{\tau^2\beta^{-1} \Tr(B^\top PB +R)T^{\beta-1}\log^\alpha(T)} \convP 1,
\end{equation*}
with $\beta = 1/2$ therefore achieving the optimal rate \citep{simchowitz2020naive} of
$\mathcal{R}(U,T) = \logO_p(T^{-1/2})$.
\end{thm*}

\subsection{Proof Outline}
\begin{proof}
We are interested in the cost 

\[\sum_{t=1}^{T} x_t^\top Qx_t + u_t^\top Ru_t \quad \text{with $u_t = \Kh_t x_t + \eta_t$}.\]
Recall the \cref{eq: StateExpansion} from \cref{lemma: StateExpansion} that
\[    x_{t} = \sum_{p=0}^{t-1}(A+B \Kh_{t-1})\cdots(A+B \Kh_{p+1})(B \eta_p+\varepsilon_p) +(A+B \Kh_{t-1})\cdots(A+B K_0)x_0.\]


Notice that the state $x_t$ has the same expression as if the system had noise $\tilde{\varepsilon}_t = B\eta_t + \varepsilon_t$ and controller $\tilde{u}_t = \Kh_t x_t$. We wish to switch to the new system because there are some existing tools with controls in the form of $\tilde{u}_t = \Kh_t x_t$.

We will first show in \cref{subsection: Cost difference induced by transformation} that the difference between the original cost and transformed cost is 
\begin{equation*}
    \sum_{t=1}^{T} u_t^\top  Ru_t -  \tilde{u}_t^\top R\tilde{u}_t =  \frac{\tau^2}{\beta}T^{\beta}\log^\alpha(T)\Tr(R)(1+o_p(1))
,\end{equation*}
and then prove in \cref{subsection: Cost of transferred system} the new system cost is
\begin{equation*}
    \sum_{t=1}^{T} x_t^\top Qx_t + \tilde{u}_t^\top R\tilde{u}_t = T \sigma^2\Tr (P) 
    +  \frac{\tau^2}{\beta} T^\beta \log^\alpha(T) \Tr(B^\top P B) (1+o_p(1))
.\end{equation*}
Combining the above two equations, we conclude that
\begin{align*}
    \mathcal{J}(U,T)
    &= \frac{1}{T}
    \left[\sum_{t=1}^{T} x_t^\top Qx_t + u_t^\top Ru_t\right] \\
    &= \sigma^2\Tr (P) + \tau^2\beta^{-1} \Tr(B^\top PB +R)T^{\beta-1}\log^\alpha(T) (1+ o_p(1))
.\end{align*}
Based on similar analysis we prove in \cref{subsection: Optimal average cost} that
\begin{align*}
    \mathcal{J}(U^*,T)
    &= \sigma^2\Tr (P) + \calO_p(T^{-1/2}\log(T))
.\end{align*}
Recall that we choose 
$\beta \in [1/2,1]$,
and $\alpha>3/2$ when $\beta=1/2$, which means $T^{\beta-1}\log^\alpha(T)$ is of larger order than $T^{-1/2}\log(T)$. Finally we finish the proof with
\begin{align*}
    \mathcal{R}(U,T) &= \mathcal{J}(U,T)- \mathcal{J}(U^*,T) \\
    &= \tau^2\beta^{-1} \Tr(B^\top PB +R)T^{\beta-1}\log^\alpha(T) (1+ o_p(1))
.\end{align*}
\end{proof}


\subsection{Cost difference induced by transformation}
\label{subsection: Cost difference induced by transformation}
The difference is expressed as
\begin{equation*}
\begin{split}
        \sum_{t=1}^{T} u_t^\top  Ru_t -  \tilde{u}_t^\top R\tilde{u}_t     
    =&  \sum_{t=1}^{T}(\Kh_t x_t + \eta_t)^\top  R (\Kh_t x_t + \eta_t)   - \sum_{t=1}^{T} (\Kh_t x_t)^\top  R (\Kh_t x_t) \\
    =& 2\sum_{t=1}^{T} (\Kh_t x_t)^\top  R \eta_t +  \sum_{t=1}^{T} \eta_t^\top R \eta_t 
.\end{split}
\end{equation*}
We show in \cref{eq: (Ktxt)T R etaT} that
\begin{equation*}
    \sum_{t=1}^{T} (\Kh_t x_t)^\top  R \eta_t = o\left(T^{\beta/2}\log^{\frac{\alpha+3}{2}}(T) \right) \as
,\end{equation*}
which is a direct corollary of \cref{lemma: three parts u_tx_t}.

    Next we consider the order of $\sum_{t=1}^{T} \eta_t^\top R \eta_t$.
    Since $\eta_t \sim \calN(0, \tau^2 t^{-1+\beta}\log^\alpha(t)I_d)$,
    \begin{equation*}
    \begin{split}
         \E \sum_{t=1}^{T} \eta_t^\top R \eta_t 
        &= \sum_{t=1}^{T} \Tr(\E \eta_t\eta_t^\top R )  \\
        &= \sum_{t=1}^{T} \tau^2 t^{-1+\beta}\log^\alpha(t) \Tr(R)  \\
        & \quad \text{(see the proof in \cref{eq: sum eta_t})} \\
        &= \tau^2\frac{T^{\beta}}{\beta}\log^\alpha(T)\Tr(R)(1+o(1)) 
    .\end{split}
    \end{equation*}
    While the variance of $\sum_{t=1}^{T} \eta_t^\top R \eta_t$ is $\calO(\sum_{t=1}^{T} t^{-2+2\beta}\log^{2\alpha}(t)) = \calO(T^{-1+2\beta}\log^{2\alpha}(T))$, which means the standard error $\calO(T^{-1/2+\beta}\log^{\alpha}(T))$ is of lower order than the expectation. Thus
    
    \begin{equation*}
        \sum_{t=1}^{T} \eta_t^\top R \eta_t 
        = \tau^2\frac{T^{\beta}}{\beta}\log^\alpha(T)\Tr(R)(1+o_p(1))
    .\end{equation*}

As a conclusion, the error caused by this transformation is of order $\logO_p(T^\beta)$, and the dominating term is $\sum_{t=1}^{T} \eta_t^\top R \eta_t$. 
\begin{equation}
    \label{eq: regret first part}
    \sum_{t=1}^{T} u_t^\top  Ru_t -  \tilde{u}_t^\top R\tilde{u}_t = 
    \tau^2\frac{T^{\beta}}{\beta}\log^\alpha(T)\Tr(R)(1+o_p(1))
.\end{equation}

\subsection{Cost of transformed system}
\label{subsection: Cost of transferred system}

Next we proceed as if our system was  $x_t$ with system noise $\tilde\varepsilon_t = B \eta_t+\varepsilon_t$ and controller $\tilde{u}_t = \Kh_t x_t$. 
The key idea of the following proof is from Appendix C of \citet{fazel2018global}.


We are interested in the cost

\[ \sum_{t=1}^{T} x_t^\top Qx_t + \tilde{u}_t^\top R\tilde{u}_t  \quad \text{with $\tilde{u}_t = \Kh_t x_t $},\]

which can be written as 
\begin{equation}
\label{eq: regret decomposition}
\begin{split}
    \sum_{t=1}^{T} x_t^\top Qx_t + \tilde{u}_t^\top R\tilde{u}_t 
    =& \sum_{t=1}^{T} x_t^\top Qx_t + (\Kh_t x_t)^\top R\Kh_t x_t \\
    =& \sum_{t=1}^{T} x_t^\top (Q + \Kh_t^\top R \Kh_t)x_t  \\
    =& \sum_{t=1}^{T} \left[x_t^\top (Q + \Kh_t^\top R \Kh_t)x_t + x_{t+1}^\top P x_{t+1} - x_{t}^\top P x_{t}
    \right] 
    + x_1^\top Px_1 - x_{T+1}^\top P x_{T+1} \\
    =& \sum_{t=1}^{T} \left[x_t^\top (Q + \Kh_t^\top R \Kh_t)x_t + ((A+B\Kh_t)x_t +\tilde\varepsilon_{t})^\top P ((A+B\Kh_t)x_t +\tilde\varepsilon_{t}) - x_{t}^\top P x_{t}\right] \\
    & \quad + \logO_p(1) \quad \text{(by \cref{lemma: Hi prob bounds in theorem 2}) } \\
    =& \sum_{t=1}^{T} \Big[ x_t^\top (Q + \Kh_t^\top R \Kh_t)x_t + x_t^\top (A+B\Kh_t)^\top P (A+B\Kh_t)x_t  - x_{t}^\top P x_{t} \\
    & \quad+ 2  \tilde\varepsilon_{t}^\top P (A+B\Kh_t)x_t + \tilde\varepsilon_{t}^\top P \tilde\varepsilon_{t}
    \Big]
    +\logO_p(1)   
.\end{split}
\end{equation}
We constructed the specific form of the first term on purpose. The following lemma translates the first term into a quadratic term with respect to $\Kh_t - K$.
\begin{lemma}
\label{lem: useful lemma from fazel}
For any $\Kh$ with suitable dimension,
\begin{equation*}
\begin{split}
    &x^\top (Q + \Kh^\top R \Kh)x + x^\top (A+B\Kh)^\top P (A+B\Kh)x - x^\top P x \\
    & \qquad = x^\top (\Kh-K)^\top( R + B^\top P B) (\Kh-K)x
.\end{split}
\end{equation*}
\end{lemma}
The proof can be found in \cref{The proof of lem: useful lemma from fazel}.
As a result
\begin{equation*}
\begin{split}
    \sum_{t=1}^{T} x_t^\top Qx_t + \tilde{u}_t^\top R\tilde{u}_t 
    =& \sum_{t=1}^{T} x_t^\top (\Kh_t-K)^\top( R + B^\top P B) (\Kh_t-K)x_t \\
    &\quad + 2  \tilde\varepsilon_{t}^\top P (A+B\Kh_t)x_t + \tilde\varepsilon_{t}^\top P \tilde\varepsilon_{t}
    +\logO_p(1)   
.\end{split}
\end{equation*}

Now we have three terms, and we will examine them in order.

\begin{enumerate}
    \item The first term we consider is $\sum_{t=1}^{T} x_t^\top (\Kh_t-K)^\top( R + B^\top P B) (\Kh_t-K)x_t$. Recall from \cref{lemma: Hi prob bounds in theorem 2} that
    \begin{equation*}
        \norm{x_t}, \norm{u_t} = \calO(\log^{1/2}(t)) \as
    \end{equation*}
    and 
    \begin{equation*}
        \norm{\Kh_{t} - K} = \calO(t^{-\frac{\beta}{2}} \log^{\frac{-\alpha + 1}{2}}(t)) \as
    \end{equation*}
    As a result
    \begin{equation*}
    \begin{split}
         &\sum_{t=1}^{T} x_t^\top (\Kh_t-K)^\top( R + B^\top P B) (\Kh_t-K)x_t \\
        \le & \sum_{t=1}^{T} \norm{x_t}^2 \norm{\Kh_t-K}^2 \norm{ R + B^\top P B} \\
        = & \sum_{t=1}^{T} \calO(\log(t)) \calO(t^{-\beta} \log^{-\alpha + 1}(t))
        \as \\
        = & \calO(T^{1-\beta}\log^{-\alpha + 2}(T)) \as
        \qquad (\text{by  \cref{eq: sum eta_t}})
    \end{split}
    \end{equation*}
    
    \item The second term we consider is $\sum_{t=1}^{T} \tilde\varepsilon_{t}^\top P (A+B\Kh_t)x_t$. Similar as before, we notice that $\tilde\varepsilon_{t} = \varepsilon_{t} + B\eta_t \independent (A+B\Kh_t)x_t$. Then 
    \begin{equation*}
        \E \sum_{t=1}^{T} \tilde\varepsilon_{t}^\top P (A+B\Kh_t)x_t = 0
    .\end{equation*}
    Next consider 
    \begin{equation*}
    \begin{split}
        &\E (\sum_{t=1}^{T} \tilde\varepsilon_{t}^\top P (A+B\Kh_t)x_t)^2  \\
        =&\sum_{t=1}^{T} \E (\tilde\varepsilon_{t}^\top P (A+B\Kh_t)x_t)^2 \\
        \le & \sum_{t=1}^{T} \E \norm{\tilde\varepsilon_{t}}^2 \norm{P}^2 \norm{(A+B\Kh_t)}^2 \norm{x_t}^2 \\
        & \text{ ($\norm{\Kh_t} \le C_K$ based on \cref{alg:myAlg} design)}  \\
        \le & \sum_{t=1}^{T}  \norm{P}^2 (\norm{A} + \norm{B} C_K)^2 \E \norm{\tilde\varepsilon_{t}}^2 \E \norm{x_t}^2 \\
        = & \calO(1) \E \sum_{t=1}^{T}  \norm{x_t}^2 \\
        & \text{ (because of \cref{lem:bound_covariance} } \E \sum_{t=1}^{T} \norm{x_t}^2 = \calO(T\log^2(T) )) \\
        = & \calO(T\log^2(T))
    .\end{split}
    \end{equation*}
    Thus
    \begin{equation}
    \label{eq: regret 2nd item}
        \sum_{t=1}^{T} \tilde\varepsilon_{t}^\top P (A+B\Kh_t)x_t = \calO_p(T^{1/2}\log(T))
    .\end{equation}
    
    \item The third term we consider is $\sum_{t=1}^{T} \tilde\varepsilon_{t}^\top P \tilde\varepsilon_{t}$. The expectation is
    \begin{equation*}
    \begin{split}
        &\E \sum_{t=1}^{T} \tilde\varepsilon_{t}^\top P \tilde\varepsilon_{t} \\
        =& \sum_{t=1}^{T} \Tr (P \E \tilde\varepsilon_{t} \tilde\varepsilon_{t}^\top ) \\
        =& \sum_{t=1}^{T} \Tr (P (\sigma^2 I_n + \tau^2 t^{\beta-1}\log^\alpha(t) B B^\top  )) \\
        =& T \sigma^2\Tr (P) + \frac{\tau^2}{\beta} T^\beta \log^\alpha(T) \Tr(B^\top P B) (1+o(1))
        \quad
        \text{(By \cref{eq: sum eta_t})}
    .\end{split}
    \end{equation*}
    On the other hand, the variance is the sum of variances for each single summand with total order $\calO(T)$. As a result, when $\beta > 1/2$ or $\beta = 1/2, \alpha > 0$
    \begin{equation}
    \label{eq: regret 3rd item}
        \sum_{t=1}^{T} \tilde\varepsilon_{t}^\top P \tilde\varepsilon_{t}
        = T \sigma^2\Tr (P) + \frac{\tau^2}{\beta} T^\beta \log^\alpha(T) \Tr(B^\top P B) (1+o_p(1))
    .\end{equation}
    
\end{enumerate}
Summing up all three parts we have: when $\beta > 1/2$, or $\beta = 1/2, \alpha > 1$,
\begin{equation}
    \sum_{t=1}^{T} x_t^\top Qx_t + \tilde{u}_t^\top R\tilde{u}_t = T \sigma^2\Tr (P) 
    +  \frac{\tau^2}{\beta} T^\beta \log^\alpha(T) \Tr(B^\top P B) (1+o_p(1))
.\end{equation}
Taking the transformation part into consideration (\cref{eq: regret first part}):
\begin{equation*}
    \sum_{t=1}^{T} u_t^\top  Ru_t -  \tilde{u}_t^\top R\tilde{u}_t = 
    \tau^2\frac{T^{\beta}}{\beta}\log^\alpha(T)\Tr(R)(1+o_p(1))
.\end{equation*}
Finally we have when $\beta > 1/2$, or $\beta = 1/2, \alpha > 1$

\begin{align*}
    \mathcal{J}(U,T)
    &= \frac{1}{T}
    \left[\sum_{t=1}^{T} x_t^\top Qx_t + u_t^\top Ru_t\right] \\
    &= \sigma^2\Tr (P) + \tau^2\beta^{-1} \Tr(B^\top PB +R)T^{\beta-1}\log^\alpha(T) (1+ o_p(1))
.\end{align*}
Finally we only need to prove that the optimal average cost can be expressed as:
\begin{align*}
    \mathcal{J}(U^*,T)
    &= \sigma^2\Tr (P) + \calO_p(T^{-1/2}\log(T))
.\end{align*}

\subsection{Optimal average cost}
\label{subsection: Optimal average cost}
Denote the states and actions following policy $U^*(H_t) = Kx_t$ as $x'_t$ and $u'_t$. Following \cref{eq: regret decomposition} we know that
\begin{equation*}
\begin{split}
    &\sum_{t=1}^{T} (x'_t)^\top Qx'_t + (u'_t)^\top Ru'_t \\
    =& \sum_{t=1}^{T} \Big[ (x'_t)^\top (Q + K^\top R K)x'_t + (x'_t)^\top (A+BK)^\top P (A+BK)x'_t  - (x'_t)^\top P x'_{t} \\
    &+ 2  \varepsilon_{t}^\top P (A+B\Kh_t)x_t + \varepsilon_{t}^\top P \varepsilon_{t}
    \Big]
    +\logO_p(1)   
.\end{split}
\end{equation*}
Following \cref{lem: useful lemma from fazel}, since our $\Kh$ is exactly $K$:
\begin{equation*}
    (x'_t)^\top (Q + K^\top R K)x'_t + (x'_t)^\top (A+BK)^\top P (A+BK)x'_t  - (x'_t)^\top P x'_{t} = 0
\end{equation*}
The remaining terms can be considered in exactly same way as \cref{eq: regret 2nd item} and \cref{eq: regret 3rd item}, which turn out to be:
\begin{equation*}
    \sum_{t=1}^{T} \tilde\varepsilon_{t}^\top P (A+B\Kh_t)x_t = \calO_p(T^{1/2}\log(T))
,\end{equation*}
and
\begin{equation*}
    \sum_{t=1}^{T} \varepsilon_{t}^\top P \varepsilon_{t}
        = T \sigma^2\Tr (P) + \calO_p(T^{1/2})
.\end{equation*}
Finally we arrive at the conclusion that
\begin{align*}
    \mathcal{J}(U^*,T) 
    &= \frac1T \left(\sum_{t=1}^{T} (x'_t)^\top Qx'_t + (u'_t)^\top Ru'_t\right) \\
    &= \frac1T \left(\calO_p(T^{1/2}\log(T)) + T \sigma^2\Tr (P) + \calO_p(T^{1/2})\right) \\
    &= \sigma^2\Tr (P) + \calO_p(T^{-1/2}\log(T)) 
.\end{align*}

\section{The proof of \cref{thm:main CLT}}
\label{The proof of thm:main CLT}

\begin{thm*}
\cref{alg:myAlg} applied to a system described by \cref{eq:LinearModel} under \cref{asm:InitialStableCondition} satisfies, as $t \to \infty$,
\begin{equation*}
    \vvector \left[    
    \begin{bmatrix}
        \Ah_t - A,\Bh_t- B
    \end{bmatrix} D_t\right] \convD 
    \calN(0, \sigma^2 I_{\statedim(n+d)} )
.\end{equation*}
\end{thm*}
\begin{proof}
One can find the definition of $D_t$ in \cref{eq:D_t Definition}.
The proof heavily relies on the following theorems from \citet{anderson1992asymptotic}. For better understanding, we directly state those theorems with the same notation as our paper.

\begin{theorem}[Theorems 1 and 3 in \citet{anderson1992asymptotic}]
\label{thm: more general CLT}
Let $\left\{x_i, u_i, \varepsilon_i \right\}$, $i=0,1 \cdots$, be a sequence of random vectors described by \cref{eq:LinearModel} under \cref{asm:InitialStableCondition}, and let $\left\{\calF_i\right\}$ be an increasing sequence of $\sigma$-fields such that $\left\{x_i, u_i\right\}$ is $\calF_{i-1}$ measureable and $\varepsilon_i$ is $\calF_i$ measurable. Let the matrix $D_t$ be a deterministic matrix such that 
\begin{equation}
\label{eq: more general CLT condition 1 old}
    D_t^{-1}\sum_{i=0}^{t-1} \begin{bmatrix}         x_i\\         u_i\\ \end{bmatrix} \begin{bmatrix}         x_i\\         u_i\\ \end{bmatrix}^\top  (D_t^\top)^{-1}  \convP C
,\end{equation}
where $C$ is a constant matrix, and 
\begin{equation}
\label{eq: more general CLT condition 2}
    \max_{1 \le i \le t} 
    \begin{bmatrix}         x_i\\         u_i\\ \end{bmatrix}^\top 
    (D_tD_t^\top)^{-1}
    \begin{bmatrix}         x_i\\         u_i\\ \end{bmatrix} \convP 0
.\end{equation}
Suppose further that $\E(\varepsilon_i|\calF_{i-1}) = 0 $ a.s., $\E(\varepsilon_i\varepsilon_i^\top|\calF_{i-1}) = \Sigma_i$ a.s.,
\begin{equation}
\label{eq: trivial condition 1}
    \sum_{i=0}^{t-1} \left[\Sigma_i \otimes D_t^{-1}\begin{bmatrix}         x_i\\         u_i\\ \end{bmatrix} \begin{bmatrix}         x_i\\         u_i\\ \end{bmatrix}^\top  (D_t^\top)^{-1}\right] \convP \Sigma \otimes C,
\end{equation}
where $\Sigma$ is a constant positive semi-definite matrix and 
\begin{equation}
\label{eq: trivial condition 2}
\sup_{i \ge 1} \E \left[ \varepsilon_i^\top \varepsilon_i \bm{1}_{\varepsilon_i^\top \varepsilon_i > a} | \calF_{i-1} \right] \convP 0
,\end{equation}
as $a \to \infty$. Then
\begin{equation}
    \vvector \left[    
    \begin{bmatrix}
        \Ah_t - A,\Bh_t- B
    \end{bmatrix} D_t\right] \convD \calN(0, C^{-1} \otimes \Sigma)
.\end{equation}

\end{theorem}

As we have seen in \cref{alg:myAlg} the controller $\Kh_t$ is fully determined by $\{x_i, u_i\}_{i=0}^{t-1}$. Pick
\[\calF_{t-1} =\sigma(\{x_i, u_i, \eta_i\}_{i=0}^{t}, \{\varepsilon_i\}_{i=0}^{t-1}).\]
Now we verified the design vector $\begin{bmatrix}
x_t \\
u_t
\end{bmatrix}$ at stage $t$ is $\calF_{t-1}$ measurable. Since $\varepsilon_t \iid \calN(0, \sigma ^2 I_\inputdim)$, we know that $\varepsilon_t \independent \mathcal{F}_{t-1}$, and $\left\{\varepsilon_t \right\}$ is a martingale difference sequence with respect to an increasing sequence of $\sigma$-fields $\left\{\mathcal{F}_t \right\}$. 
\cref{eq: trivial condition 1} holds by the fact that all variances $\Sigma_i = \sigma^2 I_\inputdim$ and \cref{eq: more general CLT condition 1 old}. For \cref{eq: trivial condition 2}, notice that we can remove the $\sup$ since every term has the same value, so the conclusion follows from a standard property of Gaussian distributions.


Actually, \cref{eq: more general CLT condition 1 old} is already shown in \cref{thm:main tool}.
\cref{eq: more general CLT condition 2} requires less effort to prove as we defined $D_t$ by
\begin{equation}
\label{eq: D_t defn}
        D_t := 
        t^{\beta/2}\log^{\alpha/2}(t)
\left[
    \begin{array}{cc}
    I_n & 0\\
    K & I_d\\
    \end{array}
\right]
\left[
    \begin{array}{cc}
    C_t^{1/2} & 0\\
    0 & \sqrt{\frac{\tau^2}{\beta}} I_d\\
    \end{array}
\right]
.\end{equation}
As a result, \cref{eq: more general CLT condition 2} is not surprising since $z_t$ should be only of constant order. 

\subsection{The proof of \cref{eq: more general CLT condition 2}}
\label{5th part of proving main thm}


Since 
\begin{align*}
    D_tD_t^\top
    = t^{\beta}\log^{\alpha}(t) 
    \left[
    \begin{array}{cc}
    I_n & 0\\
    K & I_d\\
    \end{array}
\right]
\left[
    \begin{array}{cc}
    C_t & 0\\
    0 & \frac{\tau^2}{\beta} I_d\\
    \end{array}
\right]
\left[
    \begin{array}{cc}
    I_n & K^\top\\
    0 & I_d\\
    \end{array}
\right],
\end{align*}
we have
\begin{align}
\label{eq: DtDt -1 order}
\begin{split}
    (D_tD_t^\top)^{-1} 
    =& t^{-\beta}\log^{-\alpha}(t) 
    \left[
    \begin{array}{cc}
    I_n & -K^\top\\
    0 & I_d\\
    \end{array}
\right]
\left[
    \begin{array}{cc}
    C_t^{-1} & 0\\
    0 & \frac{\tau^2}{\beta} I_d\\
    \end{array}
\right]
\left[
    \begin{array}{cc}
    I_n & 0\\
    -K & I_d\\
    \end{array}
\right]
\\
=& \calO(t^{-\beta}\log^{-\alpha}(t)) \quad \text{(by \cref{eq: Ct inverse order})}.
\end{split}
\end{align}
Recall that \cref{eq: more general CLT condition 2} is
\begin{equation*}
    \max_{1\le i \le t} 
    \begin{bmatrix}         x_i\\         u_i\\ \end{bmatrix}^\top 
    (D_tD_t^\top)^{-1}
    \begin{bmatrix}         x_i\\         u_i\\ \end{bmatrix} \convP 0
.\end{equation*}
It suffices to show 
\[t^{-\beta/2}\log^{-\alpha/2}(t)\max_{1\le i \le t}\norm{x_i} \convP 0 \;\text{ and  }\;t^{-\beta/2}\log^{-\alpha/2}(t)\max_{1\le i \le t}\norm{u_i} \convP 0.\]
Actually we already shown in \cref{lemma: Hi prob bounds in theorem 2} that 
    \begin{equation*}
        \norm{x_t}, \norm{u_t} = \calO(\log^{1/2}(t)) \as \end{equation*}
This is a uniform bound over $t$, thus a direct corollary is
    \begin{equation*}
       \max_{1\le i \le t}\norm{x_i} , \max_{1\le i \le t}\norm{u_i}  = \calO(\log^{1/2}(t)) \as 
    \end{equation*}
That immediately implies
\[t^{-\beta/2}\max_{1\le i \le t}\norm{x_i} \asConv 0 \;\text{ and }\; t^{-\beta/2}\max_{1\le i \le t}\norm{u_i} \asConv 0.\]

\end{proof}


\section{The proof of \cref{thm:prediction CLT parametric}}
\label{The proof of thm:prediction CLT parametric}
Here we state and prove a generalization of \cref{thm:prediction CLT parametric} that allows for the case when $\beta=1$ and $\alpha\le 0$.

\begin{thm*}
\cref{alg:myAlg} applied to a system described by \cref{eq:LinearModel} under \cref{asm:InitialStableCondition} satisfies, as $t \to \infty$,
\begin{align}
        \left(
    x_t^\top
    \left( \sum_{p=0}^{\infty}(A+BK) ^{p}
    \left(I_n + 1_{\{\beta=1,\alpha=0\}}\frac{\tau^2}{\sigma^2}BB^\top\right)
    \left((A+BK) ^{p}\right)^\top\right)^{-1}x_t  
    +
     \beta \sigma^2
     \lnorm{w_t}^2
    \right)^{-1/2} \nonumber\\
    \hspace{-2mm}
     \cdot\, t^{1/2}  \left((\Ah_t - A)x_t + (\Bh_t- B)u_t\right) 
     \convD \calN(0,I_n).
\end{align}
\end{thm*}
\begin{proof}
We can generalize the input noise $\eta_t$ to $\xi_t$ which is any random vector independent of the data before $t$: $\{\varepsilon_i, \eta_i\}_{i=0}^{t-1}$. Hereafter, $u_t = \Kh_t x_t + \xi_t$ (but $u_i$ for $i < t$ is still $\Kh_i x_i + \eta_i$).

The proof will proceed by showing that $(\Ah_t, \Bh_t)$ acts as if it were independent of $(x_t, u_t)$, and then effectively conditioning on $(x_t, u_t)$ and using $(\Ah_t, \Bh_t)$'s asymptotic distribution from \cref{thm:main CLT}.

Define $\rhoL := \frac{2 + \rho(L)}{3}$ as in \cref{lemma: Hi prob bounds in theorem 2}.
Define replacements of $x_t$ and $u_t$ which are independent of $\Ah_{t-\myfloor{-\frac{\log(t)}{\log(\rho_L)}}}$ and $\Bh_{t-\myfloor{-\frac{\log(t)}{\log(\rho_L)}}}$:
\begin{equation}
    \label{eq: tilde x t definition}
    \tilde{x}_t := \sum_{p=t-\myfloor{-\frac{\log(t)}{\log(\rho_L)}}}^{t-1}(A+BK)^{t-p-1} ( B\eta_p + \varepsilon_p)
,\end{equation}
and
\begin{equation}
    \label{eq: tilde u t definition}
    \tilde{u}_t := K\tilde{x}_t + \xi_t= K\sum_{p=t-\myfloor{-\frac{\log(t)}{\log(\rho_L)}}}^{t-1}(A+BK)^{t-p-1} ( B\eta_p + \varepsilon_p) + \xi_t
.\end{equation}
We can show that the difference between $\tilde{x}_t, \tilde{u}_t$ and $x_t, u_t$ is very small:
\begin{lemma}
\label{lem: difference between real and substitutes x u}
\begin{equation*}
    x_t = \tilde{x}_t +  O(t^{-\frac{\beta}{2}} \log^{\frac{-\alpha + 2}{2}}(t)) \as
\end{equation*}
\begin{equation*}
    u_t = \tilde{u}_t +  O(t^{-\frac{\beta}{2}} \log^{\frac{-\alpha + 2}{2}}(t)) \as
\end{equation*}
\end{lemma}
The proof can be found in \cref{The proof of lem: difference between real and substitutes x u}. At the same time, the difference between $\Ah_{t-\myfloor{-\frac{\log(t)}{\log(\rho_L)}}}$,  $\Bh_{t-\myfloor{-\frac{\log(t)}{\log(\rho_L)}}}$ and $\Ah_t, \Bh_t$ is also small:
\begin{lemma}
\label{lem: difference between real and substitutes A B}
\begin{equation*}
    \Ah_{t} = \Ah_{t-\myfloor{-\frac{\log(t)}{\log(\rho_L)}}} + \calO_p(t^{-\beta}\log^{-\alpha+3/2}(t))
.\end{equation*}
\begin{equation*}
    \Bh_{t} = \Bh_{t-\myfloor{-\frac{\log(t)}{\log(\rho_L)}}} + \calO_p(t^{-\beta}\log^{-\alpha+3/2}(t))
.\end{equation*}
\end{lemma}
The proof can be found in \cref{The proof of lem: difference between real and substitutes A B}.
These substitutions are very close to our original concern, and they have the good independence property:
\begin{equation*}
    \left(\Ah_{t-\myfloor{-\frac{\log(t)}{\log(\rho_L)}}} - A, \Bh_{t-\myfloor{-\frac{\log(t)}{\log(\rho_L)}}}- B\right) \independent (\tilde{x}_t, \tilde{u}_t)
.\end{equation*}
This is because $\Ah_{t-\myfloor{-\frac{\log(t)}{\log(\rho_L)}}}$ and $\Bh_{t-\myfloor{-\frac{\log(t)}{\log(\rho_L)}}}$ are only functions of the system up to time $t-\myfloor{-\frac{\log(t)}{\log(\rho_L)}}-1$, while $\tilde{x}_t$ and $\tilde{u}_t$ are independent with event before time $t-\myfloor{-\frac{\log(t)}{\log(\rho_L)}}$ by definitions in \cref{eq: tilde x t definition,eq: tilde u t definition}.
Our initial target is to identify the distribution of $(\Ah_t - A)x_t + (\Bh_t- B)u_t$. We will start from its substitution 
\begin{equation*}
\begin{split}
    &(\Ah_{t-\myfloor{-\frac{\log(t)}{\log(\rho_L)}}} - A)\tilde{x}_t + (\Bh_{t-\myfloor{-\frac{\log(t)}{\log(\rho_L)}}}- B)\tilde{u}_t \\
    =& ((\Ah_{t-\myfloor{-\frac{\log(t)}{\log(\rho_L)}}} - A) + K (\Bh_{t-\myfloor{-\frac{\log(t)}{\log(\rho_L)}}}- B))\tilde{x}_t + (\Bh_{t-\myfloor{-\frac{\log(t)}{\log(\rho_L)}}}- B)\xi_t
.\end{split}
\end{equation*}
Because of this independence after substitution, the first term is independent with the second term, and their asymptotic distribution can be described by \cref{eq: fast slow rate CLT}. 
\begin{lemma}
\label{lem: CLT of substitution}
For any $\xi_t$ independent of the data  before $t$: $\{\varepsilon_i, \eta_i\}_{i=0}^{t-1}$: 
\begin{align*}
    &\left(
    \tilde{x}_t^\top 
   \left( \sum_{p=0}^{\infty}L ^{p}\left(I_n + 1_{\{\beta=1,\alpha=0\}}\frac{\tau^2}{\sigma^2}BB^\top\right)(L ^{p})^\top\right)^{-1}
    \tilde{x}_t  
    +
     \frac{\beta \sigma^2}{\tau^2}
     t^{1-\beta} \log^{-\alpha}(t)
     \lnorm{\xi_t}^2
    \right)^{-1/2} \\
    &
    \hspace{2cm} \cdot t^{1/2}  \left[(\Ah_{t-\myfloor{-\frac{\log(t)}{\log(\rho_L)}}} - A)\tilde{x}_t + 
    (\Bh_{t-\myfloor{-\frac{\log(t)}{\log(\rho_L)}}}- B)(K\tilde{x}_t+\xi_t)\right] 
    \convD 
    \calN(0,I_n).
\end{align*}
\end{lemma}
The proof of \cref{lem: CLT of substitution} can be found in \cref{The proof of lem: CLT of substitution}.
 With the help of \cref{lem: difference between real and substitutes x u} and \cref{lem: difference between real and substitutes A B}
, we can change all the replacements back to the original form:
\begin{lemma}
\label{lem: CLT original}
For any $\xi_t$ independent of the data before $t$: $\{\varepsilon_i, \eta_i\}_{i=0}^{t-1}$,
\begin{align*}
    &\left(
    x_t^\top       
    \left( \sum_{p=0}^{\infty}L ^{p}\left(I_n + 1_{\{\beta=1,\alpha=0\}}\frac{\tau^2}{\sigma^2}BB^\top\right)(L ^{p})^\top\right)^{-1}
    x_t  
    +
     \frac{\beta \sigma^2}{\tau^2}
     t^{1-\beta} \log^{-\alpha}(t)
     \lnorm{\xi_t}^2
    \right)^{-1/2} \\
    &
    \hspace{2cm} \cdot t^{1/2}  \left[(\Ah_{t} - A)x_t + 
    (\Bh_{t}- B)(\Kh_tx_t+\xi_t)\right] 
    \convD 
    \calN(0,I_n)
.\end{align*}
\end{lemma}
The proof of \cref{lem: CLT original} can be found in 
\cref{The proof of lem: CLT original}.
Since $\eta_t$ is independent with$\{\varepsilon_i, \eta_i\}_{i=0}^{t-1}$, which satisfies the condition of $\xi_t$, we can restate the result with $\eta_t$ replaced by $\xi_t$:
\begin{align*}
    &\left(
    x_t^\top       
    \left( \sum_{p=0}^{\infty}L ^{p}\left(I_n + 1_{\{\beta=1,\alpha=0\}}\frac{\tau^2}{\sigma^2}BB^\top\right)(L ^{p})^\top\right)^{-1}
    x_t  
    +
     \frac{\beta \sigma^2}{\tau^2}
     t^{1-\beta} \log^{-\alpha}(t)
     \lnorm{\eta_t}^2
    \right)^{-1/2} \\
    &
    \hspace{2cm} \cdot t^{1/2}  \left[(\Ah_{t} - A)x_t + 
    (\Bh_{t}- B)(\Kh_tx_t+\eta_t)\right] 
    \convD 
    \calN(0,I_n)
.\end{align*}
Finally, we have the desired conclusion using $\eta_t =  \tau\sqrt{t^{\beta-1}\log^\alpha(t)}\,w_t$:
\begin{align*}
    \left(
    x_t^\top       
    \left( \sum_{p=0}^{\infty}L ^{p}\left(I_n + 1_{\{\beta=1,\alpha=0\}}\frac{\tau^2}{\sigma^2}BB^\top\right)(L ^{p})^\top\right)^{-1}
    x_t  
    +
          \beta \sigma^2
     \lnorm{w_t}^2
    \right)^{-1/2} \\
    \cdot  \; t^{1/2}  \left((\Ah_t - A)x_t + (\Bh_t- B)u_t\right)
    \convD 
    \calN(0,I_n)
.\end{align*}
\end{proof}


\section{The proof of Corollaries}
\subsection{The proof of \cref{corr: K CLT parametric}}
\label{subsection: The proof of K CLT parametric}
\begin{corr*}
Assume $A+BK$ is full rank.
\cref{alg:myAlg} applied to a system described by \cref{eq:LinearModel} under \cref{asm:InitialStableCondition} satisfies
\begin{equation*}
    \sqrt{\frac{\tau^2}{\sigma^2\beta}}t^{\beta/2} \log^{\alpha/2}(t)
    \left(\left(\frac{dK}{d[A,B]}\right) 
    \left(\begin{bmatrix}
    -K^\top \\
    I_d
    \end{bmatrix}
    \otimes 
    I_n
    \right)
    \right)^{-1}
    \vvector\left(\Kh_t - K\right)
    \convD 
    \calN(0, I_{nd})
.\end{equation*}
\end{corr*}
\begin{proof}
Before we prove this result, we should first examine that the matrix $\left(\frac{dK}{d[A,B]}\right) 
    \left(\begin{bmatrix}
    -K^\top \\
    I_d
    \end{bmatrix}
    \otimes 
    I_n
    \right)$ is indeed invertible. Since$\left(\begin{bmatrix}
    -K^\top \\
    I_d
    \end{bmatrix}
    \otimes 
    I_n
    \right)$ has an identity matrix component $I_{dn}$, it is sufficient to show that $\frac{dK}{d[A,B]}$ is full rank.
\subsubsection{$\frac{dK}{d[A,B]}$ is full rank}
We can ignore the effect of $K_0$ and consider $\Kh_t$ to be the same as certainty equivalent controller $\Kt_t$ which is directly calculated by plugging $\Ah_{t-1}, \Bh_{t-1}$ into DARE \cref{eq:ControllerK,eq:riccati}. This is because $\Kh_t = K_0$ only happens finitely often and thus does not affect asymptotic properties; see \cref{section: The proof of one_epoch_estimate_withMyalg}.

Before we start, we need to define how we solve $\frac{dK}{d[A,B]} \in \mathbb{R}^{nd \times n(n+d)}$ and then prove that $\frac{dK}{d[A,B]}$ is indeed a full rank matrix.
Lemmas 3.1 and B.1 from \citet{simchowitz2020naive} gives the  relationship between the derivatives of $K, P, A, B$:
\begin{equation}
    \label{eq: dK}
    dK = -(R+B^\top P B)^{-1}(dB^\top P (A+BK) + B^\top P (dA +dB K) + B^\top dP (A+BK))
,\end{equation}
where $dP$ can be solved from
\begin{equation}
    \label{eq: dP}
    (A+BK)^\top dP (A+BK) -dP + (dA +dBK)^\top P (A+BK) + (A+BK)^\top P (dA +dBK) = 0
.\end{equation}
Now we can solve $\frac{dK}{d[A,B]}$ by \cref{eq: dK} and \cref{eq: dP}. Denote the kernel space of the derivative matrix $\frac{dK}{d[A,B]}$ as $\mathcal{S}$. It suffices to show that $\mathcal{S}$'s dimension is $n(n+d) - nd = n^2$, which implies $\frac{dK}{d[A,B]}$ is full rank with rank $nd$. The equivalent definition of kernel space $\mathcal{S}$ is the small perturbation $\vvector[dA, dB]$ such that $K$ does not change ($dK = 0$):
\begin{equation*}
    dK = \frac{dK}{d[A,B]}\vvector(dA, dB) = 0.
\end{equation*}
Any vector in kernel space $\mathcal{S}$ can be considered as $\vvector[dA, dB]$ which satisfies $dK = 0$ in \cref{eq: dK}, and that means:
\begin{equation}
\label{eq: kernel space eq1}
    dB^\top P (A+BK) + B^\top P (dA +dB K) + B^\top dP (A+BK) = 0
.\end{equation}
On the other hand, \cref{eq: dP} describes a linear recursive relationship between $dP$ and $dA +dB K$, so that we can solve $dP$ with the infinite summation:
\begin{align*}
    dP =& (A+BK)^\top dP (A+BK) + (dA +dBK)^\top P (A+BK) + (A+BK)^\top P (dA +dBK) \\
    =& ((A+BK)^\top)^2 dP (A+BK)^2 \\
    &+ 
    (A+BK)^\top\left((dA +dBK)^\top P (A+BK) + (A+BK)^\top P (dA +dBK)\right) (A+BK) \\
    &+
    (dA +dBK)^\top P (A+BK) + (A+BK)^\top P (dA +dBK)\\
    & \text{(recursively plugging in the first equation)} \\
    =& 
    \sum_{i=0}^\infty
    ((A+BK)^\top)^i
    \left((dA +dBK)^\top P (A+BK) + (A+BK)^\top P (dA +dBK)\right)
    (A+BK)^i
.\end{align*}
Also recall that $A+BK$ is assumed to be full rank matrix, and we can show that $P$ is also full rank; see \cref{section: The proof of one_epoch_estimate_withK}. Thus we can explicitly solve $dB$ from \cref{eq: kernel space eq1} as a linear equation with regard to $dA + dB K$:
\begin{equation*}
    dB^\top  =   -(P (A+BK))^{-1}(B^\top P (dA +dB K) + B^\top dP (A+BK) )
.\end{equation*}
This tells us the kernel space $\mathcal{S}$ is the image of a function of its linear subspace $dA +dB K \in \mathbb{R}^{n^2}$, which means $dim(\mathcal{S}) \le n^2$. Notice by kernel space definition its dimension should be at least $dim(\mathcal{S}) \ge n(n+d) - nd = n^2$, where the equality is achieved when $\frac{dK}{d[A,B]}$ has full rank $nd$. Combining these two equations we have $dim(\mathcal{S}) = n^2$. Finally we arrived at the desired conclusion that dimension of $\frac{dK}{d[A,B]} \in \mathbb{R}^{nd \times n(n+d)}$'s kernel space $\mathcal{S}$ is exactly $n^2$, which means $\frac{dK}{d[A,B]}$ is full rank.

Next we describe the rest of the proof:

\subsubsection{Proof by the Delta method}

By Taylor expansion and the consistency of $[\Ah_t, \Bh_t]$ (see \cref{prop:one_epoch_estimate}), we have
\begin{equation*}
    \vvector\left(\Kh_t - K\right)
    = \left(\frac{dK}{d[A,B]}\right) \vvector\left[\Ah_t- A, \Bh_t-B\right] 
    (1 + o_p(1)) 
.\end{equation*}
From \cref{remark: fast convergence rate} we know
\begin{equation*}
    \Ah_t- A = (\Bh_t-B)(-K) (1+o_p(1))
.\end{equation*}
Then 
\begin{equation*}
    \vvector\left(\Kh_t - K\right)
    = \left(\frac{dK}{d[A,B]}\right) \vvector\left((\Bh_t-B)
    \begin{bmatrix}
    -K, & I_d
    \end{bmatrix}\right) 
    (1 + o_p(1))
.\end{equation*}
which can be written as
\begin{equation*}
    \vvector\left(\Kh_t - K\right)
    = \left(\frac{dK}{d[A,B]}\right) \left(\begin{bmatrix}
    -K^\top \\
    I_d
    \end{bmatrix}
    \otimes 
    I_n
    \right)
    \vvector\left(\Bh_t-B\right) 
    (1 + o_p(1)) 
.\end{equation*}
By \cref{eq: fast slow rate CLT},
\begin{equation*}
    \sqrt{\frac{\tau^2}{\sigma^2\beta}}t^{\beta/2} \log^{\alpha/2}(t)\vvector\left(\Bh_t-B\right) \convD \calN(0, I_{nd})
.\end{equation*}
Combining the above two equations, finally we have
\begin{equation*}
\sqrt{\frac{\tau^2}{\sigma^2\beta}}t^{\beta/2} \log^{\alpha/2}(t)
    \vvector\left(\Kh_t - K\right)
    \convD \left(\frac{dK}{d[A,B]}\right) 
    \left(\begin{bmatrix}
    -K^\top \\
    I_d
    \end{bmatrix}
    \otimes 
    I_n
    \right)
    \calN(0, I_{nd})
.\end{equation*}
From the fact that $\frac{dK}{d[A,B]}$ is full rank and that $\left(\begin{bmatrix}
    -K^\top \\
    I_d
    \end{bmatrix}
    \otimes 
    I_n
    \right)$ has an identity matrix component $I_{dn}$, we can take matrix inverse and get
\begin{equation*}
    \sqrt{\frac{\tau^2}{\sigma^2\beta}}t^{\beta/2} \log^{\alpha/2}(t)
    \left(\left(\frac{dK}{d[A,B]}\right) 
    \left(\begin{bmatrix}
    -K^\top \\
    I_d
    \end{bmatrix}
    \otimes 
    I_n
    \right)
    \right)^{-1}
    \vvector\left(\Kh_t - K\right)
    \convD 
    \calN(0, I_{nd})
.\end{equation*}
\end{proof}

\subsection{The proof of \cref{corr:regret}}
\label{The proof of corr:regret}
\begin{corr*}
The average regret of the controller $U$ defined by \cref{alg:myAlg} applied through time horizon $T$ to a system described by \cref{eq:LinearModel} under \cref{asm:InitialStableCondition} satisfies, as $t \to \infty$ and $T \to \infty$,
\begin{equation}
\label{eq:regret my alg observable 2}
\frac{\mathcal{R}(U,T)}{\tau^2\beta^{-1} \Tr(\Bh_{t}^\top \Ph_t \Bh_{t} +R)T^{\beta-1}\log^\alpha(T)} \convP 1
.\end{equation}
\end{corr*}

\begin{proof}
    This is a direct corollary from \cref{thm:regret}, which states
    \begin{equation*}
\frac{\mathcal{R}(U,T)}{\tau^2\beta^{-1} \Tr(B^\top PB +R)T^{\beta-1}\log^\alpha(T)} \convP 1,
\end{equation*} 
and
from \cref{prop:one_epoch_estimate} and \cref{cor: Ph controlled with Ah and Bh} which implies the consistency of $\Bh_t$ and $\Ph_t$. By Slutsky's theorem we can replace the parameters $B$ and  $P$ in \cref{eq:regret my alg observable 2} with $\Bh_t$ and $\Ph_t$.
\end{proof}

\subsection{The proof of \cref{thm:main}}
\label{The proof of thm:main}
\begin{corr*}
\cref{alg:myAlg} applied to a system described by \cref{eq:LinearModel} under \cref{asm:InitialStableCondition} satisfies
\begin{equation*}
    \Tr\left( 
    \begin{bmatrix}
        \Ah_t - A,\Bh_t- B
    \end{bmatrix} 
    \sum_{i=0}^{t-1}
    \begin{bmatrix}
            x_i\\
            u_i\\
    \end{bmatrix}
    \begin{bmatrix}
            x_i\\
            u_i\\
    \end{bmatrix}^\top
    \begin{bmatrix}
        \Ah_t - A,\Bh_t- B
    \end{bmatrix} ^\top 
    \right)
    \convD \sigma^2\chi^2_{n(n+d)}
.\end{equation*}
\end{corr*}
\begin{proof}
For notational simplicity denote 
$\hat{\Theta}_t :=     \begin{bmatrix}
        \Ah_t, \Bh_t
    \end{bmatrix}$
and 
$\Theta :=     \begin{bmatrix}
        A, B
    \end{bmatrix}$
    . By \cref{thm:main CLT} we know
\begin{equation}
\label{eq: thm3 result}
    \vvector \left((\hat{\Theta}_t -\Theta)D_t \right) \convD
    \calN(0, \sigma^2  I_{\statedim(n+d)} )
.\end{equation}
Potentially we can derive an ellipsoid "confidence region" with the above formula by
\begin{equation}
\label{eq: ellipsoid 1}
    \Tr\left( (\hat{\Theta}_t -\Theta) \left(D_tD_t^\top\right) (\hat{\Theta}_t -\Theta)^\top \right)
    \convD \sigma^2 \chi^2_{n(n+d)}
.\end{equation}
However, since a true confidence region should not require any knowledge on oracle parameters, we need to replace $D_tD_t^\top$ with some observable expression, which turns out to be:

\begin{equation*}
    \Tr\left( (\hat{\Theta}_t -\Theta) \left(\sum_{i=0}^{t-1}\begin{bmatrix}         x_i\\         u_i\\ \end{bmatrix} \begin{bmatrix}         x_i\\         u_i\\ \end{bmatrix}^\top \right) (\hat{\Theta}_t -\Theta)^\top \right)
    \convD  \sigma^2 \chi^2_{n(n+d)}
.\end{equation*}
Next we will explain why it is valid to replace $D_tD_t^\top$ by $\sum_{i=0}^{t-1}\begin{bmatrix}         x_i\\         u_i\\ \end{bmatrix} \begin{bmatrix}         x_i\\         u_i\\ \end{bmatrix}^\top$. We know from \cref{eq: ellipsoid 1} that
\begin{equation*}
    \Tr\left( (\hat{\Theta}_t -\Theta) \left(D_t I_{n+d} D_t^\top\right) (\hat{\Theta}_t -\Theta)^\top \right)
    \convD \sigma^2 \chi^2_{n(n+d)}
,\end{equation*}
and we can replace $I_{n+d}$ by $D_t^{-1}
\sum_{i=0}^{t-1}
\begin{bmatrix}
        x_i\\
        u_i\\
\end{bmatrix}
\begin{bmatrix}
        x_i\\
        u_i\\
\end{bmatrix}^\top  
(D_t^\top)^{-1}
+ o_p(1)$  thanks to \cref{thm:main tool}. As a result,
\begin{equation*}
    \Tr\left( (\hat{\Theta}_t -\Theta)D_t \left(D_t^{-1}
\sum_{i=0}^{t-1}
\begin{bmatrix}
        x_i\\
        u_i\\
\end{bmatrix}
\begin{bmatrix}
        x_i\\
        u_i\\
\end{bmatrix}^\top  
(D_t^\top)^{-1}
+ o_p(1)\right) D_t^\top(\hat{\Theta}_t -\Theta)^\top \right)
    \convD \sigma^2 \chi^2_{n(n+d)}
.\end{equation*}
By \cref{eq: thm3 result},
$\vvector \left((\hat{\Theta}_t -\Theta)D_t\right)$ is of constant order, and thus the $o_p(1)$ can be ignored. Finally, we have
\begin{equation*}
    \Tr\left( (\hat{\Theta}_t -\Theta) \left(\sum_{i=0}^{t-1}\begin{bmatrix}         x_i\\         u_i\\ \end{bmatrix} \begin{bmatrix}         x_i\\         u_i\\ \end{bmatrix}^\top \right) (\hat{\Theta}_t -\Theta)^\top \right)
    \convD  \sigma^2 \chi^2_{n(n+d)}
.\end{equation*}
\end{proof}

\subsection{The proof of \cref{corr: K confidence region}}
\label{subsection: The proof of K confidence region}
\begin{corr*}
\cref{alg:myAlg} applied to a system described by \cref{eq:LinearModel} under \cref{asm:InitialStableCondition} satisfies
\begin{equation}
    \vvector(
        \Kh_t - K
    ) ^\top 
        \left(
        \left(\frac{dK}{d[A, B]}\right)_t
    \left(
\sum_{i=0}^{t-1}
    \begin{bmatrix}
            x_i\\
            u_i\\
    \end{bmatrix}
    \begin{bmatrix}
            x_i\\
            u_i\\
    \end{bmatrix}^\top
\otimes I_n\right)^{-1}
    \left(\frac{dK}{d[A, B]}\right)_t^\top
    \right)^{-1}
    \vvector(
        \Kh_t - K
    ) 
    \convD \sigma^2\chi^2_{nd}
,\end{equation}
where $\left(\frac{dK}{d[A, B]}\right)_t \in \mathbb{R}^{nd \times n(n+d)}$ is defined as $\frac{dK}{d[A, B]}$ evaluated at $\Ah_{t-1}, \Bh_{t-1}$. 
\end{corr*}

\begin{proof}
Again, let us denote $\hat{\Theta}_t :=     \begin{bmatrix}
        \Ah_t, \Bh_t
    \end{bmatrix}$
and 
$\Theta :=     \begin{bmatrix}
        A, B
    \end{bmatrix}$. Starting from \cref{thm:main CLT}
\begin{equation*}
    \vvector \left((\hat{\Theta}_t -\Theta)D_t\right) \convD
    \calN(0, \sigma^2 I_{\statedim(n+d)} )
,\end{equation*}
we need to transfer $D_t$ to its observable version in terms of the Gram matrix. More specifically, we need to find another matrix $E_t$ which is observable and satisfies:
\begin{itemize}
    \item $D_t^{-1}E_t \convP I_{n+d}$ because we want to use Slutsky's theorem.
    \item $E_tE_t^\top = \sum_{i=0}^{t-1}
    \begin{bmatrix}
            x_i\\
            u_i\\
    \end{bmatrix}
    \begin{bmatrix}
            x_i\\
            u_i\\
    \end{bmatrix}^\top$ because 
    $D_t^{-1}
\sum_{i=0}^{t-1}
\begin{bmatrix}
        x_i\\
        u_i\\
\end{bmatrix}
\begin{bmatrix}
        x_i\\
        u_i\\
\end{bmatrix}^\top  
(D_t^\top)^{-1} \convP I_{n+d}$.
\end{itemize}
For now let us assume we have already found such matrix $E_t$, and thus we can replace $D_t$ with $E_t$:
\begin{equation*}
    \vvector \left(    
       (\hat\Theta_t - \Theta)E_t
        \right) \convD 
    \calN(0, \sigma^2 I_{\statedim(n+d)} )
.\end{equation*}
That is:
\begin{equation*}
    (E_t^\top \otimes I_n) \vvector \left(    
       \hat\Theta_t - \Theta
        \right) \convD 
    \calN(0, \sigma^2 I_{\statedim(n+d)} )
.\end{equation*}
Further denote $F_t := E_t^\top \otimes I_n$, and then
\begin{equation}
\label{eq: F_t CLT}
    F_t \vvector \left(    
       \hat\Theta_t - \Theta
        \right) \convD 
    \calN(0, \sigma^2 I_{\statedim(n+d)} )
.\end{equation}
By Taylor expansion and the consistency of $\hat\Theta_t$ (see \cref{prop:one_epoch_estimate}), we have
\begin{equation*}
    \vvector\left(\Kh_t - K\right)
    = \left(\frac{dK}{d\Theta}\right)_t \vvector\left(\hat\Theta_t - \Theta\right) 
    (1 + o_p(1)) 
.\end{equation*}
Since we will prove $D_t^{-1}E_t \convP I_{n+d}$ in \cref{Finding a valid E_t}, $E_t$ is asymptotically invertible, which means we can take inverse of $F_t = E_t^\top \otimes I_n$ in asymptotic equations:
\begin{equation*}
    \vvector\left(\Kh_t - K\right)
    = \left(\frac{dK}{d\Theta}\right)_t (F_t)^{-1}
    F_t \vvector\left(\hat\Theta_t - \Theta\right) 
    (1 + o_p(1)) 
.\end{equation*}
We have already shown in \cref{subsection: The proof of K CLT parametric} that $\frac{dK}{d\Theta}$ is full rank, in the same way we can prove that $\left(\frac{dK}{d\Theta}\right)_t$ is almost surely full rank (the only difference is that we replaced $A, B$ with $\Ah_{t-1}, \Bh_{t-1}$). Recall the QR decomposition, we can re-express $\left(\frac{dK}{d\Theta}\right)_t(F_t)^{-1}$ as $\left(\frac{dK}{d\Theta}\right)_t(F_t)^{-1} = Q_tU_t$, where $Q_t \in \mathbb{R}^{nd \times nd}$ is an invertible matrix, and $U_t \in \mathbb{R}^{nd \times n(n+d)}$ satisfies $U_tU_t^\top = I_{nd}$. This implies that
\begin{equation*}
    \vvector\left(\Kh_t - K\right)
    = Q_tU_tF_t \vvector\left(\hat\Theta_t - \Theta\right) 
    (1 + o(1)) \as
\end{equation*}
From this and \cref{eq: F_t CLT} we know
\begin{equation*}
    Q_t^{-1}\vvector\left(\Kh_t - K\right)
    = U_tF_t \vvector\left(\hat\Theta_t - \Theta\right) 
    (1 + o(1)) 
    \convD \calN(0, \sigma^2 I_{nd} )
.\end{equation*}
That is,
\begin{equation*}
    \vvector\left(\Kh_t - K\right)^\top (Q_t^\top)^{-1}Q_t^{-1}\vvector\left(\Kh_t - K\right)
    \convD \sigma^2\chi^2_{nd}
.\end{equation*}
\begin{equation*}
    \vvector\left(\Kh_t - K\right)^\top (Q_t U_tU_t^\top Q_t^\top)^{-1}\vvector\left(\Kh_t - K\right)
    \convD \sigma^2\chi^2_{nd}
.\end{equation*}
Recall that $\left(\frac{dK}{d\Theta}\right)_t(F_t)^{-1} = Q_tU_t$, and thus
\begin{equation*}
    \vvector\left(\Kh_t - K\right)^\top 
    \left(
    \left(\frac{dK}{d\Theta}\right)_t
    (F_t^\top F_t)^{-1}
    \left(\frac{dK}{d\Theta}\right)_t^\top
    \right)^{-1}\vvector\left(\Kh_t - K\right)
    \convD \sigma^2\chi^2_{nd}
.\end{equation*}
By definition 
\begin{align*}
    F_t^\top F_t =& (E_t^\top \otimes I_n)^\top(E_t^\top \otimes I_n) \\
    =&(E_t \otimes I_n)(E_t^\top \otimes I_n) \\
    =& E_t E_t^\top \otimes I_n \\
    =& 
    \sum_{i=0}^{t-1}
\begin{bmatrix}
        x_i\\
        u_i\\
\end{bmatrix}
\begin{bmatrix}
        x_i\\
        u_i\\
\end{bmatrix}^\top \otimes I_n.
\end{align*}
Finally we can say
\begin{equation*}
    \vvector\begin{bmatrix}
        \Kh_t - K
    \end{bmatrix} ^\top 
        \left(
        \left(\frac{dK}{d\Theta}\right)_t
    \left(
    \sum_{i=0}^{t-1}
\begin{bmatrix}
        x_i\\
        u_i\\
\end{bmatrix}
\begin{bmatrix}
        x_i\\
        u_i\\
\end{bmatrix}^\top \otimes I_n\right)^{-1}
    \left(\frac{dK}{d\Theta}\right)_t^\top
    \right)^{-1}
    \vvector\begin{bmatrix}
        \Kh_t - K
    \end{bmatrix} 
    \convD \sigma^2\chi^2_{nd}
.\end{equation*}

The only remaining task is to find a valid $E_t$ which satisfies $D_t^{-1}E_t \convP I_{n+d}$ and $E_tE_t^\top = \sum_{i=0}^{t-1}
    \begin{bmatrix}
            x_i\\
            u_i\\
    \end{bmatrix}
    \begin{bmatrix}
            x_i\\
            u_i\\
    \end{bmatrix}^\top$.
Although we already have \cref{thm:main tool}, $E_t = \left(\sum_{i=0}^{t-1}
    \begin{bmatrix}
            x_i\\
            u_i\\
    \end{bmatrix}
    \begin{bmatrix}
            x_i\\
            u_i\\
    \end{bmatrix}^\top\right)^{1/2}$ is still not necessarily a valid choice, because we can only show $D_t^{-1}\left(\sum_{i=0}^{t-1}
    \begin{bmatrix}
            x_i\\
            u_i\\
    \end{bmatrix}
    \begin{bmatrix}
            x_i\\
            u_i\\
    \end{bmatrix}^\top\right)^{1/2}$ is asymptotically an orthogonal matrix, but not identity matrix.

\subsubsection{Finding a valid $E_t$}
\label{Finding a valid E_t}
Recall \cref{eq: uu decomp} that
\begin{equation*}
    \sum_{i=0}^{t-1}u_iu_i^\top/t^\beta\log^{\alpha}(t) = 
            KM_tK ^\top + \Delta_t K^\top  + K\Delta_t^\top  + \frac{\tau^2}\beta  I_d + o_p(1)
.\end{equation*}
Now denote 
\begin{equation}
\label{eq: definition Delta u}
    \Delta_u := \sum_{i=0}^{t-1}u_iu_i^\top/ t^\beta\log^{\alpha}(t)
    - \left(KM_tK ^\top + \Delta_t K^\top  + K\Delta_t^\top\right) = \frac{\tau^2}\beta  I_d + o_p(1)
,\end{equation}
which is asymptotically proportional to the identity matrix, and is also symmetric. 
Recall that $D_t$ is defined as 
\[D_t := 
        t^{\beta/2}\log^{\alpha/2}(t)
\left[
    \begin{array}{cc}
    I_n & 0\\
    K & I_d\\
    \end{array}
\right]
\left[
    \begin{array}{cc}
    C_t^{1/2} & 0\\
    0 & \sqrt{\frac{\tau^2}{\beta}} I_d\\
    \end{array}
\right].\] 
We will verify that the following construction of $E_t$ is a valid choice:
\[E_t := 
        t^{\beta/2}\log^{\alpha/2}(t)
\left[
    \begin{array}{cc}
    I_n & 0\\
    K & I_d\\
    \end{array}
\right]
\left[
    \begin{array}{cc}
    (M_t - \Delta_t^\top \Delta_u^{-1}\Delta_t )^{1/2} & 
     \Delta_t^\top \Delta_u^{-1/2}\\
    0 & \Delta_u^{1/2}\\
    \end{array}
\right].\] 
We shall examine the two conditions $D_t^{-1}E_t \convP I_{n+d}$ and $E_tE_t^\top = \sum_{i=0}^{t-1}
    \begin{bmatrix}
            x_i\\
            u_i\\
    \end{bmatrix}
    \begin{bmatrix}
            x_i\\
            u_i\\
    \end{bmatrix}^\top$ in order.
~\paragraph{Proving $D_t^{-1}E_t \convP I_{n+d}$}
It suffices to show:
\begin{equation*}
    \left[
    \begin{array}{cc}
    C_t^{-1/2} & 0\\
    0 & \sqrt{\frac{\beta}{\tau^2}} I_d\\
    \end{array}
\right]
\left[
    \begin{array}{cc}
    (M_t - \Delta_t^\top \Delta_u^{-1}\Delta_t )^{1/2} & 
    \Delta_t^\top\Delta_u^{-1/2} \\
    0 & \Delta_u^{1/2}\\
    \end{array}
\right] \convP I_{n+d}
.\end{equation*}
\cref{eq: Ct inverse order,eq: Ct -1 Mt convP I_n,eq: Delta t order,eq: definition Delta u} states that
\begin{itemize}
    \item $C_t^{-1} = \calO(t^{\beta-1}\log^\alpha(t))$
    \item $M_t = C_t (1+o_p(1))$
    \item $\Delta_t = \calO_p(t^{1-3\beta/2}\log^{\frac{-3\alpha + 3}{2}}(t))$
    \item $\Delta_u = \frac{\tau^2}\beta  I_d + o_p(1)$
\end{itemize}
With these facts, $C_t^{-1/2}\Delta_t^\top\Delta_u^{-1/2} =  \calO_p(t^{1/2-\beta}\log^{\frac{-2\alpha + 3}{2}}(t)) \convP 0$ and $\sqrt{\frac{\tau^2}{\beta}} I_d \Delta_u^{1/2} \convP I_d$ are immediate.
It only remains to show that $C_t^{-1/2} (M_t - \Delta_t^\top \Delta_u^{-1}\Delta_t )^{1/2} \convP I_n$.
Notice 
\begin{equation*}
    C_t^{-1/2} (M_t - \Delta_t^\top \Delta_u^{-1}\Delta_t )^{1/2}
    = (C_t^{-1}M_t - C_t^{-1}\Delta_t^\top \Delta_u^{-1}\Delta_t)^{1/2}
,\end{equation*}
and \cref{eq: Ct -1 Mt convP I_n} shows that $C_t^{-1}M_t \convP I_n$. It only remains to show 
\begin{equation*}
    C_t^{-1}\Delta_t^\top \Delta_u^{-1}\Delta_t \convP 0
,\end{equation*}
which is true because when $\beta > 1/2$ or $\beta = 1/2$ and $\alpha > 3/2$:
\begin{align*}
    &C_t^{-1}\Delta_t^\top \Delta_u^{-1}\Delta_t \\
    =& \calO(t^{\beta-1}\log^\alpha(t))\calO_p(t^{1-3\beta/2}\log^{\frac{-3\alpha + 3}{2}}(t))
    \calO_p(1)
    \calO_p(t^{1-3\beta/2}\log^{\frac{-3\alpha + 3}{2}}(t)) \\
    =& \calO_p(t^{-2\beta+1}\log^{-2\alpha+3}(t)) \\
    =& o_p(1)
.\end{align*}
~\paragraph{Proving $E_tE_t^\top = \sum_{i=0}^{t-1}
    \begin{bmatrix}
            x_i\\
            u_i\\
    \end{bmatrix}
    \begin{bmatrix}
            x_i\\
            u_i\\
    \end{bmatrix}^\top$}
    
\begin{align*}
    E_tE_t^\top =& 
        t^{\beta}\log^{\alpha}(t)
\left[
    \begin{array}{cc}
    I_n & 0\\
    K & I_d\\
    \end{array}
\right]
\left[
    \begin{array}{cc}
    (M_t - \Delta_t^\top \Delta_u^{-1}\Delta_t )^{1/2} & 
    \Delta_t^\top\Delta_u^{-1/2} \\
    0 & \Delta_u^{1/2}\\
    \end{array}
\right] \\
& \hspace{1cm} \cdot \left[
    \begin{array}{cc}
    (M_t - \Delta_t^\top \Delta_u^{-1}\Delta_t )^{1/2} & 
    0\\
    \Delta_u^{-1/2}\Delta_t  & \Delta_u^{1/2}\\
    \end{array}
\right]
\left[
    \begin{array}{cc}
    I_n & K^\top\\
    0 & I_d\\
    \end{array}
\right]\\
=& t^{\beta}\log^{\alpha}(t)
\left[
    \begin{array}{cc}
    I_n & 0\\
    K & I_d\\
    \end{array}
\right]\left[
    \begin{array}{cc}
    M_t & \Delta_t^\top\\
    \Delta_t & \Delta_u\\
    \end{array}
\right]\left[
    \begin{array}{cc}
    I_n & K^\top \\
    0 & I_d\\
    \end{array}
\right] \\
=&  
t^{\beta}\log^{\alpha}(t)
\left[
    \begin{array}{cc}
    M_t & M_tK ^\top + \Delta_t^\top  \\
    KM_t+ \Delta_t & KM_tK ^\top + \Delta_t K^\top  + K\Delta_t^\top  + \Delta_u  \\
    \end{array}
\right] \\
&\text{(By Definitions \cref{eq: defn Mt,eq: defn Delta t,eq: definition Delta u}}) \\
=& \sum_{i=0}^{t-1} \begin{bmatrix}         x_i\\         u_i\\ \end{bmatrix} \begin{bmatrix}         x_i\\         u_i\\ \end{bmatrix}^\top
.\end{align*}
We will re-use the following equation later:
\begin{equation}
    \label{eq: Gram matrix my symbol}
    \sum_{i=0}^{t-1} \begin{bmatrix}         x_i\\         u_i\\ \end{bmatrix} \begin{bmatrix}         x_i\\         u_i\\ \end{bmatrix}^\top  =
    t^{\beta}\log^{\alpha}(t)
\left[
    \begin{array}{cc}
    I_n & 0\\
    K & I_d\\
    \end{array}
\right]\left[
    \begin{array}{cc}
    M_t & \Delta_t^\top\\
    \Delta_t & \Delta_u\\
    \end{array}
\right]\left[
    \begin{array}{cc}
    I_n & K^\top \\
    0 & I_d\\
    \end{array}
\right]
\end{equation}

\end{proof}

\subsection{The proof of \cref{thm: prediction CLT}}
\label{The proof of thm: prediction CLT}
\begin{corr*}
\cref{alg:myAlg} applied to a system described by \cref{eq:LinearModel} under \cref{asm:InitialStableCondition} satisfies: 
\begin{align*}
\begin{split}
    \left(
    \sigma^2
\begin{bmatrix}
        x_t \\
        u_{t}
    \end{bmatrix}^\top
    \left(
    \sum_{i=0}^{t-1}
    \begin{bmatrix}
            x_i\\
            u_i\\
    \end{bmatrix}
    \begin{bmatrix}
            x_i\\
            u_i\\
    \end{bmatrix}
    ^\top
    \right)^{-1}
    \begin{bmatrix}
        x_t \\
        u_{t}
    \end{bmatrix}
    \right)^{-1/2} \left((\Ah_{t} - A)x_t + 
    (\Bh_{t}- B)u_{t}\right) 
    \convD \calN(0, I_n)
.\end{split}
\end{align*}
where $u_t = \Kh_t x_t + \xi_t$ for any $\xi_t$ independent of the data before $t$: $\{\varepsilon_i, \eta_i\}_{i=0}^{t-1}$.
\end{corr*}
\begin{proof}

This one final lemma connects \cref{lem: CLT original} to our desired conclusion by changing the parametric expression to the observable one:
\begin{lemma}
\label{lem: variance equivalence}
For any $\xi_t$ independent of the data before $t$: $\{\varepsilon_i, \eta_i\}_{i=0}^{t-1}$,
\begin{align*}
    \left(
    x_t^\top   
    \left( \sum_{p=0}^{\infty}L ^{p}\left(I_n + 1_{\{\beta=1,\alpha=0\}}\frac{\tau^2}{\sigma^2}BB^\top\right)(L ^{p})^\top\right)^{-1}
    x_t  
    +
     \frac{\beta \sigma^2}{\tau^2}
     t^{1-\beta} \log^{-\alpha}(t)
     \lnorm{\xi_t}^2
    \right)^{-1/2} \\
    \cdot \;t^{1/2}
    \left(
    \sigma^2
\begin{bmatrix}
        x_t \\
        u_{t}
    \end{bmatrix}^\top
    \left(
    \sum_{i=0}^{t-1}
    \begin{bmatrix}
            x_i\\
            u_i\\
    \end{bmatrix}
    \begin{bmatrix}
            x_i\\
            u_i\\
    \end{bmatrix}
    ^\top
    \right)^{-1}
    \begin{bmatrix}
        x_t \\
        u_{t}
    \end{bmatrix}
    \right)^{1/2} \convP 1
.\end{align*}
\end{lemma}
The proof of \cref{lem: variance equivalence} can be found in \cref{The proof of lem: variance equivalence}. Finally, we can say
\begin{align*}
    \left(
    \sigma^2
\begin{bmatrix}
        x_t \\
        u_{t}
    \end{bmatrix}^\top
    \left(
    \sum_{i=0}^{t-1}
    \begin{bmatrix}
            x_i\\
            u_i\\
    \end{bmatrix}
    \begin{bmatrix}
            x_i\\
            u_i\\
    \end{bmatrix}
    ^\top
    \right)^{-1}
    \begin{bmatrix}
        x_t \\
        u_{t}
    \end{bmatrix}
    \right)^{-1/2} 
    \left((\Ah_{t} - A)x_t + 
    (\Bh_{t}- B)u_t\right)
    \convD \calN(0, I_n)
.\end{align*}

\end{proof}

\section{The proof of Propositions}
\label{sec: The proof of Propositions}
\subsection{The proof of \cref{prop:one_epoch_estimate}}
\label{section: The proof of one_epoch_estimate}
\begin{prop*}[Similar to Proposition C.1 in \citet{dean2018regret}]
Let $x_0 \in \R^{\statedim}$ be any initial state. Assume \cref{asm:InitialStableCondition} is satisfied. When applying $\cref{alg:myAlg}$, 
  \begin{equation*}
      \max\left\{ \norm{\Ah_t - A}, \norm{\Bh_t - B}\right\} = 
    \calO(t^{-\frac{\beta}{2}} \log^{\frac{-\alpha + 1}{2}}(t)) \as
  \end{equation*}
\end{prop*}

\subsubsection{Proof Outline}
\begin{proof}
We shall see that all the properties we derived in this section only require the safety condition \cref{alg:myAlg} Line \ref{line:check} without any other requirement on the controller $\Kh_t$, and thus also apply to \cref{alg:myAlg} with logarithmic updates; see \cref{remark: Logarithmically-update estimates}. 

According to \cref{alg:myAlg} Line \ref{line:check}, we keep our controller $\Kh_t$ bounded $\norm{\Kh_t} \le C_K$, which means the next state can not be too far from the previous state. At the same time, whenever the state is too large ($\norm{x_t} > C_x\log(t)$), it is tuned down by safe controller $K_0$. Overall speaking, the state $x_t$ is always controlled with at most $\log(t)$ growth. We will see in \cref{lemma:lwm} that when state growth is controlled, we have a decent bound on $\Ah_t, \Bh_t$.

In other words, as long as we still run \cref{alg:myAlg} Line \ref{line:check} at every time step, which is enough to "control" the system by itself, any $\Ah_t, \Bh_t$ generated with Line \ref{line:ols} satisfies 
  \begin{equation*}
      \max\left\{ \norm{\Ah_t - A}, \norm{\Bh_t - B}\right\} = 
    \calO(t^{-\frac{\beta}{2}} \log^{\frac{-\alpha + 1}{2}}(t)) \as
  \end{equation*}
regardless of the estimation result before time $t$. 

\cref{lemma:lwm} follows from a result by \citet{simchowitz2018learning} on the estimation of linear response time-series. We present that result in the context of our problem. Let $\Theta := [A, B]$, and define $z_t := 
\begin{bmatrix}
x_t \\
u_t
\end{bmatrix}$. Then, the OLS estimator \cref{eq: AhBh estimator} is
\begin{align}
\label{eq:ols_M}
(\Ah_T, \Bh_T) = \hat{\Theta}_T \in \arg \min_{\Theta} \sum_{t = 0}^{T - 1} \frac{1}{2}\ltwonorm{x_{t + 1} - \Theta z_t}^2.
\end{align}

We know that the accuracy of the OLS estimator is related to the covariance structure of the predictors, which are $\{z_t\}_{t = 0}^{T}$ in our context. To capture such covariance structure, we need the following definiton:
\begin{defn}[BMSB condition]
\label{def:BMSB condition}
The $\{\calF_t\}_{t \geq 0}$-adapted process $\{z_t\}_{t = 0}^T$ is said to satisfy the $(k, \nu, \xi)$-\emph{block martingale small-ball} (BMSB) condition if for any $0 \le j \le T-k$ and $v \in \mathcal{S}^{\statedim + \inputdim-1} := \{x \in \mathbb{R}^{\statedim + \inputdim}: \norm{x} = 1\}$, one has that 

\begin{align*}
\frac{1}{k}\sum_{i = 1}^{k} \P\left( |\langle v, z_{j + i}\rangle| \geq \nu {|\calF_j}\right) \geq \xi  \as
\end{align*}
\end{defn}

This condition is used for characterizing the size of the minimum eigenvalue of the matrix $\sum_{t = 0}^{T - 1} z_t z_t^\top$.
A larger $\nu$ guarantees a larger lower bound of the minimum eigenvalue. In the context of our problem the result by \citet{simchowitz2018learning} translates as follows.

\begin{lemma}[A slightly different version of Theorem C.2 in \citet{dean2018regret}]
\label{lemma:lwm}
For $\delta \in (0,\frac{(n+d)\xi^2}{2}]$, for every $T$, $k$, $\nu$, and $\xi$ such that $\{z_t\}_{t = 0}^T$ satisfies the $(k, \nu, \xi)$-BMSB and 
\begin{equation}
\label{eq:condition of T for AB bound}
T/k \ge \frac{10(n+d)}{\xi^2}   \log\left(\frac{100(n+d)\sum_{t = 0}^{T-1}\Tr(\E  z_t z_t^\top)}{T \nu^{2}\xi^2\delta^{1 + \frac1{n+d}}}    \right) 
.\end{equation}

the estimate $\hat{\Theta}_T$ defined in \cref{eq:ols_M} satisfies the following statistical rate
\begin{equation}
\label{eq:hiprob AB bound}
\P\left(\lnorm{\hat{\Theta}_T-\Theta} >\frac{90\sigma}{\xi\nu}\sqrt{\frac{n+d}{T}\left(1  + \log\left(\frac{10(n+d)\sum_{t = 0}^{T-1}\Tr(\E  z_t z_t^\top)}{T\delta^{1 + \frac1{n+d}} \nu^{2}\xi}    \right) \right)} \right) \le  3\delta.
\end{equation}

\end{lemma}
The proof of \cref{lemma:lwm} can be found in \cref{The proof of lemma:lwm}.

We will show that $\sum_{t=1}^T \Tr(\E  z_t z_t^\top)$ grows linearly with $T$ (ignoring logarithmic terms), which means in \cref{eq:condition of T for AB bound} the LHS grows faster than the RHS, and is thus always satisfied if $T$ is large enough. \cref{lemma:lwm} is saying that for any $T$ larger than some constant, we can control the $L_2$ norm of the system parameter estimate $\hat{\Theta}_T$, which implies we can control the $L_2$ norm of both $\Ah_T$ and $\Bh_T$. 

Still there is one more gap from our \cref{prop:one_epoch_estimate}, which requires \emph{uniform} control on $\Ah_T$ and $\Bh_T$. Fortunately, we have the blessing that this high-probability bound is in the log scale w.r.t $\delta$. Because of that, we can choose a series of decaying $\delta_T = 1/T^2$ for each different estimate $\hat{\Theta}_T$, so that $\sum_{T=C}^\infty 1/T^2 \le 1/C$ and we can achieve a uniform high probability bound on $\Ah_T$ and $\Bh_T$ for all $T  > C$, which directly leads to the desired conclusion once we plug in appropriate values for $k$, $\nu$, and $\xi$:
  \begin{equation*}
      \max\left\{ \norm{\Ah_t - A}, \norm{\Bh_t - B}\right\} = 
    \calO(t^{-\frac{\beta}{2}} \log^{\frac{-\alpha + 1}{2}}(t)) \as
  \end{equation*}
  To sum up, there are three main steps in our proof of \cref{prop:one_epoch_estimate}:
\begin{itemize}
    \item Verify $\{z_t\}_{t = 0}^T$ satisfies the $(k, \nu, \xi)$-BMSB condition in our setting.
    \item Replace $\Tr(\E  z_t z_t^\top)$ in \cref{lemma:lwm} by an explicit upper bound in terms of $T$.
    \item Prove a uniform high probability bound for $\Ah_T$ and $\Bh_T$ by choosing with $\delta_T = 1/T^2$ with \cref{lemma:lwm}.
\end{itemize}

\end{proof}

\subsubsection{Verifying $\{z_t\}_{t = 0}^T$ satisfies the $(k, \nu, \xi)$-BMSB condition} 
In order to apply \cref{lemma:lwm}, we need to find $k$, $\nu$, and $\xi$ such that $\{z_t\}_{t = 0}^T$ satisfies the
$(k, \nu, \xi)$-BMSB condition.

\begin{lemma}[Similar to Lemma C.3 in \citet{dean2018regret}]
\label{lem:bmsb}
If we assume \cref{asm:InitialStableCondition}, then apply \cref{alg:myAlg}, the process $\{z_t\}_{t \geq 0}^T $ satisfies the 
$(k, \nu, \xi)$-BMSB condition for
\begin{align*} 
(k, \nu, \xi) = \left(1, \sqrt{\sigma_{\eta,T}^2\min\left(\frac{1}{2}, \frac{\sigma ^2}{2 \sigma ^2 C_K^2  + \tau^2} \right)} , \frac{3}{10}\right),
\end{align*}
where $\sigma_{\eta,T}^2 = \tau^2 T^{\beta-1}\log^\alpha(T)$.
\end{lemma}

See \cref{The proof of lem:bmsb} for the proof of \cref{lem:bmsb}. 


\subsubsection{Upper bound of $\Tr(\E  z_t z_t^\top)$ in terms of $T$} 
The benefit of a non-random upper bound of $\Tr(\E  z_t z_t^\top)$ w.r.t $T$ is two-fold.
\begin{itemize}
    \item We can know exactly how large our $T$ should be for \cref{eq:condition of T for AB bound} to hold.
    \item Furthermore, we can also substitute the upper bound in to \cref{eq:hiprob AB bound}. 
\end{itemize} 

\cref{lem:bound_covariance} shows that we have an upper bound of $\Tr(\E  z_t z_t^\top)$ that is $\logO(T)$.
\begin{lemma}[Similar to Lemma C.4 in \citet{dean2018regret}]
\label{lem:bound_covariance}
If we assume \cref{asm:InitialStableCondition}, then apply \cref{alg:myAlg}, the process $\{z_t\}_{t \geq 0}^T $ satisfies
\begin{equation}
\begin{split}
\label{eq:lemma3Conclusion}
\sum_{t = 0}^{T - 1} \Tr \left( \E z_t z_t^\T\right) = \calO(T\log^2(T))
\end{split}
.\end{equation}
\end{lemma}
See \cref{The proof of lem:bound_covariance} for the proof of \cref{lem:bound_covariance}. 


\subsubsection{Uniform upper bound for $\max\left\{ \norm{\Ah_t - A}, \norm{\Bh_t - B}\right\}$}
With \cref{lem:bmsb} and \cref{lem:bound_covariance} in hand, we can translate \cref{lemma:lwm} into our problem setting. Fixing $\delta \in (0,\frac{(n+d)\xi^2}{2}]$, we already proved by \cref{lem:bmsb} that the process $z_t = \begin{bmatrix}
x_t \\
u_t
\end{bmatrix}$
satisfies the
\begin{align}
\label{eq:BMSB condition}
(k, \nu, \xi) = \left(1, \sqrt{\sigma_{\eta,T}^2\min\left(\frac{1}{2}, \frac{\sigma ^2}{2 \sigma ^2 C_K^2  + \tau^2} \right)} , \frac{3}{10}\right)\text{ BMSB condition.}
\end{align}

If we choose $\delta = \frac1{3T^2}$ and $T$ such that \cref{eq:condition of T for AB bound} 
holds with $(k,\nu,\xi)$ in \cref{eq:BMSB condition}, we can apply \cref{lemma:lwm}. By \cref{eq:lemma3Conclusion}, we only need $T$ to satisfy


\begin{align*}
T/k 
\ge& \frac{10(n+d)}{\xi^2}   \log\left(\frac{100(n+d)\logO(T)}{T \nu^{2}\xi^2\delta^{1 + \frac1{n+d}}}    \right) 
  \\
=&   \calO(1)\log \left(\frac{ \logO(T)}{T \sigma_{\eta,T}^2 \frac{\sigma ^2}{2 \sigma ^2 C_K^2  + \tau^2} T^{-2(1 + \frac1{n+d})}} \right)  \qquad (\xi = \frac{3}{10} \text{ is fixed constant}) \\ 
=&   \logO(1)  \qquad \text{(Recall that } \sigma_{\eta,T}^2 = T^{\beta-1}\log^\alpha(T) )
.\end{align*}


Since $T$ is growing faster than $\logO(1)$, the above condition is essentially saying that our $T$ should be larger than some constant $\calO(1)$. Suppose that is the case, then following \cref{lemma:lwm} and \cref{lem:bound_covariance}, the estimate $\hat{\Theta}_T$ defined in \cref{eq:ols_M} satisfies the following statistical rate
\begin{align*}
& \P\left(\lnorm{\hat{\Theta}_T-\Theta} >\frac{90\sigma}{\xi\nu}\sqrt{\frac{n+d}{T}\left(1  + \log\left(\frac{10(n+d)\logO(T)}{T\delta^{1 + \frac1{n+d}} \nu^{2}\xi}    \right) \right)} \right) \\
\leq& \P\left(\lnorm{\hat{\Theta}_T-\Theta} >\frac{90\sigma}{\xi\nu}\sqrt{\frac{n+d}{T}\left(1  + \log\left(\frac{10(n+d)\sum_{t = 1}^T\Tr(\E  z_t z_t^\top)}{T\delta^{1 + \frac1{n+d}} \nu^{2}\xi}    \right) \right)} \right)  \\
\le&  3\delta.
\end{align*}

Notice that $\hat{\Theta}_T = [\Ah_T, \Bh_T]$, and we know that $\max\left\{ \substack{\norm{\Ah_T - A},\\ \norm{\Bh_T - B}}\right\} \le \ltwonorm{\hat{\Theta}_T - \Theta}$. That is to say
\[\P\left(\max\left\{ \substack{\norm{\Ah_T - A},\\ \norm{\Bh_T - B}}\right\} >\frac{90\sigma}{\xi\nu}\sqrt{\frac{n+d}{T}\left(1  + \log\left(\frac{10(n+d)\logO(1)}{T\delta^{1 + \frac1{n+d}} \nu^{2}\xi}    \right) \right)} \right) \le 3\delta.\]

Next we substitute $k= 1$, $\xi = \frac{3}{10}$, $\nu =  \sqrt{\sigma_{\eta,T}^2\min\left(\frac{1}{2}, \frac{\sigma ^2}{2 \sigma ^2 C_K^2  + \tau^2} \right)}$, and $\delta = \frac1{3T^2}$ into the previous equation

\begin{align*}
    &\P \left(\max\left\{ \substack{\norm{\Ah_T - A},\\ \norm{\Bh_T - B}}\right\} > \frac{\calO(1)  }{ \sqrt{\sigma_{\eta,T}^2}} \sqrt{\frac{n+d}{T}\left(1  + \log\left(\frac{\logO(T)}{T (3T^{-2})^{1 + \frac1{n+d}} \sigma_{\eta,T}^2}    \right) \right)}\right) \le \frac1{T^2}
.\end{align*}
By merging all constant parameters in to the $\calO$ style expression, and noticing that $\sigma_{\eta,T}^2 = \tau^2T^{\beta-1}\log^{\alpha}(T)$, where $\beta \in [1/2,1)$, we have for any $T >  \calO(1)$:

\[\P \left(\max\left\{ \substack{\norm{\Ah_T - A},\\ \norm{\Bh_T - B}}\right\} > \calO( T^{\frac{1-\beta}{2}}\log^{-\alpha/2}(T)) \sqrt{\frac{n+d}{T}\calO(\log(T))}\right) \le \frac1{T^2},\]
which implies
\[ \qquad \P \left(\max\left\{ \substack{\norm{\Ah_T - A},\\ \norm{\Bh_T - B}}\right\} > \calO( T^{-\frac{\beta}{2}} 
\log^{\frac{-\alpha+1}{2}}(T))
\right) \le \frac1{T^2}.\]

Notice that 
\begin{align*}
    \sum_{T=C+1}^{\infty} \frac{1}{T^2} 
    \le \sum_{T=C+1}^{\infty} \frac{1}{T(T-1)} 
    \le \sum_{T=C+1}^{\infty} \frac{1}{T-1} - \frac{1}{T} 
    = \frac{1}{C} 
.\end{align*}
Therefore we can derive a uniform confidence bound on the estimation error of parameters $\Ah_t$ and $\Bh_t$: For any integer $C >  \calO(1)$:
\begin{align*}
    &\P \left(\exists t > C , \;s.t. \;\max\left\{ \substack{\norm{\Ah_t - A},\\ \norm{\Bh_t - B}}\right\} >  \calO( T^{-\frac{\beta}{2}} 
\log^{\frac{-\alpha+1}{2}}(T))   \right) \\
    \le& \sum_{t=C+1}^\infty \P \left( \max\left\{ \substack{\norm{\Ah_t - A},\\ \norm{\Bh_t - B}}\right\} >  \calO( T^{-\frac{\beta}{2}} 
\log^{\frac{-\alpha+1}{2}}(T))   \right) \\
    \le& \frac{1}{C} 
.\end{align*}
Notice that this is a uniform upper bound for all $t > C$. Recall \cref{defn: Big O notation} \cref{itm: big O as defn}, where we define $X_n = \calO(a_n) \as$ as: 
for almost every $\omega \in \Omega$, there exists a number $C(\omega)$ such that $\abs{X_n(\omega)} \le C(\omega)a_n$, where $\Omega$ denotes the sample space of $\{X_n\}_n$. The previous equation is telling us the union of such event $\omega$ happens with at least probability $1-1/C$, and by taking $C \to \infty$ that is exactly the definition of $\calO(a_n) \as$, and thus:
\begin{equation*}
    \max\left\{ \norm{\Ah_t - A}, \norm{\Bh_t - B}\right\} = 
    \calO(t^{-\frac{\beta}{2}} \log^{\frac{-\alpha + 1}{2}}(t)) \as
\end{equation*}
The same bound holds for logarithmic updates. The reason is that for time $t$, the closest estimation update will always be within $t/c$ time steps of $t$, which does not change the order: 
\begin{equation*}
    \calO((t/c)^{-\frac{\beta}{2}} \log^{\frac{-\alpha + 1}{2}}(t/c))
    = \calO(t^{-\frac{\beta}{2}} \log^{\frac{-\alpha + 1}{2}}(t)) 
.\end{equation*}

\subsection{The proof of \cref{prop:one_epoch_estimate_withK}}

\label{section: The proof of one_epoch_estimate_withK}

\begin{prop*}
Let $x_0 \in \R^{\statedim}$ be any initial state. Assume \cref{asm:InitialStableCondition} is satisfied. When applying $\cref{alg:myAlg}$, 
\begin{equation*}
\max\left\{ \norm{\Ah_t - A}, \norm{\Bh_t - B}, \norm{\Kt_{t+1} - K}\right\}
     = \calO(t^{-\frac{\beta}{2}} \log^{\frac{-\alpha + 1}{2}}(t)) \as
\end{equation*}
\end{prop*}

\subsubsection{Proof Outline}
\begin{proof}
When the problem parameters $(A, B, Q, R)$ are known the optimal policy is given by linear feedback, $u_t = K x_t$, where $K = - (R + B^\top P B)^{-1}B^\top P A$ and $P$ is the (positive definite) solution to the discrete Riccati equation
\begin{align}
\label{eq:discrete_riccati}
  P = A^\top P A - A^\top P B (R + B^\top P B)^{-1} B^\top P A + Q
.\end{align}
In the following context any time we mention $\Ph_t$ and $\Kt_{t+1}$, we are refering to the corresponding certainty equivalent responses.
\begin{align*}
  \Ph_t = \Ah_t^\top \Ph_t \Ah_t - \Ah_t^\top \Ph_t \Bh_t (R + \Bh_t^\top \Ph_t \Bh_t)^{-1} \Bh_t^\top \Ph_t \Ah_t + Q
.\end{align*}
\begin{align*}
\Kt_{t+1} = - (R + \Bh_t^\top \Ph_t \Bh_t)^{-1}\Bh_t^\top \Ph_t \Ah_t
.\end{align*}
Since we already controlled the estimation error of $\Ah_t-A$ and $\Bh_t-B$, one natural thing to ask is that, if we have control over $\Ah_t-A$ and $\Bh_t-B$, do we have control over $\Kt_{t+1}-K$?  This can be achieved by two steps:
\begin{enumerate}
    \item Show that we can control $\Kt_{t+1}$ once $\Ah_t$, $\Bh_t$, and $\Ph_t$  are controlled.
    \item Show that we can control $\Ph_t$ once $\Ah_t$ and $\Bh_t$ are controlled.
\end{enumerate}
\end{proof}

\subsubsection{Show that we can control $\Kt_{t+1}$ once $\Ah_t$, $\Bh_t$, and $\Ph_t$  are controlled}

This is already stated by Proposition 1 in \citet{mania2019certainty}.
Denote the quantity
$$\Gamma_1 := 1 + \max\{\norm{A}, \norm{B}, \norm{P}, \norm{K}\}.$$ 
\begin{prop}[Proposition 1 in \citet{mania2019certainty}]
  \label{prop:stability_perturb}
  Let $\epsilon > 0$ such
  that $\norm{\Ah - A} \leq \epsilon$ and $\norm{\Bh - B} \leq \epsilon$. Also, let $\norm{\Phat - P} \leq \epsilon_P$ such that $\epsilon_P \geq \epsilon$. Assume $\smin(R) \ge 1$ we have
  \begin{align*} 
    \norm{\Kh - K} \leq 7 \Gamma_1^3 \, \epsilon_P .
  \end{align*}
\end{prop}
The $\smin(R)$ represents the minimum eigenvalue of $R$.
we can discard the constraint of $\smin(R) \ge 1$ by the following observation. If we replace our $Q$ and $R$ by $Q/\smin(R)$ and $R/\smin(R)$, then the corresponding solution $P$ for \cref{eq:discrete_riccati} will be $P/\smin(R)$. Notice that changing $Q$ and $R$ by the same proportion does not change the LQR problem. With that being said, our  LS estimator $\Ah_t$, $\Bh_t$, and the nominal controller $\Kt_t$ will remain the same. By this transformation the minimum eigenvalue condition is satisfied, and we only need to control 
\[\norm{\Phat - P}/\smin(R) \le \epsilon_P\]
such that $\epsilon_P \geq \epsilon$, and we will have $\norm{\Kh - K} \leq 7 \Gamma_2^3 \, \epsilon_P$, where $\Gamma_2 := 1 + \max\{\norm{A}, \norm{B}, \norm{P}/\smin(R), \norm{K}\}$. Here we can replace this denominator $\smin(R)$ by any constant smaller than $\smin(R)$, and the whole story would still work. Since later we will also require $\smin(P) \ge 1$, we can choose the shared denominator to be $\min\{\smin(R), \smin(P)\}$. To sum up we have the following corollary of \cref{prop:stability_perturb}.

\begin{corr}
\label{corr:stability_perturb_tmp}
Let $\epsilon > 0$ such
that $\norm{\Ah - A} \leq \epsilon$ and $\norm{\Bh - B} \leq \epsilon$. Also, let $\norm{\Phat - P} \le \min\{\smin(R), \smin(P)\}\epsilon_P$ such that $\epsilon_P \geq \epsilon$. Then we have
\begin{align*} 
\norm{\Kh - K} \leq 7 \Gamma_3^3 \, \epsilon_P .
\end{align*}  
where $\Gamma_3 := 1 + \max\{\norm{A}, \norm{B}, \norm{P}/\min\{\smin(R), \smin(P)\}, \norm{K}\}$.
\end{corr}

Now we only need to prove that $\norm{\Phat - P} = \calO(\epsilon)$ given $\norm{\Ah - A} \leq \epsilon$ and  $\norm{\Bh - B} \leq \epsilon$.

\subsubsection{Show that we can control $\Ph_t$ once $\Ah_t$ and $\Bh_t$ are controlled}

Consider a general square matrix $M$. In order to quantify the decay rate of $\norm{M^k}$, we define
 \begin{align*}
\transient{M}{ \rho} := \sup \left\{\norm{M^k} \rho^{-k} \colon k\geq 0 \right\}.
 \end{align*}
 In other words, $\transient{M}{\rho}$ is the smallest value such that $\norm{M^k} \leq \transient{M}{\rho} \rho^k$ for all $k \geq 0$.
 We note that $\transient{M}{\rho}$ might be infinite, depending on the value of $\rho$, and it is always greater than or equal to one. If $\rho$ is larger than $\rho(M)$, we are guaranteed to have a finite $\transient{M}{\rho}$ (this is a consequence of Gelfand's formula). In particular, if $M$ is a stable matrix, we can choose $\rho < 1$ such that $\transient{M}{\rho}$ is finite. Also, we note that $\transient{M}{\rho}$ is a decreasing function of $\rho$; if $\rho \geq \norm{M}$, we have $\transient{M}{\rho} = 1$.

Recall that  $L := A + BK$. 
 The following proposition that upper bounds $\norm{\Ph - P}$ holds in a more general LQG setting where the matrix $Q$ is unknown:
\begin{prop}[Proposition 2 in \citet{mania2019certainty}]
  \label{prop:93}
  Let $\gamma \geq \rho(L)$ and also let $\epsilon$ be such that $\norm{\Ah - A}$, $\norm{\Bh - B}$, and $\norm{\Qh - Q}$ are at most $\epsilon$. Let $\norm{\cdot}_+ = \norm{\cdot} + 1$. We assume that $R \succ 0$, $(A, B)$ is stabilizable, $(Q^{1/2}, A)$ observable, and $\smin(P) \geq 1$.  
 \begin{align*}
\norm{\Phat - P} \leq \calO(1)\, \epsilon \, \frac{\transient{L}{\gamma}^2}{1 - \gamma^2} \norm{A}_+^2 \norm{P}_+^2 \norm{B}_+ \norm{R^{-1}}_+,
 \end{align*}
as long as
  \begin{align*}
  \epsilon \leq \calO(1) \frac{(1 - \gamma^2)^2}{\transient{L}{\gamma}^4} \norm{A}_+^{-2} \norm{P}_+^{-2} \norm{B}_+^{-3} \norm{R^{-1}}_+^{-2} \min \left\{ \norm{L}_+^{-2}, \norm{P}_+^{-1} \right\}.
  \end{align*}
\end{prop}
Here $\calO(1)$ are pure constants without dependence of any other parameters.
We already assumed in \cref{asm:InitialStableCondition} that $(A,B)$ stabilizable, but we have not defined `observable' yet. An equivalent statement of observable can be found here.
\begin{lemma}[Lemma 2.1 in \citep{payne1973discrete}]
The pair $(C,A)$ is observable if and only if $Ax = \lambda x$, $Cx = 0$ imply $x = 0$
\label{lem:observable}
\end{lemma}
Since we already assumed $Q$ is positive definite, $Qx = 0$ imply $x = 0$, and thus $(Q^{1/2}, A)$ is observable. In the LQAC setting we know $Q$ exactly, so we can remove the estimation bound condition on $Q$.


Now we can restate \cref{prop:93} in the LQAC setting:

\begin{corr}
\label{cor: Ph controlled with Ah and Bh}
  Let $\epsilon$ such that $\norm{\Ah_t - A}$, and $\norm{\Bh_t - B}$ are at most $\epsilon$. Let $\norm{\cdot}_+ = \norm{\cdot} + 1$. We assume that $R \succ 0$, $(A, B)$ is stabilizable,  and $\smin(P) \geq 1$.  
 \begin{align*}
\norm{\Phat_t - P} \leq \calO(1)\, \epsilon \, \frac{\transient{L}{\rho(L)}^2}{1 - \rho(L)^2} \norm{A}_+^2 \norm{P}_+^2 \norm{B}_+ \norm{R^{-1}}_+ = \calO(\epsilon)
 .\end{align*}
as long as
  \begin{align*}
  \epsilon \leq \calO(1) \frac{(1 - \rho(L)^2)^2}{\transient{L}{\rho(L)}^4} \norm{A}_+^{-2} \norm{P}_+^{-2} \norm{B}_+^{-3} \norm{R^{-1}}_+^{-2} \min \left\{ \norm{L}_+^{-2}, \norm{P}_+^{-1} \right\} = \calO(1).
  \end{align*}
\end{corr}
Here, the upper bound condition on $\epsilon$ is to ensure that $\Ah_t, \Bh_t$ is stabilizable, so that $\Ph_t$ is well defined.
Furthermore, following the paragraph after Proposition 2 in \citet{mania2019certainty}, the assumption $\smin(P) \geq 1$ can be made without loss of generality when the other assumptions are satisfied. The reason is that, when $R \succ 0$ and $(Q^{1/2}, A)$ observable, the value function matrix $P$ is guaranteed to be positive definite. 
Similar to how we got \cref{corr:stability_perturb_tmp}, by replacing $Q$, $R$ and $P$ with $Q/\min\{\smin(R), \smin(P)\}$, $R/\min\{\smin(R), \smin(P)\}$ and $P/\min\{\smin(R), \smin(P)\}$, we can remove the constraint $\smin(P) \geq 1$.

\begin{corr}
\label{cor:prop93}
  Suppose $\norm{\Ah_t - A} \le \epsilon$ and $\norm{\Bh_t - B} \le \epsilon$. Let $\norm{\cdot}_+ = \norm{\cdot} + 1$. We assume that $R \succ 0$ and $(A, B)$ is stabilizable.  
 \begin{align*}
\norm{\Phat_t - P} 
\leq& \min\{\smin(R), \smin(P)\}\calO(1)\, \epsilon \, \frac{\transient{L}{\rho(L)}^2}{1 - \rho(L)^2} \norm{A}_+^2 \lnorm{\frac{P}{\min\{\smin(R), \smin(P)\}}}_+^2 \norm{B}_+ \lnorm{\left(\frac{R}{\min\{\smin(R), \smin(P)\}}\right)^{-1}}_+ \\
=& \calO(\epsilon)
 .\end{align*}
as long as
  \begin{align*}
  \epsilon 
  \leq& \calO(1) \frac{(1 - \rho(L)^2)^2}{\transient{L}{\rho(L)}^4} \norm{A}_+^{-2} \lnorm{\frac{P}{\min\{\smin(R), \smin(P)\}}}_+^{-2} \\
  &\norm{B}_+^{-3} \lnorm{\left(\frac{R}{\min\{\smin(R), \smin(P)\}}\right)^{-1}}_+^{-2} \min \left\{ \norm{L}_+^{-2}, \lnorm{\frac{P}{\min\{\smin(R), \smin(P)\}}}_+^{-1} \right\} \\
  =& \calO(1).
  \end{align*}
\end{corr}

\subsubsection{Combining the two results together}
With \cref{corr:stability_perturb_tmp} and \cref{cor:prop93} the following corollary is straightforward.
\begin{corr}
  \label{cor:stability_perturb}
  Let $\epsilon > 0$ such
  that $\epsilon \le \calO(1)$, $\norm{\Ah_t - A} \leq \epsilon$ and $\norm{\Bh_t - B} \leq \epsilon$. Then, we have
  \begin{align*}
    \norm{\Kt_{t+1} - K} \leq 7 \Gamma^3 \, \epsilon_P = \calO(\epsilon) .
  \end{align*}
  Here $\Gamma := 1 + \max\{\norm{A}, \norm{B}, \norm{P}/\min\{\smin(R), \smin(P)\}, \norm{K}\}$.

\end{corr}
\begin{proof}
With \cref{cor:prop93} we can find $\epsilon_P$ such that $\norm{\Phat_t - P} = \calO(\epsilon)$. Thus, the condition of \cref{corr:stability_perturb_tmp} is satisfied.
\end{proof}
\subsubsection{Concluding the proof of \cref{prop:one_epoch_estimate_withK}}
\begin{proof}
With \cref{prop:one_epoch_estimate} and \cref{cor:stability_perturb}, it is straightforward to give a new corollary with uniform control on all $\norm{\Ah_t - A}$, $\norm{\Bh_t - B}$, and $\norm{\Kt_{t+1} - K}$. Recall that we already proved the high probability bound in \cref{prop:one_epoch_estimate} that 
\begin{equation*}
          \max\left\{ \norm{\Ah_t - A}, \norm{\Bh_t - B}\right\} = 
    \calO(t^{-\frac{\beta}{2}} \log^{\frac{-\alpha + 1}{2}}(t)) \as
\end{equation*}
Basically, to satisfy the constraint in \cref{cor:stability_perturb}, we only need our bound (named $\epsilon$) in \cref{prop:one_epoch_estimate} to satisfy
\[\epsilon = \calO(t^{-\frac{\beta}{2}} \log^{\frac{-\alpha + 1}{2}}(t)) \le \calO(1) \as\]
which is always true when $t$ is large enough. (This also ensures $\Ah_t, \Bh_t$ to be stabilizable so that $K_0$ is only used finitely many times.)
That means,
\begin{equation*}
    \norm{\Kt_{t+1} - K} = \calO(\epsilon) = \calO(t^{-\frac{\beta}{2}} \log^{\frac{-\alpha + 1}{2}}(t)) \as
\end{equation*}
Finally, we can say
\begin{equation*}
\max\left\{ \norm{\Ah_t - A}, \norm{\Bh_t - B}, \norm{\Kt_{t+1} - K}\right\}
     = \calO(t^{-\frac{\beta}{2}} \log^{\frac{-\alpha + 1}{2}}(t)) \as
\end{equation*}

\end{proof}

\subsection{The proof of \cref{prop:one_epoch_estimate_withMyalg}}
\label{section: The proof of one_epoch_estimate_withMyalg}

\begin{prop*}
Let $x_0 \in \R^{\statedim}$ be any initial state. Assume \cref{asm:InitialStableCondition} is satisfied. When applying \cref{alg:myAlg}
\begin{equation*}
    \max\left\{ \norm{\Ah_t - A}, \norm{\Bh_t - B}, \norm{\Kh_{t+1} - K}\right\} = \calO(t^{-\frac{\beta}{2}} \log^{\frac{-\alpha + 1}{2}}(t)) \as
\end{equation*}

\end{prop*}

\begin{proof}
Going thorough the whole \cref{alg:myAlg}, 
there are two conditions that might cause the difference between $\Kh_{t}$ and $\Kt_{t}$:
\begin{enumerate}
    \item $\norm{\Kt_{t}} > C_K$, and
    \item $\norm{x_{t}} > C_{x,t} = C_x\log(t)$.
\end{enumerate}
Our objective is to show that, with probability $1$, $\Kh_{t} \neq \Kt_{t}$ will happen only finitely often.
\paragraph{The first case $\norm{\Kt_{t}} > C_K$}
The first case is when $\norm{\Kt_{t}} > C_K$, this will not happen infinitely often. The first case 
$\norm{\Kt_{t}} > C_K$ can only happen when 
\begin{equation}
    \label{eq: first Kh Kt diff condition}
    \norm{\Kt_{t} - K} \ge \norm{\Kt_{t}} - \norm{K} > C_K - \norm{K}
.\end{equation}
By \cref{prop:one_epoch_estimate_withK}, we know that $\norm{\Kt_{t} - K}$ is exponentially decaying:
\begin{equation*}
    \max\left\{\norm{\Kt_{t} - K}\right\} = \calO(t^{-\frac{\beta}{2}} \log^{\frac{-\alpha + 1}{2}}(t)) \as
\end{equation*}
As a result, \cref{eq: first Kh Kt diff condition} will hold only finitely many times, \as

\paragraph{The second case $\norm{x_{t}} > C_{x,t} = C_x\log(t)$}
To examine how often this would happen, we need to dig into more details of the decomposition of $\norm{x_t}$. Recall the previously derived formula from \cref{lemma: StateExpansion}:
\[x_{t} = \sum_{p=0}^{t-1}(A+B \Kh_{t-1})\cdots(A+B \Kh_{p+1})(B \eta_p+\varepsilon_p) +(A+B \Kh_{t-1})\cdots(A+B \Kh_0)x_0.\]

We hope to get an upper bound for $\norm{x_{t}}$. Apparently the main difficulty here is to bound the norm of $(A+B \Kh_{t-1})\cdots(A+B \Kh_{p+1})$. The following lemma serves as a key.
\begin{lemma}
\label{lem:productBound}
Suppose we have a constant square matrix $M$ with spectral radius $\rho(M) < 1$, and a sequence of uniformly bounded random variables $\{\delta_t\}_{t=0}^\infty$, satisfying $\norm{\delta_t} \asConv 0$. 
Denote the constant $\rho_M:= \frac{2 + \rho(M)}{3} < 1$. 
Then we have, for any $t, q \in \mathbb{N}$, $t > q$:
\[\norm{(M + \delta_{t-1}) \cdots (M + \delta_{q})} = \calO(\rhoM^{t-q}) \as\]
And as a direct corollary
\[\norm{M^{t-q}} = \calO(\rhoM^{t-q}) .\] 
\end{lemma}
The proof can be found in \cref{The proof of lem:productBound}.

Notice that by our \cref{alg:myAlg}, $\norm{\Kh_{t}} \le C_K$ always holds, thus there exists a uniform upper bound on $\norm{B\delta_t} := \norm{B (\Kh_{t} - K)} \le \norm{B}(C_K + \norm{K})$. 
Now we can separate the whole $\norm{(A+B \Kh_{t-1})\cdots(A+B \Kh_{p+1})}$ into two parts. If we denote $\rho_0 := \max(\frac{2+\rho(A+B K_{0})}{3}, \frac{2+\rho(A+BK)}{3})$, then with \cref{lem:productBound}, we can simultaneously bound both parts.
\begin{enumerate}
    \item The first part contains the $A+B \Kh_{k}$ where $\Kh_k = K_0$, this part of product is denoted as $I_1$. In this part, $A+B \Kh_{k} = A+B K_0$. Suppose this part has $p_1$ same items, by  \cref{{lem:productBound}} we know $I_1 \le \calO(\rho_0^{p_1}) \as$
    
    \item The second part contains the $(A+B \Kh_{k})$ where $\Kh_k = \Kt_k$ to be our true certainty equivalent controller, this part of the product is denoted as $I_2$. If we denote $\delta_k := (\Kh_{k} - K)$, then, in this part, $(A+B \Kh_{k}) = (A+B K + B\delta_k)$. Remember our conclusion in \cref{prop:one_epoch_estimate_withK} that $\norm{\Kt_k - K } \asConv 0$, thus $\norm{\delta_k} \asConv 0$, assuming this part has $p_2$ items, then since $\norm{B\delta_k} \le \norm{B}(C_K + \norm{K})$, by \cref{lem:productBound}
    \[I_2 \le \calO(\rho_0^{p_2}) \as\]
    
\end{enumerate}
We know $p_1 + p_2 = t - p -1$. Combining these two parts we have
\begin{align*}
\norm{(A+B \Kh_{t-1})\cdots(A+B \Kh_{p+1})} \le \calO(\rho_0^{t-p})  \as
\end{align*}
Finally we have the bound on $x_t$:
\begin{align*}
\norm{x_t} =& \lnorm{\sum_{p=0}^{t-1}(A+B \Kh_{t-1})\cdots(A+B \Kh_{p+1})(B \eta_p+\varepsilon_p) +(A+B \Kh_{t-1})\cdots(A+B K_0)x_0}   \\
\le& \left(\sum_{p=0}^{t-1}\lnorm{(A+B \Kh_{t-1})\cdots(A+B \Kh_{p+1})}\lnorm{B \eta_p+\varepsilon_p} +\lnorm{(A+B \Kh_{t-1})\cdots(A+B K_0)}\lnorm{x_0}\right)   \\
\le& \sum_{p=0}^{t-1}\calO(\rho_0^{t-p})\lnorm{B \eta_p+\varepsilon_p} +\calO(\rho_0^{t})\lnorm{x_0}   \as \\
\end{align*}

Then
\begin{equation*}
    \norm{x_t} = \calO\left(\sum_{p=0}^{t-1}\rho_0^{t-p}\lnorm{B \eta_p+\varepsilon_p} +\rho_0^{t}\lnorm{x_0}\right) \as
\end{equation*}
By Gaussian tail bounds (see \cref{lemma: Hi prob bounds in theorem 2}), we know that 
\begin{equation*}
    \norm{B\eta_t+\varepsilon_t} = \calO(\log^{1/2}(t)) \as
\end{equation*}
Then
\begin{equation*}
    \norm{x_t} = \calO\left(\sum_{p=0}^{t-1}\rho_0^{t-p} \log^{1/2}(t)\right) +o(1)\as
\end{equation*}
Because $\rho_0^{t-p}$ is geometric sequence,
\begin{equation*}
    \norm{x_t} \le \calO(\log^{1/2}(t)) \as
\end{equation*}
Thus for almost any $\omega \in \Omega$, $\norm{x_{t}} > C_{x,t} = C_x\log(t)$ will happen only finitely many times.

Finally, because two conditions $\norm{\Kt_{t}} > C_K$ and $\norm{x_{t}} > C_{x,t} = C_x\log(t)$ will happen only finitely many times, $\Kh_t$ and $\Kt_t$ eventually are the same. Following \cref{prop:one_epoch_estimate_withK},
\begin{equation*}
\begin{split}
    \max\left\{ \norm{\Ah_t - A}, \norm{\Bh_t - B}, \norm{\Kh_{t+1} - K}\right\} = \calO(t^{-\frac{\beta}{2}} \log^{\frac{-\alpha + 1}{2}}(t)) \as
\end{split}
\end{equation*}

\end{proof}

\section{The proof of lemmas}
\subsection{Lemmas in \cref{The proof of thm:main tool}}
\subsubsection{The proof of \cref{lemma: Hi prob bounds in theorem 2}}
\label{The proof of lemma: Hi prob bounds in theorem 2}
\begin{lemma*}
~
\begin{itemize}
    \item
    \begin{equation}
    \label{eq:bound on eta_p 2}
    \norm{\varepsilon_t}, \norm{\eta_t} = \calO(\log^{1/2}(t))  \as
    \end{equation}
    
    \item 
    \begin{equation}
    \label{eq: stochastic bound on B eta p plus  varepsilon p 2}
    \norm{B\eta_t+\varepsilon_t}  = \calO(\log^{1/2}(t))  \as
    \end{equation}
        \end{itemize}
    Assume \cref{eq:uniform high probability bound for Kh}, then:
    \begin{itemize}
    \item 
    \begin{equation}
    \label{eq: stochastic bound delta_t 2}
        \norm{\delta_t} = \norm{\Kh_{t} - K} =
        \calO(t^{-\frac{\beta}{2}} \log^{\frac{-\alpha + 1}{2}}(t) ) \as
    \end{equation}
    
    \item 
    \begin{equation}
    \label{eq: stochastic bound Lhat product 2}
        \norm{(L + B\delta_{t-1}) \cdots (L + B\delta_{q})} =
        \calO(\rhoL^{t-q}) \as
    \end{equation}
    
    \item
    \begin{equation}
    \label{eq: bound norm xi by log t 2} 
        \norm{x_t}, \norm{u_t} = 
        \calO(\log^{1/2}(t)) \as
    \end{equation}
\end{itemize}
where $\delta_t := \Kh_{t} - K$, $L := A+BK$, and $\rhoL := \frac{2 + \rho(L)}{3}$. \textbf{Additionally, when $t=0, 1$ all these terms are bounded by $\calO(1) \as$}
\end{lemma*}

\begin{proof}
Outline:
~\paragraph{The proof of \cref{eq:bound on eta_p 2} and \cref{eq: stochastic bound on B eta p plus  varepsilon p 2}}
The following lemma give the proof that \cref{eq:bound on eta_p 2} and \cref{eq: stochastic bound on B eta p plus  varepsilon p 2} holds with probability at least $1-\delta$, which can be shown by the tail bound for i.i.d Gaussian random variables.

\begin{lemma}
\label{lemma: stochastic bound on B eta p plus  varepsilon p}
For the noise $\eta_t \iid \calN(0, \tau^2 t^{1-\beta}\log^\alpha(t))$ and $\varepsilon_t \iid \calN(0, \sigma^2)$, we have that for any $\delta \in (0,1)$, with probability $1-\delta$, the following two equations holds for any $t \ge 1 $:

\begin{equation*}
    \norm{\varepsilon_t}, \norm{\eta_t}, \norm{B\eta_t+\varepsilon_t} \le \calO(1) \log^{1/2}(t^2/\delta) 
.\end{equation*}
\end{lemma}
We will prove \cref{lemma: stochastic bound on B eta p plus  varepsilon p} shortly.
By \cref{defn: Big O notation} \cref{itm: big O as defn}, this implies
\begin{equation*}
    \norm{\varepsilon_t}, \norm{\eta_t} \le \calO(\log^{1/2}(t))  \as
\end{equation*}
and
\begin{equation*}
    \norm{B\eta_t+\varepsilon_t}  \le \calO(\log^{1/2}(t))  \as
\end{equation*}
~\paragraph{The proof of \cref{eq: stochastic bound delta_t 2} and \cref{eq: stochastic bound Lhat product 2}}
\cref{eq: stochastic bound delta_t 2} directly follows from \cref{eq:uniform high probability bound for Kh}. 
\cref{eq: stochastic bound Lhat product 2} follows from \cref{lem:productBound} given that we have $\delta_t \asConv 0$ from \cref{prop:one_epoch_estimate_withMyalg}:
\[ \norm{(L + B\delta_{t-1}) \cdots (L + B\delta_{q})} \le \calO(\rhoL^{t-q}) \as\]
~\paragraph{The proof of \cref{eq: bound norm xi by log t 2}}
Finally we need to prove \cref{eq: bound norm xi by log t 2} that
\[ \norm{x_t}, \norm{u_t} = \calO(\log^{1/2}(t)) \as \]
With the fact from \cref{lemma: StateExpansion} that
\begin{align*}
 x_{t} =& \sum_{p=0}^{t-1}(A+B \Kh_{t-1})\cdots(A+B \Kh_{p+1})(B \eta_p+\varepsilon_p) +(A+B \Kh_{t-1})\cdots(A+B K_0)x_0 \\
 =& \sum_{p=0}^{t-1}(L + B\delta_{t-1}) \cdots (L + B\delta_{p+1})(B \eta_p+\varepsilon_p) +(L + B\delta_{t-1}) \cdots (L + B\delta_{0})x_0 
,\end{align*}
combined with the conclusion of \cref{eq: stochastic bound Lhat product 2} and \cref{eq: stochastic bound on B eta p plus  varepsilon p 2}, we derive a norm bound on $x_t$:
\begin{align*}
 \norm{x_{t}} 
 \le& \sum_{p=0}^{t-1}\norm{(L + B\delta_{t-1}) \cdots (L + B\delta_{p+1})}\norm{B \eta_p+\varepsilon_p} + \norm{(L + B\delta_{t-1}) \cdots (L + B\delta_{0})}\norm{x_0} \as \\
 =& \sum_{p=0}^{t-1}\calO(\rhoL^{t-p})\norm{B \eta_p+\varepsilon_p} + \calO(\rhoL^{t})\norm{x_0} \as\\
  =& \sum_{p=0}^{t-1}\calO(\rhoL^{t-p})\calO(\log^{1/2}(p)) + o(1) \as \\
\le& \sum_{p=0}^{t-1}\calO(\rhoL^{t-p})\calO(\log^{1/2}(t)) + o(1) \as\\
=& \calO(\log^{1/2}(t)) \as
\end{align*}
Recall that we have already shown \cref{eq:bound on eta_p 2}:
\[\norm{\eta_t} = \calO(\log^{1/2}(t)) \as \]
That means
\begin{align*}
&\norm{u_i} \\
= &\norm{(A+B\Kh_i)x_i + \eta_i} \\
\le& (\norm{A}+\norm{B}C_K)\norm{x_i} + \norm{\eta_i} \\
= & \calO(\log^{1/2}(t)) \as
\end{align*}
\end{proof}

~\paragraph{The proof of \cref{lemma: stochastic bound on B eta p plus  varepsilon p}}
\label{The proof of lemma: stochastic bound on B eta p plus  varepsilon p}
\begin{proof}
For any Gaussian variable $X \sim \calN(0, \sigma^2)$,
\[\P(X > t\sigma) \le e^{-t^2/2}, \]
and
\[\P(X^2 > t^2\sigma^2) = 2\P(X > t\sigma) \le 2e^{-t^2/2}.\]
For any multivariate normal vector sequence $X_t \sim \calN(0, \sigma^2I_n)$,
\begin{equation*}
 \P(\norm{X_t}^2 > nt\sigma^2) = \P\left(\sum_{i=1}^n X_{t,i}^2 > nt\sigma^2\right) \le \sum_{i=1}^n \P(X_i^2 > t\sigma^2)\le 2ne^{-t/2}   
.\end{equation*}
That means for any constant $c > 0$,
\begin{equation*}
    \P(\norm{X_t}^2 >n2\log(ct^2/\delta)\sigma^2) \le 2ne^{-2\log(ct^2/\delta)/2} = \frac{2n\delta}{ct^2}.
\end{equation*}
We can sum up all choices of $t$ to get a uniform bound.
A well known equation states that $\sum_{t=1}^\infty 1/t^2 = \frac{\pi^2}6$. Then 
\begin{equation*}
    \P(\exists t \ge 1: \,\norm{X_t}^2 >2n\sigma^2\log(ct^2/\delta))
    \le \sum_{t=1}^\infty \frac{2n\delta}{ct^2}
.\end{equation*}
We can choose $c = \frac{1}{2n}\sum_{t=1}^\infty 1/t^2 = \frac{\pi^2}{6\cdot 2n}$, so that
\begin{equation*}
    \P(\exists t \ge 1: \,\norm{X_t}^2 >2n\sigma^2\log(ct^2/\delta))
    \le \delta
.\end{equation*}
That is to say, with probability at least $1-\delta$, we have for any  $t \ge 1$,
\begin{equation*}
    \norm{X_t} \le    \calO(1) \log^{1/2}(ct^2/\delta) = \calO(1) (\log(t^2/\delta) + \log(c))^{1/2}
.\end{equation*}
Since $\log(c)$ can be dominated by $\log(t^2/\delta)$, the above equation can simply be written as
\begin{equation*}
    \norm{X_t} \le    \calO(1) \log^{1/2}(t^2/\delta)
.\end{equation*}
This bound holds for $\varepsilon_t$ which has constant variance and is also true for $\eta_t$ which has shrinking variance. Thus, with probability at least $1-\delta$:
\begin{equation*}
    \norm{\varepsilon_t}, \norm{\eta_t} \le    \calO(1) 
    \log^{1/2}(t^2/\delta)
.\end{equation*}
Consider the fact that $\norm{B\eta_t+\varepsilon_t} \le \norm{B}\norm{\eta_t} + \norm{\varepsilon_t}$, which means $\norm{B\eta_t+\varepsilon_t}$ can still be bounded by:
\begin{equation*}
    \norm{B\eta_t+\varepsilon_t} \le \calO(1) \log^{1/2}(t^2/\delta)
.\end{equation*}

\end{proof}


\subsubsection{The proof of \cref{lemma: StateExpansion}}
\label{The proof of lemma: StateExpansion}
\begin{lemma*}
\begin{equation*}
    x_{t} = \sum_{p=0}^{t-1}(A+B \Kh_{t-1})\cdots(A+B \Kh_{p+1})(B \eta_p+\varepsilon_p) +(A+B \Kh_{t-1})\cdots(A+B K_0)x_0
.\end{equation*}
\begin{equation*}
u_{t} = \sum_{p=0}^{t-1}\Kh_t(A+B \Kh_{t-1})\cdots(A+B \Kh_{p+1})(B \eta_p+\varepsilon_p) +\Kh_t(A+B \Kh_{t-1})\cdots(A+B K_0)x_0 + \eta_t   
.\end{equation*}
Here when $p = t-1$, $(A+B \Kh_{t-1})\cdots(A+B \Kh_{p+1}) := I_n$.
\end{lemma*}
\begin{proof}
Consider the following relationship: 
\[u_t = \Kh_tx_t + \eta_t.\]
\begin{align*}
    x_t =& A x_{t-1} + B u_{t-1} + \varepsilon_{t-1} \\
    =& A x_{t-1} + B (\Kh_{t-1}x_{t-1} + \eta_{t-1}) + \varepsilon_{t-1} \\
    =& (A+B \Kh_{t-1})x_{t-1} + B \eta_{t-1} + \varepsilon_{t-1} 
.\end{align*}
Iteratively do this calculation to the end:
\begin{equation*}
    x_{t} = \sum_{p=0}^{t-1}(A+B \Kh_{t-1})\cdots(A+B \Kh_{p+1})(B \eta_p+\varepsilon_p) +(A+B \Kh_{t-1})\cdots(A+B K_0)x_0
.\end{equation*}
\begin{equation*}
u_{t} = \sum_{p=0}^{t-1}\Kh_t(A+B \Kh_{t-1})\cdots(A+B \Kh_{p+1})(B \eta_p+\varepsilon_p) +\Kh_t(A+B \Kh_{t-1})\cdots(A+B K_0)x_0 + \eta_t   
.\end{equation*}
\end{proof}

\subsubsection{The proof of \cref{lemma: four components xtxt}}
\label{The proof of lemma: four components xtxt}

\begin{lemma*}
Assume \cref{eq:uniform high probability bound for Kh}, then

\begin{enumerate}

\item 
\begin{align*}
    &\sum_{i=1}^{t-1}\sum_{p=0}^{i-1}\sum_{q=0}^{i-1}\left[(A+B K)^{i-p-1}\right](B \eta_p+\varepsilon_p)(B \eta_q+\varepsilon_q)^\top 
\left[(A+B K)^{i-q-1}\right]^\top  \\
& \qquad= t\sum_{p=0}^{\infty}L ^{p}(L ^{p})^\top \sigma^2 + t^\beta\frac{\tau^2}{\beta}\log^\alpha(t)(1+o_p(1))\sum_{q=0 }^{\infty}L ^{q}BB^\top [L ^{q}]^\top \\
& \qquad = t^\beta \log^\alpha(t) (C_t +o_p(1))
.\end{align*}

\item
\begin{equation*}
\begin{aligned}
 &\sum_{i=1}^{t-1}\sum_{p=0}^{i-1}\sum_{q=0}^{i-1}\left[(A+B \Kh_{i-1})\cdots(A+B \Kh_{p+1}) - (A+B K)^{i-p-1}\right]\\
 &\qquad \cdot(B \eta_p+\varepsilon_p)(B \eta_q+\varepsilon_q)^\top 
\left[(A+B K)^{i-q-1}\right]^\top  =  \calO_p(t^{1-\beta/2}\log^{\frac{-\alpha + 3}{2}}(t))
.\end{aligned}
\end{equation*}

\item
\begin{align*}
 &\sum_{i=1}^{t-1}\sum_{p=0}^{i-1}\sum_{q=0}^{i-1}\left[ (A+B K)^{i-p-1}\right](B \eta_p+\varepsilon_p)(B \eta_q+\varepsilon_q)^\top \\
 &\qquad \cdot
\left[(A+B \Kh_{t-1})\cdots(A+B \Kh_{q+1}) - (A+B K)^{i-q-1}\right]^\top  = \calO_p(t^{1-\beta/2}\log^{\frac{-\alpha + 3}{2}}(t))
.\end{align*}

\item 
\begin{align*}
    &\sum_{i=1}^{t-1}\sum_{p=0}^{i-1}\sum_{q=0}^{i-1}\left[(A+B \Kh_{t-1})\cdots(A+B \Kh_{p+1}) - (A+B K)^{i-p-1}\right] \\
 &\qquad \cdot (B \eta_p+\varepsilon_p)(B \eta_q+\varepsilon_q)^\top \left[(A+B \Kh_{t-1})\cdots(A+B \Kh_{q+1}) - (A+B K)^{i-q-1}\right]^\top  = \calO_p(t^{1-\beta/2}\log^{\frac{-\alpha + 3}{2}}(t))
.\end{align*}
\end{enumerate}
\end{lemma*}

\begin{proof}
The first step is to show the order of 2nd, 3rd and 4th part because they follow by the same method, especially the second part is just a transpose of the third part. Then we can focus on analyzing the first part, which is replacing all controllers $\Kh_t$ by optimal controller $K$.

\paragraph{Second Part}
With \cref{lemma: Hi prob bounds in theorem 2} in hand, now we are in good shape to start our proof with the second part showing
\begin{align*}
 &\sum_{i=1}^{t-1}\sum_{p=0}^{i-1}\sum_{q=0}^{i-1}\left[(A+B \Kh_{i-1})\cdots(A+B \Kh_{p+1}) - (A+B K)^{i-p-1}\right]\\
  &\qquad (B \eta_p+\varepsilon_p)(B \eta_q+\varepsilon_q)^\top 
\left[(A+B K)^{i-q-1}\right]^\top  =  \calO_p(t^{1-\beta/2}\log^{\frac{-\alpha + 3}{2}}(t))
.\end{align*}

Since we have already shown the uniform bound of $(B \eta_p+\varepsilon_p)(B \eta_q+\varepsilon_q)^\top $ in \cref{lemma: Hi prob bounds in theorem 2}, and that $ \left[(A+B K)^{i-q-1}\right]^\top$ has an exponential decay rate, the main difficulty in bounding the second part is to give a tight bound on $\left[(A+B \Kh_{i-1})\cdots(A+B \Kh_{p+1}) - (A+B K)^{i-p-1}\right]$. 


Recall the conclusion of \cref{lemma: Hi prob bounds in theorem 2}:
\begin{equation}
\label{eq:L star product bound}
\norm{(L + B\delta_{i-1}) \cdots (L + B\delta_{p+1})} = \calO(\rho_L ^{i-p}) \as
\end{equation}
Thus 
\begin{equation}
\label{eq: difference Khat product and K product}
\begin{split}
&\norm{(A+B \Kh_{i-1})\cdots(A+B \Kh_{p+1}) - (A+B K)^{i-p-1}} \\
= &\norm{(L + B\delta_{i-1}) \cdots (L + B\delta_{p+1}) - L^{i-p-1}} \\
\le & \norm{B\delta_{i-1} (L + B\delta_{i-2}) \cdots (L + B\delta_{p+1})} + 
\norm{LB\delta_{i-2} (L + B\delta_{i-3}) \cdots (L + B\delta_{p+1})} + 
\cdots 
\norm{L^{i-p-2}B\delta_{p+1}} \\
&(\text{For example, } (L + B\delta_3)(L + B\delta_2)(L + B\delta_1)- L^3 = \delta_3(L + B\delta_2)(L + B\delta_1) + L \delta_2 (L + B\delta_1) + L^2 B\delta_1 )\\
\le & \norm{B\delta_{i-1}} \norm{(L + B\delta_{i-2}) \cdots (L + B\delta_{p+1})} + 
\norm{B\delta_{i-2}} \norm{L(L + B\delta_{i-3}) \cdots (L + B\delta_{p+1})} + 
\cdots 
\norm{B\delta_{p+1}}\norm{L^{i-p-2}} \\
\le &  \calO(\rho_L ^{i-p})(\norm{\delta_{i-1}}+ \cdots + \norm{\delta_{p+1}})  \as \qquad (\text{using \cref{eq:L star product bound}})
\end{split}
\end{equation}
Now the L2 norm of the second term can be bounded as

\begin{equation}
\label{eq:second part step1}
\begin{split}
&\Bigg\|\sum_{i=1}^{t-1}\sum_{p=0}^{i-1}\sum_{q=0}^{i-1}\left[(A+B \Kh_{i-1})\cdots(A+B \Kh_{p+1}) - (A+B K)^{i-p-1}\right]\\
& \qquad \cdot (B \eta_p+\varepsilon_p)(B \eta_q+\varepsilon_q)^\top 
\left[(A+B K)^{i-q-1}\right]^\top \Bigg\|\\
& \qquad \le \sum_{i=1}^{t-1}\sum_{p=0}^{i-1}\sum_{q=0}^{i-1}
\lnorm{\left[(A+B \Kh_{i-1})\cdots(A+B \Kh_{p+1}) - (A+B K)^{i-p-1}\right]}\\
& \qquad \qquad \cdot\lnorm{(B \eta_p+\varepsilon_p)(B \eta_q+\varepsilon_q)^\top }
\lnorm{\left[(A+B K)^{i-q-1}\right]^\top }\\
& \qquad \le \sum_{i=1}^{t-1}\sum_{p=0 }^{i-1}\sum_{q=0 }^{i-1}\calO(\rho_L^{i-p})(\norm{\delta_{i-1}}+ \cdots + \norm{\delta_{p+1}})\calO(\rho_L^{i-q}) \norm{(B\eta_p+\varepsilon_p)(B\eta_q+\varepsilon_q)^\top } \as\\
& \qquad \le \sum_{i=1}^{t-1}\sum_{p=0 }^{i-1}\sum_{q=0 }^{i-1}\calO(\rho_L^{i-p})(\norm{\delta_{i-1}}+ \cdots + \norm{\delta_{p+1}})\calO(\rho_L^{i-q}) (\norm{B\eta_p+\varepsilon_p}^2+\norm{B\eta_q+\varepsilon_q}^2) \as
\end{split}
\end{equation}
%
At first glance it seems like there is no way this would generate the desired bound, because the $\norm{\delta_{i-1}}+ \cdots + \norm{\delta_{p+1}}$ term could diverge when $i$ is large. However, thanks to the exponentially decaying term $\calO(\rho_L^{i-p})$, we can avoid this by changing the order of summation:
\begin{equation}
\label{eq:part2 basic tool}
\begin{split}
\sum_{p=0 }^{i-1}\calO(\rho_L^{i-p})(\norm{\delta_{i-1}}+ \cdots + \norm{\delta_{p+1}}) 
= &\sum_{p=0 }^{i-1}\calO(\rho_L^{i-p})\sum_{j=p+1}^{i-1}\norm{\delta_j} \\
= &\sum_{p=0 }^{i-1}\sum_{j=p+1}^{i-1}\calO(\rho_L^{i-p})\norm{\delta_j}  \\
= &\sum_{j=1 }^{i-1}\sum_{p=0}^{j-1}\calO(\rho_L^{i-p})\norm{\delta_j} \qquad \text{(exchange the order of summation)} \\
= &\sum_{j=1 }^{i-1}\norm{\delta_j}\sum_{p=0}^{j-1}\calO(\rho_L^{i-p}) \\
= &\sum_{j=1 }^{i-1}\norm{\delta_j}\calO(\rho_L^{i-j}) \\
\end{split}
\end{equation}

The final form is almost the same as the beginning, except that the summation of $\delta_i$ disappears. 
Restart from \cref{eq:second part step1}, and remember to use \cref{eq:part2 basic tool}
(\textbf{Additionally, when $p=0, 1$, $\calO(\log(p))$ is meant to be $\calO(1) \as$}):
\begin{align}
\label{eq: second part final step}
&\Bigg\|\sum_{i=1}^{t-1}\sum_{p=0}^{i-1}\sum_{q=0}^{i-1}\left[(A+B \Kh_{i-1})\cdots(A+B \Kh_{p+1}) - (A+B K)^{i-p-1}\right] \nonumber\\
&\qquad \cdot(B \eta_p+\varepsilon_p)(B \eta_q+\varepsilon_q)^\top 
\left[(A+B K)^{i-q-1}\right]^\top \Bigg\| \nonumber\\
& \qquad \le  \sum_{i=1}^{t-1}\sum_{p=0 }^{i-1}\sum_{q=0 }^{i-1}\calO(\rho_L^{i-p})(\norm{\delta_{i-1}}+ \cdots + \norm{\delta_{p+1}})\calO(\rho_L^{i-q}) (\norm{B\eta_p+\varepsilon_p}^2+\norm{B\eta_q+\varepsilon_q}^2) \as \quad \text{(by \cref{lemma: Hi prob bounds in theorem 2})} \nonumber\\
& \qquad \le  \sum_{i=1}^{t-1}\sum_{p=0 }^{i-1}\sum_{q=0 }^{i-1}\calO(\rho_L^{i-p})(\norm{\delta_{i-1}}+ \cdots + \norm{\delta_{p+1}})\calO(\rho_L^{i-q}) (\calO(\log(p)) + \calO(\log(q))) \as \nonumber\\
& \qquad \le  \sum_{i=1}^{t-1}\sum_{p=0 }^{i-1}\sum_{q=0 }^{i-1}\calO(\rho_L^{i-p})(\norm{\delta_{i-1}}+ \cdots + \norm{\delta_{p+1}})\calO(\rho_L^{i-q}) \calO(\log(t)) \as \nonumber\\
& \qquad = \calO(\log(t))\sum_{i=1}^{t-1}\sum_{p=0 }^{i-1}\calO(\rho_L^{i-p})(\norm{\delta_{i-1}}+ \cdots + \norm{\delta_{p+1}}) \as \nonumber\\
& \qquad =\calO(\log(t)) \sum_{i=1}^{t-1}\sum_{j=1 }^{i-1}\norm{\delta_j}\calO(\rho_L^{i-j}) \as \nonumber\\
& \qquad = \calO(\log(t))\sum_{j=1}^{t-1}\norm{\delta_{j}}\sum_{i=j+1}^{t-1}\calO(\rho_L^{i-j}) \as \qquad \text{(by \cref{eq:part2 basic tool})} \nonumber\\
& \qquad =  \calO(\log(t))\sum_{j=1}^{t-1}\norm{\delta_{j}} \as \nonumber\\
& \qquad = \calO(\log(t))\left(\sum_{j=1}^{t-1}  \calO(j^{-\frac{\beta}{2}} \log^{\frac{-\alpha + 1}{2}}(j)) \right) \as \qquad \text{(by \cref{eq: sum eta_t})}  \nonumber\\
& \qquad = 
\calO(\log(t)) \calO(t^{1-\frac{\beta}{2}}\log^{\frac{-\alpha + 1}{2}}(t)) \as \nonumber\\
& \qquad = 
\calO(t^{1-\beta/2}\log^{\frac{-\alpha + 3}{2}}(t)) \as
\end{align}
We know that for any matrix $A$, $\norm{A} \le \norm{A}_F \le \sqrt{r} \norm{A}$, where $r$ is the rank of matrix $A$. Thus \cref{eq: second part final step} implies an upper bound on the Frobenius norm, and the Frobenius norm implies entry-wise upper bound:
\begin{align*}
 &\sum_{i=1}^{t-1}\sum_{p=0}^{i-1}\sum_{q=0}^{i-1}\left[(A+B \Kh_{i-1})\cdots(A+B \Kh_{p+1}) - (A+B K)^{i-p-1}\right]\\
 &\qquad \cdot(B \eta_p+\varepsilon_p)(B \eta_q+\varepsilon_q)^\top 
\left[(A+B K)^{i-q-1}\right]^\top  =  \calO(t^{1-\beta/2}\log^{\frac{-\alpha + 3}{2}}(t)) \as
\end{align*}

\paragraph{Third Part}
This part is the transpose of the second part, thus shares the same result with the second part.

\paragraph{Fourth Part}
We wish to show that
\begin{align*}
&\Bigg\|\sum_{i=1}^{t-1}\sum_{p=0}^{i-1}\sum_{q=0}^{i-1}\left[(A+B \Kh_{i-1})\cdots(A+B \Kh_{p+1}) - (A+B K)^{i-p-1}\right]\\
&\qquad \cdot(B \eta_p+\varepsilon_p)(B \eta_q+\varepsilon_q)^\top 
\left[(A+B \Kh_{i-1})\cdots(A+B \Kh_{q+1}) - (A+B K)^{i-q-1}\right]^\top \Bigg\|\\
& \qquad =\calO_p(t^{1-\beta/2}\log^{\frac{-\alpha + 3}{2}}(t))
.\end{align*}
By \cref{lem:productBound} we have
\[
\norm{(A+B \Kh_{i-1})\cdots(A+B \Kh_{q+1})} = \calO(\rho_L ^{i-q}) \as,\]
and 
\[
\norm{(A+B K)^{i-q-1}} = \calO(\rho_L ^{i-q})  \as\]
Thus,
\begin{align*}
    (A+B \Kh_{i-1})\cdots(A+B \Kh_{q+1}) - (A+B K)^{i-q-1} = \calO(\rho_L ^{i-q}) \as
\end{align*}
Combining this with \cref{eq: difference Khat product and K product},
\begin{align*}
&\norm{\sum_{i=1}^{t-1}\sum_{p=0}^{i-1}\sum_{q=0}^{i-1}\left[(A+B \Kh_{i-1})\cdots(A+B \Kh_{p+1}) - (A+B K)^{i-p-1}\right]\\
&\qquad \cdot (B \eta_p+\varepsilon_p)(B \eta_q+\varepsilon_q)^\top 
\left[(A+B \Kh_{i-1})\cdots(A+B \Kh_{q+1}) - (A+B K)^{i-q-1}\right]^\top }\\
& \qquad \le \sum_{i=1}^{t-1}\sum_{p=0 }^{i-1}\sum_{q=0 }^{i-1}\calO(\rho_L^{i-p})(\norm{\delta_{i-1}}+ \cdots + \norm{\delta_{p+1}}) \norm{(B\eta_p+\varepsilon_p)(B\eta_q+\varepsilon_q)^\top }\calO(\rho_L^{i-q}) \as\\
& \qquad \le \sum_{i=1}^{t-1}\sum_{p=0 }^{i-1}\sum_{q=0 }^{i-1}\calO(\rho_L^{i-p})(\norm{\delta_{i-1}}+ \cdots + \norm{\delta_{p+1}})\calO(\rho_L^{i-q}) (\norm{B\eta_p+\varepsilon_p}^2+\norm{B\eta_q+\varepsilon_q}^2) \as
,\end{align*}
which is exactly the same as the final line of \cref{eq:second part step1}. Then following the same proof procedure as in the second part we can get the same order as in the second part: $\calO(t^{1-\beta/2}\log^{\frac{-\alpha + 3}{2}}(t)) \as$
\paragraph{Summarize second, third, and fourth parts}
To sum up, all three parts are bounded by the same order $\calO(t^{1-\beta/2}\log^{\frac{-\alpha + 3}{2}}(t)) \as$

\paragraph{First Part}
It remains to show
\begin{align*}
    &\sum_{i=1}^{t-1}\sum_{p=0}^{i-1}\sum_{q=0}^{i-1}\left[(A+B K)^{i-p-1}\right](B \eta_p+\varepsilon_p)(B \eta_q+\varepsilon_q)^\top 
\left[(A+B K)^{i-q-1}\right]^\top  \\
& \qquad = t\sum_{p=0}^{\infty}L ^{p}(L ^{p})^\top \sigma^2 + t^\beta\frac{\tau^2}{\beta}\log^\alpha(t)\sum_{q=0 }^{\infty}L ^{q}BB^\top [L ^{q}]^\top (I_n+o_p(1))
.\end{align*}

Recall $L = A+BK$. We divide the left hand side into two separate parts:
\begin{itemize}
    \item The part where $p \neq q$. We will show this part is dominated by the $p = q$ part and is only of order $\calO_p(t^{1/2})$.
    \[G_t \defineas \sum_{i=1}^{t-1}\sum_{p\neq q }^{i-1}L ^{i-p-1}(B\eta_p+\varepsilon_p)(B\eta_q+\varepsilon_q)^\top [L ^{i-q-1}]^\top = \calO_p(t^{1/2}).\]
    
    \item The part where $p = q$. We will show that
    \begin{align*}
        &\sum_{i=1}^{t-1}\sum_{p = q =0}^{i-1}L ^{i-p-1}(B\eta_p+\varepsilon_p)(B\eta_q+\varepsilon_q)^\top [L ^{i-q-1}]^\top  \\
    & \qquad  = t\sum_{p=0}^{\infty}L ^{p}(L ^{p})^\top \sigma^2 + t^\beta\frac{\tau^2}{\beta}\log^\alpha(t)\sum_{q=0 }^{\infty}L ^{q}BB^\top [L ^{q}]^\top (I_n+o_p(1)).
    \end{align*}
\end{itemize}
Let us first consider the part where $p \neq q$. 
We will show the order of $G_t$ by considering its expectation and variance. Since $G_t$ is a summation of cross terms and $\E (B\eta_p+\varepsilon_p)= 0 $, $\E(G_t) = 0$. 
Now it remains to consider the variance 
\begin{equation*}
\begin{split}
\E(\norm{G_t}_F^2) =& \E(\Tr(G_t^2)) \\
=& \E \Bigg(\Tr\Bigg(\sum_{p\neq q }^{t-1}\sum_{i=p \lor q+1}^{t-1}\sum_{j=p \lor q+1}^{t-1}L ^{i-p-1}(B\eta_p+\varepsilon_p)(B\eta_q+\varepsilon_q)^\top [L ^{i-q-1}]^\top  \\
& \qquad \cdot L ^{j-p-1}(B\eta_p+\varepsilon_p)(B\eta_q+\varepsilon_q)^\top [L ^{j-q-1}]^\top \Bigg) \Bigg)\\
& +\E \Bigg(\Tr\Bigg(\sum_{p\neq q }^{t-1}\sum_{i=p \lor q+1}^{t-1}\sum_{j=p \lor q+1}^{t-1}L ^{i-p-1}(B\eta_p+\varepsilon_p)(B\eta_q+\varepsilon_q)^\top [L ^{i-q-1}]^\top  \\
&  \qquad \cdot L ^{j-q-1}(B\eta_q+\varepsilon_q)(B\eta_p+\varepsilon_p)^\top [L ^{j-p-1}]^\top  \Bigg) \Bigg)\\
&\qquad \text{(terms with odd power go away in expectation) } 
.\end{split}
\end{equation*}
It is sufficient to consider the first term in the previous expression, and the other term can be analyzed in exactly the same way. Notice the following relationship on any square matrix $A$ with dimension $n$ 
\[\mathbf{Tr}^2(A) \le n\norm{A}_F^2 \le n \cdot n\norm{A}^2.\]
That is
\begin{equation*}
    \Tr(A) \le n\norm{A}
.\end{equation*}
Then
\begin{align*}
& \E \Bigg(\Tr\Bigg(\sum_{p\neq q }^{t-1}\sum_{i=p \lor q+1}^{t-1}\sum_{j=p \lor q+1}^{t-1}L ^{i-p-1}(B\eta_p+\varepsilon_p)(B\eta_q+\varepsilon_q)^\top [L ^{i-q-1}]^\top  \\
& \qquad \cdot L ^{j-p-1}(B\eta_p+\varepsilon_p)(B\eta_q+\varepsilon_q)^\top [L ^{j-q-1}]^\top \Bigg) \Bigg) \\
& \qquad \le n\E \Bigg\|\sum_{p\neq q }^{t-1}\sum_{i=p \lor q+1}^{t-1}\sum_{j=p \lor q+1}^{t-1}L ^{i-p-1}(B\eta_p+\varepsilon_p)(B\eta_q+\varepsilon_q)^\top [L ^{i-q-1}]^\top  \\
& \qquad \qquad \cdot L ^{j-p-1}(B\eta_p+\varepsilon_p)(B\eta_q+\varepsilon_q)^\top [L ^{j-q-1}]^\top  \Bigg\|\\
& \qquad \le \E\sum_{p\neq q }^{t-1}\sum_{i=p \lor q+1}^{t-1}\sum_{j=p \lor q+1}^{t-1} \calO(\rho_L^{i-p}\rho_L^{i-q})\norm{B\eta_p+\varepsilon_p}_2^2\norm{B\eta_q+\varepsilon_q}_2^2O(\rho_L^{j-p}\rho_L^{j-q}) 
 \qquad \text{(by \cref{lem:productBound})} \\
& \qquad = \calO\left( \sum_{p\neq q }^{t-1}\sum_{i=p \lor q+1}^{t-1}\sum_{j=p \lor q+1}^{t-1}\rho_L^{2i-p-q}\rho_L^{2j-p-q} \right)  \\
& \qquad = \calO\left(\sum_{p > q}^{t-1}\rho_L^{2(p-q)} \right) \quad \text{(WLOG consider the part where } p>q \text{)} \\
& \qquad = \calO\left(\sum_{q=0}^{t-1}1 \right) \\
& \qquad = \calO(t)
.\end{align*}
Thus the entry-wise standard error of $G_t$ is of order $\calO(t^{1/2})$. Combining this with the fact that $\E G_t = 0$, we have
\begin{equation}
    \label{eq: Gt order}
    G_t \defineas \sum_{i=1}^{t-1}\sum_{p\neq q }^{i-1}L ^{i-p-1}(B\eta_p+\varepsilon_p)(B\eta_q+\varepsilon_q)^\top [L ^{i-q-1}]^\top = \calO_p(t^{1/2}).
\end{equation}

and it remains to consider
\[R \defineas \sum_{i=1}^{t-1}\sum_{p=0 }^{i-1}L ^{i-p-1}(B\eta_p+\varepsilon_p)(B\eta_p+\varepsilon_p)^\top [L ^{i-p-1}]^\top .\]
Consider the expectation of $R$: $\E(R) = R_0$, where
\[R_0 \defineas \sum_{i=0}^{t-1}\sum_{p=0 }^{i-1}L ^{i-p-1}(p^{\beta-1}\log^\alpha(p)BB^\top \tau^2 + I_n\sigma^2)[L ^{i-p-1}]^\top 
.\]

Let us first show $R - R_0 = \calO_p(t ^{1/2})$, and after that we only need to consider $R_0$, which is the dominating term. We know that $B\eta_p+\varepsilon_p$ has a finite fourth moment, so the sum of the variances of each element of $R-R_0$ can be written as
\begin{align*}
\E \norm{R-R_0}_F^2
=&\E(\Tr((R-R_0)^2)) \\
\le& \E(\Tr (\sum_{p=0 }^{t-1}\sum_{i=p+1}^{t-1}\sum_{j=p +1}^{t-1}L ^{i-p-1}[(B\eta_p+\varepsilon_p)(B\eta_p+\varepsilon_p)^\top -(p^{\beta-1}\log^\alpha(p)BB^\top \tau^2 + I_m\sigma^2)]\\
&\qquad \cdot [L ^{i-p-1}]^\top  L ^{j-p-1}[(B\eta_p+\varepsilon_p)(B\eta_p+\varepsilon_p)^\top -(p^{\beta-1}\log^\alpha(p)BB^\top \tau^2 + I_m\sigma^2)][L ^{j-p-1}]^\top )) \\
\le& n\E \norm{\sum_{p=0 }^{t-1}\sum_{i=p+1}^{t-1}\sum_{j=p +1}^{t-1}L ^{i-p-1}[(B\eta_p+\varepsilon_p)(B\eta_p+\varepsilon_p)^\top -(p^{\beta-1}\log^\alpha(p)BB^\top \tau^2 + I_m\sigma^2)]\\
&\qquad \cdot [L ^{i-p-1}]^\top  L ^{j-p-1}[(B\eta_p+\varepsilon_p)(B\eta_p+\varepsilon_p)^\top -(p^{\beta-1}\log^\alpha(p)BB^\top \tau^2 + I_m\sigma^2)][L ^{j-p-1}]^\top  }\\
\le& \calO(\E\sum_{p=0 }^{t-1}\sum_{i=p +1}^{t-1}\sum_{j=p +1}^{t-1} \rho_L^{2i-2p}\norm{(B\eta_p+\varepsilon_p)(B\eta_p+\varepsilon_p)^\top -(p^{\beta-1}\log^\alpha(p)BB^\top \tau^2 + I_m\sigma^2)}^2\rho_L^{2j-2p}  )\\
=& \calO( \sum_{p=0}^{t-1}\E\norm{(B\eta_p+\varepsilon_p)(B\eta_p+\varepsilon_p)^\top -(p^{\beta-1}\log^\alpha(p)BB^\top \tau^2 + I_m\sigma^2)}^2 )  \\
=& \calO(t) 
.\end{align*}
Thus $R - R_0 = \calO_p(t ^{1/2})$. Now we only need to focus on:
\[R_0 = \sum_{i=1}^{t-1}\sum_{p=0 }^{i-1}L ^{i-p-1}(p^{\beta-1}\log^\alpha(p)BB^\top \tau^2 + I_m\sigma^2)[L ^{i-p-1}]^\top .\]
Again, when $p=0,1$, $p^{\beta-1}\log^\alpha(p)$ should be considered as 1.
Let us start from the identity matrix part $\sum_{i=1}^{t-1}\sum_{p=0 }^{i-1}L ^{i-p-1}I_m\sigma^2[L ^{i-p-1}]^\top $.
\begin{equation*}
\begin{split}
\sum_{i=1}^{t-1}\sum_{p=0 }^{i-1}L ^{i-p-1}[L ^{i-p-1}]^\top = &\sum_{i=1}^{t-1}\sum_{q=0 }^{i-1}L ^{q}[L ^{q}]^\top  \\
= & \sum_{i=1}^{t-1}(\sum_{p=0 }^{\infty}L ^{p}(L ^{p})^\top  - \sum_{q=i}^{\infty}L ^{q}[L ^{q}]^\top ) \\
=& t\sum_{p=0}^{\infty}L ^{p}(L ^{p})^\top  - \sum_{i=1}^{t-1}\sum_{q=i}^{\infty}L ^{q}[L ^{q}]^\top 
.\end{split}
\end{equation*}
Notice
\begin{equation*}
\begin{split}
\norm{\sum_{i=1}^{t-1}\sum_{q=i}^{\infty}L ^{q}[L ^{q}]^\top } \le & \sum_{i=1}^{t-1}\sum_{q=i}^{\infty}\calO(\rho_L^{2q}) \\
= & \sum_{i=1}^{t-1}\calO(\rho_L^{2i}) \\
= & \calO(1)
.\end{split}
\end{equation*}
Thus
\begin{equation*}
    \sum_{i=1}^{t-1}\sum_{p=0 }^{i-1}L ^{i-p-1}[L ^{i-p-1}]^\top  = t\sum_{p=0}^{\infty}L ^{p}(L ^{p})^\top  + \calO(1)
.\end{equation*}
On the other hand  (\textbf{when $p=0, 1$, $p^{\beta-1}\log^\alpha(p)$ is meant to be $1$}),
\begin{equation}
\label{eq: noname_1}
\begin{split}
& \sum_{i=1}^{t-1}\sum_{p=0 }^{i-1}L ^{i-p-1}p^{\beta-1}\log^\alpha(p)BB^\top [L ^{i-p-1}]^\top  \\
& \qquad = \sum_{p=0}^{t-2}\sum_{i=p+1}^{t-1}L ^{i-p-1}p^{\beta-1}\log^\alpha(p)BB^\top [L ^{i-p-1}]^\top  \\
& \qquad = \sum_{p=0}^{t-2}p^{\beta-1}\log^\alpha(p)\sum_{q=0 }^{t-p-2}L ^{q}BB^\top [L ^{q}]^\top  \\
& \qquad = \sum_{p=0}^{t-2}p^{\beta-1}\log^\alpha(p)\left(\sum_{q=0 }^{\infty}L ^{q}BB^\top [L ^{q}]^\top  - \sum_{q=t-p-1 }^{\infty}L ^{q}BB^\top [L ^{q}]^\top \right) \\
& \qquad = \sum_{p=0}^{t-2}p^{\beta-1}\log^\alpha(p)\left(\sum_{q=0 }^{\infty}L ^{q}BB^\top [L ^{q}]^\top  - \calO(\rhoL ^{2(t-p-1)})\right) \\
& \qquad = \sum_{p=0}^{t-2}p^{\beta-1}\log^\alpha(p)\sum_{q=0 }^{\infty}L ^{q}BB^\top [L ^{q}]^\top  + \sum_{p=0}^{t-2}p^{\beta-1}\log^\alpha(p)\calO\left(\rhoL ^{2(t-p-1)}\right)\\
& \qquad \le \sum_{p=0}^{t-2}p^{\beta-1}\log^\alpha(p)\sum_{q=0 }^{\infty}L ^{q}BB^\top [L ^{q}]^\top  + \sum_{p=0}^{t-2}\calO(1)\calO\left(\rhoL ^{2(t-p-1)}\right)\\
& \qquad = \sum_{p=0}^{t-2}p^{\beta-1}\log^\alpha(p)\sum_{q=0 }^{\infty}L ^{q}BB^\top [L ^{q}]^\top  + \calO(1)
.\end{split}
\end{equation}
Now it remains to calculate $\sum_{p=0}^{t-2}p^{\beta-1}\log^\alpha(p)$.
Let us consider a more general case $\sum_{p=0}^{t}p^{\gamma}\log^\alpha(p)$ where $\gamma > -1$ and $\alpha$ is any real number. It is clear that this summation goes to infinity when $t \to \infty$. 
Recall the Stolz--Cesàro theorem: 
\begin{theorem}[Stolz--Cesàro]
\label{thm: Stolz-Cesaro theorem}
Let $\{a_t\}_{t\ge 1}$ and $\{b_t\}_{t\ge1}$ be two sequences of real numbers. Assume that $\{b_t\}_{t\ge1}$ is a strictly monotone and divergent sequence and the following limit exists:
\begin{equation*}
    \lim_{t \to \infty} \frac{a_{t+1}-a_t}{b_{t+1}-b_t} = l
\end{equation*}
Then, the limit
\begin{equation*}
    \lim_{t \to \infty} \frac{a_t}{b_t} = l
\end{equation*}
\end{theorem}
In \cref{thm: Stolz-Cesaro theorem}, we choose $a_t$ and $b_t$ to be $\sum_{p=0}^{t}p^{\gamma}\log^\alpha(p) $ and $t^{\gamma+1}\log^\alpha(t)$, respectively.
\begin{align*}
    \lim_{t\to\infty}\frac{a_{t} - a_{t-1}}{b_t - b_{t-1}} =& \lim_{t\to\infty}\frac{t^{\gamma}\log^\alpha(t)}{t^{\gamma+1}\log^\alpha(t) - (t-1)^{\gamma+1}\log^\alpha(t-1)} \\
    =& \lim_{t\to\infty}\frac{1}{t - (\frac{t-1}t)^{\gamma}(t-1) \left(\frac{\log(t-1)}{\log(t)}\right)^\alpha} \\
    =& \lim_{t\to\infty}\frac{1/t}{1 - (1-\frac{1}t)^{\gamma+1} \left(1+\frac{\log(t-1)-\log(t)}{\log(t)}\right)^\alpha}  \\
    =& \lim_{t\to\infty}\frac{1/t}{1 - (1 - \frac{\gamma+1}t + o(\frac1t)) \left(1+\frac{-\frac1t + o(\frac1t)}{\log(t)}\right)^\alpha} \\
    =& \lim_{t\to\infty}\frac{1/t}{1 - (1 - \frac{\gamma+1}t + o(\frac1t)) \left(1+o(\frac1t)\right)^\alpha} \\
    =& \lim_{t\to\infty}\frac{1/t}{1 - (1 - \frac{\gamma+1}t + o(\frac1t)) e^{\alpha\log\left(1+o(\frac1t)\right)}} \\
    =& \lim_{t\to\infty}\frac{1/t}{1 - (1 - \frac{\gamma+1}t + o(\frac1t)) e^{\alpha o(\frac1t)}} \\
    =& \lim_{t\to\infty}\frac{1/t}{1 - (1 - \frac{\gamma+1}t + o(\frac1t)) \left(1+o(\frac\alpha{t})\right)} \\
    =& \lim_{t\to\infty}\frac{1/t}{\frac{\gamma+1}t + o(\frac1t) } \\
    =& \frac{1}{\gamma+1} 
.\end{align*}
By \cref{thm: Stolz-Cesaro theorem}, we know
\begin{align*}
    \lim_{t\to\infty}\frac{a_{t} }{b_t } = \lim_{t\to\infty}\frac{\sum_{p=0}^{t}p^{\gamma}\log^\alpha(p)}{t^{\gamma+1}\log^\alpha(t)} = \frac{1}{\gamma+1} 
\end{align*}
That is to say, for any $\gamma > -1$:
\begin{align}
\label{eq: sum eta_t}
\begin{split}
        \sum_{p=0}^{t}p^{\gamma}\log^\alpha(p) 
    &  = \frac{1}{\gamma+1} t^{\gamma+1}\log^\alpha(t) (1+o(1))
.\end{split}
\end{align}

Following \cref{eq: noname_1,eq: sum eta_t},
\begin{equation*}
\begin{split}
    &\sum_{i=1}^{t-1}\sum_{p=0 }^{i-1}L ^{i-p-1}p^{\beta-1}\log^\alpha(p)BB^\top [L ^{i-p-1}]^\top \\
    =& \sum_{p=0}^{t}p^{\beta-1}\log^\alpha(p)\sum_{q=0 }^{\infty}L ^{q}BB^\top [L ^{q}]^\top  + \calO(1) \\
    = & \frac{t^\beta}\beta \log^\alpha(t)(1+o(1))\sum_{q=0 }^{\infty}L ^{q}BB^\top [L ^{q}]^\top  + \calO(1) \\
    = & \frac{t^\beta}\beta \log^\alpha(t)\sum_{q=0 }^{\infty}L ^{q}BB^\top [L ^{q}]^\top (I_n+o(1))
.\end{split}   
\end{equation*}
To sum up,
\[R_0 = t\sum_{p=0}^{\infty}L ^{p}(L ^{p})^\top \sigma^2 + \frac{t^\beta}\beta \log^\alpha(t)\sum_{q=0 }^{\infty}L ^{q}BB^\top [L ^{q}]^\top (I_n+o(1)) .\]
Recall that $R - R_0 = \calO_p(t^{1/2})$, so
\begin{align*}
    R &= \sum_{i=1}^{t-1}\sum_{p=0 }^{i-1}L ^{i-p-1}(B\eta_p+\varepsilon_p)(B\eta_p+\varepsilon_p)^\top [L ^{i-p-1}]^\top \\
 &= 
t\sum_{p=0}^{\infty}L ^{p}(L ^{p})^\top \sigma^2 + t^\beta\frac{\tau^2}{\beta}\log^\alpha(t)\sum_{q=0 }^{\infty}L ^{q}BB^\top [L ^{q}]^\top (I_n+o_p(1)).
\end{align*}
Recall \cref{eq: Gt order}:
\[\sum_{i=1}^{t-1}\sum_{p\neq q }^{i-1}L ^{i-p-1}(B\eta_p+\varepsilon_p)(B\eta_q+\varepsilon_q)^\top [L ^{i-q-1}]^\top = \calO_p(t^{1/2}).\]
Finally we proved the order of the first part:
\begin{align*}
    &\sum_{i=1}^{t-1}\sum_{p=0}^{i-1}\sum_{q=0}^{i-1}\left[(A+B K)^{i-p-1}\right](B \eta_p+\varepsilon_p)(B \eta_q+\varepsilon_q)^\top 
\left[(A+B K)^{i-q-1}\right]^\top  \\
& \qquad = \sum_{i=1}^{t-1}\sum_{p=0 }^{i-1}\sum_{q=0 }^{i-1}L ^{i-p-1}(B\eta_p+\varepsilon_p)(B\eta_q+\varepsilon_q)^\top [L ^{i-q-1}]^\top \\
& \qquad= t\sum_{p=0}^{\infty}L ^{p}(L ^{p})^\top \sigma^2 + t^\beta\frac{\tau^2}{\beta}\log^\alpha(t)\sum_{q=0 }^{\infty}L ^{q}BB^\top [L ^{q}]^\top (I_n+o_p(1)) \\
& \qquad= t^\beta \log^\alpha(t) (C_t +o_p(1)) \qquad \text{(by $C_t$ definition \cref{eq: Ct definition})} 
.\end{align*}

\end{proof}


\subsubsection{The proof of \cref{lemma: Term with starting point xtxt}}
\label{The proof of lemma: Term with starting point xtxt}
\begin{lemma*}
Assume \cref{eq:uniform high probability bound for Kh}, then
\begin{enumerate}
    \item $\sum_{i=0}^{t-1}\left[ (A+B \Kh_{i-1})\cdots(A+B K_{0})x_0\right]
    \left[\sum_{q=0}^{i-1}(A+B \Kh_{i-1})\cdots(A+B \Kh_{q+1})(B\eta_q+\varepsilon_q)\right]^T = \logO(1) \as$
    
    \item $\sum_{i=0}^{t-1}\left[ (A+B \Kh_{i-1})\cdots(A+B K_{0})x_0\right]
    \left[ (A+B \Kh_{i-1})\cdots(A+B K_{0})x_0\right]^T = \calO(1) \as$
\end{enumerate}
    
\end{lemma*}

\begin{proof}
This can be proved using a similar technique as in \cref{The proof of lemma: four components xtxt}. \textbf{Recall that when $q=0, 1$, $\log^\alpha(q)$ is taken to be $1$.}
\begin{align*}
&\lnorm{\sum_{i=1}^{t-1}\sum_{q=0}^{i-1} (A+B \Kh_{i-1})\cdots(A+B K_{0})x_0(B\eta_q+\varepsilon_q)^\top 
    \left[(A+B \Kh_{i-1})\cdots(A+B \Kh_{q+1})\right]^\top} \\
& \qquad \le \sum_{i=1}^{t-1}\sum_{q=0}^{i-1} \norm{(L + B\delta_{t-1}) \cdots (L + B\delta_{0})}\norm{x_0}\norm{B\eta_q+\varepsilon_q}
    \norm{(L + B\delta_{t-1}) \cdots (L + B\delta_{q+1})}^\top \\
& \qquad \le  \sum_{i=1}^{t-1}\sum_{q=0}^{i-1} \calO(\rho_L^i)\norm{x_0}\norm{B\eta_q+\varepsilon_q} \calO(\rho_L^{i-q}) \as \qquad \text{(by \cref{lemma: Hi prob bounds in theorem 2})} \\
& \qquad \le  \sum_{i=1}^{t-1}\sum_{q=0}^{i-1} \calO(\rho_L^{2i-q})\calO(1)\calO(\log^{1/2}(q)) \as \quad \text{(by \cref{lemma: Hi prob bounds in theorem 2})}\\
& \qquad \le  \sum_{i=1}^{t-1}\sum_{q=0}^{i-1} \calO(\rho_L^{2i-q})\logO(1)   \as \\
& \qquad =  \sum_{i=1}^{t-1} \calO(\rho_L^{i})\logO(1)   \as \\
& \qquad \le  \logO(1)  \as
\end{align*}

Also,
\begin{align*}
&\lnorm{\sum_{i=1}^{t-1}\left[ (A+B \Kh_{i-1})\cdots(A+B K_{0})x_0\right]
    \left[ (A+B \Kh_{i-1})\cdots(A+B K_{0})x_0\right]^T} \\
& \qquad \le  \sum_{i=1}^{t-1} \calO(\rho_L^i)\norm{x_0}^2  \calO(\rho_L^{i}) \as \qquad \text{(by \cref{lemma: Hi prob bounds in theorem 2})}\\
& \qquad \le  \sum_{i=1}^{t-1} \calO(\rho_L^{2i})  \as \\
& \qquad \le  \calO(1)   \as
\end{align*}

\end{proof}

\subsubsection{The proof of \cref{lemma:AtBtConvergeIp}}
\label{The proof of lemma:AtBtConvergeIp}
\begin{lemma*}
Assume we have two matrix sequences $\{A_t\}_{t=1}^\infty$ and $\{B_t\}_{t=1}^\infty$, where $A_t$ and $B_t$ are $p \times p$ positive definite matrices, and
\[A_t^{2}B_t^2 \convP I_p.\]
Then
\[A_tB_t \convP I_p.\]
\end{lemma*}
\begin{proof}
The basic idea is to utilize the equivalence of entry-wise convergence and F-norm convergence and the fact that the F-norm is invariant under orthogonal transformation. We know that positive definite matrices can be diagonalized by orthogonal transformation, and these diagonal matrices are easier to deal with.
Starting from our only equation 
\[A_t^{2}B_t^2 \convP I_p.\]
Entry-wise convergence implies F-norm convergence:
\[\norm{A_t^{2}B_t^2 - I_p}_F \convP 0.\]
By the positive definiteness of $A_t$ and $B_t$, we can assume they have the diagnolization $A_t = U_{At}\Lambda_{At}U_{At}^\top$ and $B_t = U_{Bt}\Lambda_{Bt}U_{Bt}^\top$, where $\Lambda_{At}$ and $\Lambda_{Bt}$ are diagonal matrices with diagonal values $\lambda_{Ai, t}$ and $\lambda_{Bi, t}$ ($i=1,2, \cdots, p$), and $U_{At}$ and $U_{Bt}$ are orthogonal matrices. With this transformation, we have
\[\norm{U_{At}\Lambda_{At}^2U_{At}^\top U_{Bt}\Lambda_{Bt}^2U_{Bt}^\top - I_p}_F \convP 0.\]
Since orthogonal transformation does not affect F-norm, on RHS inside the F-norm, we can multiply $U_{At}^\top$ on the left and $U_{Bt}$ and on the right and get
\[\norm{\Lambda_{At}^2U_{At}^\top U_{Bt}\Lambda_{Bt}^2 - U_{At}^\top U_{Bt}}_F \convP 0.\]
Because F-norm convergence to zero is equivalent to entry-wise convergence to zero,

\[\Lambda_{At}^2U_{At}^\top U_{Bt}\Lambda_{Bt}^2 - U_{At}^\top U_{Bt} \convP 0.\]

Denote $T_t := U_{At}^\top U_{Bt}$, then

\[\Lambda_{At}^2T_t\Lambda_{Bt}^2 - T_t\convP 0.\]

If we consider the $ij$th element of the above equation:

\[\lambda_{Ai, t}^2 T_{ij} \lambda_{Bj, t}^2 - T_{ij} \convP 0,\]

which is
\[(\lambda_{Ai, t}\lambda_{Bj, t}-1)(\lambda_{Ai, t}\lambda_{Bj, t}+1) T_{ij}  \convP 0.\]
Since by positive definiteness we have $\lambda_{Ai, t}, \lambda_{Bj, t} > 0$ , the above equation implies
\[(\lambda_{Ai, t}\lambda_{Bj, t}-1)T_{ij}  \convP 0.\]
This holds for every $i, j$ pair. If we write out this equation back to matrix form, we would get
\[\Lambda_{At}T_t\Lambda_{Bt} - T_t\convP 0.\]
By the same trick this is equivalent to the F-norm form
\[\norm{\Lambda_{At}T_t\Lambda_{Bt} - T_t}_F \convP 0,\]
\[\norm{\Lambda_{At} U_{At}^\top U_{Bt} \Lambda_{Bt} - U_{At}^\top U_{Bt} }_F \convP 0.\]
On RHS inside the F-norm, we can multiply $U_{At}$ on the left and $U_{Bt}^\top$ and on the right and get
\[\norm{U_{At}\Lambda_{At} U_{At}^\top U_{Bt} \Lambda_{Bt} U_{Bt}^T - I_p }_F \convP 0.\]
Plug in our definition $A_t = U_{At}\Lambda_{At}U_{At}^\top$ and $B_t = U_{Bt}\Lambda_{Bt}U_{Bt}^\top$:
\[\norm{A_tB_t - I_p }_F \convP 0.\]
And this implies 
\[A_tB_t \convP I_p.\]
\end{proof}

\subsubsection{The proof of \cref{lemma: three parts u_tx_t}}
\label{The proof of lemma: three parts u_tx_t}
\begin{lemma*}
Assume \cref{eq:uniform high probability bound for Kh}, then
\begin{enumerate}
    \item $
    \sum_{i=0}^{t-1}(\Kh_i-K )x_ix_i^\top =  
    \calO(t^{1-\beta/2}\log^{\frac{-\alpha + 3}{2}}(t)) \as
    $
    \item $ \sum_{i=0}^{t-1}\eta_ix_i^\top = 
    o\left(t^{\beta/2}\log^{\frac{\alpha+3}{2}}(t) \right) \as $
\end{enumerate}
\end{lemma*}

\begin{proof}
~\paragraph{First part $\sum_{i=0}^{t-1}(\Kh_i-K )x_ix_i^\top$}
By \cref{lemma: Hi prob bounds in theorem 2} we have a uniform bound for $\delta_i = \Kh_i-K$ and $x_i$. We can derive the result in the first part by directly plugging in the bound for $\norm{\delta_i} $ and $\norm{x_i}$.

By \cref{lemma: Hi prob bounds in theorem 2} 
\begin{equation*}
\begin{aligned}
&\norm{x_i} \le \calO(\log^{1/2}(t))  \as \\
\end{aligned}    
\end{equation*}

Thus
\begin{equation*}
\begin{split}
\lnorm{\sum_{i=0}^{t-1}(\Kh_i-K )x_ix_i^\top }
=& \sum_{i=0}^{t-1}\norm{\delta_i}\norm{x_ix_i^\top } \\
\le& \calO(\log(t))\sum_{i=0}^{t-1}\norm{\delta_i} \as \qquad \text{(by \cref{lemma: Hi prob bounds in theorem 2}}  )\\
\le& \calO(\log(t))\sum_{i=0}^{t-1}\calO(i^{-\beta/2}\log^{\frac{-\alpha+1}{2}}(i)) \as \qquad \text{(by \cref{lemma: Hi prob bounds in theorem 2}}  )\\
\le&  \calO(\log(t) t^{1-\beta/2}\log^{\frac{-\alpha + 1}{2}}(t)) \as
\qquad \text{(by \cref{eq: sum eta_t}}  )\\
\le&  \calO(t^{1-\beta/2}\log^{\frac{-\alpha + 3}{2}}(t)) \as
\end{split}
\end{equation*}
which means (by bounding entry-wise terms by the operator norm)
\[\sum_{i=0}^{t-1}(\Kh_i-K )x_ix_i^\top  =\calO(t^{1-\beta/2}\log^{\frac{-\alpha + 3}{2}}(t)) \as\]



~\paragraph{Second Part $\sum_{i=0}^{t-1}\eta_ix_i^\top$}

Following Lemma 2 (iii) from \citet{lai1982least}:
\begin{lemma}
\label{lem: Lai and Wei martingale}
Let $\{\epsilon_n\}$ be a martingale difference sequence with respect to an increasing sequence of $\sigma$-fields $\{\calF_n\}$ such that $\sup_n \E(\varepsilon_n^2|\calF_{n-1}) < \infty$ a.s. Let $v_n$ be an $\calF_{n-1}$-measurable random variable for every $n$. Then 
\begin{equation*}
    \sum_{i=1}^n v_i\epsilon_i < \infty \text{ a.s. on } 
    \{\sum_{i=1}^\infty v_i^2 < \infty\}
.\end{equation*}
And for any $\eta > 1/2$
\begin{equation*}
    \sum_{i=1}^n v_i\epsilon_i = 
    o\left(
    (\sum_{i=1}^n v_i^2)^{1/2} \log^\eta(\sum_{i=1}^n v_i^2)
    \right) 
    \text{ a.s. on } 
    \{\sum_{i=1}^\infty v_i^2 = \infty\}
.\end{equation*}
\end{lemma}
As a result, with probability 1
\begin{equation}
    \label{eq: Lai and Wei martingale}
\begin{split}
    \sum_{i=1}^n v_i\epsilon_i 
    =& 
    o\left(
    (\sum_{i=1}^n v_i^2)^{1/2} \log(\sum_{i=1}^n v_i^2)
    \right)1_{\sum_{i=1}^\infty v_i^2 = \infty} + \calO(1)1_{\sum_{i=1}^\infty v_i^2 < \infty} \as \\
    =& 
    o\left(
    (\sum_{i=1}^n v_i^2)^{1/2} \log(\sum_{i=1}^n v_i^2)
    \right) + \calO(1) \as 
\end{split}
\end{equation}
We can apply \cref{lem: Lai and Wei martingale} to our context by noticing
\begin{equation*}
    \sum_{i=0}^{t-1}\eta_ix_i^\top  
    = \sum_{i=0}^{t-1}\eta_ii^{\frac{1-\beta}2}\log^{-\alpha/2}(i)(i^{\frac{\beta-1}2}\log^{\alpha/2}(i)x_i^\top)
.\end{equation*}
Here we normalized all $\eta_i$ to have a fixed normal distribution $\eta_ii^{\frac{1-\beta}2}\log^{-\alpha/2}(i) \sim \calN(0, \tau^2I_d)$. Apply \cref{eq: Lai and Wei martingale} entry-wise, where $v_i$ corresponds to a fixed entry of $i^{\frac{\beta-1}2}\log^{\alpha/2}(i)x_i^\top$ and $\epsilon_i$ corresponds to a fixed entry of $\eta_ii^{\frac{1-\beta}2}\log^{-\alpha/2}(i)$. $v_i$ is bounded by $i^{\frac{\beta-1}2}\log^{\alpha/2}(i)\norm{x_i}$. Thus
\begin{equation*}
    \sum_{i=0}^{t-1}\eta_ix_i^\top  = o\left(V_t^{1/2}\log(V_t)\right) + \calO(1) \as,
\end{equation*}
where $V_t := \sum_{i=0}^{t-1} (i^{\frac{\beta-1}2}\log^{\alpha/2}(i)\norm{x_i})^2$. Applying the bounds in \cref{lemma: Hi prob bounds in theorem 2} (\textbf{recall that when $i=0, 1$, $i^{\beta-1}\log^\alpha(i)$ is taken to be $1$}):
\begin{equation*}
\begin{split}
    V_t =& \sum_{i=0}^{t-1} (i^{\frac{\beta-1}2}\log^{\alpha/2}(i)\norm{x_i})^2 \\
    =&\sum_{i=0}^{t-1} 
    i^{-1+\beta}\log^{\alpha}(i)
    \calO(\log(t)) \as \quad \text{(by \cref{lemma: Hi prob bounds in theorem 2}}  )\\
    =& \calO(t^{\beta}\log^{\alpha}(t))
    \calO(\log(t)) \as \quad \text{(by \cref{eq: sum eta_t}}  )\\
    =& \calO(t^{\beta}\log^{\alpha+1}(t)) \as
\end{split}
\end{equation*}
Thus,
\begin{align*}
    \sum_{i=0}^{t-1}\eta_ix_i^\top =& o\left(V_t^{1/2}\log(V_t)\right) + \calO(1) \\
    =& o\left(\calO(t^{\beta}\log^{\alpha+1}(t))^{1/2}\log(\calO(t^{\beta}\log^{\alpha+1}(t)))\right) + \calO(1) \\
    =& o\left(\calO(t^{\beta/2}\log^{\frac{\alpha+1}{2}}(t)\log(t))\right) + \calO(1) \\
    =& o\left(t^{\beta/2}\log^{\frac{\alpha+3}{2}}(t) \right) \as
\end{align*}

\end{proof}

In exactly the same way, we can show that 
\begin{equation}
\label{eq: (Ktxt)T R etaT}
    \sum_{i=1}^{t} (\Kh_i x_i)^\top  R \eta_i = o\left(t^{\beta/2}\log^{\frac{\alpha+3}{2}}(t) \right) \as
\end{equation}
We first standardize $\eta_i$
\begin{equation*}
    \sum_{i=1}^{t} (\Kh_i x_i)^\top  R \eta_i =
    \sum_{i=0}^{t-1}
    (i^{\frac{\beta-1}2}\log^{\alpha/2}(i)(\Kh_i x_i)^\top  R)
    \eta_ii^{\frac{1-\beta}2}\log^{-\alpha/2}(i)
,\end{equation*}
and then $v_i$ is bounded by 
\begin{align*}
    &i^{\frac{\beta-1}2}\log^{\alpha/2}(i)\lnorm{(\Kh_i x_i)^\top  R} \\
    & \qquad \le i^{\frac{\beta-1}2}\log^{\alpha/2}(i)\lnorm{\Kh_i}\norm{R} \norm{x_i} \\
    & \qquad \le  i^{\frac{\beta-1}2}\log^{\alpha/2}(i)C_K\norm{R} \norm{x_i} \quad \text{(by \cref{alg:myAlg}'s design)}
,\end{align*}
which is different from $v_i$ in $\sum_{i=0}^{t-1}\eta_ix_i^\top$ by a constant factor $C_K\norm{R}$. The rest of the proof is all the same.

\subsubsection{The proof of \cref{lemma: six parts u_tu_t}}
\label{The proof of lemma: six parts u_tu_t}

\begin{lemma*}
Assume \cref{eq:uniform high probability bound for Kh}, then
\begin{enumerate}
    \item $\sum_{i=0}^{t-1} \delta_ix_ix_i^\top \delta_i^\top = \calO(t^{1-\beta}\log^{-\alpha+2}(t)) \as$
    \item $\sum_{i=0}^{t-1}\delta_ix_i\eta_i^\top = (\sum_{i=0}^{t-1} \eta_ix_i^\top \delta_i^\top)^\top =  
    o\left(\log^{2}(t)\right) \as$
    \item $\sum_{i=0}^{t-1}\eta_i\eta_i^\top  = t^\beta\frac{\tau^2}{\beta}\log^\alpha(t)(I_d + o_p(1)) $
\end{enumerate}
\end{lemma*}

\begin{proof}
    
    ~\paragraph{First part $\sum_{i=0}^{t-1} \delta_ix_ix_i^\top \delta_i^\top$}
    Recall the conclusion from \cref{lemma: Hi prob bounds in theorem 2}: $\norm{x_t} = \calO(\log^{1/2}(t)) \as$
     and $\norm{\delta_t} = \calO(t^{-\frac{\beta}{2}} \log^{\frac{-\alpha + 1}{2}}(t)) \as$
    \begin{align*}
        \lnorm{\sum_{i=1}^{t-1} \delta_ix_ix_i^\top \delta_i^\top }
        \le&\sum_{i=1}^{t-1} \norm{\delta_i}^2\norm{x_i}^2 \\    
        \le&  \calO(\log(t))\sum_{i=1}^{t-1}\calO(i^{-\beta}\log^{-\alpha+1}(i))  \as \qquad(\text{by \cref{lemma: Hi prob bounds in theorem 2}})\\
        =& \calO(t^{1-\beta}\log^{-\alpha+2}(t))\as \qquad (\text{by \cref{eq: sum eta_t}})
    \end{align*}
    This implies (by bounding the entries by the operator norm, and including the $i=0$ term as $\calO(1)$):
    \[\sum_{i=0}^{t-1} \delta_ix_ix_i^\top \delta_i^\top  = \calO(t^{1-\beta}\log^{-\alpha+2}(t)) \as\]

    ~\paragraph{Second part $\sum_{i=0}^{t-1} \eta_ix_i^\top \delta_i^\top$} The representative of the third term is $\sum_{i=0}^{t-1} \eta_ix_i^\top \delta_i^\top$.  The proof idea is similar to that in \cref{lemma: three parts u_tx_t} when we prove the bound for $\sum_{i=0}^{t-1} \eta_ix_i^\top $. Here we have an extra shrinking term $\delta_i$ which makes things easier.
    
    Again, we can apply \cref{lem: Lai and Wei martingale} to our context by noticing
\begin{equation*}
    \sum_{i=0}^{t-1}\eta_ix_i^\top \delta_i^\top 
    = \sum_{i=0}^{t-1}\eta_ii^{\frac{1-\beta}2}\log^{-\alpha/2}(i)(i^{\frac{\beta-1}2}\log^{\alpha/2}(i)x_i^\top\delta_i^\top)
.\end{equation*}
Here we normalized all $\eta_i$ to have a fixed normal distribution. Apply \cref{lem: Lai and Wei martingale} entry-wise, where $v_i$ corresponds to a fixed entry of $i^{\frac{\beta-1}2}\log^{\alpha/2}(i)x_i^\top\delta_i^\top$ and $\epsilon_i$ corresponds to a fixed entry of the normalized $\eta_i$. Our $v_i$ is bounded by $i^{\frac{\beta-1}2}\log^{\alpha/2}(i)\norm{x_i}\norm{\delta_i}$. Thus, 
\begin{equation*}
    \sum_{i=0}^{t-1}\eta_ix_i^\top\delta_i^\top  = o\left(V_t^{1/2}\log(V_t)\right) + \calO(1)
.\end{equation*}
where $V_t := \sum_{i=0}^{t-1} (i^{\frac{\beta-1}2}\log^{\alpha/2}(i)\norm{x_i}\norm{\delta_i})^2$. Apply the  high probability bound in \cref{lemma: Hi prob bounds in theorem 2} and we have
\begin{equation*}
\begin{split}
    V_t =& \sum_{i=1}^{t-1} (i^{\frac{\beta-1}2}\log^{\alpha/2}(i)\norm{x_i}\norm{\delta_i})^2
    \\
    =&\sum_{i=1}^{t-1} 
    i^{-1+\beta}\log^{\alpha}(i)
    \calO(\log(t))\calO(t^{-\beta} \log^{-\alpha + 1}(t)) \as 
    \qquad \text{(by \cref{lemma: Hi prob bounds in theorem 2}}  )\\
    =& \calO(t^{\beta}\log^{\alpha}(t))
    \calO(\log(t))
    \calO(t^{-\beta} \log^{-\alpha + 1}(t)) \as 
    \qquad \text{(by \cref{eq: sum eta_t}}  )\\
    =& \calO(\log^{2}(t)) \as
\end{split}
\end{equation*}
That is to say, $V_t = \calO(\log^{2}(t)) \as$ (adding the $i=0$ term as $\calO(1)$). Thus,
\begin{align*}
    \sum_{i=0}^{t-1}\eta_ix_i^\top\delta_i^\top
    =& o\left(V_t^{1/2}\log(V_t)\right) + \calO(1) \as \\
    =& o\left(\calO(\log^{2}(t))^{1/2}\log(\calO(\log^{2}(t)))\right) + \calO(1) \as\\
    =& o\left(o(\log^{2}(t))\right) + \calO(1) \as\\
    =& o\left(\log^{2}(t)\right)\as
\end{align*}

    \paragraph{Third part $\sum_{i=0}^{t-1} \eta_i\eta_i^\top$}
    By \cref{eq: sum eta_t}:
    \[\E (\sum_{i=0}^{t-1} \eta_i\eta_i^\top) 
    = \sum_{i=0}^{t-1} \tau^2 i^{\beta-1}\log^\alpha(i)I_d 
    = t^\beta\frac{\tau^2}{\beta}\log^\alpha(t)(I_d  + o(1)).\]
    With a little abuse of notation we use $\Var(\cdot)$ as entry-wise variance of a matrix. Again, $i=0,1$ terms are meant to be $\calO(1)$. 
\begin{align*}
    \begin{split}
        \Var(\sum_{i=0}^{t-1} \eta_i\eta_i^\top) 
=& \sum_{i=0}^{t-1} \Var(\eta_i\eta_i^\top) \\ 
=& \calO\left(\sum_{i=0}^{t-1} i^{2(\beta-1)}\log^{2\alpha}(i)\right) \\
\le& \calO\left(\sum_{i=0}^{t-1} i^{2(\beta-1)}\log^{2\max\{0,\alpha\}}(i)\right) \\
\le& \calO\left(\sum_{i=0}^{t-1} i^{2(\beta-1)}\log^{2\max\{0,\alpha\}}(t)\right) \\
=&\logO\left( \sum_{i=0}^{t-1} i^{2(\beta-1)}\right) \\
=& \logO(t^{2\beta-1}) .
    \end{split}
\end{align*}
    When $\beta > 1/2$ the last equation follows by \cref{eq: sum eta_t} and when $\beta = 1/2$ it is summation of harmonic series which is $\logO(1)$.
    Thus the standard error is only of order $\logO(t^{\beta-1/2})$, which is smaller than $\E (\sum_{i=0}^{t-1} \eta_i\eta_i^\top) $.
    That is to say, 
   \[\sum_{i=0}^{t-1}\eta_i\eta_i^\top  
   = t^\beta\frac{\tau^2}{\beta}\log^\alpha(t)(I_d  + o_p(1)).\]


\end{proof}

\subsection{Lemmas in \cref{The proof of thm:regret}}
\subsubsection{The proof of \cref{lem: useful lemma from fazel}}
\label{The proof of lem: useful lemma from fazel}
\begin{lemma*}
For any $\Kh$ with suitable dimension,
\begin{equation*}
\begin{split}
    &x^\top (Q + \Kh^\top R \Kh)x + x^\top (A+B\Kh)^\top P (A+B\Kh)x - x^\top P x \\
    & \qquad = x^\top (\Kh-K)^\top( R + B^\top P B) (\Kh-K)x
.\end{split}
\end{equation*}
\end{lemma*}
Recall $P$ is the middle step described by the DARE. It should satisfy \cref{eq:ControllerK}
\begin{equation*}
K = - (R + B^\top P B)^{-1}B^\top P A   
.\end{equation*}
As a result,
\begin{equation}
\label{eq: P property 1}
    (R +B^\top P B)K + B^\top P A  = 0
.\end{equation}
Also it is well known that \citep{Jamieson2018Lecture2}:
\begin{equation}
    \label{eq: P property 2}
    Q + K^\top R K + (A+BK)^\top P(A+BK) = P
.\end{equation}
Let $\Kh$ be another controller, then we have the following useful equation stated by  \cref{lem: useful lemma from fazel}.
\begin{equation*}
\begin{split}
    &x^\top (Q + \Kh^\top R \Kh)x + x^\top (A+B\Kh)^\top P (A+B\Kh)x - x^\top P x \\
    & \qquad = x^\top (Q + (\Kh-K+K)^\top R (\Kh-K+K))x \\
    & \qquad \qquad + x^\top (A+B(\Kh-K)+BK )^\top P (A+B(\Kh-K)+BK)x \\
    & \qquad \qquad- x^\top P x \\
    & \qquad = x^\top (Q + K^\top RK + (A+BK)^\top P(A+BK)) x \\
    & \qquad \qquad + 2x^\top (\Kh-K)^\top (  RK + B^\top P (A+BK))x   \\
    & \qquad \qquad + x^\top (\Kh-K)^\top( R + B^\top P B) (\Kh-K)x   \\
    & \qquad \qquad- x^\top P x \\
    & \qquad = x^\top (Q + K^\top RK + (A+BK)^\top P(A+BK)) x  - x^\top P x\\
    & \qquad \qquad + 2x^\top (\Kh-K)^\top (  (R+B^\top P B)K +  B^\top P A)x   \\
    & \qquad \qquad + x^\top (\Kh-K)^\top( R + B^\top P B) (\Kh-K)x   \\
    & \qquad = x^\top (\Kh-K)^\top( R + B^\top P B) (\Kh-K)x \qquad \text{(by \cref{eq: P property 1,eq: P property 2}})
.\end{split}
\end{equation*}

\subsection{Lemmas in \cref{The proof of thm:prediction CLT parametric}}
\subsubsection{The proof of \cref{lem: difference between real and substitutes x u}}
\label{The proof of lem: difference between real and substitutes x u}
\begin{lemma*}
\begin{equation*}
    x_t = \tilde{x}_t +  O(t^{-\frac{\beta}{2}} \log^{\frac{-\alpha + 2}{2}}(t)) \as
\end{equation*}
\begin{equation*}
    u_t = \tilde{u}_t +  O(t^{-\frac{\beta}{2}} \log^{\frac{-\alpha + 2}{2}}(t)) \as
\end{equation*}
where
\begin{equation}
\label{eq: tilde xt defn}
    \tilde{x}_t := \sum_{p=t-\myfloor{-\frac{\log(t)}{\log(\rho_L)}}}^{t-1}(A+BK)^{t-p-1} ( B\eta_p + \varepsilon_p)
,\end{equation}
and
\begin{equation*}
    \tilde{u}_t := K\tilde{x}_t + \xi_t= K\sum_{p=t-\myfloor{-\frac{\log(t)}{\log(\rho_L)}}}^{t-1}(A+BK)^{t-p-1} ( B\eta_p + \varepsilon_p) + \xi_t
.\end{equation*}
\end{lemma*}

\begin{proof}
Recall \cref{lemma: StateExpansion} states that
\begin{equation*}
x_{t} = \sum_{p=0}^{t-1}(A+B \Kh_{t-1})\cdots(A+B \Kh_{p+1})(B \eta_p+\varepsilon_p) +(A+B \Kh_{t-1})\cdots(A+B K_0)x_0
.\end{equation*}
Similarly, we can rewrite $x_t$ as if starting from time $t - \myfloor{-\frac{\log(t)}{\log(\rho_L)}}$:
\begin{equation}
\label{eq: xt another decomposition}
x_{t} = \sum_{p=t-\myfloor{-\frac{\log(t)}{\log(\rho_L)}}}^{t-1}(A+B \Kh_{t-1})\cdots(A+B \Kh_{p+1})(B \eta_p+\varepsilon_p) + (A+B \Kh_{t-1})\cdots(A+B \Kh_{t-\myfloor{-\frac{\log(t)}{\log(\rho_L)}}})x_{t-\myfloor{-\frac{\log(t)}{\log(\rho_L)}}}.
\end{equation}
By \cref{lemma: Hi prob bounds in theorem 2}, we know
\begin{equation*}
\begin{split}
    (A+B \Kh_{t-1})\cdots(A+B \Kh_{t-\myfloor{-\frac{\log(t)}{\log(\rho_L)}}})
    \le& \calO(\rhoL^{-\log(t)/\log(\rho_L)}) \as\\
    =& \calO(e^{-\log(t)}) \as\\
    =& \calO(t^{-1})  \as
\end{split}
\end{equation*}
and
\begin{equation*}
\norm{x_t}, \norm{u_t} \le \calO(\log^{1/2}(t)) \as
\end{equation*}
Thus
\begin{equation*}
(A+B \Kh_{t-1})\cdots(A+B \Kh_{t-\myfloor{-\frac{\log(t)}{\log(\rho_L)}}})x_{t-\myfloor{-\frac{\log(t)}{\log(\rho_L)}}} = \calO(t^{-1}\log^{1/2}(t)) \as
\end{equation*}
Next, comparing \cref{eq: tilde xt defn} with \cref{eq: xt another decomposition}, we still need to bound the difference between $(A+B \Kh_{t-1})\cdots(A+B \Kh_{p+1})$ and $(A+BK)^{t-p-1}$. Again by \cref{lemma: Hi prob bounds in theorem 2},
\begin{equation*}
\begin{split}
&\lnorm{\sum_{p=t-\myfloor{-\frac{\log(t)}{\log(\rho_L)}}}^{t-1} 
\left[
(A+B \Kh_{t-1})\cdots(A+B \Kh_{p+1}) - (A+B K)^{t-p-1}
\right]
( B\eta_p + \varepsilon_p)
} \\
& \qquad \le
\sum_{p=t-\myfloor{-\frac{\log(t)}{\log(\rho_L)}}}^{t-1} 
\calO(\rho_L ^{t-p})(\norm{\delta_{t-1}}+ \cdots + \norm{\delta_{p+1}}) 
\calO(\log^{1/2}(t)) \as \qquad \text{(by \cref{eq: difference Khat product and K product,eq:bound on eta_p})}\\
& \qquad =
\sum_{p=t-\myfloor{-\frac{\log(t)}{\log(\rho_L)}} }^{t-1}\norm{\delta_{p+1}}\calO(\rho_L^{t-p})
\calO(\log^{1/2}(t)) \as \qquad \text{(by \cref{eq:part2 basic tool})}\\
& \qquad \le
 \calO((t/2)^{-\frac{\beta}{2}} \log^{\frac{-\alpha + 1}{2}}(t/2))
\sum_{p=t-\myfloor{-\frac{\log(t)}{\log(\rho_L)}} }^{t-1}\calO(\rho_L^{t-p})
\calO(\log^{1/2}(t)) \as \\
& \qquad \qquad
\text{(by \cref{eq: stochastic bound delta_t} and that asymptotically $p > t/2$)}
\\
& \qquad =
 \calO(t^{-\frac{\beta}{2}} \log^{\frac{-\alpha + 1}{2}}(t))
\calO(\log^{1/2}(t)) \as \\
& \qquad =
 \calO(t^{-\frac{\beta}{2}} \log^{\frac{-\alpha + 2}{2}}(t)
 ) \as 
\end{split}
\end{equation*}
This is larger than $\calO(t^{-1}\log^{1/2}(t))$. To summarize, 
\begin{equation*}
    x_t = \tilde{x}_t + 
     \calO(t^{-\frac{\beta}{2}} \log^{\frac{-\alpha + 2}{2}}(t)
 ) \as
\end{equation*}
Since $u_t - \tilde{u}_t = K(x_t - \tilde{x}_t)$,
\begin{equation*}
u_t = \tilde{u}_t +  O(t^{-\frac{\beta}{2}} \log^{\frac{-\alpha + 2}{2}}(t)) \as
\end{equation*}

\end{proof}

\subsubsection{The proof of \cref{lem: difference between real and substitutes A B}}
\label{The proof of lem: difference between real and substitutes A B}
\begin{lemma*}
\begin{equation*}
    \Ah_{t} = \Ah_{t-\myfloor{-\frac{\log(t)}{\log(\rho_L)}}} + \calO_p(t^{-\beta}\log^{-\alpha+3/2}(t))
.\end{equation*}
\begin{equation*}
    \Bh_{t} = \Bh_{t-\myfloor{-\frac{\log(t)}{\log(\rho_L)}}} + \calO_p(t^{-\beta}\log^{-\alpha+3/2}(t))
.\end{equation*}
\end{lemma*}
\begin{proof}
We can bound the distance of neighboring estimators by the following recursive LS formula. Denote $\hat\Theta_{t} := [\Ahat_{t}, \Bhat_{t}]$, $z_i := \begin{bmatrix}
x_i \\
u_i
\end{bmatrix}$, $ H_t := (\sum_{i=0}^{t-1} z_{i}z_i^\top)^{-1}$.
Then the LS estimator \cref{eq: AhBh estimator} is
\begin{equation*}
\hat\Theta_{t} = \sum_{i=0}^{t-1} z_{i+1}z_i^\top(\sum_{i=0}^{t-1} z_{i}z_i^\top)^{-1}
= \sum_{i=0}^{t-1} z_{i+1}z_i^\top H_t
.\end{equation*}
For simplicity, denote $a_t := \myfloor{-\frac{\log(t)}{\log(\rho_L)}}$, then our objective is to bound the difference $ \hat\Theta_{t}-\hat\Theta_{t-a_t}$.
\begin{equation*}
\hat\Theta_{t-a_t}  = \sum_{i=0}^{t-a_t-1} z_{i+1}z_i^\top H_{t-a_t}
.\end{equation*}
As a result,
\begin{equation*}
\hat\Theta_{t} = (\hat\Theta_{t-a_t}H_{t-a_t}^{-1} + 
\sum_{i=t-a_t}^{t-1}z_{i+1}z_i^\top) H_t
.\end{equation*}
And
\begin{equation}
\label{eq: theta t - theta t-at}
\begin{split}
\hat\Theta_{t}-\hat\Theta_{t-a_t} =& \left(\hat\Theta_{t-a_t}(H_{t-a_t}^{-1}-H_t^{-1}) + \sum_{i=t-a_t}^{t-1}z_{i+1}z_i^\top\right) H_t \\
=& \left(-\hat\Theta_{t-a_t}\left(\sum_{i=t-a_t}^{t-1} z_{i}z_i^\top\right) + \sum_{i=t-a_t}^{t-1}z_{i+1}z_i^\top\right) H_t \\
=& \left(-\hat\Theta_{t-a_t}\left(\sum_{i=t-a_t}^{t-1} z_{i}z_i^\top\right) + \sum_{i=t-a_t}^{t-1}(\Theta z_i + \varepsilon_i)z_i^\top\right) H_t \\
=& (\Theta-\hat\Theta_{t-a_t})\left(\sum_{i=t-a_t}^{t-1} z_{i}z_i^\top\right)H_t + \sum_{i=t-a_t}^{t-1}\varepsilon_{i}z_i^\top H_t 
.\end{split}
\end{equation}
Following \cref{eq: Dt -1 Gram Dt -1,eq: DtDt -1 order}, 
\begin{equation}
\label{eq: Ht order}
H_t = \calO_p(t^{-\beta}\log^{-\alpha}(t)) 
.\end{equation}
Next will bound the first and second term separately.

~\paragraph{First term $(\Theta-\hat\Theta_{t-a_t})(\sum_{i=t-a_t}^{t-1} z_{i}z_i^\top)H_t$}
By \cref{lemma: Hi prob bounds in theorem 2}, \
\begin{equation*}
z_t  = \calO(\log^{1/2}(t)) \as
\end{equation*}
Recall that from \cref{eq:final Conclusion,eq: DtDt -1 order}, $\Theta-\hat\Theta_{t-a_t} = \calO_p(t^{-\beta/2}\log^{-\alpha/2}(t))$.
As a result,
\begin{equation}
\label{eq: tilde estimate part 1}
\begin{split}
(\Theta-\hat\Theta_{t-a_t})(\sum_{i=t-a_t}^{t-1} z_{i}z_i^\top)H_t 
=& \calO_p\left(t^{-\beta/2}\log^{-\alpha/2}(t)\right)\calO_p(a_t \log(t) t^{-\beta}\log^{-\alpha}(t)) \\
=& \calO_p(t^{-3\beta/2}\log^{-3\alpha/2+2}(t))
.\end{split}
\end{equation}
We will see that this order is smaller than the second term, so that the second term is dominating. 

~\paragraph{Second term $ \sum_{i=t-a_t}^{t-1}\varepsilon_{i}z_i^\top H_t$}
Consider the variance of the $jk$-th element of $ \sum_{i=t-a_t}^{t-1}\varepsilon_{i}z_i^\top$, which is applicable to any choice of $j$ and $k$. Fix $j$, $k$. Define $\calF_{t-1}$ as the filtration which contains every variable except for $\varepsilon_{t-1,j}$. We know that $\varepsilon_{t-1,j} \independent \calF_{t-1}$ and $\varepsilon_{t-1,j} \sim \calN(0, \sigma^2)$.
\begin{align*}
    &\Var\left(\sum_{i=t-a_t}^{t-1}\varepsilon_{ij}(z_{i} )_k\right) \\
    & \quad =\Var\left(\E\left(\sum_{i=t-a_t}^{t-1}\varepsilon_{ij}(z_{i} )_k \Bigg| \calF_{t-1}\right)\right) + \E\left(\Var\left(\sum_{i=t-a_t}^{t-1}\varepsilon_{ij}(z_{i} )_k \Bigg| \calF_{t-1}\right)\right) \\
& \quad =\Var\left(\sum_{i=t-a_t}^{t-2}\varepsilon_{ij}(z_{i} )_k \right) + \E\left((z_{t-1} )_k^2\sigma^2\right) \\ 
& \quad = \sigma^2 \sum_{i=t-a_t}^{t-1}\E\left((z_{i} )_k^2\right)  \qquad \text{(by recursively conditioning on $\calF_{t-2}, \cdots, \calF_{t-a_t} $)}\\
& \quad \le \sigma^2 \sum_{i=t-a_t}^{t-1}\E\norm{z_{i}}^2 \\
& \quad \le \sigma^2 a_t \calO(\log^2(t)) \qquad \text{(by \cref{eq: E zt 2 final})} \\
& \quad \le \sigma^2  \calO(\log^3(t))  \qquad \left(\text{by } a_t := \myfloor{-\frac{\log(t)}{\log(\rho_L)}}\right)
\end{align*}
Since $\E\left(\sum_{i=t-a_t}^{t-1}\varepsilon_{ij}(z_{i}^\top )_k\right) = 0$, we have $\sum_{i=t-a_t}^{t-1}\varepsilon_{ij}(z_{i}^\top )_k = \calO_p(\log^{3/2}(t))$, which implies
\begin{equation*}
    \sum_{i=t-a_t}^{t-1}\varepsilon_{i}z_{i}^\top = \calO_p(\log^{3/2}(t)).
\end{equation*}
By \cref{eq: Ht order},
\begin{equation}
\label{eq: tilde estimate part 2}
    \sum_{i=t-a_t}^{t-1}\varepsilon_{i}z_{i}^\top H_t = \calO_p(\log^{3/2}(t))\calO_p(t^{-\beta}\log^{-\alpha}(t)) 
    = \calO_p(t^{-\beta}\log^{-\alpha + 3/2}(t)) .
\end{equation}

This is larger than the first term. Combining \cref{eq: theta t - theta t-at,eq: tilde estimate part 1,eq: tilde estimate part 2} we have
\begin{equation*}
\begin{split}
\hat\Theta_{t}-\hat\Theta_{t-a_t} = \calO_p(t^{-\beta}\log^{-\alpha+3/2}(t))
.\end{split}
\end{equation*}
\end{proof}

\subsubsection{The proof of \cref{lem: CLT of substitution}}
\label{The proof of lem: CLT of substitution}
\begin{lemma*}
For any $\xi_t$ independent of the data  before $t$: $\{\varepsilon_i, \eta_i\}_{i=0}^{t-1}$: 
\begin{align*}
    &\left(
    \tilde{x}_t^\top   
    \left( \sum_{p=0}^{\infty}L ^{p}\left(I_n + 1_{\{\beta=1,\alpha=0\}}\frac{\tau^2}{\sigma^2}BB^\top\right)(L ^{p})^\top\right)^{-1}
    \tilde{x}_t  
    +
     \frac{\beta \sigma^2}{\tau^2}
     t^{1-\beta} \log^{-\alpha}(t)
     \lnorm{\xi_t}^2
    \right)^{-1/2} \\
    &
    \qquad \cdot t^{1/2}  \left((\Ah_{t-\myfloor{-\frac{\log(t)}{\log(\rho_L)}}} - A)\tilde{x}_t + 
    (\Bh_{t-\myfloor{-\frac{\log(t)}{\log(\rho_L)}}}- B)(K\tilde{x}_t+\xi_t)\right)
    \convD 
    \calN(0,I_n).
\end{align*}
\end{lemma*}
\begin{proof}
We will start from finding the conditional distribution of
\begin{equation*}
    (\Ah_{t-\myfloor{-\frac{\log(t)}{\log(\rho_L)}}} - A)\tilde{x}_t + (\Bh_{t-\myfloor{-\frac{\log(t)}{\log(\rho_L)}}}- B)\tilde{u}_t \bigg\vert \tilde{x}_t = x, \tilde{u}_t = Kx + \xi
.\end{equation*}
where $x$ and $\xi$ are constants. This should be easy because 
$
    \Ah_{t-\myfloor{-\frac{\log(t)}{\log(\rho_L)}}}- A, \Bh_{t-\myfloor{-\frac{\log(t)}{\log(\rho_L)}}}-B \independent \tilde{x}_t, \tilde{u}_t
$, which means we can directly apply the asymptotic normality result from \cref{thm:main CLT}. Recall \cref{eq: fast slow rate CLT} that
\begin{equation*}
       t^{\beta/2} \log^{\alpha/2}(t)
   \vvector \left(   
\begin{bmatrix}
    \Ah_t - A + (\Bh_t- B)K, &\Bh_t- B
\end{bmatrix} 
\left[
\begin{array}{cc}
C_t^{1/2} & 0\\
0 & \sqrt{\frac{\tau^2}{\beta}} I_d\\
\end{array}
\right]
\right) \convD 
\calN(0, \sigma^2 I_{\statedim(n+d)} )
,\end{equation*}
where $C_t = t^{1-\beta}\log^{-\alpha}(t)    
\sum_{p=0}^{\infty}L ^{p}\left(\sigma^2I_n + 1_{\{\beta=1,\alpha=0\}}\tau^2BB^\top\right)(L ^{p})^\top
(I_n + o_p(1))$ (by \cref{eq: Ct order}). Here there are two different convergence speeds and we need to consider them separately. 
More precisely,
\begin{align*}
   \vvector \left(
\left[
\begin{array}{cc}
(\Ah_t - A + (\Bh_t- B)K)
t^{\beta/2} \log^{\alpha/2}(t)
C_t^{1/2}\sigma^{-1} & 
(\Bh_t- B)t^{\beta/2} \log^{\alpha/2}(t)\sqrt{\frac{\tau^2}{\sigma^2 \beta}} I_d\\
\end{array}
\right]
\right) \\
\convD 
\calN(0, 
I_{n+d}\otimes I_n)
.\end{align*}
That is to say, for any constant vector $x$ and $\xi_t$ independent of data before $t$, we have
\begin{align*}
   &\vvector \Bigg(   
\left[
\begin{array}{cc}
(\Ah_t - A + (\Bh_t- B)K)
t^{\beta/2} \log^{\alpha/2}(t)C_t^{1/2}\sigma^{-1}
& (\Bh_t- B)t^{\beta/2} \log^{\alpha/2}(t)\sqrt{\frac{\tau^2}{\sigma^2 \beta}} I_d\\
\end{array}
\right]
\\
&\qquad \cdot \begin{bmatrix}
    t^{-\beta/2} \log^{-\alpha/2}(t)C_t^{-1/2}\sigma
    x \\
    t^{(1-\beta)/2} \log^{-\alpha/2}(t) 
    \sqrt{\frac{\sigma^2 \beta}{\tau^2}} \xi_t
\end{bmatrix}
\Bigg/
\lnorm{\begin{bmatrix}
    t^{-\beta/2} \log^{-\alpha/2}(t)C_t^{-1/2}\sigma
    x \\
    t^{(1-\beta)/2} \log^{-\alpha/2}(t) 
    \sqrt{\frac{\sigma^2 \beta}{\tau^2}} \xi_t
\end{bmatrix}}
\Bigg)
 \convD 
\calN(0, I_n)
.\end{align*}
The above equation holds because we are multiplying independent unit vector to the left hand side, so the result is still a normal distribution. Simplifying the equation:
\begin{align*}
    \left(
    x^\top   
    \left( \sum_{p=0}^{\infty}L ^{p}\left(I_n + 1_{\{\beta=1,\alpha=0\}}\frac{\tau^2}{\sigma^2}BB^\top\right)(L ^{p})^\top\right)^{-1}
    x  
    +
     \frac{\beta \sigma^2}{\tau^2}
     t^{1-\beta} \log^{-\alpha}(t)\norm{\xi_t}^2
    \right)^{-1/2} \\
    \cdot t^{1/2}  \left[(\Ah_{t} - A)x + (\Bh_{t}- B)(Kx+\xi_t)\right] \convD
    \calN(0,I_n)
.\end{align*}
We can replace $t$ with $t-\myfloor{-\frac{\log(t)}{\log(\rho_L)}}$:
\begin{align*}
    &\left(
    x^\top   
    \left( \sum_{p=0}^{\infty}L ^{p}\left(I_n + 1_{\{\beta=1,\alpha=0\}}\frac{\tau^2}{\sigma^2}BB^\top\right)(L ^{p})^\top\right)^{-1}
    x  
    +
     \frac{\beta \sigma^2}{\tau^2}
     \left(t-\myfloor{-\frac{\log(t)}{\log(\rho_L)}}\right)^{1-\beta} \log^{-\alpha}\left(t-\myfloor{-\frac{\log(t)}{\log(\rho_L)}}\right)
     \lnorm{\xi_t}^2
    \right)^{-1/2} \\
    & \qquad \cdot
    \left(t-\myfloor{-\frac{\log(t)}{\log(\rho_L)}}\right)^{1/2}  \left[(\Ah_{t-\myfloor{-\frac{\log(t)}{\log(\rho_L)}}} - A)x + 
    (\Bh_{t-\myfloor{-\frac{\log(t)}{\log(\rho_L)}}}- B)(Kx+\xi_t)\right]  \convD 
    \calN(0,I_n)
.\end{align*}
Because $\left(t-\myfloor{-\frac{\log(t)}{\log(\rho_L)}}\right)^{1/2}t^{-1/2} \to 1$, we can drop the first three instances of $\myfloor{-\frac{\log(t)}{\log(\rho_L)}}$:
\begin{align*}
    &\left(
    x^\top   
    \left( \sum_{p=0}^{\infty}L ^{p}\left(I_n + 1_{\{\beta=1,\alpha=0\}}\frac{\tau^2}{\sigma^2}BB^\top\right)(L ^{p})^\top\right)^{-1}
    x  
    +
     \frac{\beta \sigma^2}{\tau^2}
     t^{1-\beta} \log^{-\alpha}(t)
     \lnorm{\xi_t}^2
    \right)^{-1/2} \\
    & \qquad \cdot
    t^{1/2}  \left[(\Ah_{t-\myfloor{-\frac{\log(t)}{\log(\rho_L)}}} - A)x + 
    (\Bh_{t-\myfloor{-\frac{\log(t)}{\log(\rho_L)}}}- B)(Kx+\xi_t)\right] 
    \convD 
    \calN(0,I_n)
.\end{align*}
Here we actually used the fact that for $c_t,a_t,b_t > 0$, when $a_t/b_t \to 1$, then $(c_t+a_t)/(c_t+b_t) \to 1$. This is because
\begin{align*}
    \labs{\frac{c_t+a_t}{c_t+b_t} - \frac{a_t}{b_t}} 
    = \labs{\frac{(b_t-a_t)c_t}{(c_t+b_t)b_t}} 
    \le  \labs{\frac{b_t-a_t}{b_t}} 
    \to 0.
\end{align*}
In our specific context $c_t$ is the constant $x^\top   
\left( \sum_{p=0}^{\infty}L ^{p}\left(I_n + 1_{\{\beta=1,\alpha=0\}}\frac{\tau^2}{\sigma^2}BB^\top\right)(L ^{p})^\top\right)^{-1}
x$.

Since $\tilde{x}_t \independent \Ah_{t-\myfloor{-\frac{\log(t)}{\log(\rho_L)}}} - A, \Bh_{t-\myfloor{-\frac{\log(t)}{\log(\rho_L)}}}- B$, we can replace $x$ with $\tilde{x}_t$ by conditioning on $\tilde{x}_t = x$, replace all $x$ with $\tilde{x}_t$, and finally remove the conditioning since they all converge in distribution to standard normal and $\tilde{x}_t$ asymptotically have same distribution.

\begin{align}
\label{eq: tilde prediction clt}
\begin{split}
&\left(
    \tilde{x}_t^\top   
    \left( \sum_{p=0}^{\infty}L ^{p}\left(I_n + 1_{\{\beta=1,\alpha=0\}}\frac{\tau^2}{\sigma^2}BB^\top\right)(L ^{p})^\top\right)^{-1}
    \tilde{x}_t  
    +
     \frac{\beta \sigma^2}{\tau^2}
     t^{1-\beta} \log^{-\alpha}(t)
     \lnorm{\xi_t}^2
    \right)^{-1/2} \\
    &
    \qquad \cdot t^{1/2}  \left((\Ah_{t-\myfloor{-\frac{\log(t)}{\log(\rho_L)}}} - A)\tilde{x}_t + 
    (\Bh_{t-\myfloor{-\frac{\log(t)}{\log(\rho_L)}}}- B)(K\tilde{x}_t+\xi_t)\right)
    \convD 
    \calN(0,I_n).
\end{split}
\end{align}

\end{proof}

\subsubsection{The proof of \cref{lem: CLT original}}
 \label{The proof of lem: CLT original}
    \begin{lemma*}
For any $\xi_t$ independent of the data before $t$: $\{\varepsilon_i, \eta_i\}_{i=0}^{t-1}$,
\begin{align*}
    &\left(
    x_t^\top   
    \left( \sum_{p=0}^{\infty}L ^{p}\left(I_n + 1_{\{\beta=1,\alpha=0\}}\frac{\tau^2}{\sigma^2}BB^\top\right)(L ^{p})^\top\right)^{-1}
    x_t  
    +
     \frac{\beta \sigma^2}{\tau^2}
     t^{1-\beta} \log^{-\alpha}(t)
     \lnorm{\xi_t}^2
    \right)^{-1/2} \\
    & \qquad \cdot
    t^{1/2}  \left[(\Ah_{t} - A)x_t + 
    (\Bh_{t}- B)(\Kh_tx_t+\xi_t)\right] 
    \convD 
    \calN(0,I_n)
.\end{align*}

\end{lemma*}
 \begin{proof}
 Since we already proved \cref{lem: CLT of substitution}, the only thing we need to do is to replace $\tilde{x}_t$ with $x_t$, $K$ with $\Kh_t$, and $\Ah_{t-\myfloor{-\frac{\log(t)}{\log(\rho_L)}}}$,  $\Bh_{t-\myfloor{-\frac{\log(t)}{\log(\rho_L)}}}$ with $ \Ah_t$, $\Bh_t$.
 
 ~\paragraph{Replacing $\tilde{x}_t$ with $x_t$}
 First, we can replace 
 $$\left(
    \tilde{x}_t^\top   
    \left( \sum_{p=0}^{\infty}L ^{p}\left(I_n + 1_{\{\beta=1,\alpha=0\}}\frac{\tau^2}{\sigma^2}BB^\top\right)(L ^{p})^\top\right)^{-1}
    \tilde{x}_t  
    +
     \frac{\beta \sigma^2}{\tau^2}
     t^{1-\beta} \log^{-\alpha}(t)
     \lnorm{\xi_t}^2
    \right)$$
    with 
    
    $\left(
    x_t^\top   
    \left( \sum_{p=0}^{\infty}L ^{p}\left(I_n + 1_{\{\beta=1,\alpha=0\}}\frac{\tau^2}{\sigma^2}BB^\top\right)(L ^{p})^\top\right)^{-1}
    x_t  
    +
     \frac{\beta \sigma^2}{\tau^2}
     t^{1-\beta} \log^{-\alpha}(t)
     \lnorm{\xi_t}^2
    \right)$ in \cref{eq: tilde prediction clt} because $\tilde{x}_t = x_t + o_p(1)$ by \cref{lem: difference between real and substitutes x u}.
\begin{align*}
\begin{split}
&\left(
    x_t^\top   
    \left( \sum_{p=0}^{\infty}L ^{p}\left(I_n + 1_{\{\beta=1,\alpha=0\}}\frac{\tau^2}{\sigma^2}BB^\top\right)(L ^{p})^\top\right)^{-1}
    x_t  
    +
     \frac{\beta \sigma^2}{\tau^2}
     t^{1-\beta} \log^{-\alpha}(t)
     \lnorm{\xi_t}^2
    \right)^{-1/2} \\
    &
    \qquad \cdot t^{1/2}  \left((\Ah_{t-\myfloor{-\frac{\log(t)}{\log(\rho_L)}}} - A)\tilde{x}_t + 
    (\Bh_{t-\myfloor{-\frac{\log(t)}{\log(\rho_L)}}}- B)(K\tilde{x}_t+\xi_t)\right)
    \convD 
    \calN(0,I_n).
\end{split}
\end{align*}
 
 Since $x_t^\top   
 \left( \sum_{p=0}^{\infty}L ^{p}\left(I_n + 1_{\{\beta=1,\alpha=0\}}\frac{\tau^2}{\sigma^2}BB^\top\right)(L ^{p})^\top\right)^{-1}
 x_t$ is bounded away from $0$ with high probability ($x_t$ has the component $\varepsilon_{t-1}$),
\begin{equation*}
\left(
    x_t^\top   
    \left( \sum_{p=0}^{\infty}L ^{p}\left(I_n + 1_{\{\beta=1,\alpha=0\}}\frac{\tau^2}{\sigma^2}BB^\top\right)(L ^{p})^\top\right)^{-1}
    x_t  
    +
     \frac{\beta \sigma^2}{\tau^2}
     t^{1-\beta} \log^{-\alpha}(t)
     \lnorm{\xi_t}^2
    \right)^{-1/2} = \calO_p(1)
.\end{equation*}
By \cref{lem: difference between real and substitutes x u}, $\tilde{x}_t = x_t + O_p(t^{-\frac{\beta}{2}} \log^{\frac{-\alpha + 2}{2}}(t))$.
Recall \cref{prop:one_epoch_estimate_withMyalg} states that $\norm{\Ah_{t-\myfloor{-\frac{\log(t)}{\log(\rho_L)}}}-A}, \norm{\Bh_{t-\myfloor{-\frac{\log(t)}{\log(\rho_L)}}}-B},  \norm{\Kh_{t}-K} = \calO_p(t^{-\beta/2}\log^{\frac{-\alpha + 1}{2}}(t))$.
Thus, the error induced by replacing the remaining $\tilde{x}_t$ with $x_t$ in \cref{eq: tilde prediction clt} is 
\begin{equation*}
    \calO_p(1)t^{1/2}
    \calO_p(t^{-\frac{\beta}{2}} \log^{\frac{-\alpha + 2}{2}}(t))
    \calO_p(t^{-\frac\beta2}
    \log^{\frac{-\alpha + 1}{2}}(t)) 
    =\calO_p(t^{1/2-\beta} \log^{-\alpha + 3/2}(t)) 
.\end{equation*}
Under our condition $\beta > 1/2$ or $\beta = 1/2, \alpha > 3/2$, this error is of order $o_p(1)$, which is negligible.
Now we can replace all $\tilde{x}_t$ with $x_t$:
\begin{align*}
    &\left(
    x_t^\top   
    \left( \sum_{p=0}^{\infty}L ^{p}\left(I_n + 1_{\{\beta=1,\alpha=0\}}\frac{\tau^2}{\sigma^2}BB^\top\right)(L ^{p})^\top\right)^{-1}
    x_t  
    +
     \frac{\beta \sigma^2}{\tau^2}
     t^{1-\beta} \log^{-\alpha}(t)
     \lnorm{\xi_t}^2
    \right)^{-1/2} \\
    & \qquad \cdot t^{1/2}  
    \left[
    (\Ah_{t-\myfloor{-\frac{\log(t)}{\log(\rho_L)}}} - A)x_t + 
    (\Bh_{t-\myfloor{-\frac{\log(t)}{\log(\rho_L)}}}- B)(Kx_t+\xi_t)\right] 
    \convD 
    \calN(0,I_n)
.\end{align*}
 ~\paragraph{Replacing $K$ by $\Kh_t $}
Since $\norm{\Bh_{t-\myfloor{-\frac{\log(t)}{\log(\rho_L)}}}-B}, \norm{\Kh_{t}-K} = \calO_p(t^{-\beta/2}\log^{\frac{-\alpha + 1}{2}}(t))$ (see \cref{prop:one_epoch_estimate_withMyalg}), and $x_t = \calO_p(\log^{1/2}(t))$, the final difference is still of order $\calO_p(t^{1/2-\beta} \log^{-\alpha + 3/2}(t)) = o_p(1)$. Thus
\begin{align*}
    &\left(
    x_t^\top   
    \left( \sum_{p=0}^{\infty}L ^{p}\left(I_n + 1_{\{\beta=1,\alpha=0\}}\frac{\tau^2}{\sigma^2}BB^\top\right)(L ^{p})^\top\right)^{-1}
    x_t  
    +
     \frac{\beta \sigma^2}{\tau^2}
     t^{1-\beta} \log^{-\alpha}(t)
     \lnorm{\xi_t}^2
    \right)^{-1/2} \\
    & \qquad \cdot
    t^{1/2}  
    \left[
    (\Ah_{t-\myfloor{-\frac{\log(t)}{\log(\rho_L)}}} - A)x_t + 
    (\Bh_{t-\myfloor{-\frac{\log(t)}{\log(\rho_L)}}}- B)(\Kh_tx_t+\xi_t)\right] 
    \convD 
    \calN(0,I_n)
.\end{align*}

 ~\paragraph{Replacing $\Ah_{t-\myfloor{-\frac{\log(t)}{\log(\rho_L)}}}$,  $\Bh_{t-\myfloor{-\frac{\log(t)}{\log(\rho_L)}}}$ with $ \Ah_t$, $\Bh_t$}
By \cref{lem: difference between real and substitutes A B}, 
\begin{equation*}
    \Ah_{t} - \Ah_{t-\myfloor{-\frac{\log(t)}{\log(\rho_L)}}}, \Bh_{t} - \Bh_{t-\myfloor{-\frac{\log(t)}{\log(\rho_L)}}} = 
    \calO_p(t^{-\beta}\log^{-\alpha+3/2}(t))
.\end{equation*}
Notice the $x_t$ and $\xi_t$ are multiplied by 
$$\left(
    x_t^\top   
    \left( \sum_{p=0}^{\infty}L ^{p}\left(I_n + 1_{\{\beta=1,\alpha=0\}}\frac{\tau^2}{\sigma^2}BB^\top\right)(L ^{p})^\top\right)^{-1}
    x_t  
    +
     \frac{\beta \sigma^2}{\tau^2}
     t^{1-\beta} \log^{-\alpha}(t)
     \lnorm{\xi_t}^2
    \right)^{-1/2},$$ 
    thus their order is only $\calO_p(1)$. The difference induced by replacing $\Ah_{t-\myfloor{-\frac{\log(t)}{\log(\rho_L)}}}, \Bh_{t-\myfloor{-\frac{\log(t)}{\log(\rho_L)}}}$ with $\Ah_{t}, \Bh_{t} $ is of order $\calO_p(t^{1/2-\beta}\log^{-\alpha+3/2}(t))$. When $\beta > 1/2$ or $\beta = 1/2, \alpha > 3/2$, this error is of order $o_p(1)$. Finally, after replacement we have
\begin{align*}
    &\left(
    x_t^\top   
    \left( \sum_{p=0}^{\infty}L ^{p}\left(I_n + 1_{\{\beta=1,\alpha=0\}}\frac{\tau^2}{\sigma^2}BB^\top\right)(L ^{p})^\top\right)^{-1}
    x_t  
    +
     \frac{\beta \sigma^2}{\tau^2}
     t^{1-\beta} \log^{-\alpha}(t)
     \lnorm{\xi_t}^2
    \right)^{-1/2} \\
    & \qquad \cdot
    t^{1/2}  \left[(\Ah_{t} - A)x_t + 
    (\Bh_{t}- B)(\Kh_tx_t+\xi_t)\right] 
    \convD 
    \calN(0,I_n)
.\end{align*}
\end{proof}

\subsubsection{The proof of \cref{lem: variance equivalence}}
\label{The proof of lem: variance equivalence}
\begin{lemma*}
For any $\xi_t$ independent of the data before $t$: $\{\varepsilon_i, \eta_i\}_{i=0}^{t-1}$,
\begin{align*}
    \left(
    x_t^\top   
    \left( \sum_{p=0}^{\infty}L ^{p}\left(I_n + 1_{\{\beta=1,\alpha=0\}}\frac{\tau^2}{\sigma^2}BB^\top\right)(L ^{p})^\top\right)^{-1}
    x_t  
    +
     \frac{\beta \sigma^2}{\tau^2}
     t^{1-\beta} \log^{-\alpha}(t)
     \lnorm{\xi_t}^2
    \right)^{-1/2} \\
    \cdot t^{1/2}
    \left(
    \sigma^2
\begin{bmatrix}
        x_t \\
        u_{t}
    \end{bmatrix}^\top
    \left(
    \sum_{i=0}^{t-1}
    \begin{bmatrix}
            x_i\\
            u_i\\
    \end{bmatrix}
    \begin{bmatrix}
            x_i\\
            u_i\\
    \end{bmatrix}
    ^\top
    \right)^{-1}
    \begin{bmatrix}
        x_t \\
        u_{t}
    \end{bmatrix}
    \right)^{1/2} \convP 1
.\end{align*}

\end{lemma*}
\begin{proof}
    By 
    $u_t = \Kh_tx_t+\xi_t$, it suffices to show
\begin{align*}
    &\left(
    x_t^\top   
    \left( \sum_{p=0}^{\infty}L ^{p}\left(I_n + 1_{\{\beta=1,\alpha=0\}}\frac{\tau^2}{\sigma^2}BB^\top\right)(L ^{p})^\top\right)^{-1}
    x_t  
    +
     \frac{\beta \sigma^2}{\tau^2}
     t^{1-\beta} \log^{-\alpha}(t)
     \lnorm{\xi_t}^2
    \right)^{-1} \\
    & \qquad \cdot
    t^{1/2}
    \left(
    \sigma^2
\begin{bmatrix}
        x_t \\
        \Kh_tx_t+\xi_t
    \end{bmatrix}^\top
   \left(
    \sum_{i=0}^{t-1}
    \begin{bmatrix}
            x_i\\
            u_i\\
    \end{bmatrix}
    \begin{bmatrix}
            x_i\\
            u_i\\
    \end{bmatrix}
    ^\top
    \right)^{-1}
    \begin{bmatrix}
        x_t \\
        \Kh_tx_t+\xi_t
    \end{bmatrix}
    \right) \convP 1
.\end{align*}
By \cref{eq: Gram matrix my symbol}:
\begin{equation*}
        \sum_{i=0}^{t-1}
    \begin{bmatrix}
            x_i\\
            u_i\\
    \end{bmatrix}
    \begin{bmatrix}
            x_i\\
            u_i\\
    \end{bmatrix}
    ^\top
    /t^\beta \log^{\alpha}(t)
=
    \left[
\begin{array}{cc}
I_n & 0\\
K & I_d\\
\end{array}
\right]\left[
\begin{array}{cc}
M_t & \Delta_t^\top\\
\Delta_t & \Delta_u\\
\end{array}
\right]\left[
\begin{array}{cc}
I_n & K^\top \\
0 & I_d\\
\end{array}
\right]
.\end{equation*}
Thus
\begin{align*}
\begin{split}
    &\begin{bmatrix}
        x_t \\
        \Kh_tx_t+\xi_t
    \end{bmatrix}^\top
    \left(
    \sum_{i=0}^{t-1}
    \begin{bmatrix}
            x_i\\
            u_i\\
    \end{bmatrix}
    \begin{bmatrix}
            x_i\\
            u_i\\
    \end{bmatrix}
    ^\top
    \right)^{-1}
    \begin{bmatrix}
        x_t \\
        \Kh_tx_t+\xi_t
    \end{bmatrix} t^\beta \log^{\alpha}(t)
    \\
    & \qquad =
\begin{bmatrix}
        x_t \\
        \Kh_tx_t+\xi_t
    \end{bmatrix}^\top
\left[
\begin{array}{cc}
I_n & K^\top \\
0 & I_d\\
\end{array}
\right]^{-1}
\left[
\begin{array}{cc}
M_t & \Delta_t^\top\\
\Delta_t & \Delta_u\\
\end{array}
\right]^{-1}
\left[
\begin{array}{cc}
I_n & 0\\
K & I_d\\
\end{array}
\right]^{-1}
    \begin{bmatrix}
        x_t \\
        \Kh_tx_t+\xi_t
    \end{bmatrix}\\
& \qquad = 
    \begin{bmatrix}
        x_t \\
        \Kh_tx_t+\xi_t
    \end{bmatrix}^\top
\left[
\begin{array}{cc}
I_n & -K^\top \\
0 & I_d\\
\end{array}
\right]
\left[
\begin{array}{cc}
(M_t-\Delta_t^\top \Delta_u^{-1} \Delta_t)^{-1} 
& 
-(M_t-\Delta_t^\top \Delta_u^{-1} \Delta_t)^{-1}\Delta_t^\top \Delta_u^{-1}\\
-((M_t-\Delta_t^\top \Delta_u^{-1} \Delta_t)^{-1}\Delta_t^\top \Delta_u^{-1})^\top & (\Delta_u - \Delta_t M_t^{-1} \Delta_t^\top)^{-1}
\end{array}
\right]\\
& \qquad \qquad \cdot 
\left[
\begin{array}{cc}
I_n & 0\\
-K & I_d\\
\end{array}
\right]
    \begin{bmatrix}
        x_t \\
        \Kh_tx_t+\xi_t
    \end{bmatrix}
    \qquad 
     \text{(by block matrix inversion)}\\
& \qquad =
    \begin{bmatrix}
        x_t \\
        \Kh_tx_t+\xi_t-Kx_t
    \end{bmatrix}^\top
    \left[
\begin{array}{cc}
(M_t-\Delta_t^\top \Delta_u^{-1} \Delta_t)^{-1} 
& 
-(M_t-\Delta_t^\top \Delta_u^{-1} \Delta_t)^{-1}\Delta_t^\top \Delta_u^{-1}\\
-((M_t-\Delta_t^\top \Delta_u^{-1} \Delta_t)^{-1}\Delta_t^\top \Delta_u^{-1})^\top & 
(\Delta_u - \Delta_t M_t^{-1} \Delta_t^\top)^{-1}\\
\end{array}
\right] \\
& \qquad \qquad \cdot
    \begin{bmatrix}
        x_t \\
        \Kh_tx_t+\xi_t-Kx_t
    \end{bmatrix} \\
& \qquad =
x_t^\top (M_t-\Delta_t^\top \Delta_u^{-1} \Delta_t)^{-1} x_t -
2x_t^\top (M_t-\Delta_t^\top \Delta_u^{-1} \Delta_t)^{-1}\Delta_t^\top \Delta_u^{-1} (\Kh_tx_t+\xi_t-Kx_t) \\
& \qquad \qquad +
(\Kh_tx_t+\xi_t-Kx_t)^\top(\Delta_u - \Delta_t M_t^{-1} \Delta_t^\top)^{-1}(\Kh_tx_t+\xi_t-Kx_t)
.\end{split}
\end{align*}
By \cref{eq:Cov xx}, \cref{eq: Delta t order}, \cref{eq: definition Delta u}:
\begin{align}
\label{eq: Mt Delta orders}
\begin{split}
      M_t =&    
        \log^{-\alpha}(t)t^{1-\beta} 
        \left(\sum_{p=0}^{\infty}L ^{p}\left(\sigma^2I_n + 1_{\{\beta=1,\alpha=0\}}\tau^2BB^\top\right)(L ^{p})^\top
        \right)
        (I_n + o(1)) \\
          M_t^{-1} =&    
        \log^{\alpha}(t)t^{-1+\beta}          \left(\sum_{p=0}^{\infty}L ^{p}\left(\sigma^2I_n + 1_{\{\beta=1,\alpha=0\}}\tau^2BB^\top\right)(L ^{p})^\top
        \right)^{-1}(I_n + o(1)) \\
    \Delta_t =& \calO_p(t^{1-3\beta/2}\log^{\frac{-3\alpha + 3}{2}}(t)) \\
    \Delta_u =& \frac{\tau^2}\beta  (I_d + o_p(1))  
.\end{split}
\end{align}
As a result, when $\beta > 1/2$ or $\beta = 1/2, \alpha > 3/2$
\begin{align*}
    \Delta_t^\top \Delta_u^{-1} \Delta_t =& \calO_p(t^{2-3\beta}\log^{-3\alpha + 3}(t))
    = o_p(t^{1-\beta}\log^{-\alpha}(t))
    \\
    (M_t - \Delta_t^\top \Delta_u^{-1} \Delta_t)^{-1} 
    =& M_t^{-1} (I_n-o_p(1))^{-1}= M_t^{-1} (I_n+o_p(1))
    \\
    \Delta_t M_t^{-1} \Delta_t^\top 
    =& \calO_p(t^{1-3\beta/2}\log^{\frac{-3\alpha + 3}{2}}(t)) \calO_p(t^{\beta-1} \log^{\alpha}(t)) \calO_p(t^{1-3\beta/2}\log^{\frac{-3\alpha + 3}{2}}(t))\\
    =& \calO_p(t^{1-2\beta}\log^{-2\alpha + 3}(t)) =  o_p(1)
    \\
    (\Delta_u - \Delta_t M_t^{-1} \Delta_t^\top)^{-1}
    =&\Delta_u^{-1}(I_d+ o_p(1))
.\end{align*}
Notice by \cref{lemma: Hi prob bounds in theorem 2}, $\Kh_t-K = \calO_p(t^{-\frac{\beta}{2}} \log^{\frac{-\alpha + 1}{2}}(t))$. Then
\begin{align*}
    &\begin{bmatrix}
        x_t \\
        \Kh_tx_t+\xi_t
    \end{bmatrix}^\top
    \left(
    \sum_{i=0}^{t-1}
    \begin{bmatrix}
            x_i\\
            u_i\\
    \end{bmatrix}
    \begin{bmatrix}
            x_i\\
            u_i\\
    \end{bmatrix}
    ^\top
    \right)^{-1}
    \begin{bmatrix}
        x_t \\
        \Kh_tx_t+\xi_t
    \end{bmatrix} t^\beta \log^{\alpha}(t)
    \\
    & \qquad =
    x_t^\top (M_t-\Delta_t^\top \Delta_u^{-1} \Delta_t)^{-1} x_t +
2x_t^\top (M_t-\Delta_t^\top \Delta_u^{-1} \Delta_t)^{-1}\Delta_t^\top \Delta_u^{-1} (\Kh_tx_t+\xi_t-Kx_t) \\
& \qquad \qquad +
(\Kh_tx_t+\xi_t-Kx_t)^\top(\Delta_u - \Delta_t M_t^{-1} \Delta_t^\top)^{-1}(\Kh_tx_t+\xi_t-Kx_t) \\
& \qquad =
x_t^\top M_t^{-1}(I_n+o_p(1)) x_t +
2x_t^\top M_t^{-1}(I_n+o_p(1))\calO_p(t^{1-3\beta/2}\log^{\frac{-3\alpha + 3}{2}}(t)) \Delta_u^{-1} (\calO_p(t^{-\frac{\beta}{2}} \log^{\frac{-\alpha + 1}{2}}(t))x_t+\xi_t) \\
& \qquad \qquad +
(\calO_p(t^{-\frac{\beta}{2}} \log^{\frac{-\alpha + 1}{2}}(t))x_t+\xi_t)^\top
\Delta_u^{-1}(I_d+ o_p(1)) 
(\calO_p(t^{-\frac{\beta}{2}} \log^{\frac{-\alpha + 1}{2}}(t))x_t+\xi_t) 
.\end{align*}
~\paragraph{Quadratic terms of $x_t$}
Let us first consider all those quadratic terms of $x_t$:
\begin{itemize}
    \item $ x_t^\top M_t^{-1}(I_n+o_p(1)) x_t.$
    \item 
    \begin{align*}
        &2x_t^\top M_t^{-1}(I_n+o_p(1))\calO_p(t^{1-3\beta/2}\log^{\frac{-3\alpha + 3}{2}}(t)) \Delta_u^{-1} \calO_p(t^{-\frac{\beta}{2}} \log^{\frac{-\alpha + 1}{2}}(t))x_t \\
        & \qquad =  2x_t^\top M_t^{-1}(I_n+o_p(1))\calO_p(t^{1-2\beta}\log^{\frac{-4\alpha + 4}{2}}(t)) x_t \\
        & \qquad = x_t^\top M_t^{-1}o_p(1) x_t
    .\end{align*}
    \item 
    \begin{align*}
        &x_t^\top \calO_p\left(t^{-\frac{\beta}{2}} \log^{\frac{-\alpha + 1}{2}}(t)\right) \Delta_u^{-1}(I_d+ o_p(1)) \calO_p\left(t^{-\frac{\beta}{2}} \log^{\frac{-\alpha + 1}{2}}(t)\right)x_t \\
        & \qquad = x_t^\top \calO_p\left(t^{-\beta} \log^{-\alpha + 1}(t)\right) x_t \\
        & \qquad = x_t^\top M_t^{-1} t^{1-\beta}\log^{-\alpha}(t)\calO_p\left(t^{-\beta} \log^{-\alpha + 1}(t)\right) x_t \qquad \text{(by \cref{eq: Mt Delta orders})}\\
        & \qquad = x_t^\top M_t^{-1} \calO_p\left(t^{1-2\beta} \log^{-2\alpha + 1}(t)\right) x_t \\
        & \qquad = x_t^\top M_t^{-1} o_p(1) x_t 
    .\end{align*}

\end{itemize}


Thus the later two items are dominated by the first term, and the quadratic terms of $x_t$ can be summarized by $x_t^\top M_t^{-1}(I_n+o_p(1)) x_t = x_t^\top M_t^{-1} x_t(1+o_p(1))$. 
~\paragraph{Quadratic terms of $\xi_t$} That is already in a simple single item form, so we just keep it as $\xi_t^\top \Delta_u^{-1} (I_d+o_p(1))\xi_t = \xi_t^\top \Delta_u^{-1}\xi_t (1+o_p(1))$.
~\paragraph{Cross terms between $x_t$ and $\xi_t$}
Finally consider the cross terms of $x_t$ and $\xi_t$:
\begin{align*}
    & 2x_t^\top M_t^{-1}(I_n+o_p(1))\calO_p(t^{1-3\beta/2}\log^{\frac{-3\alpha + 3}{2}}(t)) \Delta_u^{-1} \xi_t 
    +
     2\calO_p(t^{-\frac{\beta}{2}} \log^{\frac{-\alpha + 1}{2}}(t))x_t^\top
\Delta_u^{-1}(I_d+ o_p(1)) \xi_t\\
& \qquad = 2x_t^\top \calO_p(t^{-\frac{\beta}{2}} \log^{\frac{-\alpha + 3}{2}}(t)) \xi_t
+ 2x_t^\top \calO_p(t^{-\frac{\beta}{2}} \log^{\frac{-\alpha + 1}{2}}(t)) \xi_t \qquad \text{(by \cref{eq: Mt Delta orders})}\\
& \qquad =  x_t^\top \calO_p(t^{-\frac{\beta}{2}}\log^{\frac{-\alpha + 3}{2}}(t)) \xi_t \\
& \qquad =  x_t^\top o_p(t^{\frac{\beta-1}{2}}\log^{\frac{\alpha}{2}}(t)) \xi_t \qquad \text{(because $\beta > 1/2$ or $\beta = 1/2$ and $\alpha > 3/2$)}\\
& \qquad = x_t^\top M_t^{-1/2}o_p(1) \Delta_u^{-1/2}\xi_t \qquad \text{(by \cref{eq: Mt Delta orders})}\\
& \qquad \le o_p(1) \norm{x_t^\top M_t^{-1/2}} \norm{\Delta_u^{-1/2}\xi_t } 
 \\
& \qquad \le o_p(1) \left(x_t^\top M_t^{-1} x_t + \xi_t^\top \Delta_u^{-1} \xi_t \right)
,\end{align*}
which is dominated by the quadratic part.
To sum up, we have
\begin{equation*}
\begin{split}
    &\begin{bmatrix}
        x_t \\
        \Kh_tx_t+\xi_t
    \end{bmatrix}^\top
    \begin{bmatrix}
        \sum_{i=0}^{t-1}x_ix_i^\top & \sum_{i=1}^{t-1}x_iu_i^\top \\
        \sum_{i=0}^{t-1}u_ix_i^\top & \sum_{i=1}^{t-1}u_iu_i^\top \\
    \end{bmatrix}^{-1}
    \begin{bmatrix}
        x_t \\
        \Kh_tx_t+\xi_t
    \end{bmatrix} t^\beta \log^{\alpha}(t)
    \\
& \qquad =
(x_t^\top M_t^{-1} x_t +
\xi_t^\top\Delta_u ^{-1}\xi_t)(1+o_p(1)) \\
& \qquad =
\left(x_t^\top \log^{\alpha}(t)t^{-1+\beta}          \left(\sum_{p=0}^{\infty}L ^{p}\left(\sigma^2I_n + 1_{\{\beta=1,\alpha=0\}}\tau^2BB^\top\right)(L ^{p})^\top
        \right)^{-1}  x_t
+
\xi_t^\top \frac{\beta}{\tau^2}\xi_t\right)(1+o_p(1)) 
.\end{split}
\end{equation*}
In other words
\begin{equation*}
\begin{split}
           & 
    t\sigma^2
\begin{bmatrix}
        x_t \\
        \Kh_tx_t+\xi_t
    \end{bmatrix}^\top
    \begin{bmatrix}
        \sum_{i=0}^{t-1}x_ix_i^\top & \sum_{i=1}^{t-1}x_iu_i^\top \\
        \sum_{i=0}^{t-1}u_ix_i^\top & \sum_{i=1}^{t-1}u_iu_i^\top \\
    \end{bmatrix}^{-1}
    \begin{bmatrix}
        x_t \\
        \Kh_tx_t+\xi_t
    \end{bmatrix}\\
& \qquad =
\left(    x_t^\top   
\left( \sum_{p=0}^{\infty}L ^{p}\left(I_n + 1_{\{\beta=1,\alpha=0\}}\frac{\tau^2}{\sigma^2}BB^\top\right)(L ^{p})^\top\right)^{-1}
x_t  
    +
     \frac{\beta \sigma^2}{\tau^2}
     t^{1-\beta} \log^{-\alpha}(t)
     \lnorm{\xi_t}^2\right)(1+o_p(1))
.\end{split}
\end{equation*}

\end{proof}

\subsection{Lemmas in \cref{section: The proof of one_epoch_estimate}}

\subsubsection{The proof of \cref{lemma:lwm}}
\label{The proof of lemma:lwm}

\begin{lemma*}[A slightly different version of Theorem C.2 in \citet{dean2018regret}]
Fixing $\delta \in (0,\frac{(n+d)\xi^2}{2}]$, for every $T$, $k$, $\nu$, and $\xi$ such that $\{z_t\}_{t = 0}^T$ satisfies the $(k, \nu, \xi)$-BMSB and 
\begin{equation*}
T/k \ge \frac{10(n+d)}{\xi^2}   \log\left(\frac{100(n+d)\sum_{t = 1}^T\Tr(\E  z_t z_t^\top)}{T \nu^{2}\xi^2\delta^{1 + \frac1{n+d}}}    \right) 
.\end{equation*}

the estimate $\hat{\Theta}_T$ defined in \cref{eq:ols_M} satisfies the following statistical rate
\begin{equation*}
\P\left[\lnorm{\hat{\Theta}_T-\Theta}_{2} >\frac{90\sigma}{\xi\nu}\sqrt{\frac{n+d}{T}\left(1  + \log\left(\frac{10(n+d)\sum_{t = 1}^T\Tr(\E  z_t z_t^\top)}{T\delta^{1 + \frac1{n+d}} \nu^{2}\xi}    \right) \right)} \right] \le  3\delta.
\end{equation*}

\end{lemma*}

First let us review the main theorem in \citep{simchowitz2018learning}. \cref{lemma:lwm} is actually a corollary of that.
   To capture the excitation behavior observed in the case of linear systems we introduce a general martingale small-ball condition which quantifies the growth of the covariates $X_t$ for vectors (notice that this is different from \cref{def:BMSB condition}).
	
	\begin{defn}[BMSB condition 2]\label{def:bmsb}  
	    
	     Given an $\{\calF_t\}_{t \ge 1}$-adapted random process $\{X_t\}_{t \ge 1}$ taking values in $\R^d$, we say that it satisfies the $(k,\Gamsb,\xi)$-matrix block martingale small-ball (BMSB) condition for $\Gamsb \succ 0$ if, for any $w\in \calS^{d-1}$ and $j \ge 0$, $\frac{1}{k}\sum_{i=1}^{k} \P( |\langle w, X_{j+i}\rangle | \ge \sqrt{w^\top \Gamsb w} | \calF_{j}) \ge \xi \as$ 
    \end{defn}



\begin{theorem}[Theorem 2.4 in \citet{simchowitz2018learning}]
\label{main_thm_simchowitz2018} Fix $\delta \in (0,1)$, $T \in \N$ and $0 \prec \Gamsb \preceq \Gambar$. Then if $\{z_t,x_{t+1}\}_{t \ge 0} \in (\R^{d+n} \times \R^n)^{T}$ is a random sequence such that (a) $x_{t+1} = \Theta z_t + \varepsilon_t$, where $\varepsilon_t | \calF_t$ is $\sigma^2$-sub-Gaussian and mean zero, (b) $z_0,\dots,z_{T-1}$ satisfies the $(k,\Gamsb,\xi)$-small ball condition, and (c) such that $\P[\sum_{t = 0}^{T-1}  z_t z_t^\top \npreceq T\Gambar] \le \delta$. Then if 
\begin{align*}
T \geq  \frac{10k}{\xi^2}\left(\log \left(\frac{1}{\delta}\right) + 2(d+n)\log(10/ \xi) +  \log \det (\Gambar \Gamsb^{-1}) \right),
\end{align*}
we have $\hat{\Theta}_T$ defined in \cref{eq:ols_M} satisfies the following statistical rate
\begin{align*}
\P\left[\lnorm{\hat\Theta_T-\Theta} >\frac{90\sigma}{\xi}\sqrt{\frac{n + (n+d)\log \frac{10}{\xi} + \log \det \Gambar \Gamsb^{-1} + \log\left(\frac{1}{\delta}\right)}{T\sigma_{\min}(\Gamsb)}} \right] \le  3\delta.
\end{align*}
\end{theorem}

Now the main task is to translate this theorem to \cref{lemma:lwm}. First we need to derive the (a), (b), (c) three conditions from the assumptions in \cref{lemma:lwm}. Let us check the conditions one by one.
\paragraph{Condition (a)}
\cref{main_thm_simchowitz2018} states the model should be in the form of $x_t = \Theta z_t + \varepsilon_t$, where $\varepsilon_t | \calF_t$ is $\sigma^2$-sub-Gaussian and mean zero. 
It is obvious that the system noise satisfy the sub-Gaussian and mean zero condition.

\paragraph{Condition (b)}
$z_1,\dots,z_T$ satisfies the $(k,\Gamsb,\xi)$-small ball condition. 



Based on \cref{def:bmsb}, if we pick $\Gamma_{sb} = \nu^2 I_{\statedim + \inputdim}$, then the condition becomes 
\begin{equation}
\label{eq: condition b}
    \frac{1}{k}\sum_{i=1}^{k} \P( |\langle w, z_{j+i}\rangle| \ge \sqrt{w^\top \Gamsb w} = \nu| \calF_{j}) \ge \xi \as
\end{equation}
Since we already assume $\{z_t\}_{t = 0}^T$ satisfies the $(k, \nu, \xi)$-BMSB (see \cref{def:BMSB condition}) in \cref{lemma:lwm}, \cref{eq: condition b} holds by definition.

\paragraph{Condition (c)}
We need to show that $\P[\sum_{t = 0}^{T-1}  z_t z_t^\top \npreceq T\Gambar] \le \delta$ for some choice $\Gambar$. Let us take 
\begin{equation}
\label{eq: defn Gambar}
    \Gambar = \frac{(n+d)\E\{\sum_{t = 0}^{T-1}  z_t z_t^\top\}}{T\delta}  \succ 0.
\end{equation}

First we need to show that $\Gambar = \frac{(n+d)\E\{\sum_{t = 0}^{T-1}  z_t z_t^\top\}}{T\delta} \succeq \Gamma_{sb}$, and we can prove this from \cref{eq: condition b}:
\[\text{For any $0 \le j \le T-k$,} \quad \frac{1}{k}\sum_{i=1}^{k} \P( |\langle w, z_{j+i}\rangle| \ge  \nu| \calF_{j}) \ge \xi.\] 
From a high level perspective, this equation allows us to have a lower bound on the minimum eigenvalue of $\E\{\sum_{t = 0}^{T-1}  z_t z_t^\top\}$, and then we can choose a $\delta$ small enough so that $\Gambar \succeq \Gamma_{sb} = \nu^2 I_{\statedim + \inputdim}$. 
By Markov inequality, for any $0 \le j \le T-k$,
\[\frac{\frac{1}{k}\sum_{i=1}^{k}\E |\langle w, z_{j+i}\rangle|}{\nu} \ge \xi.\]
This is equivalent to
\[\left(\frac{1}{k}\sum_{i=1}^{k}\E |\langle w, z_{j+i}\rangle|\right)^2 \ge \xi^2\nu^2.\]
By Cauchy--Schwarz inequality:
\[\frac{1}{k} \sum_{i=1}^{k}\E |\langle w, z_{j+i}\rangle|^2 \ge \frac{1}{k} \sum_{i=1}^{k}\E^2 |\langle w, z_{j+i}\rangle| \ge \left(\frac{1}{k}\sum_{i=1}^{k}\E |\langle w, z_{j+i}\rangle|\right)^2 \ge \xi^2\nu^2.\]
Thus $\frac{1}{k} \sum_{i=1}^{k}\E |\langle w, z_{jk+i}\rangle|^2 \ge \xi^2\nu^2$. By summing up this inequality with $j = 0, 1, \cdots, \lfloor\frac{T-1}{k}\rfloor - 1$, we have
\[\frac{1}{\lfloor\frac{T-1}{k}\rfloor}\sum_{j=0}^{\lfloor\frac{T-1}{k}\rfloor - 1}\frac{1}{k}\left(\sum_{i=1}^{k}\E |\langle w, z_{jk+i}\rangle|^2\right) \ge \xi^2\nu^2.\]
We can clean up the summation by merging $\sum_j$ and $\sum_i$ into one summation:
\[\frac{1}{k\lfloor\frac{T-1}{k}\rfloor}\sum_{t=1}^{k\lfloor\frac{T-1}{k}\rfloor}\E |\langle w, z_{t}\rangle|^2 \ge \xi^2\nu^2.\]

Recall that $w$ is any vector in $\calS^{d-1}$, so the above equation can be translated into
\begin{align*}
\xi^2\nu^2 \le& \min_{w \in \calS^{d-1}}\frac{1}{k\lfloor\frac{T-1}{k}\rfloor}\sum_{t=1}^{k\lfloor\frac{T-1}{k}\rfloor}\E |\langle w, z_{t}\rangle|^2 
\\
=&\; \E\left(\frac{1}{k\lfloor\frac{T-1}{k}\rfloor}\sum_{t=1}^{k\lfloor\frac{T-1}{k}\rfloor} z_{t}z_{t}^T\right)w \\
=& \min_{w \in \calS^{d-1}} w^T \E\left(\frac{1}{k\lfloor\frac{T-1}{k}\rfloor}\sum_{t=1}^{k\lfloor\frac{T-1}{k}\rfloor} z_{t}z_{t}^T\right)w \\
\le&\; \sigma_{\min}\left(\E\left(\sum_{t = 0}^{T-1}  z_t z_t^\top\right)/(k\lfloor\frac{T-1}{k}\rfloor)\right) 
.\end{align*}

This means
\begin{align*}
    \lambdamin{\Gambar} =& \lambdamin{\frac{(n+d)\E\left(\sum_{t = 0}^{T-1}  z_t z_t^\top\right)}{T\delta} } \\
    =& \lambdamin{\frac{(n+d)\E\left(\sum_{t = 0}^{T-1}  z_t z_t^\top/(k\lfloor\frac{T-1}{k}\rfloor)\right)}{T\delta} (k\lfloor\frac{T-1}{k}\rfloor)} \\
    \ge& \frac{(n+d)\xi^2\nu^2}{T\delta} k\lfloor\frac{T-1}{k}\rfloor \qquad  \\
    \ge& \frac{(n+d)\xi^2\nu^2}{T\delta} \frac{T}{2} \qquad  \text{(achieved when T is even and  $k = T/2$)}\\
    =& \frac{(n+d)\xi^2\nu^2}{2\delta} 
.\end{align*}
We wish to have $\frac{(n+d)\xi^2\nu^2}{2\delta} \ge \nu^2$ so that $\lambdamin{\Gambar} \ge \nu^2$ and $\Gambar \succeq \Gamma_{sb} = \nu^2 I_{n+d}$. One sufficient condition is 
\begin{equation*}
\delta \le \frac{(n+d)\xi^2}{2}
.\end{equation*}

Next we need to show $\P[\sum_{t = 0}^{T-1}  z_t z_t^\top \npreceq T\Gambar] \le \delta$. For simplicity denote $Z_T = \sum_{t = 0}^{T-1}  z_t z_t^\top$, which is a positive semi-definite matrix. 
\begin{align*}
    \P[\sum_{t = 0}^{T-1}  z_t z_t^\top \npreceq T\Gambar] 
    =& \P[Z_T \npreceq \frac{\E\{Z_T\}(n+d)}{\delta} ] \qquad \text{(by \cref{eq: defn Gambar})} \\
    =& \P[\E^{-1/2}\left(Z_T\right)Z_T\E^{-1/2}\left(Z_T\right) \npreceq \frac{I_{\statedim + \inputdim}(n+d)}{\delta}] \\
    =& \P[\lambda_{\max}  \{\E^{-1/2}\left(Z_T\right)Z_T\E^{-1/2}\left(Z_T\right)\} \ge \frac{(n+d)}{\delta}] \\
    \le& \P[\Tr  \{\E^{-1/2}\left(Z_T\right)Z_T\E^{-1/2}\left(Z_T\right)\} \ge \frac{(n+d)}{\delta}] \\
    \le& \E[\Tr  \{\E^{-1/2}\left(Z_T\right)Z_T\E^{-1/2}\left(Z_T\right)\}] \delta/(n+d)  \quad \text{ (by Markov inequality)}\\
    =& \Tr [\E \{\E^{-1/2}\left(Z_T\right)Z_T\E^{-1/2}\left(Z_T\right)\}] \delta/(n+d)  \\
    =& \Tr [I_{n+d}] \delta/(n+d)  \\
    =& \delta  
.\end{align*}

\paragraph{Result}
Now that we verified all conditions of \cref{main_thm_simchowitz2018}, we can now translate the conclusion of \cref{main_thm_simchowitz2018} into our setting. \cref{main_thm_simchowitz2018} requires

\begin{align*}
T \geq&  \frac{10k}{\xi^2}\left(\log \left(\frac{1}{\delta}\right) + 2(d+n)\log(10/ \xi) +  \log \det (\Gambar \Gamsb^{-1}) \right) 
.\end{align*}

First by our choice of $\Gamsb$ and $\Gambar$ we have

\begin{equation}
\label{eq:Gambar Gamsb inv}
\begin{aligned}
\log \det (\Gambar \Gamsb^{-1}) =& \log \det \left(\frac{(n+d)\E\{\sum_{t = 0}^{T-1}  z_t z_t^\top\}}{T\delta} \nu^{-2}\right) \\
=& \log\left( \left(\frac{(n+d)}{T\delta \nu^{2}}\right)^{n+d} \det \left(\E\{\sum_{t = 0}^{T-1}  z_t z_t^\top\} \right)\right) \\
\le& \log\left(\left(\frac{(n+d)}{T\delta \nu^{2}}\right)^{n+d}  \left(\sum_{t = 0}^{T-1}\Tr(\E  z_t z_t^\top) \right)^{n+d} \right) \\
=& (n+d)\log\left(\frac{(n+d)}{T\delta \nu^{2}}  \sum_{t = 0}^{T-1}\Tr(\E  z_t z_t^\top)  \right) 
.\end{aligned}
\end{equation}
With this in hand, we know that

\begin{align*}
    &\frac{10k}{\xi^2}\left(\log \left(\frac{1}{\delta}\right) + 2(d+n)\log(10/ \xi) +  \log \det (\Gambar \Gamsb^{-1}) \right)  \\
    & \qquad \le \frac{10k}{\xi^2}\left(\log \left(\frac{1}{\delta}\right) + 2(d+n)\log(10/ \xi) +  (n+d)\log\left(\frac{(n+d)}{T\delta \nu^{2}}  \sum_{t = 0}^{T-1}\Tr(\E  z_t z_t^\top)  \right)\right) \\
    & \qquad = \frac{10(n+d)k}{\xi^2}\left(\log \left(\delta^{-\frac1{n+d}}\right) + \log(100/ \xi^2) +  \log\left(\frac{(n+d)}{T\delta \nu^{2}}  \sum_{t = 0}^{T-1}\Tr(\E  z_t z_t^\top)  \right)\right) \\
    & \qquad = \frac{10(n+d)k}{\xi^2}   \log\left(\frac{100(n+d)\sum_{t = 0}^{T-1}\Tr(\E  z_t z_t^\top)}{T \nu^{2}\xi^2\delta^{1 + \frac1{n+d}}}    \right)
.\end{align*}
Thus one sufficient condition for the requirement in \cref{main_thm_simchowitz2018} is
\begin{align*}
T/k \ge \frac{10(n+d)}{\xi^2}   \log\left(\frac{100(n+d)\sum_{t = 0}^{T-1}\Tr(\E  z_t z_t^\top)}{T \nu^{2}\xi^2\delta^{1 + \frac1{n+d}}}    \right) 
.\end{align*}

Finally we need to translate the conclusion of \cref{main_thm_simchowitz2018}:
\begin{align*}
\P\left[\lnorm{\hat\Theta_T-\Theta} >\frac{90\sigma}{\xi}\sqrt{\frac{n + (n+d)\log \frac{10}{\xi} + \log \det \Gambar \Gamsb^{-1} + \log\left(\frac{1}{\delta}\right)}{T\sigma_{\min}(\Gamsb)}} \right] \le  3\delta.
\end{align*}
By \cref{eq:Gambar Gamsb inv} and $\Gamma_{sb} = \nu^2 I_{n+d}$ we have
\begin{align*}
\P\left[\lnorm{\hat\Theta_T-\Theta} >\frac{90\sigma}{\xi}\sqrt{\frac{n + (n+d)\log \frac{10}{\xi} + (n+d)\log\left(\frac{(n+d)}{T\delta \nu^{2}}  \sum_{t = 0}^{T-1}\Tr(\E  z_t z_t^\top)  \right) + \log\left(\frac{1}{\delta}\right)}{T\nu^2}} \right] \le  3\delta.
\end{align*}

Notice that
\begin{align*}
&n + (n+d)\log \frac{10}{\xi} + (n+d)\log\left(\frac{(n+d)}{T\delta \nu^{2}}  \sum_{t = 0}^{T-1}\Tr(\E  z_t z_t^\top)  \right) + \log\left(\frac{1}{\delta}\right) \\
& \qquad \le  (n+d)\left(1 + \log \frac{10}{\xi} + \log\left(\frac{(n+d)}{T\delta \nu^{2}}  \sum_{t = 0}^{T-1}\Tr(\E  z_t z_t^\top)  \right) + \log\delta^{-\frac1{n+d}}\right) \\
& \qquad = (n+d)\left(1  + \log\left(\frac{10(n+d)\sum_{t = 0}^{T-1}\Tr(\E  z_t z_t^\top)}{T\delta^{1 + \frac1{n+d}} \nu^{2}\xi}    \right) \right) 
.\end{align*}
Combining this with the previous inequality we have
\begin{align*}
\P\left[\lnorm{\hat\Theta_T-\Theta} >\frac{90\sigma}{\xi\nu}\sqrt{\frac{n+d}{T}\left(1  + \log\left(\frac{10(n+d)\sum_{t = 0}^{T-1}\Tr(\E  z_t z_t^\top)}{T\delta^{1 + \frac1{n+d}} \nu^{2}\xi}    \right) \right)} \right] \le  3\delta.
\end{align*}

\subsubsection{The proof of \cref{lem:bmsb}}
\label{The proof of lem:bmsb}

\begin{lemma*}[Similar to Lemma C.3 in \citet{dean2018regret}]
If we assume \cref{asm:InitialStableCondition}, then apply \cref{alg:myAlg}, the process $\{z_t\}_{t \geq 0}^T $ satisfies the 
$(k, \nu, \xi)$-BMSB condition for
\begin{align*}
(k, \nu, \xi) = \left(1, \sqrt{\sigma_{\eta,T}^2\min\left(\frac{1}{2}, \frac{\sigma ^2}{2 \sigma ^2 C_K^2  + \tau^2} \right)} , \frac{3}{10}\right),
\end{align*}
where $\sigma_{\eta,T}^2 = \tau^2 T^{\beta-1}\log^\alpha(T)$.
\end{lemma*}

\begin{proof}

By \cref{def:BMSB condition} the statement means, for any $v \in \calS^{\statedim + \inputdim}$ and $0 \le t \le T-1$:
\begin{align*}
\P \left(|\langle v, z_{t+1} \rangle | \geq \sqrt{\sigma_{\eta,T}^2\min\left(\frac{1}{2}, \frac{\sigma ^2}{2 \sigma ^2 C_K^2  + \tau^2} \right)}  \Bigg| \calF_t \right) \geq  3/10
.\end{align*}

Recall that
\begin{equation*}
    x_{t+1} = Ax_t + Bu_t + \varepsilon_t
.\end{equation*}
\begin{equation*}
    u_{t+1} = \Kh_{t+1}x_{t+1} + \eta_{t+1} = \Kh_{t+1}(Ax_t + Bu_t + \varepsilon_t) + \eta_{t+1}
.\end{equation*}

Denote the filtration $\calF_t = \sigma(x_0, \eta_0, \varepsilon_0 \ldots, \eta_{t - 1}, \varepsilon_{t - 1}, \eta_t) = \sigma(x_0, u_0, x_1, \cdots, x_t, u_t)$. It is clear that the process $\{z_t\}_{t \geq 0}$ is $\{\calF_t\}_{t \geq 0}$-adapted.

Recall that $\Kh_{t+1}$ is decided by $\Ah_{t}, \Bh_{t}$ in \cref{alg:myAlg}, where our  estimator $\Ah_{t}, \Bh_{t}$ is designed to be only dependent on $x_0, u_0, x_1, \cdots, u_{t-1}, x_t$, which means 
\[\Kh_{t+1} \in \calF_t = \sigma(x_0, u_0, x_1, \cdots, x_t, u_t).\]
For all $t \geq 1$, denote
\begin{align*}
\xi_{t+1} &:= \Kh_{t+1}(Ax_t + Bu_t) \in \calF_t 
.\end{align*}

Now we are ready to prove \cref{lem:bmsb}. We have
\begin{align*}
\begin{bmatrix}
x_{t + 1}\\
u_{t + 1}
\end{bmatrix} = \begin{bmatrix}
A x_{t} + B u_{t} \\
\xi_{t + 1}
\end{bmatrix} +
\begin{bmatrix}
I_\statedim & 0 \\
\Kh_{t+1} & I_\inputdim
\end{bmatrix}\begin{bmatrix}
\varepsilon_t \\
\eta_{t + 1}
\end{bmatrix}
.\end{align*}


Given $\calF_t$, 
$\begin{bmatrix}
x_{t + 1}\\
u_{t + 1}
\end{bmatrix}$ 
only has randomness in 
$\begin{bmatrix}
I_\statedim & 0 \\
\Kh_{t+1} & I_\inputdim
\end{bmatrix}\begin{bmatrix}
\varepsilon_t \\
\eta_{t + 1}
\end{bmatrix}$, where
$\begin{bmatrix}
I_\statedim & 0 \\
\Kh_{t+1} & I_\inputdim
\end{bmatrix}$ is fixed given $\calF_t$, and 
$\begin{bmatrix}
\varepsilon_t \\
\eta_{t + 1}
\end{bmatrix}$
follows $\calN \left(0, 
\begin{bmatrix}
\sigma ^2 I_\statedim & 0\\
0 & \sigma_{\eta,t+1}^2 I_\inputdim
\end{bmatrix} \right)$. That implies

\begin{align*}
\begin{bmatrix}
x_{t + 1}\\
u_{t + 1}
\end{bmatrix} \Bigg| \calF_t  \sim&
\calN\left(
\begin{bmatrix}
A x_{t} + B u_{t} \\
\xi_{t + 1}
\end{bmatrix},
\begin{bmatrix}
\sigma ^2 I_\statedim & \sigma ^2 \Kh_{t+1}^\top \\
\sigma ^2 \Kh_{t+1} & \sigma ^2 \Kh_{t+1}\Kh_{t+1}^\top  + \sigma_{\eta,t+1}^2 I_\inputdim
\end{bmatrix}
\right)
.\end{align*}

Denote $\mu_{z, t+1}$ and $\Sigma_{z, t+1}$ as the mean and covariance of this multivariate normal distribution. Recall that we denoted $z_{t+1} = \begin{bmatrix}
x_{t+1} \\
u_{t+1}
\end{bmatrix}$.
Let $v \in \calS^{\statedim + \inputdim}$ and then $\langle v, z_{t+1} \rangle\Bigg| \calF_t  \sim \calN(\langle v, \mu_{z, t+1} \rangle, v^\top \Sigma_{z, t+1} v)$. Therefore,

\begin{equation}
\label{eq:v zt 3 over 10}
\begin{aligned}
\P \left(|\langle v, z_{t+1} \rangle | \geq \sqrt{\sigma_{\min}(\Sigma_{z, t+1})} \Bigg| \calF_t  \right) &\geq \P\left(|\langle v, z_{t+1} \rangle | \geq \sqrt{ v^\top \Sigma_{z, t+1} v}  \Bigg| \calF_t  \right)\\
&\geq \P\left(|\langle v, z_{t+1} - \mu_{z, t+1} \rangle | \geq \sqrt{ v^\top \Sigma_{z, t+1} v} \Bigg| \calF_t \right)  \\
&\geq 3/10.
\end{aligned}
\end{equation}
Here we used the fact that for any $\mu, \sigma^2 \in \R$ and $\omega \sim \calN(0, \sigma^2)$, we have:
\begin{align*}
\P(|\mu + \omega| \geq \sigma) \geq \P(|\omega| \geq \sigma) \geq 3/10.
\end{align*}

Recall in \cref{alg:myAlg}, we force all our controllers $\Kh_t$ to have norm $\norm{\Kh_t} \le C_K$, where $C_K$ is a constant. 
Then, by a simple argument based on a Schur complement (Lemma~\ref{lem:lambda_min_block_matrix}):
\begin{align*}
\sigma_{\min}(\Sigma_{z, t+1}) 
\geq &\sigma_{\eta,t}^2 \min\left(\frac{1}{2}, \frac{\sigma^2}{2 \ltwonorm{\Kh_{t+1} \sigma^2 \Kh_{t+1}^\T} + \sigma_{\eta,t}^2} \right) \\
\geq& \sigma_{\eta,t}^2 \min\left(\frac{1}{2}, \frac{\sigma ^2}{2 \sigma ^2 C_K^2  + \sigma_{\eta,t}^2} \right) \\
\ge& \sigma_{\eta,T}^2\min\left(\frac{1}{2}, \frac{\sigma ^2}{2 \sigma ^2 C_K^2  + \tau^2} \right).
\end{align*}
The desired conclusion directly follows:
\begin{align*}
&\P \left(|\langle v, z_{t+1} \rangle | \geq \sqrt{\sigma_{\eta,T}^2\min\left(\frac{1}{2}, \frac{\sigma ^2}{2 \sigma ^2 C_K^2  + \tau^2} \right)}  \Bigg| \calF_t \right) \\
& \qquad \geq 
\P \left(|\langle v, z_{t+1} \rangle | \geq \sqrt{\sigma_{\min}(\Sigma_{z, t+1})}  \Bigg| \calF_t \right) 
 \\
& \qquad \geq  3/10 \qquad \text{ by \cref{eq:v zt 3 over 10}}
.\end{align*}

\end{proof}

~\paragraph{Schur complement}
\begin{lemma}[Lemma F.1 in \citet{mania2019certainty}]
\label{lem:lambda_min_block_matrix}
Let $\Sigma$ be a $\statedim \times \statedim$ positive-definite matrix and let $K $ be a real $\inputdim \times \statedim $ matrix.
Then, for any $\sigma_u \in \R$ we have that
\begin{align*}
  \sigma_{\min}\left(\bmattwo{\Sigma}{\Sigma K ^\T}{K  \Sigma}{K  \Sigma K ^\T + \sigma_u^2 I}\right) &\geq
  \sigma_u^2 \min\left(\frac{1}{2}, \frac{\sigma_{\min}(\Sigma)}{2 \ltwonorm{K  \Sigma K ^\T} + \sigma_u^2} \right) \:.
\end{align*}
\end{lemma}

\subsubsection{The proof of \cref{lem:bound_covariance}}
\label{The proof of lem:bound_covariance}

\begin{lemma*}[Similar to Lemma C.4 in \citet{dean2018regret}]
If we assume \cref{asm:InitialStableCondition}, then apply \cref{alg:myAlg}, the process $\{z_t\}_{t \geq 0}^T $ satisfies
\begin{equation*}
\begin{split}
\sum_{t = 0}^{T - 1} \Tr \left( \E z_t z_t^\T\right) = \calO(T\log^2(T))
\end{split}
.\end{equation*}
\end{lemma*}

\begin{proof}

Now, note that
\[\Tr \left( \E z_t z_t^\T\right) = \E \left( \Tr z_t z_t^\T\right) = \E \norm{z_t}^2 = \E \left(\norm{x_t}^2 + \norm{u_t}^2\right).\]
Since $\norm{u_t} = \norm{\Kh_t x_t + \eta_t} \le \norm{\Kh_t} \norm{ x_t} + \norm{\eta_t} \le C_K\norm{ x_t} + \norm{\eta_t}$, we will show that if we can bound $\norm{ x_t}$, then we can also get a bound for $\norm{u_t}$ in the same order. Next we will focus on deriving the bound for $\norm{ x_t}$.

Define $C_{x,t} := C_x \log(t)$. Since $\rho(A+B K_0) < 1$, there exists some integer $m$ that $\norm{(A+B K_0)^m} < (\frac{\rho(A+B K_0) + 1}2)^m$. Let us denote $\rho := \frac{\rho(A+B K_0) + 1}2 < 1$ just for this \cref{lem:bound_covariance}. 

For each $ t > m+1$, one of the following two statement must be true:
\begin{itemize}
    \item $\norm{x_{t-i}} > C_{x,t-i}, (i=2,\cdots, m+1)$.
    \item  $\exists i \in \{2,\cdots, m+1\}$, which satisfies $\norm{x_{t-i}} \le C_{x,t-i}$.
\end{itemize}
We can derive an upper bound for $\norm{x_t}$ in both cases, and thus have an upper bound for every $\norm{x_t}$ by adding up those two bounds in two different cases.
\begin{enumerate}
    \item If $\norm{x_{t-i}} > C_{x,t-i}, (i=2,\cdots, m+1)$, recall that if $\norm{x_{t}} > C_{x,t}$, then we assert our controller in the next step to be probing noise: $u_{t+1} = K_0x_{t+1} + \eta_{t+1}$. By assumption we already had $\norm{x_{k}} > C_{x,k}$, for $k= t-m-1, t-m, \cdots, t-2$. That means we have a consecutive $m$ steps of probing noise with $u_{k} = K_0x_{k} + \eta_{k}$, for $k= t-m, t-(m-1), \cdots, t-1$. Now we have
    \[x_{k+1} = (A+B K_0) x_{k} + B\eta_{k} + \varepsilon_k , \quad \text{for} \quad k=t-m, t-(m-1), \cdots, t-1 .\]
    That is
    \[x_{t} = (A+B K_0)^m x_{t-m} + \sum_{k=0}^{m-1}(A+B K_0)^{k}(B\eta_{t-1-k} + \varepsilon_{t-1-k}).\]
    which implies
    \begin{equation}
    \label{eq:lemma3Cond1}
    \norm{x_{t}} \le \norm{(A+B K_0)^m} \norm{x_{t-m}} + \sum_{k=0}^{m-1}\norm{(A+B K_0)^{k}}\norm{(B\eta_{t-1-k} + \varepsilon_{t-1-k})}
    .\end{equation}

    \item If $\exists i \in \{2,\cdots, m+1\}$, which satisfies $\norm{x_{t-i}} \le C_{x,t-i}, (i=2,\cdots, m+1)$, then consider the following relationship
    \begin{align*}
        x_t &= A x_{t-1} + B u_{t-1} + \varepsilon_{t-1} \\
        &= (A+B \Kh_{t-1})x_{t-1} + B\eta_{t-1} + \varepsilon_{t-1}
    .\end{align*}
    Therefore by our algorithm design that {$\norm{\Kh_t} \le C_K$} for any $t$
    \begin{align}
    \begin{split}
    \label{eq:lemma3Cond2}
        \norm{x_t} &\le \norm{A+B \Kh_{t-1}}\norm{x_{t-1}} + \norm{B\eta_{t-1} + \varepsilon_{t-1}} \\
        &\le (\norm{A}+\norm{B}\norm{\Kh_{t-1}})\norm{x_{t-1}} + \norm{B\eta_{t-1} + \varepsilon_{t-1}} \\
        &\le (\norm{A}+\norm{B}C_K)\norm{x_{t-1}} + \norm{B\eta_{t-1} + \varepsilon_{t-1}} \\
        &\le (\norm{A}+\norm{B}C_K)^{i}\norm{x_{t-i}} + \sum_{k=0}^{i-1}(\norm{A}+\norm{B}C_K)^k\norm{B\eta_{t-1-k} + \varepsilon_{t-1-k}} \\
        &\le \max\{1, (\norm{A}+\norm{B}C_K)^{m}\}C_{x,t} + \sum_{k=0}^{m-1} (\norm{A}+\norm{B}C_K)^k\norm{B\eta_{t-1-k} + \varepsilon_{t-1-k}}   .
    \end{split}
    \end{align}
\end{enumerate}    
    By adding up \cref{eq:lemma3Cond1,eq:lemma3Cond2}, we have a bound that is applicable to both cases. Notice our previous assumption that $\norm{(A+BK_0)^m} \le \rho^m$, where $\rho < 1$, further take $\norm{(A+BK_0)^k}$, and $(\norm{A}+\norm{B}C_K)^k$ to be all bounded by a constant $M \ge 1$ for $k = 0, 1, \cdots, m$, which is of order $M = \calO(1)$ (because $m = \calO(1)$). By \cref{eq:lemma3Cond1,eq:lemma3Cond2}
    \begin{equation}   
    \label{eq:lemma3basicEq}
    \begin{aligned}
    \norm{x_{t}} &\le \norm{(A+BK_0)^m} \norm{x_{t-m}} + \sum_{k=0}^{m-1}\norm{(A+BK_0)^{k}}\norm{B\eta_{t-1-k} + \varepsilon_{t-1-k}} \\
    &\qquad + \max\{1, (\norm{A}+\norm{B}C_K)^{m}\} C_{x,t-i} + \sum_{k=0}^{m-1}(\norm{A}+\norm{B}C_K)^k\norm{B\eta_{t-k-1} + \varepsilon_{t-k-1}} \\
    &\le \rho^m \norm{x_{t-m}} + M\left(C_{x,t-i} + 2\sum_{k=0}^{m-1}  \norm{B\eta_{t-k-1} + \varepsilon_{t-k-1}}\right) \\    
    &\le \rho^m \norm{x_{t-m}} + M\left(C_{x,t} + 2\sum_{k=0}^{m-1}  \norm{B\eta_{t-k-1} + \varepsilon_{t-k-1}}\right)
    .\end{aligned}
    \end{equation}
    \cref{eq:lemma3basicEq} is very promising because it has a shrinking weight on $\norm{x_{t-m}}$. Let us use a simplified notation for the remainder:
    \[J_t := M\left(C_{x,t} + 2\sum_{k=0}^{m-1}  \norm{B\eta_{t-k-1} + \varepsilon_{t-k-1}} \right).\]
    In $\E [J_t^2] $ there are three types of components:
    \begin{itemize}
        \item $M = \calO(1)$
        \item $C_{x,t} = C_x \log(t)$
        \item $\E(\sum_{k=0}^{m-1}  \norm{B\eta_{t-k-1} + \varepsilon_{t-k-1}})^2 =  \calO(1)$.
    \end{itemize}
    Since 
    \begin{equation}
    \label{eq: E It2}
        \E [J_t^2] 
        \le M^2 \cdot 2(C_{x,t}^2 + 4\E(\sum_{k=0}^{m-1}  \norm{B\eta_{t-k-1} + \varepsilon_{t-k-1}})^2) 
        =
        \calO(\log^2(t)),
    \end{equation}
    we can control $\E\norm{x_t}^2$ by 
    \begin{equation}
    \label{eq: E xt 2}
    \begin{split}
    \E\norm{x_t}^2 &\le \E\left(\rho^m \norm{x_{t-m}} + J_t\right)^2 \\
    &= \rho^{2m} \E\norm{x_{t-m}}^2 + \E J_t^2 + 2\rho^m \E\norm{x_{t-m}}\abs{J_t} \\
    &\le \rho^{2m} \E\norm{x_{t-m}}^2 + \E J_t^2 + \frac{1-\rho^{2m}}{2} \E\norm{x_{t-m}}^2 + \frac{2\rho^{2m}}{1-\rho^{2m}}\E J_t^2 \\
    &= \frac{1+\rho^{2m}}{2} \E\norm{x_{t-m}}^2 + \frac{1+\rho^{2m}}{1-\rho^{2m}}\E J_t^2    \\
     & \qquad \text{ (because } 2ab \le a^2 +b^2   \text{ with } a^2 =  \frac{1-\rho^{2m}}{2} \norm{x_{t-m}}^2 \text{ and } b^2 = \frac{2\rho^{2m}}{1-\rho^{2m}} J_t^2).
    \end{split}
    \end{equation}
    
    By \cref{eq: E It2,eq: E xt 2}, 
    \begin{equation}
    \label{eq: E xt 2 mid}
    \begin{split}
    \E\norm{x_t}^2 &\le 
    \frac{1+\rho^{2m}}{2} \E\norm{x_{t-m}}^2 + \calO(\log^2(t)) \\
    &\le 
    (\frac{1+\rho^{2m}}{2})^2 \E\norm{x_{t-2m}}^2 + \frac{1+\rho^{2m}}{2}\calO(\log^2(t)) +  \calO(\log^2(t))\\
    &\le 
    (\frac{1+\rho^{2m}}{2})^{\lfloor \frac{t}{m}\rfloor} \E\norm{x_{t-m\lfloor \frac{t}{m}\rfloor}}^2 + \sum_{i=0}^{\lfloor \frac{t}{m}\rfloor-1}(\frac{1+\rho^{2m}}{2})^i \calO(\log^2(t)) \\
    &\le
     \E\norm{x_{t-m\lfloor \frac{t}{m}\rfloor}}^2 + \calO(\log^2(t)) \\
     & \qquad \text{(Recall that $\rho < 1$, and thus $\frac{1+\rho^{2m}}{2} < 1$)}.
    \end{split}
    \end{equation}
    
    Now it only remains to show that $\E\norm{x_{t-m\lfloor \frac{t}{m}\rfloor}}^2$ is bounded by some constant.
    Notice that
    \begin{align*}
        \E\norm{x_t}^2 
        \le& \E\left((\norm{A}+\norm{B}\norm{\Kh_t})\norm{x_{t-1}} + \norm{B}\norm{\eta_t} + \norm{\varepsilon_t}\right)^2 \\
        \le& 3\left((\norm{A}+\norm{B}C_K)^2\E\norm{x_{t-1}}^2 + \norm{B}^2\E\norm{\eta_t}^2 + \norm{\varepsilon_t}^2\right)\\
        \le& 3\left((\norm{A}+\norm{B}C_K)^2\E\norm{x_{t-1}}^2 + \norm{B}^2\tau^2 + \sigma^2\right) 
    .\end{align*}
    By iteratively applying this inequality down to $\E\norm{x_0}^2$, we know that for $ t \le m$:
    \[\E \norm{x_t}^2 = \calO(1).\]
    Thus
    following from \cref{eq: E xt 2 mid} we have
    \begin{equation*}
        \E \norm{x_t}^2 = \calO(\log^2(t)).
    \end{equation*}
    Since we already controlled the expectation of $\norm{x_t}^2$, it is straightforward to control the expectation of $\norm{u_t}^2$:

    \[u_t = \Kh_tx_t + \eta_t.\]
    \begin{align*}
        \E\norm{u_t}^2 &\le \E\norm{\Kh_tx_t + \eta_t}^2 \\
         &\le 2\E(\norm{\Kh_t}^2\norm{x_t}^2 + \norm{\eta_t}^2) \\
         &\le 2\E(C_K^2\norm{x_t}^2 + \norm{\eta_t}^2) \\
         &\le \calO(\log^2(t)) 
    .\end{align*}
    Thus,
    \begin{equation}
    \label{eq: E zt 2 final}
        \E\norm{z_t}^2 = \E\norm{x_t}^2 + \E\norm{u_t}^2 \le \calO(\log^2(t)) 
    \end{equation}
    Then we have
    
    \begin{equation*}
    \E\sum_{t=0}^{T-1}\norm{x_t}^2, \E\sum_{t=0}^{T-1}\norm{u_t}^2, \E\sum_{t=0}^{T-1}\norm{z_t}^2 \le  \calO(T\log^2(T))
    .\end{equation*}
\end{proof}

\subsection{Lemma in \cref{section: The proof of one_epoch_estimate_withMyalg}}

\subsubsection{The proof of \cref{lem:productBound}}
\label{The proof of lem:productBound}
\begin{lemma*}
Suppose we have a constant square matrix $M$ with spectral radius $\rho(M) < 1$, and a sequence of uniformly bounded random variables $\{\delta_t\}_{t=0}^\infty$, satisfying $\norm{\delta_t} \asConv 0$. 
Denote the constant $\rho_M:= \frac{2 + \rho(M)}{3} < 1$. 
Then we have, for any $t, q \in \mathbb{N}$, $t > q$:
\[\norm{(M + \delta_{t-1}) \cdots (M + \delta_{q})} = \calO(\rhoM^{t-q}) \as\]
And as a direct corollary
\[\norm{M^{t-q}} = \calO(\rhoM^{t-q}) .\] 
\end{lemma*}
\begin{proof}

Our assumption of stability only says $\rho(M)< 1$, but our analysis prefers similar exponential decay with regard to spectral norm.
First, we need a conversion between spectral radius and spectral norm. Define
 \begin{align*}
\transient{M}{ \rho} := \sup \left\{\norm{M^k} \rho^{-k} \colon k\geq 0 \right\}.
\end{align*}
For simplicity, let us denote 
\[\tau(M) := \tau\left(M, \frac{1 + \rho(M)}{2}\right).\]
and with Gelfand's Formula
\begin{align*}
    \rho (M)=\lim _{{k\to \infty }}\left\|M^{k}\right\|^{{{\frac  {1}{k}}}}
.\end{align*}
Thus $\tau(M)$ is finite because $\frac{1 + \rho(M)}{2} > \rho(M)$.
Since $\{\delta_t\}_{t=0}^\infty$ is uniformly bounded, we can assume an upper bound $U_\delta$ for $\norm{M + \delta_i}$. Let us now consider the spectral norm of $(M+\delta_{t-1})\cdots(M+\delta_{q})$. 
\begin{equation*}
\begin{split}
\norm{(M+\delta_{t-1})\cdots(M+\delta_{q})} 
\le& \sum_{m=0}^{t-q}\norm{M^{t-q-m}} \sum_{q \le k_1 < \cdots < k_m \le t-1} \prod_{j=1}^m \norm{\delta_{k_j}}\\
\le& \sum_{m=0}^{t-q}\tau(M)\left(\frac{1 + \rho(M)}{2}\right)^{t-q-m}  \sum_{q \le k_1 < \cdots < k_m \le t-1} \prod_{j=1}^m \norm{\delta_{k_j}}\\
=& \tau(M)\sum_{m=0}^{t-q}\left(\frac{1 + \rho(M)}{2}\right)^{t-q-m}  \sum_{q \le k_1 < \cdots < k_m \le t-1} \prod_{j=1}^m \norm{\delta_{k_j}}\\
=& \tau(M)
\left(\frac{1 + \rho(M)}{2}+\norm{\delta_{t-1}}\right)\cdots
\left(\frac{1 + \rho(M)}{2}+\norm{\delta_{q+1}}\right).
\end{split}
\end{equation*}

Since $\norm{\delta_t} \to 0 \as$, for every $\omega$ in the sample space $\Omega$, such that there exists some $T_1(\omega)$, whenever $t > T_1(\omega)$, $\frac{1 + \rho(M)}{2}+\norm{\delta_{t}} < \frac{2 + \rho(M)}{3} < 1$, then
\begin{align*}
\norm{(M+\delta_{t-1})\cdots(M+\delta_{q})} 
\le& \tau(M) 
\left(\frac{1 + \rho(M)}{2}+\norm{\delta_{t-1}}\right)\cdots
\left(\frac{1 + \rho(M)}{2}+\norm{\delta_{q+1}}\right) \\
\le& \tau(M)
\rho_M^{t-q-T_1(\omega)}(\frac{1 + \rho(M)}{2}+U_\delta)^{T_1(\omega)}  
.\end{align*}

Following \cref{defn: Big O notation} \cref{itm: big O as defn}:
\begin{equation*}
    \norm{(M+\delta_{t-1})\cdots(M+\delta_{q})}= \calO(\rhoM^{t-q}) \as
\end{equation*}

\end{proof}

\section{Experiment Details}
\label{section: Additional Experiments}

\subsection{Experiment Setting}
\label{subsection: Experiment on Stable System}
\subsubsection{Experiment Setting on Stable System}
\label{subsection: easy setting}
We set 
$A = \begin{bmatrix}
0.8 & 0.1 \\
0 & 0.8
\end{bmatrix}$ 
and 
$B = \begin{bmatrix}
0 \\
1 
\end{bmatrix}$, with system noise $\sigma = 1$, injected noise baseline $\tau = 1$, $Q = I_2$, $R = 1$ and initial state $x_0 = [0, 0]^\top$. As for the algorithmic hyper-parameters, we set the warning threshold for states $x_t$ at $C_x=1$ (so that $C_{x,t} = \log(t)$), the known stable controller $K_0 = [0,0]$, and the upper bound of the L2-norm for our controller $\Kh_t$ at $C_K = 5$. Note that this is conservative by about a factor of 10, since the true optimal controller in this system is
$K \approx \begin{bmatrix}
-0.10, -0.48 
\end{bmatrix}$. 
Recall that the choice of these hyper-parameters does not actually affect our theoretical coverage (as long as $C_K>\|K\|$) or regret guarantees, but in practice their values prevent the system from incurring very large regret in the first few time steps. Even for this, they are only needed because we do not assume we are given an initial controller that is very close to $K$;
in contrast, for instance, \citet{dean2018regret} started from a controller fitted with 100 samples of white noise actions. All stable system results are based on 1,000 independent runs of \cref{alg:myAlg} for $T = 10,000$ time steps.

\subsubsection{Experiment Setting on Unstable System}
\label{subsection: hard experiment setting}
The unstable system we simulate is highly unstable, and is largely the same as that in Appendix H of \citet{dean2018regret}.
We set 
$A = \begin{bmatrix}
2 & 0 & 0 \\
4 & 2 & 0 \\
0 & 4 & 2
\end{bmatrix}$ 
and 
$B = I_3$, with system noise $\sigma = 1$, injected noise baseline $\tau = 1$, $Q = 10I_3$, $R = I_3$ and initial state $x_0 = [0, 0, 0]^\top$. As for the hyper-parameters, we set the warning threshold for states $x_t$ at $C_x=1$ (so that $C_{x,t} = \log(t)$), and we examined two different choices for the known stabilizing controller: $K_0 = 
-\begin{bmatrix}
1.5 & 0 & 0 \\
0 & 1.5 & 0 \\
0 & 0 & 1.5
\end{bmatrix}$ and 
$K_0 = 
-\begin{bmatrix}
1.5 & 0 & 0 \\
3.5 & 1.5 & 0 \\
0 & 3.5 & 1.5
\end{bmatrix}$. The former choice incurs quite a bit higher regret than the latter, and hence we refer to the former as the `bad' stabilizing controller and to the latter as the `good' stabilizing controller. We set the upper bound of the L2-norm for our controller $\Kh_t$ at the level of $C_K = 1000$. Our choice of $K_0$ is different from the starting point in \citet{dean2018regret}, where they started from a $T=250$ burn in period estimate, and did not report the regret in the first 250 steps. 
All unstable system results are based on 1,000 independent runs of \cref{alg:myAlg} for $T = 5,000$ time steps.

\subsection{Experiment on Unstable System}
\label{subsection: Experiment on Unstable System}
In contrast to the stable system simulation summarized in \cref{subsection: Partial experiment result}, in this section we simulate the severely unstable system described in \cref{subsection: hard experiment setting}. In this setting, the specification of $K_0$ is critical due to the costs incurred at the early time steps, an unavoidable consequence of starting from limited information in a system that can rapidly spiral (nearly) out of control.

\subsubsection{Summary of results on unstable system}
We begin with the analogue of \cref{fig:Summary stable system} for the unstable system, given in \cref{fig:Summary unstable system}. The main takeaways are the same as the discussion in \cref{subsection: Partial experiment result}.

\begin{figure}[H]
\centering

\begin{subfigure}{.45\textwidth}
  \captionsetup{justification=centering}
  \caption{Benefit of stepwise update}
  \includegraphics[width = \linewidth]{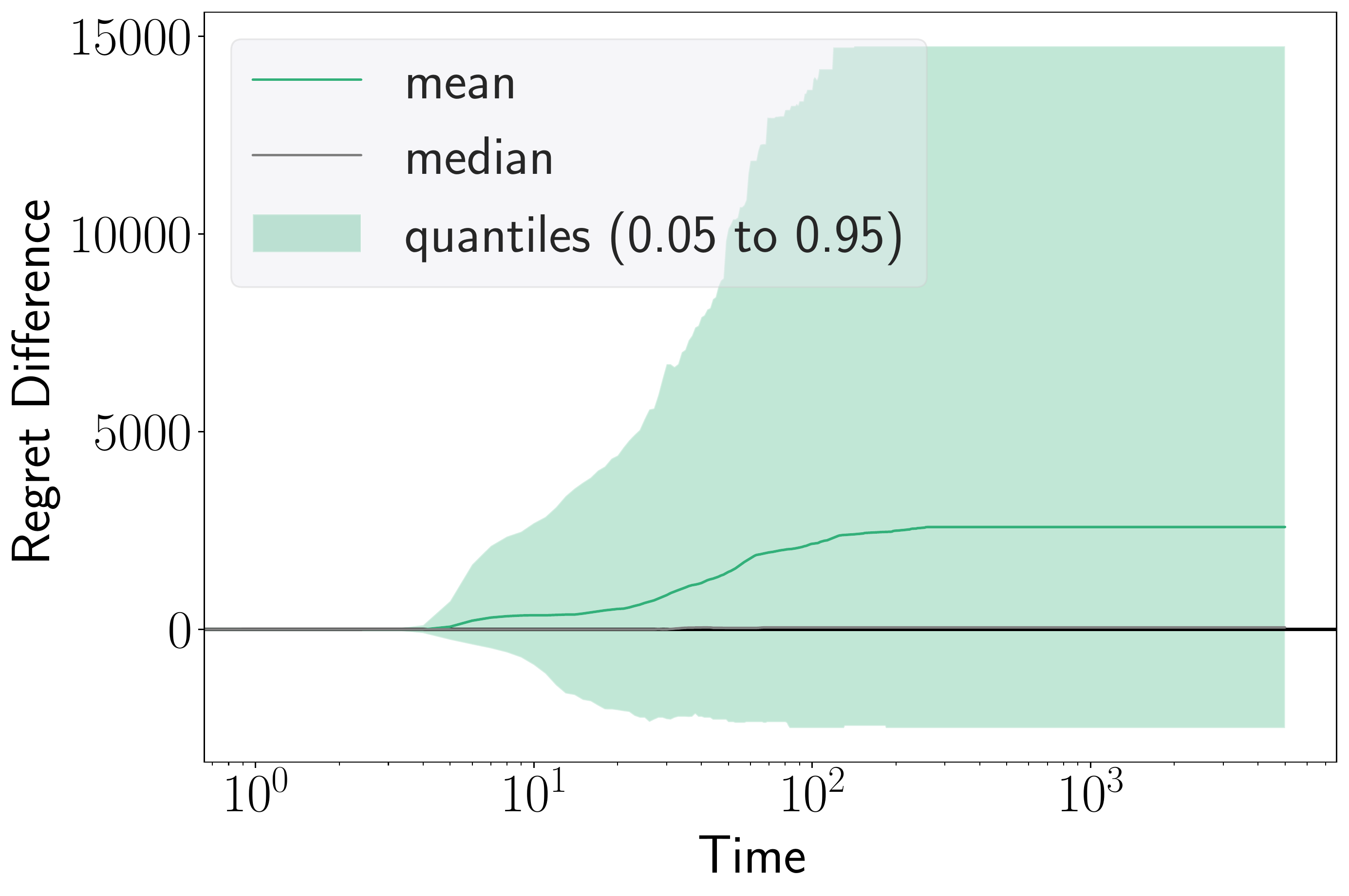}
  \label{fig:Compare Regret one half unstable}
\end{subfigure}
\begin{subfigure}{.45\textwidth}
  \caption{Regret Ratio}
  \includegraphics[width = \linewidth]{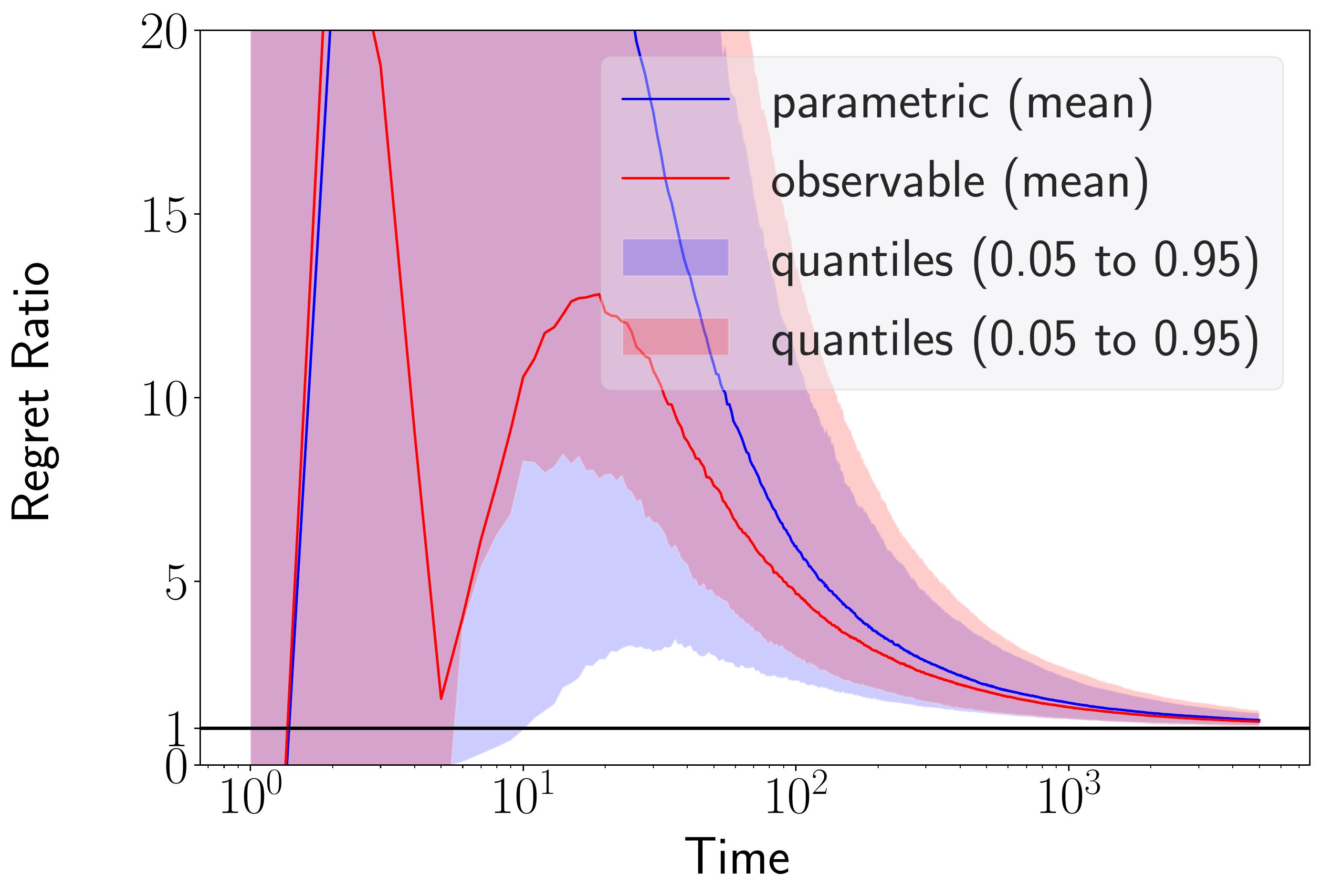}
  \label{fig:regret ratio one half unstable}
\end{subfigure}

\begin{subfigure}{.45\textwidth}
\captionsetup{justification=centering}
  \caption{Differing Convergence Rates}
  \includegraphics[width = \linewidth]{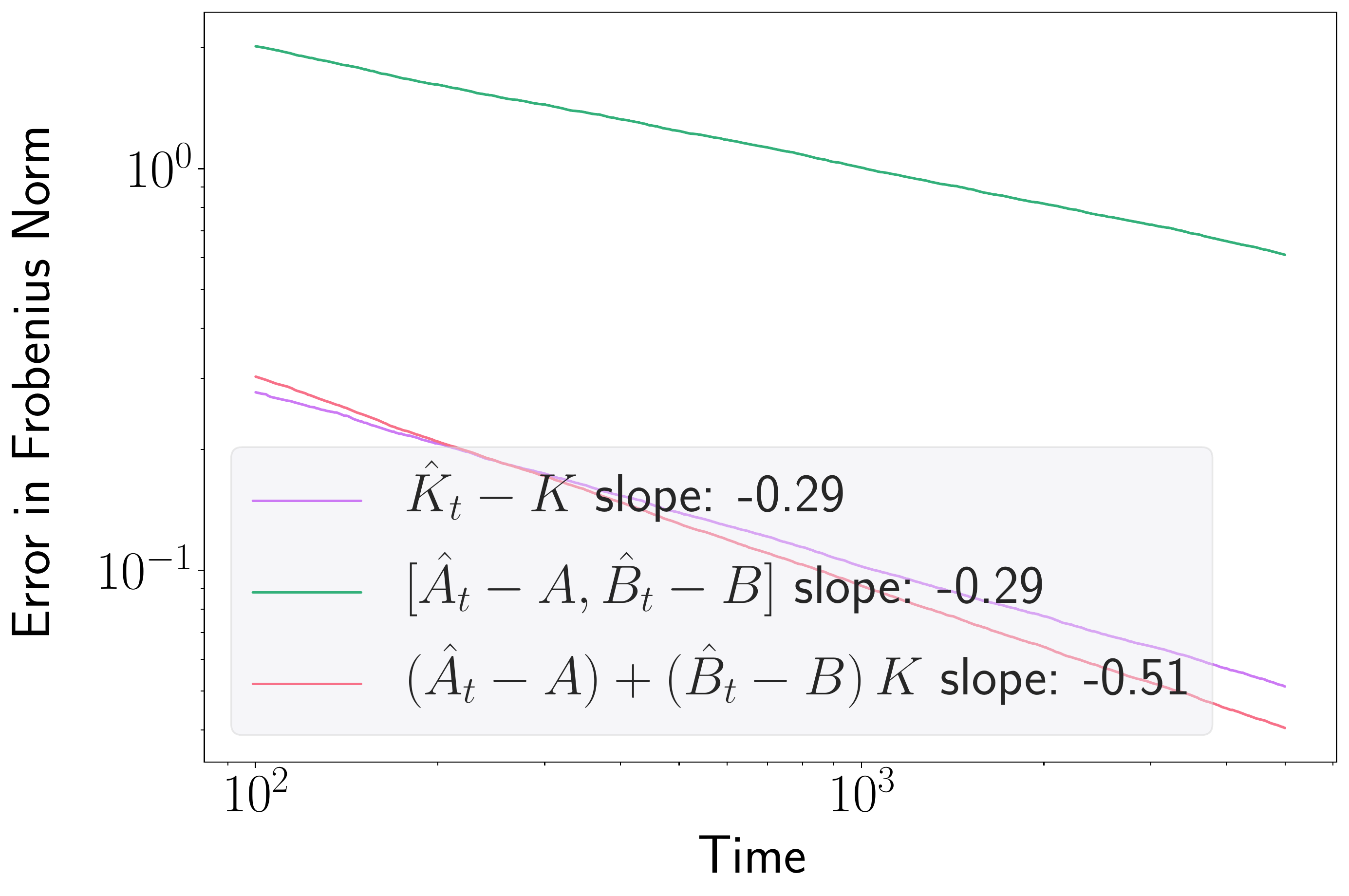}
  \label{fig:estimation rate one half unstable}
\end{subfigure}
\begin{subfigure}{.45\textwidth}
  \caption{Confidence Region Coverage}
  \includegraphics[width = \linewidth]{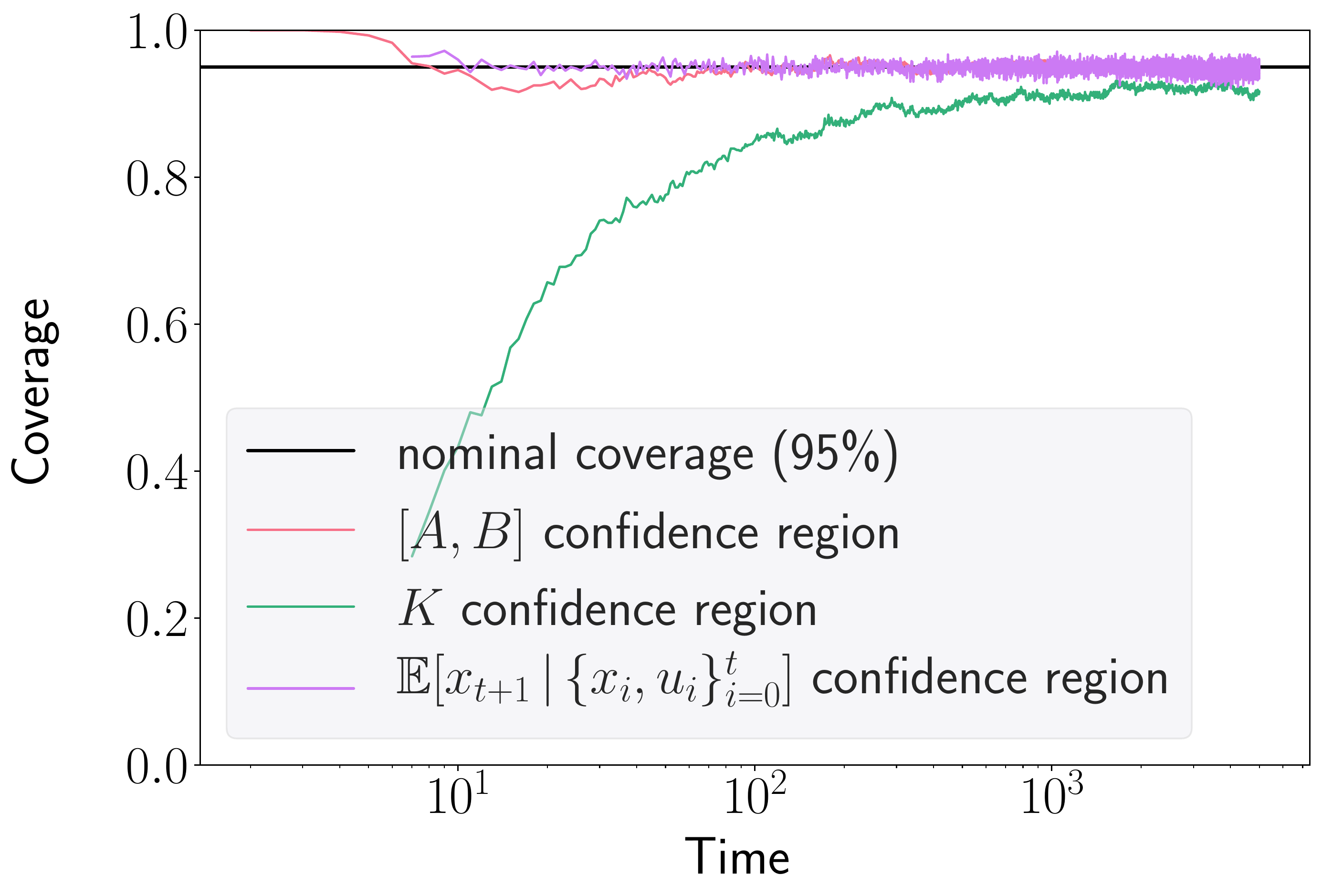}
  \label{fig:Coverage one half unstable}
\end{subfigure}

\begin{subfigure}{.45\textwidth}
  \caption{Prediction Region Coverage}
  \includegraphics[width = \linewidth]{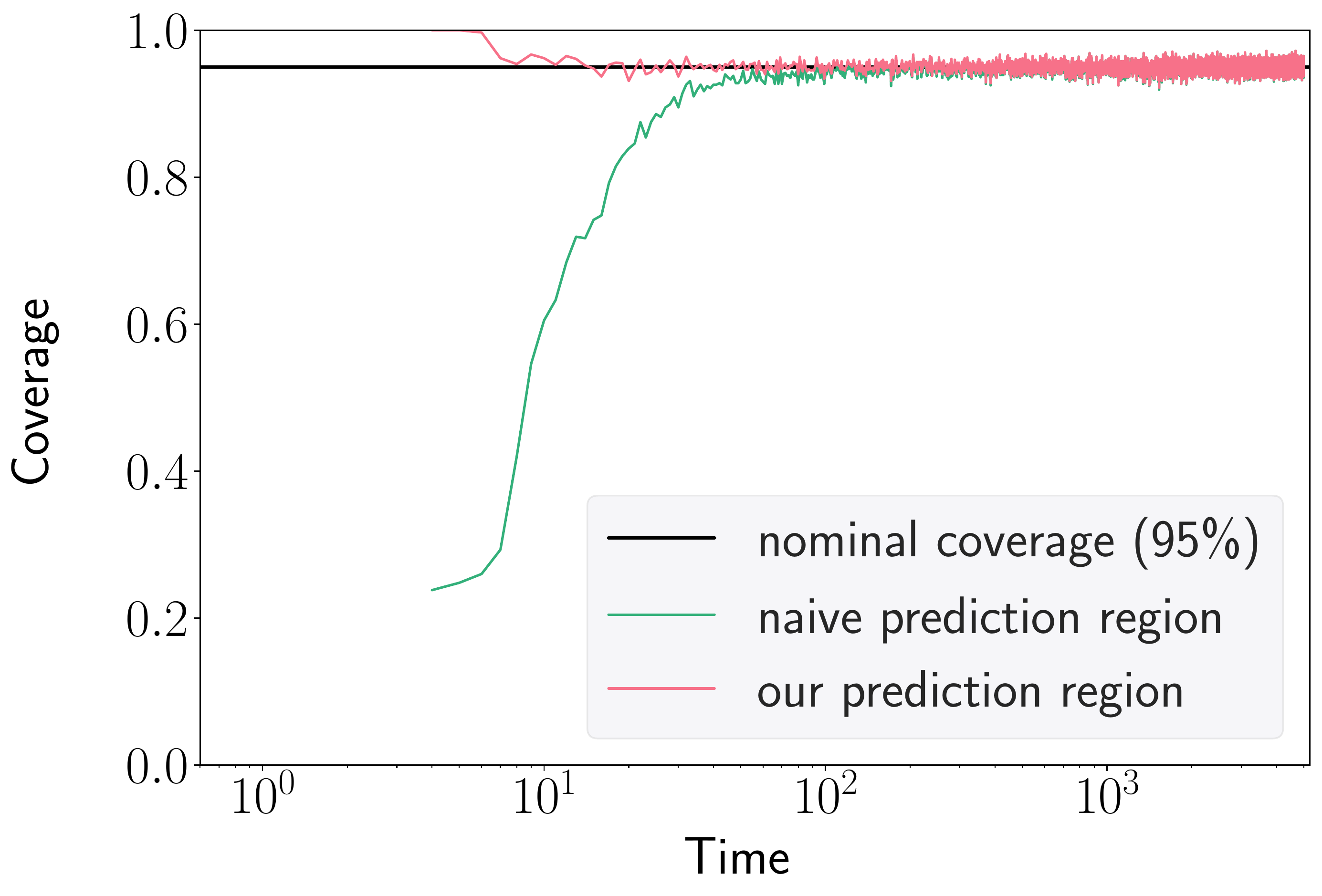}
  \label{fig:Prediction Coverage True one half unstable}
\end{subfigure}
\end{figure}

\begin{figure}[H]
\ContinuedFloat

\caption{Summary of 1000 independent experiments applying \cref{alg:myAlg} with $\beta = 0.5$, $\alpha = 2$, $C_x=1$, $C_K=5$, and $K_0 = -\begin{bmatrix}
1.5 & 0 & 0 \\
3.5 & 1.5 & 0 \\
0 & 3.5 & 1.5
\end{bmatrix}$ 
to the unstable system described in \cref{subsection: hard experiment setting}. (a) Difference between the regret of \cref{alg:myAlg} using stepwise and logarithmic updates. 
(b) The ratio of the empirical regret and our parametric or observable expressions for the regret. 
(c) The average Frobenius norm of various estimation errors considered in this paper, with slopes fitted on a log-log scale so that the estimation error is $\logO(t^{\text{slope}})$. 
The effect of $\alpha$ was removed from the slopes of $\Kh_t-K$ and $[\Ah_t-A,\Bh_t-B]$ by dividing the error by $\log^{\alpha/2}(t)$.
(d) Coverage of our 95\% confidence regions for $[A,B]$, $K$, and $\E [x_{t+1} \,|\, \{x_i, u_i\}_{i=0}^{t} ] = Ax_t + Bu_t$. 
(e) Coverage of our 95\% prediction region for $ x_{t+1} \,|\, \{x_i, u_i\}_{i=0}^{t}$, along with coverage of the naive prediction region given in \cref{eq: prediction region observable simple}.
}
    \label{fig:Summary unstable system}
\end{figure}

\subsubsection{Large Regret From Early Time Steps}
\label{subsubsection: Large Regret at Burn in Period}
For the `bad' choice of stabilizing controller $K_0 = 
-\begin{bmatrix}
1.5 & 0 & 0 \\
0 & 1.5 & 0 \\
0 & 0 & 1.5
\end{bmatrix}$, we plot the log regret in subplot (a) of \cref{fig:log regret three choices unstable system}.
We observe a rapidly increasing regret in the first roughly 200 time steps, which dominates all the regret in the remaining steps. 
We offer a brief explanation why the cost in the early time steps is very large despite assuming knowledge of a stabilizing yet sub-optimal controller $K_0$. Notice $A+BK_0 = \begin{bmatrix}
0.5 & 0 & 0 \\
4 & 0.5 & 0 \\
0 & 4 & 0.5
\end{bmatrix}$. Thus 
$(A+BK_0)^2 = \begin{bmatrix}
0.25 & 0 & 0 \\
4 & 0.25 & 0 \\
16 & 4 & 0.25
\end{bmatrix}$,
$(A+BK_0)^3 = \begin{bmatrix}
2^{-3} & 0 & 0 \\
3 & 2^{-3} & 0 \\
24 & 3 & 2^{-3}
\end{bmatrix}$,
$(A+BK_0)^4 = \begin{bmatrix}
2^{-4} & 0 & 0 \\
2 & 2^{-4} & 0 \\
24 & 2 & 2^{-4}
\end{bmatrix}$,
$(A+BK_0)^5 = \begin{bmatrix}
2^{-5} & 0 & 0 \\
1.25 & 2^{-5} & 0 \\
20 & 1.25 & 2^{-5}
\end{bmatrix}$,
$(A+BK_0)^6 = \begin{bmatrix}
2^{-6} & 0 & 0 \\
0.75 & 2^{-6} & 0 \\
15 & 0.75 & 2^{-6}
\end{bmatrix}$. So although we have a controlled system with maximum eigenvalue $0.5$, the power of $(A+BK_0)^k$ can still be very large in the bottom left corner for $k = 2,3,4,5,6$. Because of this, the randomness in the states is enlarged and propagated to several future steps. It turns out that, at the first 200 steps we used this high cost safety policy $K_0$ a lot as we do not have a good estimate of optimal controller $K$, and that is the real reason for this high burn-in period cost. As we will see later, if we change the stabilizing controller $K_0$ to be closer to the optimal $K$, the regret will be much smaller.

\subsubsection{Comparison with Thompson Sampling}
\label{subsubsection: TS unstable system}
For comparison, we implement a straightforward version of Thompson sampling as follows. 
Denote $\Theta := [A, B]$. We use a prior of
\begin{equation*}
    \vvector[\Theta _{prior}] \sim \calN(\vvector[\Theta], I_{n(n+d)})
.\end{equation*}
Using the Bayesian updating equations and denoting the least-squares estimate of $\Theta$ by $\hat{\Theta}_t = [\Ah_t,\Bh_t]$, the posterior at time $t$ is given by
\begin{equation*}
    \vvector[\Theta_t^{TS}] \sim 
    \calN\left(
    \vvector
    \left[
    \left(
    \Theta + \hat\Theta_t\sum_{i=0}^{t-1}\begin{bmatrix}         x_i\\         u_i\\ \end{bmatrix} \begin{bmatrix}         x_i\\         u_i\\ \end{bmatrix}^\top 
    \right)
    \left(
    I_{n+d}+\sum_{i=0}^{t-1}\begin{bmatrix}         x_i\\         u_i\\ \end{bmatrix} \begin{bmatrix}         x_i\\         u_i\\ \end{bmatrix}^\top 
    \right)^{-1}
    \right],     
    \left(
    I_{n+d}+\sum_{i=0}^{t-1}\begin{bmatrix}         x_i\\         u_i\\ \end{bmatrix} \begin{bmatrix}         x_i\\         u_i\\ \end{bmatrix}^\top 
    \right)^{-1}
    \otimes
    I_n
    \right)
.\end{equation*}
At each step, we draw a sample $\Theta_t^{TS}$ from this posterior and use it as the input to the DARE for calculating $\Kh_t$. Since a system is stabilizable if $\text{rank}([A-\lambda I,B])=n$ for any eigenvalue $\lambda$ of $A$ \citep{hautus1970stabilization}, the Gaussian posterior puts probability 1 on stabilizable $\Theta=[A,B]$ and hence defines a unique solution to the DARE with probability 1 as well.

We report the Thompson sampling regret in subplot (b) of \cref{fig:log regret three choices unstable system}, and see that it also suffers from rapidly increasing regret at early time points.f


\subsubsection{Improved Regret When Using `Good' $K_0$}
\label{subsubsection: Start with good choice of K_0}
\begin{figure}[!htb]
    \centering
    \includegraphics[width = 0.95\textwidth]{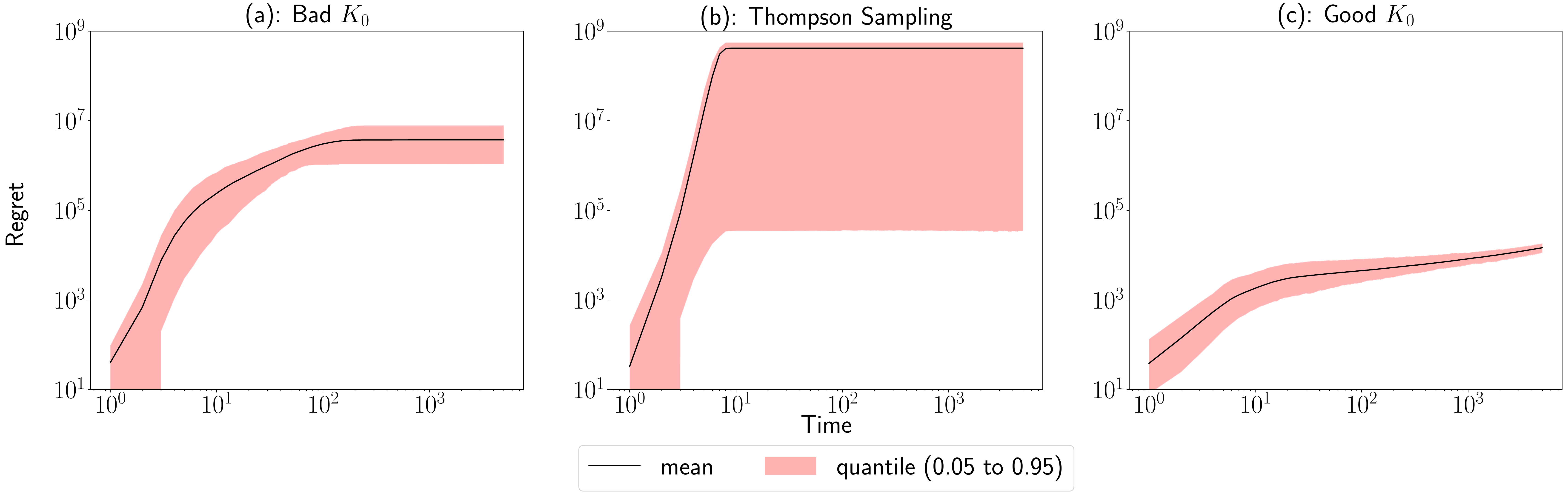}
    \caption{Regret on the log scale based on 1000 independent experiments on the unstable system for $\beta = 0.5$ and $\alpha = 0$. (a): Bad safety controller $K_0 = 
    -\begin{bmatrix}
    1.5 & 0 & 0 \\
    0 & 1.5 & 0 \\
    0 & 0 & 1.5
    \end{bmatrix}$; (b): Thompson Sampling; (c): Good safety controller $K_0 = -\begin{bmatrix}
    1.5 & 0 & 0 \\
    3.5 & 1.5 & 0 \\
    0 & 3.5 & 1.5
    \end{bmatrix}$.
    }
    \label{fig:log regret three choices unstable system}
\end{figure}

When we switch from the `bad' stabilizing controller to the `good' one specified in \cref{subsection: hard experiment setting} as
$K_0 = -\begin{bmatrix}
1.5 & 0 & 0 \\
3.5 & 1.5 & 0 \\
0 & 3.5 & 1.5
\end{bmatrix}$, we get that $A+BK_0 = \begin{bmatrix}
0.5 & 0 & 0 \\
0.5 & 0.5 & 0 \\
0 & 0.5 & 0.5
\end{bmatrix}$, which is a much better starting point than the previous $ \begin{bmatrix}
0.5 & 0 & 0 \\
4 & 0.5 & 0 \\
0 & 4 & 0.5
\end{bmatrix}$, and the regret in this setting is indeed much better (see subplot (c) of \cref{fig:log regret three choices unstable system}) and resembles that of the stable system described in \cref{subsection: easy setting}.

\subsection{Choices of $\beta$ other than $0.5$}
Our simulations consider choices of $\beta$ beyond $0.5$ and even beyond those covered by our theory. In particular, we consider $\beta = 0.1, 0.3, 0.5, 0.7, 0.9$ and observe promising evidence that some of our asymptotic coverage results may generalize to the setting of $\beta < 1/2$.

\subsubsection{Regret}
\label{apdx subsection: regret}
\begin{figure}[!htb]
\centering
\begin{subfigure}{.95\textwidth}
  \caption{Stable System}
  \includegraphics[width = \linewidth]{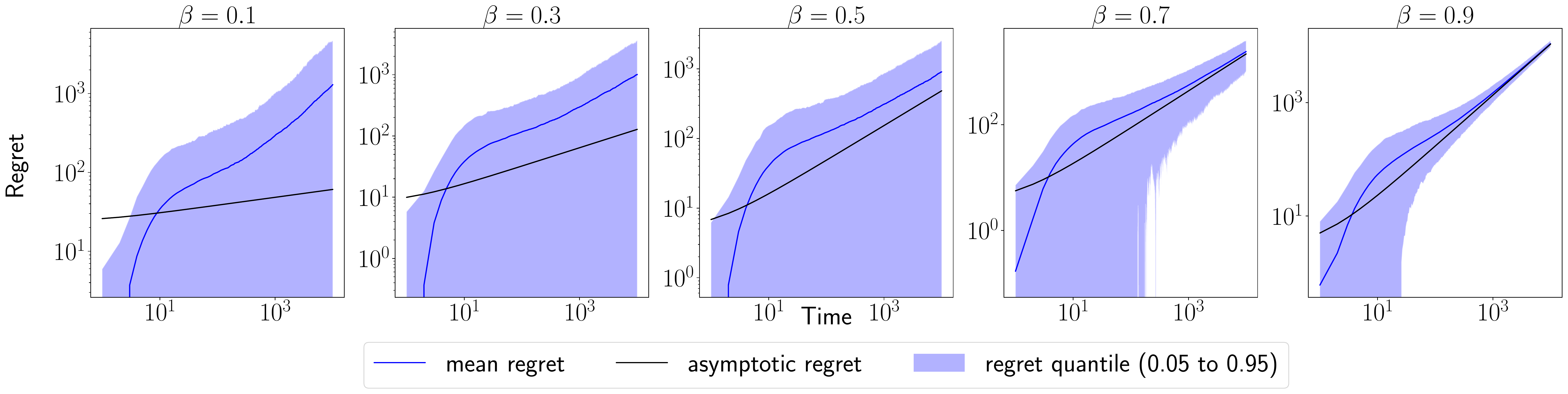}
  \label{fig:regret_stable_5beta}
\end{subfigure}

\begin{subfigure}{.95\textwidth}
  \caption{Unstable System}
  \includegraphics[width = \linewidth]{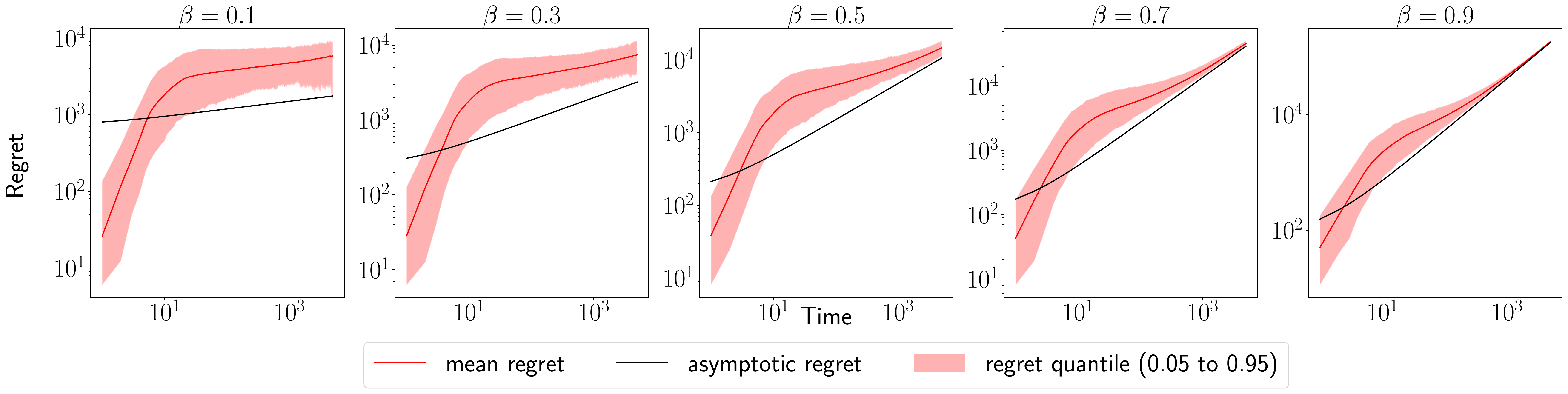}
  \label{fig:regret_unstable_5beta}
\end{subfigure}

\caption{Regret on the log scale based on 1000 independent experiments for $\beta = 0.1, 0.3, 0.5, 0.7, 0.9$ and $\alpha = 0$. (a): stable system; (b): unstable system.}
\label{fig:regret_stable_unstable}
\end{figure}

According to \cref{thm:regret} the dominating term for regret should be $ T^\beta \log^\alpha(T) \Tr((B^\top PB +R)\frac{\tau^2}{\beta})$ for any $\beta \in [1/2,1)$ and $\max\{\beta,\,\alpha -1\} > 1/2$, and that indeed matches with our experimental results (see \cref{fig:regret_stable_unstable}). The asymptotic regret expression from \cref{thm:regret} is represented as the black solid curve, which converges to the empirical regret for $\beta >0.5$, but not $\beta <0.5$.

\subsubsection{Confidence region coverage}
\label{appendix subsection: confidence coverage}
\begin{figure}[!htb]
\centering
\begin{subfigure}{.95\textwidth}
  \caption{Stable System}
  \includegraphics[width = \linewidth]{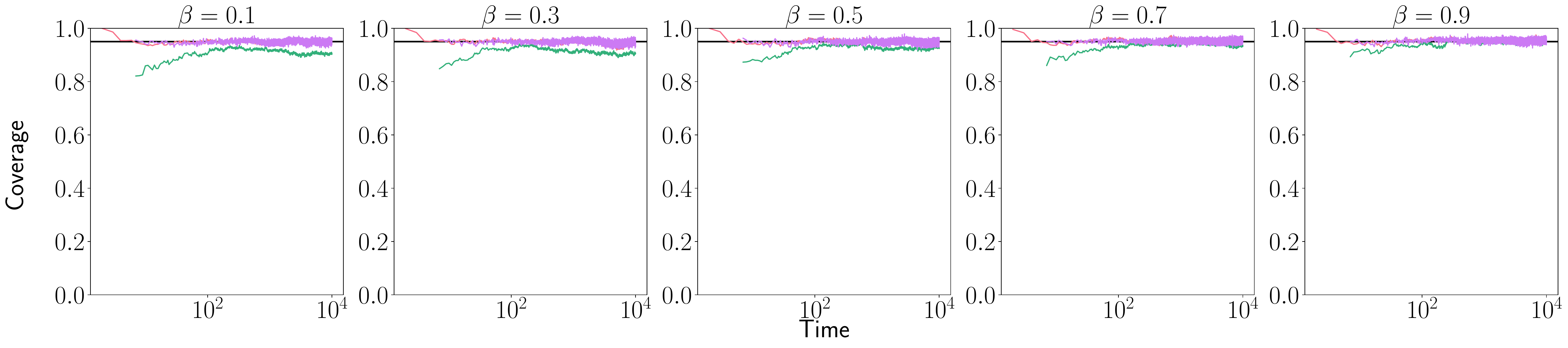}
  \label{fig:Coverage_stable_5beta}
\end{subfigure}

\begin{subfigure}{.95\textwidth}
  \caption{Unstable System}
  \includegraphics[width = \linewidth]{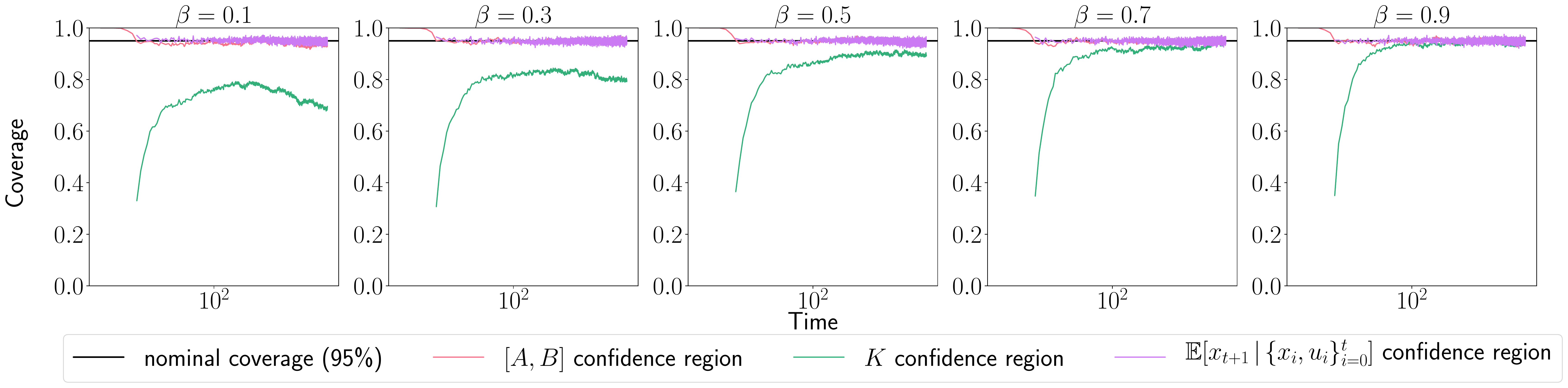}
  \label{fig:Coverage_unstable_5beta}
\end{subfigure}

\caption{Coverage on the log scale based on 1000 independent experiments for $\beta = 0.1, 0.3, 0.5, 0.7, 0.9$ and $\alpha = 0$. (a): stable system; (b): unstable system.}
\label{fig:Coverage_stable_unstable}
\end{figure}

\cref{fig:Coverage_stable_unstable} shows that the finite sample coverage of our confidence regions and prediction region closely matches the asymptotic theory from \cref{thm:main,corr: K confidence region,thm: prediction CLT} for any choice among $\beta = 0.1, 0.3, 0.5, 0.7, 0.9$, with the exception of confidence regions for $K$, which seem to only work for the $\beta\ge 0.5$ covered by our theory. 


\subsection{Algorithm design}
We now investigate how the details of \cref{alg:myAlg} (the stabilizing controller $K_0$ and the thresholds on $x_t$ and $\|\Kh_t\|$) impact the regret.



\begin{figure}[!htb]
    \centering
    \includegraphics[width = 0.95\textwidth]{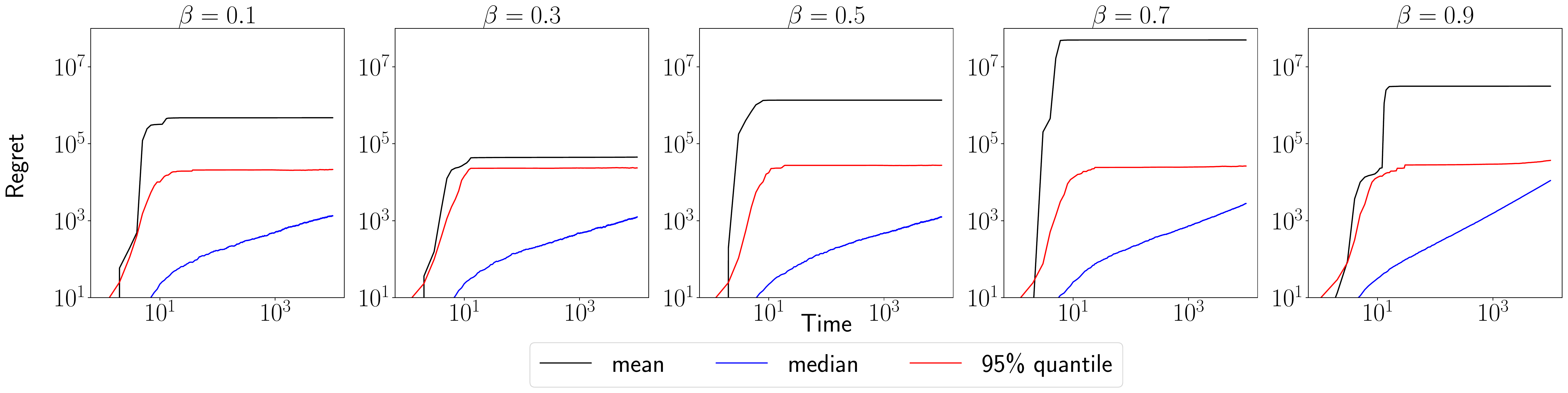}
    \caption{Regret on the log scale with no $C_{x,t}$ threshold on $\norm{x_t}$ based on 1000 independent experiments on stable system for $\beta = 0.1, 0.3, 0.5, 0.7, 0.9$ with $C_K = 5$ and $\alpha = 0$. 
    }
    \label{fig:log regret no Cx}
\end{figure}

\begin{figure}[!htb]
    \centering
    \includegraphics[width = 0.95\textwidth]{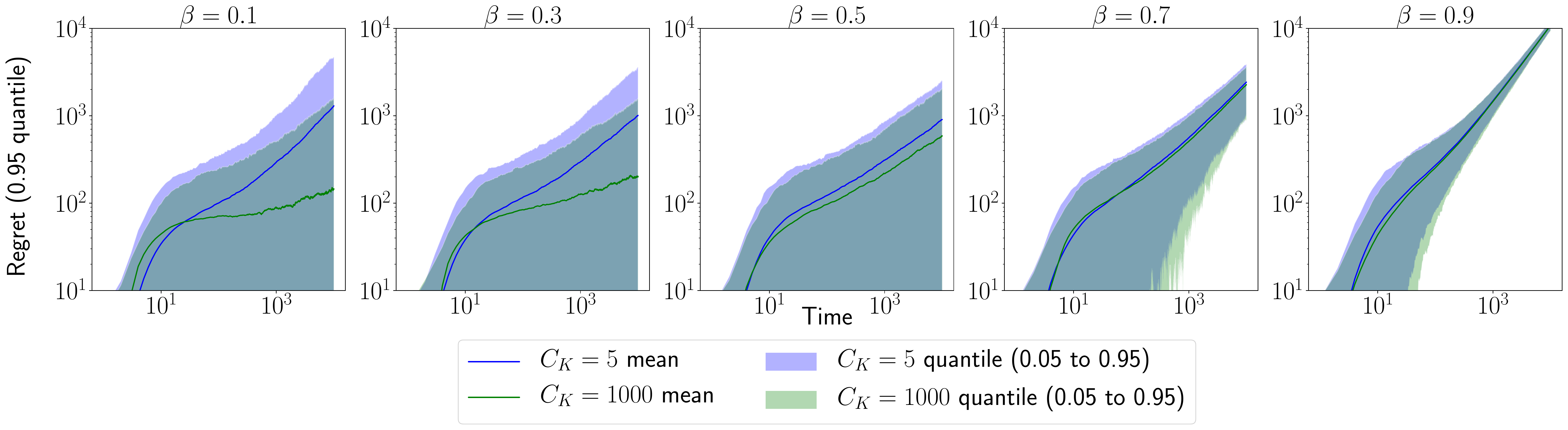}
    \caption{Regret on the log scale based on 1000 independent experiments on stable system for $\beta = 0.1, 0.3, 0.5, 0.7, 0.9$ with $\alpha = 0$ comparing $C_K = 5$ and $C_K = 1000$. 
    }
    \label{fig:log regret CK}
\end{figure}

\paragraph{The threshold $C_{x,t}$ controls extreme tail behavior}
Although we only trigger the threshold $C_{x,t}$ rarely, without it we can see some extreme behavior with low probability. In particular, when this threshold constraint is removed, we occasionally observe very large regret in early time steps due to the poor estimate $\Kh_t$, which causes instability of the system (see  \cref{fig:log regret no Cx} and compare it to the purple line and shaded region in \cref{fig:log regret CK}). The mean value is even higher than the 0.95 quantile curve because of several extremely large regrets induced by the unstable closed-loop system. And compared to when $C_{x,t}$ is used in \cref{fig:log regret CK}, the 0.95 quantile when $C_{x,t}$ is not used is considerably higher, although its median is quite similar to the mean when $C_{x,t}$ is used. 

\paragraph{Stepwise updating improves regret over logarithmic updating}
As our theory provides guarantees for \cref{alg:myAlg} with both stepwise and logarithmic updating, we run experiments to compare the regret of these two choices. 
\cref{fig:estimation rate one half,fig:estimation rate one half unstable} show the difference in regret between \cref{alg:myAlg} and the same algorithm but that only updates its estimates of the system parameters logarithmically often, i.e., at times $t=1, 2, 4, 8, \dots$ On average, we see a steady logarithmic increase in regret from switching from stepwise updates to logarithmic frequency.


\paragraph{A stabilizing controller $K_0$ closer to $K$ improves performance}
Although $K_0$ is a stabilizing controller by assumption, bad choices of $K_0$ can still make $(A+BK_0)^k$ large for some finite $k$
(see \cref{subsubsection: Large Regret at Burn in Period} for a concrete example). Thus, unsurprisingly, choosing $K_0$ to be as near as possible to the optimal controller $K$ produces smaller regret, as evidenced by 
\cref{fig:log regret three choices unstable system}.

\paragraph{Regret is robust to conservative choices of $C_K$}
To check the sensitivity of the choice of $C_K=5$ in the stable system, we also tried a looser bound $C_K =1000$. We found that the norm of $\Kh_t$ never surpassed the $C_K =1000$ bound. This larger $C_K$ made little difference for settings covered by our theory ($\beta\ge 0.5$), and surprisingly seems to actually improve the regret for smaller $\beta$ (see \cref{fig:log regret CK}).

\clearpage

\end{document}